\documentclass[10pt,twocolumn,letterpaper]{article}

\usepackage[pagenumbers]{cvpr}

\usepackage{multirow}
\usepackage{algorithm}
\usepackage{algpseudocode}
\usepackage{scalefnt}
\newcommand{\len}{\operatorname{len}}

\definecolor{cvprblue}{rgb}{0.21,0.49,0.74}
\usepackage[pagebackref,breaklinks,colorlinks,allcolors=cvprblue]{hyperref}

\def\confName{CVPR}
\def\confYear{2027}

\title{Leveraging Contrastive Learning for a Similarity-Guided Tampered Document Data Generation Pipeline}

\author{Mohamed Dhouib$^{1}$ \quad Davide Buscaldi$^{2}$ \quad Sonia Vanier$^{1}$ \quad Aymen Shabou$^{3}$\\[2pt]
{\small $^{1}$LIX, \'Ecole Polytechnique, IP Paris, France}\\
{\small $^{2}$LIPN, Universit\'e Sorbonne Paris Nord, France}\\
{\small $^{3}$DataLab Groupe, Cr\'edit Agricole S.A, France}}
\hypersetup{pdftitle={Leveraging Contrastive Learning for a Similarity-Guided Tampered Document Data Generation Pipeline},
pdfauthor={Mohamed Dhouib, Davide Buscaldi, Sonia Vanier, Aymen Shabou}}
\begin{document}
\maketitle

\begin{abstract}
Detecting tampered text in document images is a challenging task due to data scarcity, which has led previous work to propose rule-based pipelines for generating tampered documents. In this work, we identify two recurring failure modes in these pipelines: visual mismatches between source and target regions, and poorly defined crops that cut through characters or include neighboring content. Both leave highly visible artifacts that are rarely observed in real-world manipulations, undermining the ability of detectors trained on such data to learn robust, generalizable features. To address these failure modes, we introduce a crop-similarity network that compares source and target regions and a bounding-box quality assessment network that evaluates whether a crop tightly encloses the intended characters. We train the crop-similarity network using contrastive learning with a novel strategy for mining visually consistent positives. Our similarity-guided generation framework leverages both networks to produce high-quality tampered document images. To isolate the effect of the generation strategy, we train seven models under identical protocols on datasets generated from the same source images using our method and existing approaches. 
 Across architectures and open-source datasets, our framework yields consistent performance improvements, with an average relative pixel-level F1 gain of 23.1\%.

\end{abstract}


\section{Introduction}
Due to the presence of sensitive information, document images are frequent targets of malicious tampering, making the development of robust detection methods essential. Pretraining a visual model for this task is a promising approach, as pretrained models have consistently demonstrated strong performance across a wide range of visual tasks. This has become a standard strategy in natural image forgery detection~\cite{Trufor,ObjectFormer}. However, in the case of document images, large-scale human-made tampered datasets are not publicly available, and manually creating them is time-consuming and costly. This scarcity has led previous work \cite{Doctamper,newdataandopenopen} to rely on rule-based pipelines to synthetically generate tampered documents, simulating one or more of the following common tampering operations:
\begin{itemize}
    \item[(a)] \textbf{Copy-move}: a text region is copied and pasted within the same image.
    \item[(b)] \textbf{Splicing}: a text region is copied from a source image and inserted into a different target image.
    \item[(c)] \textbf{Insertion}: new text is added directly to the image.
    \item[(d)] \textbf{Inpainting}: text is removed using background-aware filling techniques.
    \item[(e)] \textbf{Coverage}: a text region is hidden by overlaying a visually similar background patch.
\end{itemize}
However, previous pipelines often exhibit two recurring failure modes. First, source and target regions may differ in visual aspects such as font or noise level. Second, selected regions may cut through characters or include neighboring content. Both introduce visible artifacts rarely observed in real-world manipulations, leading models trained on such data to overfit to these shortcuts and generalize poorly. In contrast, human-made manipulations, particularly those involving copy-move, splicing, and coverage, can result in high-quality alterations that are difficult to detect. This is mainly because humans tend to choose regions with similar backgrounds and text characteristics, allowing tampered areas to blend seamlessly into their surroundings. To our knowledge, we are the first to identify and address both failure modes. To this end, we train a crop-similarity network through contrastive learning to assess the compatibility between candidate source and target regions, and a bounding-box quality assessment network to evaluate whether a crop tightly encloses the intended characters. During generation, the first network selects visually compatible regions, while the second rejects imprecise crops.
 Our contributions are:
\begin{itemize}
    \item We identify two failure modes in existing generation pipelines: visual mismatches between source and target regions, and imprecise crop boundaries.
    \item We introduce a crop-similarity network trained through contrastive learning and a bounding-box quality assessment network to address these failure modes, and integrate them into a similarity-guided generation framework supporting five manipulation types.
     \item Through controlled comparisons using the same source images and a fixed training protocol per detector, we show that our framework consistently improves the performance of seven models across different datasets.
     \item We release \href{https://huggingface.co/datasets/MohamedDhouib1/TDoc-2.8M}{TDoc-2.8M}, a dataset of approximately 2.8 million tampered and pristine document images. The generation code, training scripts, and pretrained model weights are available through our \href{https://github.com/Mohamed-Dhouib/TDOC}{GitHub repository}.
\end{itemize}

\section{Related work}
Due to the unavailability of open-source datasets, early work~\cite{old0,old1,old2} on document tampering detection and localization was limited to training and testing on private data. Later, ~\citet{wang2022b} introduced T-SROIE, a small-scale dataset for evaluation, focusing on character-level replacements generated by the SRNet \cite{SRNET} model. However, SRNet often leaves clear visual artifacts, which can lead to overestimated real-world performance. More recently, \citet{Doctamper} introduced DocTamper, a dataset of approximately 170{,}000 samples generated by a rule-based pipeline that relies on text bounding-box height and width, as well as foreground and background colors extracted from the Sauvola \cite{SAUVOLA} algorithm. However, the tampered images often contain visual artifacts not typically found in human-made manipulations, largely due to Sauvola’s inaccurate segmentation and the rule-based pipeline’s inability to account for visual factors like font consistency and text alignment. In addition, the DTD model proposed in \citet{Doctamper} was evaluated solely on the DocTamper and T-SROIE datasets, yielding strong results that obscure its limited performance on real-world manipulations. Subsequent works~\cite{dtdv2freq,dtdv3} have built on this work and proposed architectural improvements, but follow the same evaluation strategy. More recent work has explored complementary directions~\cite{yi2026frequency,li2025ditl2,zheng2025generalized,qu2025revisiting,qu2026detect}. Recently, \citet{forensichub} introduced ForensicHub, a unified benchmark for image forensics, and the reported results show that state-of-the-art models pretrained on DocTamper exhibit limited generalization. On the other hand, \citet{newdataandopenopen} focused on concealing splicing via local post-processing and image blending. However, these operations often introduce undesirable artifacts. Recently, \citet{dataset2} proposed FSTS, which samples realistic editing operation sequences and parameter distributions to synthesize a large-scale dataset of around 290{,}000 samples. Although these distributions are closer to human-made tampering, the final synthesis pipeline remains primarily rule-based, and the resulting forgeries are still low quality. To enable robust evaluation, \citet{findit} introduced FindIt, a dataset of human-made forgeries created using a variety of tools. It was later extended into FindItAgain~\cite{finditagain}. More recently, RTM~\cite{RTM}, a diverse dataset manually crafted by experts, was proposed. Although FindItAgain and RTM both provide human-made, high-quality tampered documents, they remain limited in scale, comprising fewer than 4,000 tampered documents in total. As a result, they are better suited for evaluations or fine-tuning rather than pretraining. This motivates our work on automatic, high-quality tampered document generation.

\section{Method}
\label{main:contrastive}
To generate high-quality tampered documents that better reflect real-world scenarios, we consider the five tampering types explained above: copy-move, splicing, insertion, inpainting, and coverage. Inpainting is relatively straightforward and can be performed using background-aware filling methods such as those provided by OpenCV. For copy-move, splicing, and coverage, a crop is copied from a source image and placed into a target image. To ensure the tampering is convincing, the visual characteristics of the source and target regions must closely match. For insertion, new content such as text is added and must match the surrounding content in color, font, size, and alignment to appear realistic. To ensure visual consistency, whether between source and target regions or between rendered and surrounding text, we introduce a similarity function defined as \begin{equation}
  S(x, u) =
  \frac{\mathcal{F}_\theta(x) \cdot \mathcal{F}_\theta(u)}
       {\|\mathcal{F}_\theta(x)\| \,\|\mathcal{F}_\theta(u)\|},
\end{equation}
 where $\mathcal{F}_\theta$ is a neural network trained to extract visual features from image regions and accepts inputs of arbitrary spatial dimensions. Candidate crops can then be selected based on these similarity scores. Moreover, when the source or target crop contains text, one common source of poor tampering quality is an ill-defined bounding box that cuts through characters, leaving detectable artifacts. To avoid this, we introduce a bounding box quality function \(\mathcal{G}_\theta(\cdot)\in[0,1]\), where lower output values indicate poorly defined crops. Generating high-quality tampering requires both a high similarity score and a high bounding box quality score. Correctly training both \(\mathcal{F}_\theta\) and \(\mathcal{G}_\theta\) is therefore key to generating high-quality tampered documents. We use a novel training framework for \(\mathcal{F}_\theta \), based on contrastive learning, to capture visual similarity between two text crops or between a text crop and a blank crop. For \( \mathcal{G}_\theta \), we adopt a supervised learning approach that considers both the crop and its surrounding context to assess bounding box quality. The training procedures for both networks are detailed in the following two subsections.

\subsection{Assessing crop similarity}
\begin{figure}[t]
  \centering
  \small

  \setlength{\tabcolsep}{2pt} 
  \setlength{\fboxsep}{2pt}   

  \begin{tabular}{@{}cc@{}}
    \colorbox{gray!25}{%
      \begin{minipage}{0.44\columnwidth}
        \centering
        \parbox[c][2\baselineskip][c]{\linewidth}{%
          \centering\bfseries (A) Font Mismatch%
        }\par\vspace{1pt}
        \includegraphics[width=0.44\linewidth]{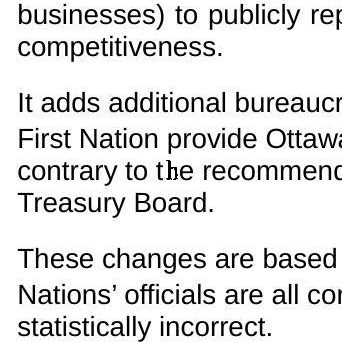}%
        \hspace{0.02\linewidth}%
        \includegraphics[width=0.44\linewidth]{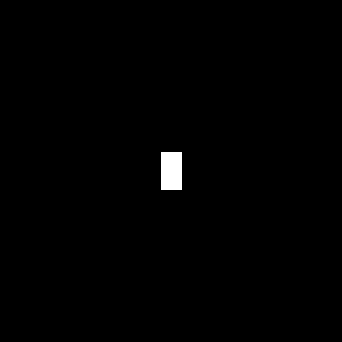}
      \end{minipage}%
    }
    &
    \colorbox{gray!25}{%
      \begin{minipage}{0.44\columnwidth}
        \centering
        \parbox[c][2\baselineskip][c]{\linewidth}{%
          \centering\bfseries (B) Blur Inconsistency%
        }\par\vspace{1pt}
        \includegraphics[width=0.44\linewidth]{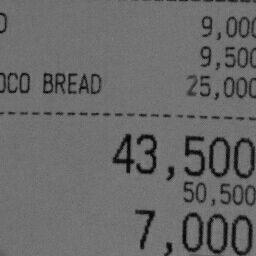}%
        \hspace{0.02\linewidth}%
        \includegraphics[width=0.44\linewidth]{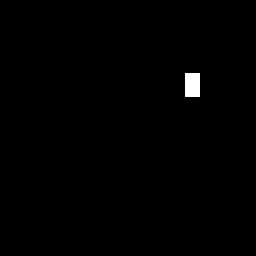}
      \end{minipage}%
    }
    \\[-1pt]
    \noalign{\vskip3pt}
    \colorbox{gray!25}{%
      \begin{minipage}{0.44\columnwidth}
        \centering
        \parbox[c][2\baselineskip][c]{\linewidth}{%
          \centering\bfseries
          (C) Background Color\\[-2pt]Mismatch%
        }\par\vspace{1pt}
        \includegraphics[width=0.44\linewidth]{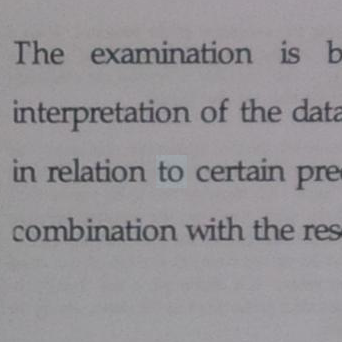}%
        \hspace{0.02\linewidth}%
        \includegraphics[width=0.44\linewidth]{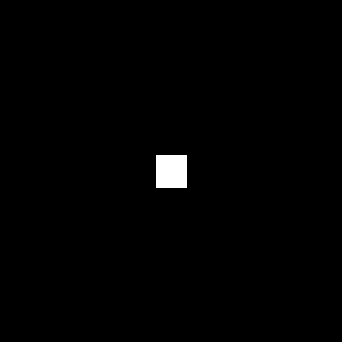}
      \end{minipage}%
    }
    &
    \colorbox{gray!25}{%
      \begin{minipage}{0.44\columnwidth}
        \centering
        \parbox[c][2\baselineskip][c]{\linewidth}{%
          \centering\bfseries
          (D) Text Horizontal\\[-2pt]Misalignment%
        }\par\vspace{1pt}
        \includegraphics[width=0.44\linewidth]{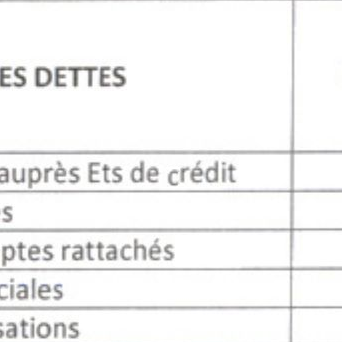}%
        \hspace{0.02\linewidth}%
        \includegraphics[width=0.44\linewidth]{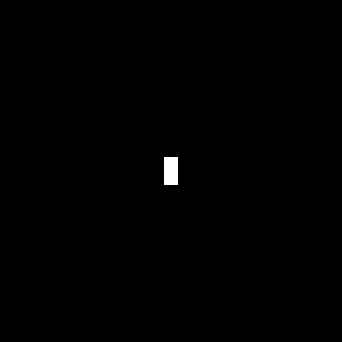}
      \end{minipage}%
    }
  \end{tabular}

  \caption{Illustration of failure cases in tampered document generation
  caused by a visual mismatch between the source and target crops,
  resulting in obvious visual artifacts.}
  \label{fig:other_failures}
\end{figure}

\vspace{0pt}
\label{f}
For copy-move and splicing, high-quality tampering requires that source and target patches share visual characteristics such as font style and size, foreground and background color, texture, and both vertical and horizontal text alignment. Additional factors, including brightness, contrast, saturation, sharpness, blur, and noise patterns, must also be consistent. Significant mismatches in any of these properties can create perceptible artifacts that reveal the tampering. As shown in \cref{fig:other_failures}, even when most features align, a single discrepancy can result in clearly visible manipulation traces. The persistent presence of these cases in the training data can compromise the training process by constituting shortcuts for the model, undermining its ability to generalize to genuine real-world manipulations. For coverage, most of these characteristics still apply, except for those related to text. The height and width of the bounding box, along with the number of characters, can serve as a good proxy for estimating the actual font size. It is, however, challenging to accurately compare the other visual characteristics, which is why we propose to train a neural network \( \mathcal{F}_\theta \) to perform this task. While a convolutional neural network is well-suited to this task, the primary challenge is obtaining appropriate training data, which motivates the use of contrastive learning.

\begin{figure}[b]
\vspace{0pt}
\small
  \centering
  \setlength\tabcolsep{4pt}
  \begin{tabular}{@{}cc@{}}
    \colorbox{gray!25}{%
      \begin{minipage}{0.44\columnwidth}\centering
        {\textbf{(A) Bad Source Crop}}\\[3pt]
        \includegraphics[width=0.44\linewidth]{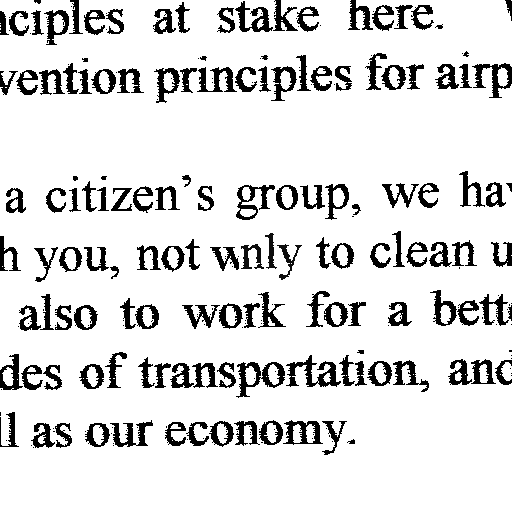}
        \quad
        \includegraphics[width=0.44\linewidth]{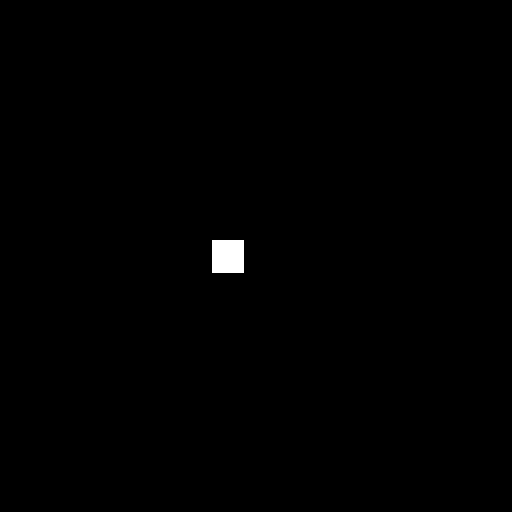}
      \end{minipage}
    }
    &
    \colorbox{gray!25}{%
      \begin{minipage}{0.44\columnwidth}\centering
        {\textbf{(B) Bad Target Crop}}\\[3pt]
        \includegraphics[width=0.44\linewidth]{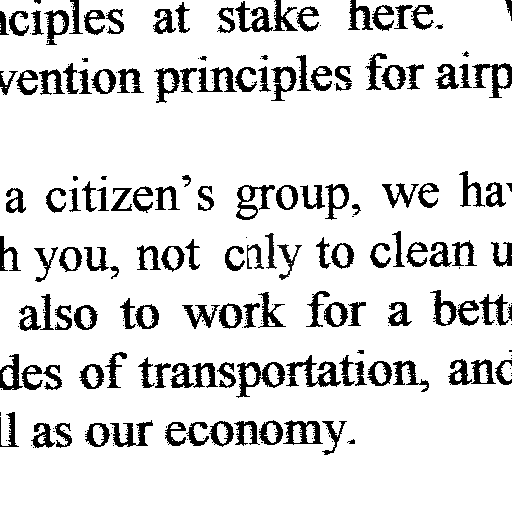}
        \quad
        \includegraphics[width=0.44\linewidth]{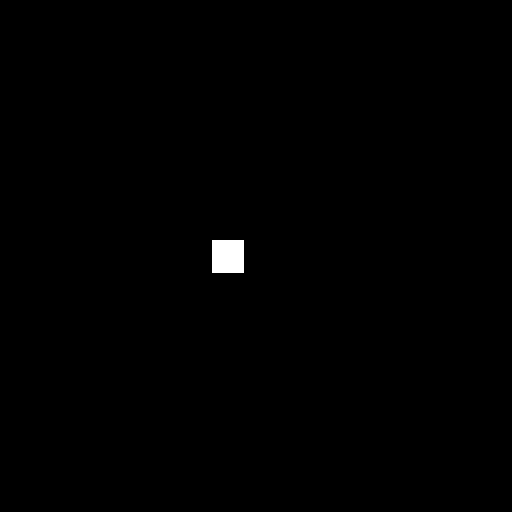}
      \end{minipage}
    }
  \end{tabular}
\caption{Illustration of failure cases in document tampering due to poor bounding boxes: (A) an ill‑defined source box cutting through characters; (B) an ill‑defined target box overlapping adjacent text.}

  \label{fig:crop_quality_failures}
  \vspace{0pt}
\end{figure}
\vspace{0pt}
\par\noindent \textbf{Defining positive pairs.} We rely on a natural property of document images that motivates a simple yet effective contrastive learning strategy: nearby same-line regions with matching width, height, and, when both are text regions, character count are generally more visually similar to one another than to regions from a distant line or another document. This provides a natural source of weak supervision for learning visual compatibility without manually annotated similarity pairs. We define a function \(\mathrm{ExtractLineSegments}(\mathrm{OCR}(I))\) that takes the per-character OCR bounding boxes \(\mathrm{OCR}(I)\) and returns pairs \(\{(b_i,\,l_i)\}_{i=1}^n\), where each \(b_i\) corresponds to either a text segment or a blank region and is assigned a line index \(l_i\). For a target crop \(b_i\), we define positives as crops from the same line whose bounding-box centers satisfy \(\|c_i-c_j\|_2<\tau_0\overline{w}\) and that have the same width and height and, when both contain text, the same number of characters. Here \(c_i\) and \(c_j\) are the centers of \(b_i\) and \(b_j\), \(\tau_0\) is a threshold, and \(\overline{w}\) is the average character width in the image. When more than one positive candidate is identified, we uniformly sample a single positive and discard the rest. Note that, at equal character count, headings and body text typically differ in width or height and are therefore not paired. For more details on \( \mathrm{ExtractLineSegments}\), please refer to \cref{appendix:getlines}.
\begin{figure}[t!]
\vspace{0pt}
\small
  \centering
  \colorbox{gray!15}{%
    \begin{minipage}{1.0\linewidth}
      \centering
      \includegraphics[width=0.45\linewidth]{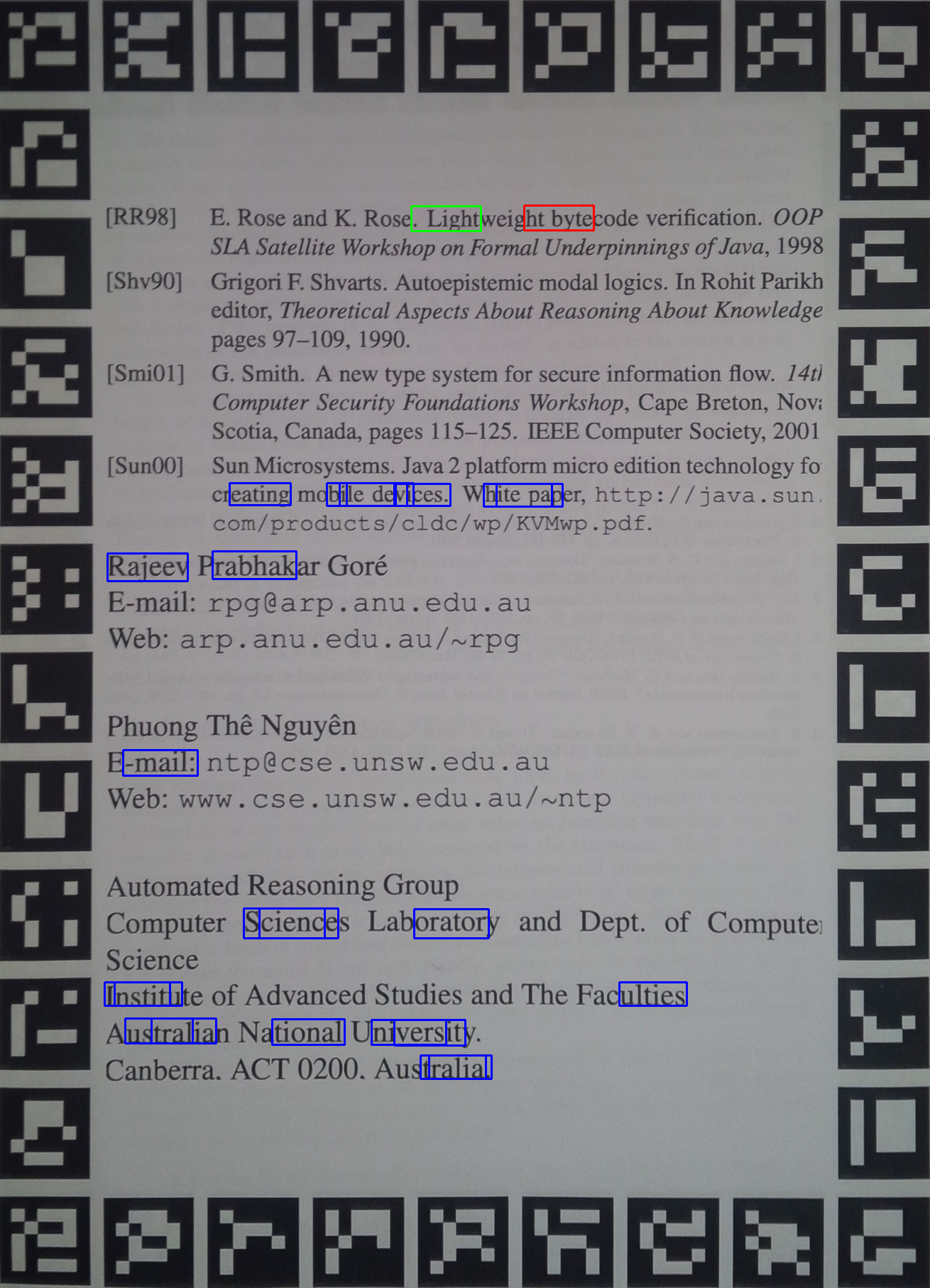}\quad
      \includegraphics[width=0.45\linewidth]{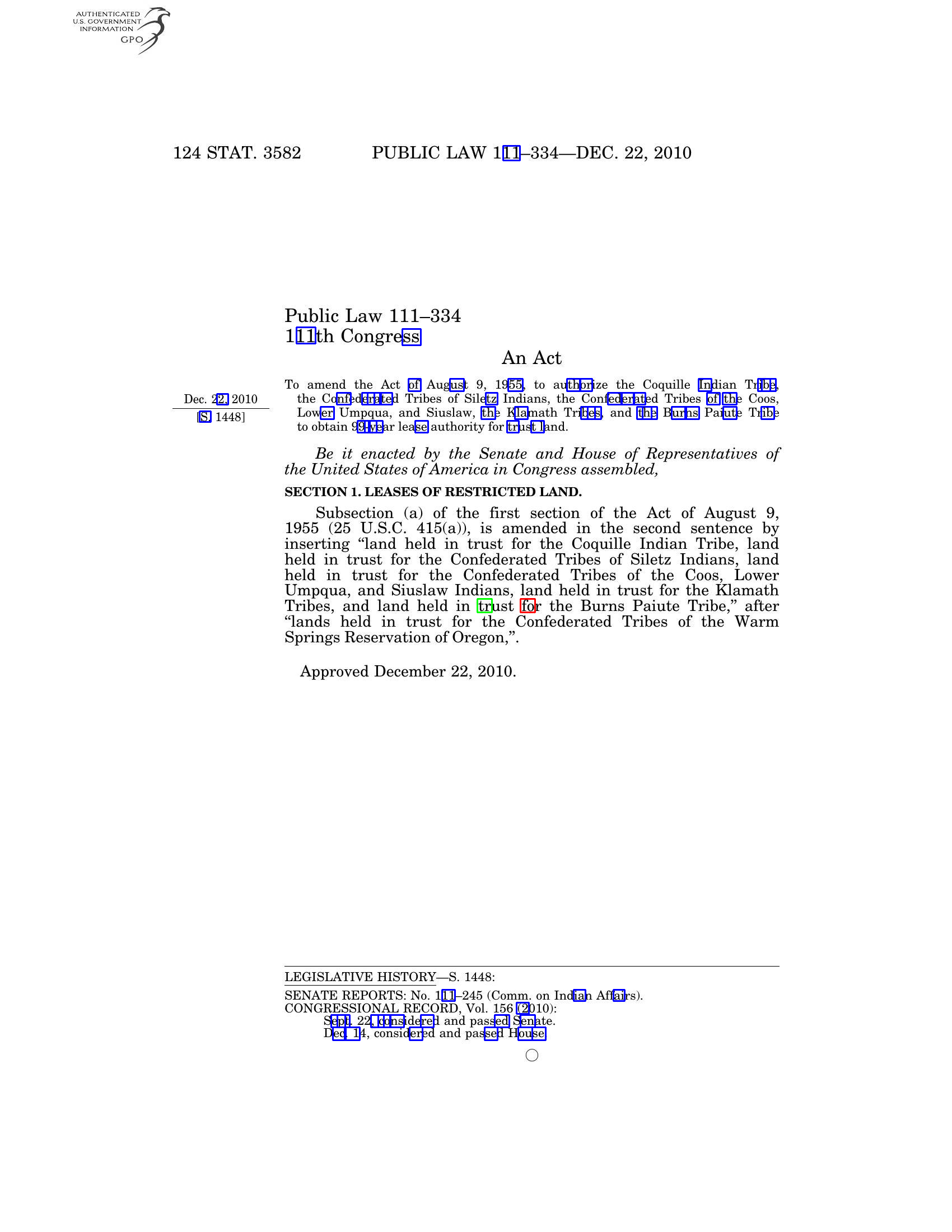}
    \end{minipage}
  }
\caption{Illustration of positive and negative pairs in our contrastive setting. The red box denotes the anchor, the green its positive, and the blue boxes the negatives.}
  \label{fig:illustration_contrastive}
  \vspace{0pt}
\end{figure}

\par\noindent \textbf{Defining negatives.} We define negatives as crops \(b_j\) whose aspect ratio satisfies \(\frac{w_j/h_j}{w_i/h_i}\in[1-\varepsilon,\,1+\varepsilon]\), with \(w\) and \(h\) denoting width and height and \(\varepsilon\) a tolerance parameter. When both crops contain text, we further require them to contain the same number of characters. For crops on the same page or on pages of the same dimensions, we compare the top and bottom edges of one crop with those of the other. The smallest of the four vertical distances must be at least $g_I$ pixels for anchor page $I$.
 A large minimum gap helps select negatives that differ visually from the anchor. \Cref{fig:illustration_contrastive} illustrates the result of this process, showing that anchors and their positive pairs share nearly identical visual attributes, while most negatives exhibit clear differences in one or more of these attributes. We note that negative examples are first collected from the same image as the source crop, and then from other images if necessary, until the desired number of negative samples is reached. In addition, we generate \( M_{\text{alt}} \) altered versions of each anchor \( b_i \) by applying random local shifts and visual transformations. These altered crops simulate visually inconsistent but structurally similar regions, and are treated as hard negatives. For more details on the applied data augmentation, please refer to \cref{appendix:metadataaug}. \\\textbf{Loss calculation.} The contrastive loss is defined as:
\begin{equation}
\resizebox{0.9\columnwidth}{!}{$\displaystyle
  \mathcal{L}_i
  = -\log
  \frac{
    \exp\bigl(\mathrm{S}(b_i, b_i^+)/\tau\bigr)
  }{
    \exp\bigl(\mathrm{S}(b_i, b_i^+)/\tau\bigr)
    + \sum_{b_i^- \in \mathcal{N}_i}
    \exp\bigl(\mathrm{S}(b_i, b_i^-)/\tau\bigr)
  },
$}
\end{equation}

\noindent where \( \tau \) is a temperature parameter, \( b_i^+ \) is a positive crop, and \( \mathcal{N}_i \) is the set of negative crops.
\\

 \noindent \textbf{Model architecture and input size.}
We implement $\mathcal{F}_\theta$ as a lightweight convolutional network with approximately 10.2 million parameters. All crops fed into $\mathcal{F}_\theta$ are resized to a height of 64 pixels. We prioritize aspect-ratio matches in the mining process, and otherwise additionally resize the width to match the corresponding anchor’s width. To robustly compare (i) two text crops and (ii) a text crop with a blank crop, we use two decoupled embedding heads: a foreground head that captures text-centric cues and a background head that models non-text regions. The similarity between two text crops is computed using both heads, whereas comparisons involving a blank crop use only the background head. This allows the model to remain sensitive to informative text cues, while maintaining a robust representation for blank regions. Full architectural details are provided in \cref{appendix:metaarch}.
\\
\textbf{Human analysis.} To validate the proposed mining strategy, we conduct a blinded study in which five annotators compare an anchor with a mined positive and a mined negative. Each annotator judges 700 cases, 100 of which are shared across all five. Across 3,500 judgments, annotators choose the mined positive as the more similar in 97.1\% of cases and agree with each other on 95\% of the shared cases. A majority vote on the shared cases flips 59\% of the negative judgments to a positive judgment and only 0.2\% the other way, suggesting that, in fewer than 1.5\% of the mined pairs, the negative is genuinely more similar to the anchor than the positive. The annotation protocol and further analysis are provided in \cref{sec:human-review}.

\begin{figure}[!t]
 \small
  \centering
  \includegraphics[width=1.0\linewidth]{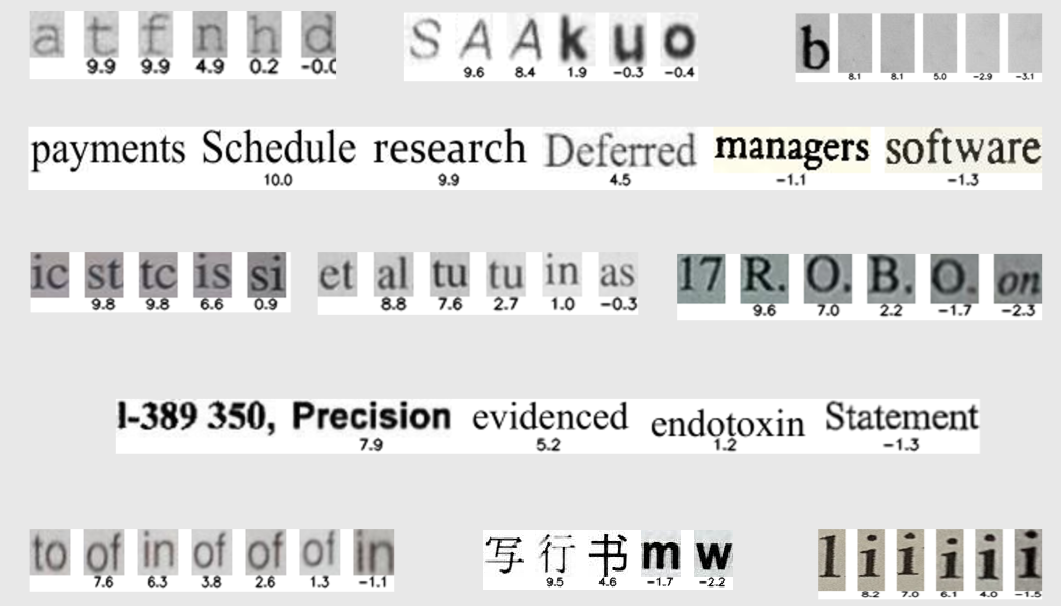}
 \caption{Similarity scores from $\mathcal{F}_\theta$ between a target crop (left) and candidate crops. The score below each crop indicates its similarity to the target crop and is scaled by ten.}

  \label{fig:example_of_output_meta}
\end{figure}

 \noindent A simplified overview of our proposed pipeline is provided in \cref{sec:contrastive-training}. Examples of similarity scores between a target crop and its candidates under \(\mathcal{F}_\theta\) are shown in \cref{fig:example_of_output_meta}. Notice that the resulting similarity scores account for all visual cues discussed earlier, including but not limited to font, brightness, saturation, and the spatial position of the text.

\vspace{0pt}
\subsection{Assessing bounding box quality}
\label{cropqualityassesmentmain}

\vspace{0pt}
Given an image \(I\) and two crops, the first from \(I\) defined by a bounding box \(b_i\) and the second from a possibly different image \(I'\) defined by \(b_j\), as shown in \cref{fig:crop_quality_failures}, even if they score highly under our visual similarity network $\mathcal{F}_\theta$, a poorly defined box, one that cuts through characters or includes parts of neighbors, will always introduce highly visible artifacts. To address these cases, one could rely on algorithms based on foreground estimation and connected-component analysis. However, these methods are slow and would introduce computational overhead to our pipeline. We therefore train a lightweight convolutional network \(\mathcal{G}_\theta\) with approximately 13.3 million parameters to predict bounding box quality, yielding up to a \(10\times\) speedup over such methods.
\\
\textbf{Training data.} For each image $I$, we extract the set of line segments, defined by their bounding boxes \((b_0,\dots,b_n)
= \mathrm{ExtractLineSegments}\bigl(\mathrm{OCR}(I)\bigr)\) and assess the quality of each bounding box following the procedure described in \cref{appendix:galgo}, then assign \(y=1\) to well-defined cases and \(y=0\) to ill-defined ones. We have observed that certain document domains yield near-perfect OCR, which results in fewer ill-defined crops, risking domain-specific overfitting. To mitigate this, for each box labeled \(y=1\) we randomly expand or contract its boundaries so that it cuts a character or includes part of a neighbor, and assign the label \(y=0\) to the resulting crop. Further details are in \cref{appendix:gprocess}.
\\
\textbf{Model input and architecture.}
In some cases, looking only at the crop does not suffice to decide if the box is well defined, so we also provide to the model its immediate neighborhood. Along with each crop
\(
c_k = I[y_0^k:\,y_1^k,\,x_0^k:\,x_1^k],
\) we extract four edge strips of thickness \(t\) pixels:
\begin{equation}
\small
\begin{array}{@{}l@{\;}l@{}}
t_k = I[y_0^k-t:y_0^k,\,x_0^k:x_1^k], &
b_k = I[y_1^k:y_1^k+t,\,x_0^k:x_1^k], \\[1pt]
\ell_k = I[y_0^k:y_1^k,\,x_0^k-t:x_0^k], &
r_k = I[y_0^k:y_1^k,\,x_1^k:x_1^k+t].
\end{array}
\end{equation}

\noindent To ensure that index 0 along the thickness axis always lies closest to the crop border, we vertically flip the top strip and horizontally flip the left strip:
\begin{equation}
\resizebox{0.9\columnwidth}{!}{$\displaystyle
  \tilde t_k = \mathrm{flipud}(t_k), \quad
  \tilde b_k = b_k, \quad
  \tilde \ell_k = \mathrm{fliplr}(\ell_k), \quad
  \tilde r_k = r_k
$}
\end{equation}
where \(\mathrm{flipud}\) reverses rows and \(\mathrm{fliplr}\) reverses columns. We then form the two stripe inputs by simple concatenation:
\begin{equation}
  \mathrm{tb}_k = \mathrm{concat}_h\bigl(\tilde t_k,\;\tilde b_k\bigr), 
  \quad
  \mathrm{lr}_k = \mathrm{concat}_v\bigl(\tilde \ell_k,\;\tilde r_k\bigr).\end{equation}
Each crop is fed to a small 2D-CNN. Each stripe is first processed by a 2D-CNN to aggregate edge context and then processed by a 1D-CNN. The obtained embeddings are concatenated and sent to a sigmoid classifier. For more details on $\mathcal{G}_\theta$'s architecture, please refer to \cref{appendix:garch}.
\\
We train $\mathcal{G}_\theta$ using cross-entropy. In \cref{appendix:evaluationcropqualityassesment}, we evaluate its ability to assess crop quality on real data and compare it with algorithm-based alternatives.
\vspace{0pt}
\subsection{Data generation}
\label{main:generation}
\vspace{0pt}
\begin{figure}[t]
\small
\vspace{0pt}
  \centering
  \includegraphics[width=1.0\linewidth]{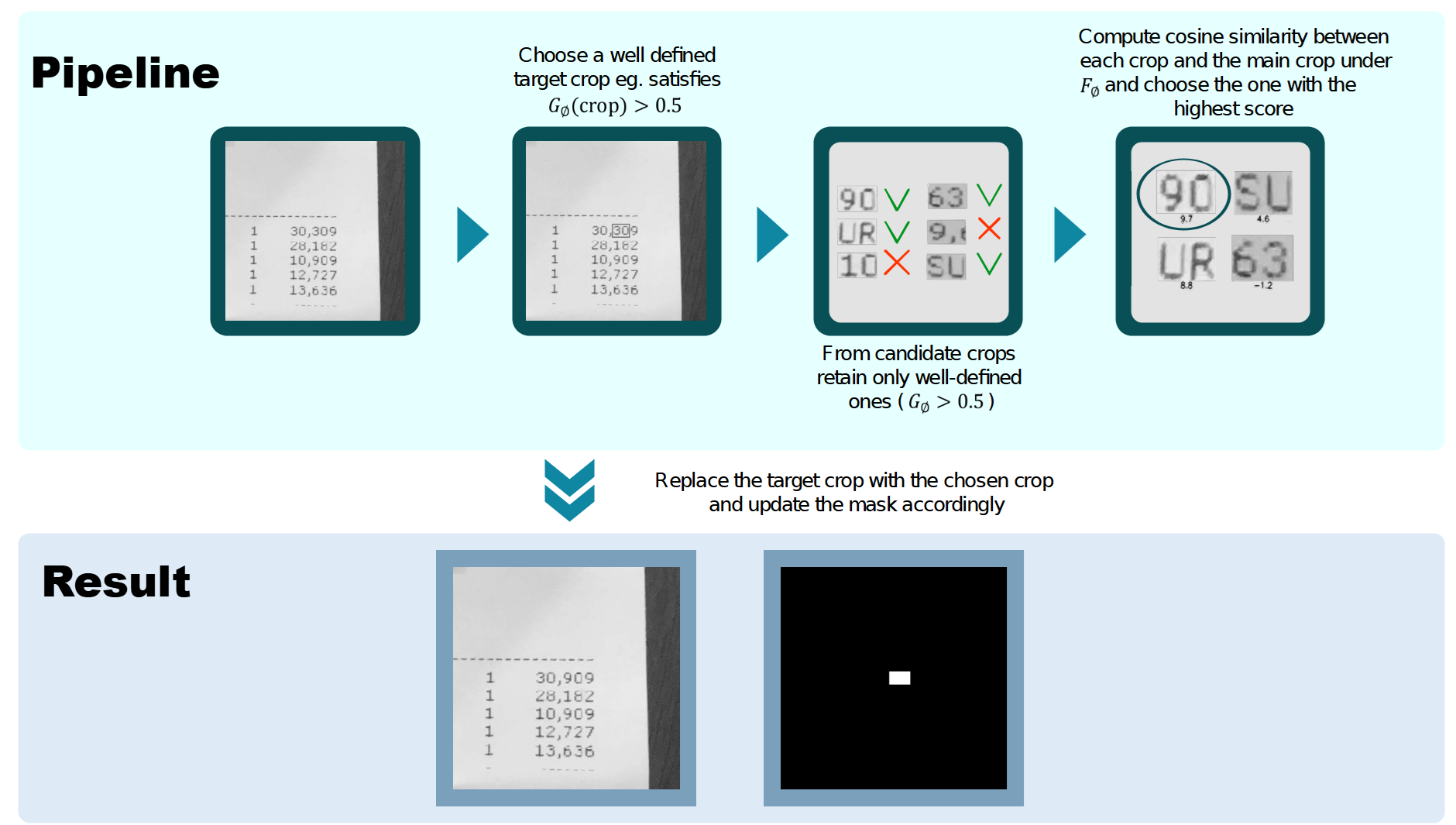}
  \caption{A simplified illustration of our tampered document generation process.}
  \label{fig:generation_process}
  \vspace{0pt}
\end{figure}

To generate realistic tampered document images, we follow a multi-stage process guided by both the crop similarity network \(\mathcal{F}_\theta\) and the crop quality network \(\mathcal{G}_\theta\). We begin by constructing a database of source crops. For each image, we apply \texttt{ExtractLineSegments} to the character-level OCR output to extract both text segments and blank regions, where each blank region is assigned the same number of characters as its nearest text neighbor of similar size. Text crops are then passed through \(\mathcal{G}_\theta\), and only those with a predicted quality score above \(\tau_2\) are retained. The selected crops are then grouped in the database by width, height, and number of characters. For any target image \(I\), we retrieve its processed crops from the database and sample up to \(n_{max}\) regions from the image to manipulate. For each selected region \(r = (x, y, w_i, h_i)\), we apply the following procedure:

\begin{itemize}
\item If \(r\) is blank and a probability \(p_{ins}\) is met, we look for the nearest text crop $c$ of the same size \((w_i,h_i)\), retrieve its character string \(t_c\), and calculate its embedding \(v_c = \mathcal{F}_\theta(c)\). For each font $f_j$ and color $col_k$ sampled around the Sauvola-estimated foreground color, we render the text into the target region and compute:
\begingroup
\fontsize{7.8pt}{8pt}\selectfont
\begin{equation}
P_{j,k}
= \mathrm{Render}(t_c,f_j,\mathrm{col}_k,\mathrm{crop}(I,r)),
\quad
v_{P_{j,k}}
= \mathcal{F}_\theta(P_{j,k})
\end{equation}
\endgroup

We score each rendering by its similarity to the reference embedding $v_c$.
\begin{equation}
  s_{j,k} =
  \frac{v_c \cdot v_{P_{j,k}}}{\|v_c\| \,\|v_{P_{j,k}}\|}.
\end{equation}
We sample a qualifying rendering with higher probability for higher scores, paste it into \(r\), and update the mask.

\item Alternatively, with probability $p_{inp}$, we apply inpainting to region $r$ and update the tampering mask accordingly.
\item Otherwise, we search for candidate crops \(c_j\) either:
\begin{itemize}
\item With probability $p_{spl}$, select from the precomputed database the crops of size $(w_i, h_i)$ whose character count matches that of $r$.

\item With probability $1 - p_{spl}$, select from the same image the crops whose character count matches that of $r$ and whose aspect ratio satisfies
\(
\frac{w_j/h_j}{w_i/h_i} \in [1/(1+\varepsilon'),\,1+\varepsilon'],
\)
then resize each to $(w_i, h_i)$.
\end{itemize}

We score each candidate by its cosine similarity to the target crop.
\begin{equation}
  s_j =
  \frac{
    \mathcal{F}_\theta(\mathrm{crop}(I, r)) \cdot \mathcal{F}_\theta(c_j)
  }{
    \bigl\|\mathcal{F}_\theta(\mathrm{crop}(I, r))\bigr\|
    \,\bigl\|\mathcal{F}_\theta(c_j)\bigr\|
  }.
\end{equation}
We sample a qualifying donor by the same rule, paste it into \(r\), and update the mask.
\end{itemize}

\noindent Because \texttt{ExtractLineSegments} provides both text and blank segments, this unified procedure enables tampering via all five categories. The generation process is illustrated in \cref{fig:generation_process}, and the full procedure is summarized in \cref{sec:data-generation}. Examples of the resulting tampered documents are shown in \cref{appendix:exemplesgeenration}, while further details on the rendering and inpainting functions are provided in \cref{appendix:render,appendix:inpaint}. 

\begin{table*}[!t]
\centering
\small
\setlength{\tabcolsep}{2.5pt}
\renewcommand{\arraystretch}{1.0}
\setlength{\aboverulesep}{0.3ex}
\setlength{\belowrulesep}{0.4ex}
\caption{Out-of-domain document tampering detection results on RTM, FindItAgain, and FindIt for several models trained on data generated using different data generation approaches. For each model and evaluation dataset, \emph{Best prior} is selected independently for Image and Pixel based on the highest F1 among the data generation methods of TextTamper~\cite{newdataandopenopen}, DocTamper~\cite{Doctamper}, and FSTS~\cite{dataset2}. P and R are taken from the same method as the selected F1. P, R, and F1 correspond to precision, recall, and F1 score, while Image and Pixel denote image-level classification and pixel-level localization performance. Note that the source images used to generate the training data are the same across models and data generation methods as described in \cref{sec:data}, and each detector uses the same training protocol across generators as described in \cref{sec:training}. Complete results for all generation methods are provided in \cref{appendix:full-controlled-results}.}
\label{tab:out-of-domain-results}
\resizebox{\textwidth}{!}{%
\begin{tabular}{ll|rrr|rrr|rrr|rrr|rrr|rrr}
\toprule
\multirow{3}{*}{\textbf{Model}} & \multirow{3}{*}{\textbf{Data Gen.}} &
\multicolumn{6}{c|}{\textbf{RTM}} &
\multicolumn{6}{c|}{\textbf{FindItAgain}} &
\multicolumn{6}{c}{\textbf{FindIt}} \\
\cmidrule(lr){3-8}\cmidrule(lr){9-14}\cmidrule(lr){15-20}
& & \multicolumn{3}{c|}{Image} & \multicolumn{3}{c|}{Pixel} &
\multicolumn{3}{c|}{Image} & \multicolumn{3}{c|}{Pixel} &
\multicolumn{3}{c|}{Image} & \multicolumn{3}{c}{Pixel} \\
& & P & R & F1 & P & R & F1 & P & R & F1 & P & R & F1 & P & R & F1 & P & R & F1 \\
\midrule
\multirow{2}{*}{PSCC-Net} & Best prior & 33.8 & 34.1 & 33.9 & 7.5 & 10.0 & 8.6 & 52.0 & 37.1 & 43.3 & 8.7 & 9.2 & 8.9 & 61.4 & 53.8 & 57.3 & 11.2 & 11.9 & 11.5 \\
 & Ours & \textbf{39.3} & \textbf{40.2} & \textbf{39.7} & \textbf{10.5} & \textbf{10.8} & \textbf{10.6} & \textbf{56.8} & \textbf{60.0} & \textbf{58.3} & \textbf{10.6} & \textbf{12.8} & \textbf{11.6} & \textbf{62.9} & \textbf{55.0} & \textbf{58.7} & \textbf{12.0} & \textbf{12.1} & \textbf{12.0} \\
\midrule
\multirow{2}{*}{CAT-Net} & Best prior & 36.5 & 32.7 & 34.5 & 8.7 & 10.1 & 9.3 & 34.4 & 31.4 & 32.8 & 10.7 & 5.2 & 7.0 & 47.8 & 41.2 & 44.3 & 10.3 & 10.1 & 10.2 \\
 & Ours & \textbf{40.0} & \textbf{35.0} & \textbf{37.3} & \textbf{9.2} & \textbf{11.0} & \textbf{10.0} & \textbf{48.0} & \textbf{34.3} & \textbf{40.0} & \textbf{11.0} & \textbf{5.6} & \textbf{7.4} & \textbf{51.5} & \textbf{42.5} & \textbf{46.6} & \textbf{13.0} & \textbf{10.5} & \textbf{11.7} \\
\midrule
\multirow{2}{*}{DTD} & Best prior & 36.5 & 52.1 & 42.9 & 10.8 & 10.0 & 10.4 & 48.1 & 37.1 & 41.9 & 9.8 & 9.9 & 9.9 & 55.6 & 56.2 & 55.9 & 13.6 & 13.9 & 13.7 \\
 & Ours & \textbf{36.6} & \textbf{73.3} & \textbf{48.9} & \textbf{14.9} & \textbf{11.4} & \textbf{12.9} & \textbf{50.0} & \textbf{40.0} & \textbf{44.4} & \textbf{12.5} & \textbf{12.4} & \textbf{12.5} & \textbf{62.3} & \textbf{66.2} & \textbf{64.2} & \textbf{14.0} & \textbf{14.4} & \textbf{14.2} \\
\midrule
\multirow{2}{*}{ADCD-Net} & Best prior & 41.8 & 48.6 & 45.0 & 13.0 & 14.1 & 13.5 & 43.9 & 51.4 & 47.4 & 11.0 & 11.5 & 11.2 & 77.6 & 65.0 & 70.8 & 17.8 & 17.3 & 17.6 \\
 & Ours & \textbf{47.0} & \textbf{51.7} & \textbf{49.2} & \textbf{24.9} & \textbf{15.7} & \textbf{19.3} & \textbf{52.3} & \textbf{65.7} & \textbf{58.2} & \textbf{13.6} & \textbf{14.9} & \textbf{14.2} & \textbf{82.8} & \textbf{66.2} & \textbf{73.6} & \textbf{19.1} & \textbf{17.9} & \textbf{18.5} \\
\midrule
\multirow{2}{*}{ASC-Former} & Best prior & 40.7 & 30.0 & 34.5 & 15.4 & 13.6 & 14.4 & 36.8 & 60.0 & 45.7 & 10.8 & 9.8 & 10.2 & 54.7 & 65.0 & 59.4 & 24.9 & 12.2 & 16.4 \\
 & Ours & \textbf{44.3} & \textbf{32.3} & \textbf{37.4} & \textbf{23.4} & \textbf{16.1} & \textbf{19.1} & \textbf{38.6} & \textbf{62.9} & \textbf{47.8} & \textbf{15.3} & \textbf{12.3} & \textbf{13.6} & \textbf{69.6} & \textbf{68.8} & \textbf{69.2} & \textbf{30.1} & \textbf{13.9} & \textbf{19.0} \\
\midrule
\multirow{2}{*}{FFDN} & Best prior & 46.1 & 46.9 & 46.5 & 19.8 & 17.1 & 18.4 & 51.4 & 51.4 & 51.4 & 13.6 & 10.6 & 11.9 & 77.8 & 61.3 & 68.5 & 28.6 & 25.5 & 27.0 \\
 & Ours & \textbf{46.4} & \textbf{51.0} & \textbf{48.6} & \textbf{28.2} & \textbf{20.6} & \textbf{23.8} & \textbf{69.0} & \textbf{57.1} & \textbf{62.5} & \textbf{31.6} & \textbf{21.4} & \textbf{25.5} & \textbf{82.3} & \textbf{70.0} & \textbf{75.7} & \textbf{36.3} & \textbf{28.3} & \textbf{31.8} \\
\midrule

\multirow{2}{*}{DITL$^2$}
& Best prior
& 57.4 & 15.9 & 24.9 & 13.9 & 12.1 & 12.9
& 40.0 & 40.0 & 40.0 & 12.1 & 9.7 & 10.8
& 65.9 & 72.5 & 69.0 & 19.3 & 14.4 & 16.5 \\
& Ours
& \textbf{61.0} & \textbf{16.5} & \textbf{26.0} & \textbf{18.3} & \textbf{12.6} & \textbf{14.9}
& \textbf{44.4} & \textbf{45.7} & \textbf{45.1} & \textbf{13.2} & \textbf{10.4} & \textbf{11.6}
& \textbf{67.0} & \textbf{73.8} & \textbf{70.2} & \textbf{20.7} & \textbf{14.7} & \textbf{17.1} \\

\bottomrule

\end{tabular}}
\end{table*}

\section{Experiments}
\subsection{Training data}
\vspace{0pt}
\label{sec:data}
To generate tampered documents and prepare training data for $F_{\theta}$ and $G_{\theta}$, we use 1.2 million documents from the CC-MAIN-2021-31-PDF-UNTRUNCATED corpus. We filter this corpus to retain only documents containing more than 512 characters, and limit the proportion of native PDFs to 5\%. Additionally, we use around 300k documents from the IIT-CDIP~\cite{IIT} dataset, and 500k documents from the DocMatix \cite{docmatrix} dataset. We further incorporate several smaller datasets, including CORD \cite{cord}, B-MOD~\cite{bmod}, DUDE~\cite{dude}, TNCR~\cite{TNCR}, M6Doc~\cite{m6doc}, SmartDoc~\cite{smartdoc}, XFUND~\cite{xfund}, and InfoVQA~\cite{infovqa}. To increase diversity, we scrape around 175k publicly available government documents using the process explained in \cref{appendix:scraping}. We upsample B-MOD, CORD and XFUND by a factor of 20, InfoVQA by a factor of 10, and TNCR by a factor of 5. For all datasets, OCR is performed using the Google Cloud Vision API. To avoid data leakage, we remove any image that fully or partially matches the source images used to construct the different splits of the evaluation datasets using the procedure detailed in \cref{appendix:outofdomain}. After filtering and upsampling, the resulting dataset contains around 2.8\,million images and is used both to generate training data for the tampering models and to train the auxiliary networks. Most of these images exhibit little to no document rotation. We address this with training-time rotation augmentation.

\subsection{Experimental setting}

\textbf{Training setup for $F_{\theta}$ and $G_{\theta}$.}
We train the crop-similarity network $\mathcal{F}_{\theta}$ and the bounding box quality assessment network $\mathcal{G}_{\theta}$ for one epoch. For $\mathcal{F}_{\theta}$, we set the number of negative samples $N = 256$ and the number of augmented anchor samples $M_{\text{alt}} = 20$. The threshold $\tau_0$ is set to 10. We set $\epsilon = 0.1$ and the temperature parameter $\tau = 0.1$. $\mathcal{G}_{\theta}$'s stripe thickness $t$ is set to 9. 
\par\noindent \textbf{Data-generation setup.}
We set the crop quality filtering threshold $\tau_2 = 0.5$, the insertion probability $p_{ins} = 0.025$, the inpainting probability $p_{inp} = 0.025$, and the splicing probability $p_{spl} = 0.10$. The maximum number of tampered regions $n_{\text{max}}$ is set to 5, and the aspect ratio tolerance $\epsilon' = 0.10$.
\par\noindent \textbf{Training setup for document tampering models.}
\label{sec:training}
 We pretrain each detector for two epochs except PSCC-Net, which we pretrain for three. We use a global batch size of 64 on the corresponding synthetic dataset, using a resolution of up to $1024 \times 1024$, as detailed in \cref{appendix:reproducibility}. We use Focal Loss ($\gamma=2$) \cite{focalloss} for the segmentation heads and, where present, the classification heads. The maximum learning rate is set to $1 \times 10^{-4}$ and we adopt a Cosine Annealing scheduler \cite{cosine_annealing}. We apply data augmentation, with details provided in \cref{appendix:dataaug}. This training setup is kept unchanged across generators for each detector, and the same seed is used for all runs.

\par\noindent\textbf{Syn2Real Protocol.} We pretrain detectors on synthetic data and evaluate them on RTM, FindItAgain, and FindIt. We report zero-shot results for all seven detectors and target-dataset fine-tuning results for the six detectors included in that experiment. We refer to the setting in which TDoc-2.8M is used as the pretraining dataset for this protocol as \emph{Syn2Real-TDoc-2.8M}.

\subsection{Results}

\textbf{Main results.} To demonstrate the superiority of our data generation approach, we train seven models, DTD~\cite{Doctamper}, ASC-Former~\cite{RTM}, CAT-Net~\cite{CAT-Net}, PSCC-Net~\cite{PSCC-Net}, ADCD-Net~\cite{adcdnet}, FFDN~\cite{dtdv2freq}, and DITL$^2$~\cite{li2025ditl2}, on datasets generated from the same source documents using four approaches: ours and the generation methods proposed in~\cite{newdataandopenopen,Doctamper,dataset2}. These are, to our knowledge, the only methods in the literature that propose automatic data-generation pipelines that simulate conventional tampering operations for document tampering detection and localization, with DocTamper~\cite{Doctamper} being the de facto reference on which most subsequent works rely for training data. We give more details about these methods in \cref{appendix:generationpreviouswork}. To isolate the effect of data generation, tampered images are derived from the same source documents described in \cref{sec:data}, and each detector uses the same configuration across generators, as explained in \cref{sec:training}. The resulting models are evaluated on three human-made document manipulation datasets, RTM~\cite{RTM}, FindItAgain~\cite{finditagain} and FindIt~\cite{findit}. This evaluation setting tests two distinct domain shifts: a manipulation-domain shift from synthetic to human-made tampering, and an image-domain shift from the training document sources to unseen evaluation sources, as detailed in \cref{sec:data}. \Cref{tab:out-of-domain-results} shows that models trained on data generated by our method achieve higher performance than those trained on data generated by previous methods. 
\begin{table}[!t]
\centering
\small
\setlength{\tabcolsep}{6pt}
\caption{Impact of the data-generation strategy on document tampering detection and localization after fine-tuning. We report F1 scores averaged across the fine-tuned models. Detailed results are provided in \cref{appendix:finetuningresultsdetails}.}

\label{tab:finetuningresults}
\begin{tabular}{l|cc|cc|cc}
\toprule
\multirow{2}{*}{\textbf{Data Gen.}}
& \multicolumn{2}{c|}{\textbf{RTM}}
& \multicolumn{2}{c|}{\textbf{FindItAgain}}
& \multicolumn{2}{c}{\textbf{FindIt}} \\
& \textbf{Img} & \textbf{Pix}
& \textbf{Img} & \textbf{Pix}
& \textbf{Img} & \textbf{Pix} \\
\midrule
FSTS      & 64.7 & 24.8 & 49.4 & 16.2 & 79.2 & 41.5 \\
DocTamper & 65.2 & 24.5 & 48.9 & 15.1 & 78.5 & 41.2 \\
TextTamper & 60.4 & 23.2 & 49.1 & 15.4 & 76.1 & 40.2 \\
Ours      & \textbf{65.7} & \textbf{25.8} & \textbf{58.0} & \textbf{21.6} & \textbf{81.1} & \textbf{42.9} \\
\bottomrule
\end{tabular}
\end{table}
This improvement is consistent across datasets and architectures, with stronger models such as FFDN, ADCD-Net and ASC-Former benefiting even more from being trained on the higher-quality data generated by our approach. Notably, these improvements also hold for DITL$^2$~\cite{li2025ditl2}, a model designed to improve cross-domain generalization, with pixel-level F1 increasing from 12.9 to 14.9 on RTM, 10.8 to 11.6 on FindItAgain, and 16.5 to 17.1 on FindIt.
The results also indicate that performance gains are more pronounced on RTM and FindItAgain, which were designed to reflect realistic manipulation scenarios, in contrast with FindIt, which was created by non-expert volunteers with heterogeneous skill levels and without explicit quality control. FFDN, for example, benefits from a 114.3\% relative pixel-level F1 score improvement on FindItAgain. A per-tampering-type analysis further shows that the largest localization gains occur for copy-move, splicing, and coverage, where visual compatibility is particularly important, as detailed in \cref{appendix:per-type}. \Cref{tab:finetuningresults} reports the average performance across models, pretrained on each synthetic dataset and then fine-tuned on each evaluation dataset's training split. Models pretrained with data generated by our approach still achieve the best results on average, confirming that our data generation approach provides a stronger starting point even when real annotations are available. Beyond these downstream performance gains, our quantitative and qualitative analyses show that manipulations generated using both $F_{\theta}$ and $G_{\theta}$ are visually more convincing and substantially harder for detectors to localize, as detailed in \cref{appendix:generateddataquality}. Threshold-independent evaluation further confirms the advantage of our generated data, with models trained on our data achieving higher pixel AUROC and AUPRC across all evaluated architectures and benchmarks, as detailed in \cref{appendix:threshold-metrics}. 
Qualitative comparisons show that detectors trained on our data identify challenging tampered regions missed by baseline-trained detectors, while precision--recall curves demonstrate a more favorable trade-off across a range of operating points, as shown in \cref{appendix:qualitative,appendix:precrecallcurves}. \Cref{appendix:datagentimeanalysis} provides an analysis of the per-image generation time of our approach, showing that $F_{\theta}$ and $G_{\theta}$ together account for less than 20 percent of the total processing time. Our improvements also remain consistent across different dataset sizes and independent random seeds, demonstrating the robustness of our approach, as detailed in \cref{smaller_scale}. We provide reproducibility details in \cref{appendix:reproducibility}.
\begin{table}[!t]
\centering
\caption{Full localization performance of ASC-Former, FFDN, ADCD-Net, and DITL$^2$ on RTM, FindItAgain, and FindIt when pretrained on the original datasets released by DocTamper~\cite{Doctamper} and FSTS~\cite{dataset2}, generator-based datasets, or our generated dataset. The generator-based datasets combine DanceText~\cite{qu2026detect}, ScaleEdit-12M~\cite{chen2026scaleedit}, AIForge-Doc~\cite{wu2026aiforgedoc,aiforgedoc2v2026}, FantasyID~\cite{Korshunov_IJCB-2_2025}, and VDE-Bench~\cite{yi2026vdebench}. Details are provided in \cref{appendix:generator-based-datasets}.}
\label{tab:doctamper-full}
\resizebox{\columnwidth}{!}{%
\setlength{\tabcolsep}{4pt}
\small
\begin{tabular}{ll|ccc|ccc|ccc|ccc}
\toprule
\multirow{2}{*}{\textbf{Benchmark}} &
\multirow{2}{*}{\textbf{Training data}} &
\multicolumn{3}{c|}{\textbf{ASC-Former}} &
\multicolumn{3}{c|}{\textbf{FFDN}} &
\multicolumn{3}{c|}{\textbf{ADCD-Net}} &
\multicolumn{3}{c}{\textbf{DITL$^2$}} \\
& & \textbf{Prec} & \textbf{Rec} & \textbf{F1}
& \textbf{Prec} & \textbf{Rec} & \textbf{F1}
& \textbf{Prec} & \textbf{Rec} & \textbf{F1}
& \textbf{Prec} & \textbf{Rec} & \textbf{F1} \\
\midrule
\multirow{4}{*}{\textbf{RTM}}
& DocTamper (CVPR~'23)
& 5.5 & 4.5 & 4.9
& 8.1 & 6.8 & 7.4
& 11.7 & 13.0 & 12.3
& 7.6 & 6.6 & 7.1 \\
& FSTS (NeurIPS~'25)
& 7.5 & 6.4 & 6.9
& 18.6 & 19.1 & 18.9
& 20.6 & 15.0 & 17.4
& 16.9 & \textbf{12.8} & 14.6 \\
& Generator-based datasets
& 15.7 & 9.5 & 11.8
& 17.1 & 11.0 & 13.4
& 15.7 & 11.1 & 13.0
& 11.2 & 6.9 & 8.5 \\
& Ours
& \textbf{23.4} & \textbf{16.1} & \textbf{19.1}
& \textbf{28.2} & \textbf{20.6} & \textbf{23.8}
& \textbf{24.9} & \textbf{15.7} & \textbf{19.3}
& \textbf{18.3} & 12.6 & \textbf{14.9} \\
\midrule
\multirow{4}{*}{\textbf{FindItAgain}}
& DocTamper (CVPR~'23)
& 8.0 & 5.7 & 6.7
& 6.8 & 7.9 & 7.3
& 7.7 & 8.1 & 7.9
& 1.0 & 5.0 & 1.6 \\
& FSTS (NeurIPS~'25)
& 7.0 & 4.5 & 5.5
& 22.9 & 18.1 & 20.2
& 11.2 & 10.0 & 10.6
& 11.2 & 7.7 & 9.2 \\
& Generator-based datasets
& 3.4 & 2.3 & 2.7
& 10.3 & 6.0 & 7.6
& 6.5 & 7.6 & 7.0
& 5.7 & 5.0 & 5.3 \\
& Ours
& \textbf{15.3} & \textbf{12.3} & \textbf{13.6}
& \textbf{31.6} & \textbf{21.4} & \textbf{25.5}
& \textbf{13.6} & \textbf{14.9} & \textbf{14.2}
& \textbf{13.2} & \textbf{10.4} & \textbf{11.6} \\
\midrule
\multirow{4}{*}{\textbf{FindIt}}
& DocTamper (CVPR~'23)
& 7.4 & 4.9 & 5.9
& 10.9 & 7.5 & 8.9
& 12.3 & 14.9 & 13.5
& \textbf{21.7} & 4.7 & 7.7 \\
& FSTS (NeurIPS~'25)
& 19.1 & 8.0 & 11.3
& 21.8 & 13.5 & 16.7
& 17.1 & 17.4 & 17.3
& 20.8 & 12.7 & 15.8 \\
& Generator-based datasets
& \textbf{31.9} & 12.4 & 17.9
& 35.8 & \textbf{29.5} & \textbf{32.3}
& \textbf{20.1} & 17.1 & \textbf{18.5}
& 20.3 & \textbf{18.0} & \textbf{19.1} \\
& Ours
& 30.1 & \textbf{13.9} & \textbf{19.0}
& \textbf{36.3} & 28.3 & 31.8
& 19.1 & \textbf{17.9} & \textbf{18.5}
& 20.7 & 14.7 & 17.1 \\
\bottomrule
\end{tabular}%
}
\vspace{-6pt}
\end{table}
 \\ \noindent\textbf{Comparison with released datasets.}
 Beyond the controlled comparison in \cref{tab:out-of-domain-results}, we compare pretraining on our generated dataset against the official datasets released by DocTamper~\cite{Doctamper} and FSTS~\cite{dataset2}, as well as recent generator-based datasets, namely DanceText~\cite{qu2026detect}, ScaleEdit-12M~\cite{chen2026scaleedit}, AIForge-Doc~\cite{wu2026aiforgedoc,aiforgedoc2v2026}, FantasyID~\cite{Korshunov_IJCB-2_2025}, and VDE-Bench~\cite{yi2026vdebench}. We combine these generator-based datasets to assess whether modern generative text-editing approaches provide stronger training supervision than our method, with details in \cref{appendix:generator-based-datasets}. As shown in \cref{tab:doctamper-full}, our data achieves the highest pixel-level F1 across all four architectures on RTM and FindItAgain. Compared with the generator-based datasets, FFDN gains 10.4 and 17.9 F1 points on these two benchmarks, respectively. On FindIt, the results are more mixed: generator-based datasets slightly outperform ours for FFDN and DITL$^2$, match ours for ADCD-Net, and perform worse for ASC-Former. These findings suggest that generator-based training data can benefit localization on the less challenging FindIt benchmark, but these improvements do not consistently transfer to RTM and FindItAgain, which contain more carefully constructed human-made forgeries. However, the image-level results in \cref{appendix:generator-image-results} show that this advantage does not hold at the image level, where models trained on our dataset outperform the generator-based datasets across all three evaluation datasets. Unlike the controlled comparison, this experiment evaluates the practical effectiveness of each training dataset rather than isolating the generation strategy.

\par\noindent \textbf{Robustness to OCR quality.} Our generation pipeline relies on character-level OCR to obtain the initial bounding boxes used by \texttt{ExtractLineSegments}. A natural concern is therefore how sensitive the overall performance of models trained on this data is to the accuracy of the OCR engine. We study this in two ways: (i) replacing the default OCR system with Tesseract, and (ii) explicitly perturbing the OCR bounding boxes. The results illustrated in \cref{tab:ocr-robustness} show that our generation pipeline is robust to OCR quality.

\begin{table}[!t]
\centering
  \small
  \setlength{\tabcolsep}{5.5pt}
  
\caption{Robustness of our data generation pipeline to OCR quality. We report the average of F1 scores across six of the seven detectors, excluding DITL$^2$. Details are provided in \cref{appendix:ocrqualityrobustness}.}
\label{tab:ocr-robustness}
\begin{tabular}{l|cc|cc|cc}
\toprule
\multirow{2}{*}{\textbf{Setting}}
& \multicolumn{2}{c|}{\textbf{RTM}}
& \multicolumn{2}{c|}{\textbf{FindItAgain}}
& \multicolumn{2}{c}{\textbf{FindIt}} \\
& \textbf{Img} & \textbf{Pix}
& \textbf{Img} & \textbf{Pix}
& \textbf{Img} & \textbf{Pix} \\
\midrule
Google OCR  & 43.5 & 16.0 & 51.9 & 14.1 & 64.7 & 17.9 \\
Tesseract OCR & 43.0 & 15.6 & 51.5 & 13.7 & 64.3 & 17.5 \\
Perturbed boxes & 42.8 & 15.3 & 51.4 & 13.3 & 64.1 & 17.2 \\
\bottomrule
\end{tabular}
\end{table}
\begin{table}[!t]
\vspace{0pt}
\centering
  \small
  \setlength{\tabcolsep}{3pt}
\caption{Ablation study on the auxiliary networks $F_{\theta}$ and $G_{\theta}$ used in the data generation process. When ablating $F_{\theta}$, Sauvola thresholding was used, as in \citet{Doctamper}. Results are averaged across evaluation datasets. Per-model and per-dataset results are provided in \cref{appendix:ablation}.}
\label{tab:ablation-f1-permodel}
\begin{tabular}{lcccccccc}
\toprule
\multirow{2}{*}{\textbf{Model}} &
\multicolumn{2}{c}{\textbf{Ours}} &
\multicolumn{2}{c}{\textbf{w/o $G_{\theta}$}} &
\multicolumn{2}{c}{\textbf{w/o $F_{\theta}$}} &
\multicolumn{2}{c}{\textbf{w/o $F_{\theta},G_{\theta}$}} \\
\cmidrule(lr){2-3}\cmidrule(lr){4-5}\cmidrule(lr){6-7}\cmidrule(lr){8-9}
& \textbf{Img} & \textbf{Pix} & \textbf{Img} & \textbf{Pix} & \textbf{Img} & \textbf{Pix} & \textbf{Img} & \textbf{Pix} \\
\midrule
CAT-Net    & \textbf{41.3} & \textbf{9.7}  & 41.1 & 9.6  & 40.6 & 9.5  & 40.5 & 9.5 \\
DTD        & \textbf{52.5} & \textbf{13.2} & 51.5 & 12.9 & 49.1 & 12.3 & 48.3 & 12.0 \\
PSCC-Net   & \textbf{52.2} & \textbf{11.4} & 51.3 & 11.0 & 45.9 & 10.3 & 43.8 & 9.8 \\
ADCD-Net   & \textbf{60.3} & \textbf{17.3} & 59.9 & 16.9 & 56.6 & 16.0 & 55.4 & 15.7 \\
ASC-Former & \textbf{51.5} & \textbf{17.2} & 50.9 & 16.8 & 49.0 & 14.5 & 48.4 & 14.1 \\
FFDN       & \textbf{62.3} & \textbf{27.0} & 61.4 & 26.6 & 57.8 & 24.0 & 55.0 & 23.5 \\
\bottomrule
\end{tabular}
\vspace{0pt}

\end{table}

\begin{table}[!b]
\centering
  \small
  \setlength{\tabcolsep}{1pt}

\setlength{\tabcolsep}{0.4mm}
\caption{Ablation of heuristic rules in our contrastive learning mining strategy and the data generation pipeline. Results are averaged across ADCD-Net, ASC-Former, and FFDN, and across evaluation datasets.}
\label{tab:ablation-main}
\begin{tabular}{l|ccc|ccc}
\toprule
\textbf{Setting} & \multicolumn{3}{c|}{\textbf{Image-level}} & \multicolumn{3}{c}{\textbf{Pixel-level}} \\
 & \textbf{Prec} & \textbf{Rec} & \textbf{F1} & \textbf{Prec} & \textbf{Rec} & \textbf{F1} \\
\midrule
Ours & \textbf{59.1} & \textbf{58.4} & \textbf{58.0} & \textbf{24.7} & \textbf{17.9} & \textbf{20.5} \\
\midrule
\multicolumn{7}{l}{\textit{Contrastive ablations}} \\
w/o hard-negative augmentation        & 58.0 & 56.6 & 56.8 & 23.9 & 16.8 & 19.5 \\
w/o cross-document negatives          & 58.5 & 57.4 & 57.2 & 24.1 & 17.2 & 19.8 \\
\midrule
\multicolumn{7}{l}{\textit{Generation ablations}} \\
w/o \texttt{ExtractLineSegments}     & 58.0 & 55.9 & 56.2 & 23.4 & 16.7 & 19.3 \\
w/o equal number of chars  & 57.8 & 56.2 & 56.0 & 22.8 & 17.3 & 19.4 \\
w/o aspect-ratio tolerance            & 58.9 & 57.7 & 57.5 & 24.4 & 17.3 & 20.0 \\
\bottomrule
\end{tabular}
\end{table}

\noindent \textbf{Ablation study.} \Cref{tab:ablation-f1-permodel} reports the impact of removing the similarity network $\mathcal{F}_{\theta}$ and the bounding-box quality network $\mathcal{G}_{\theta}$ from the generation pipeline on the performance of models trained on the resulting generated dataset. Removing $\mathcal{F}_{\theta}$ results in a consistent reduction in both image-level and pixel-level F1 scores, demonstrating that selecting visually similar regions is important for producing realistic tampering examples. Excluding $\mathcal{G}_{\theta}$ also causes a performance decline, which confirms the importance of filtering out low-quality crops that would otherwise introduce highly visible artifacts. When both networks are removed, performance drops even further, indicating that the two modules contribute complementary benefits. We also ablate several components of our contrastive mining strategy and data-generation pipeline, namely hard-negative augmentation, cross-document negative sampling, character grouping, the aspect-ratio tolerance used for copy-move, and the character-count constraints used for copy-move and splicing. As shown in \cref{tab:ablation-main}, the performance drop is moderate yet consistent, indicating that these design choices contribute meaningfully to our approach. A complementary sensitivity analysis confirms the robustness of our approach: varying the main thresholds by up to 25\% in either direction changes average pixel-level F1 by at most 0.3 points, while substantially larger variations degrade performance. Details are provided in \cref{appendix:hyper}.
\par\noindent \textbf{Generalization to AI-generated tampering.}
We evaluate against manipulations produced by FLUX-Text~\cite{Flux-text}, AnyText~\cite{Anytext}, and Qwen-Image-Edit~\cite{wu2025qwen}. Despite being trained without AI-generated manipulations, models pretrained on our data outperform the strongest prior generation baselines in eight of nine generator--architecture comparisons, improving average pixel-level F1 from 81.9 to 84.1 for FFDN and from 70.2 to 72.7 for ASC-Former.
 Further results and analysis are provided in \cref{appendix:ai-tampering}.
\section{Conclusion} We presented a similarity-guided framework for generating tampered document images, built on two lightweight networks that mitigate the identified failure modes in existing rule-based pipelines. In controlled comparisons, training on our data consistently improved seven detection models across three human-made benchmarks, with gains that persist after fine-tuning and extend to unseen AI-generated manipulations. Detecting carefully crafted human-made document tampering remains extremely challenging, and our work represents a step toward more robust detection and localization through higher-quality synthetic training data. We release our dataset of 2.8 million document images together with the generation code and model weights, and hope it provides a stronger foundation for future research.

\section*{LLM Usage Statement}

GPT-6 was used primarily to assist with correcting and improving the clarity of text originally written by the authors, as well as assist with refining author-written code. All suggestions were reviewed and further revised by the authors, who take full responsibility for the final manuscript.

{
    \small
    \bibliographystyle{ieeenat_fullname}
    \bibliography{main}
}


\clearpage

\appendix

\setcounter{topnumber}{3}
\setcounter{bottomnumber}{2}
\setcounter{totalnumber}{5}
\setcounter{dbltopnumber}{3}
\renewcommand{\topfraction}{0.92}
\renewcommand{\bottomfraction}{0.85}
\renewcommand{\textfraction}{0.06}
\renewcommand{\floatpagefraction}{0.70}
\renewcommand{\dbltopfraction}{0.92}
\renewcommand{\dblfloatpagefraction}{0.70}

\section{Contrastive Algorithm Details}
\label{sec:contrastive-training}
In \cref{main:contrastive}, we describe the contrastive learning strategy used to train $F_{\theta}$, including the procedure for mining both positive and negative pairs. An overview of the mining pipeline is provided in \cref{alg:contrastive-training}. For an anchor page $I$ of height $H$ and mean character height $h_{\mathrm{char}}(I)$, the minimum gap is $g_I=\max(256,\tau_1 h_{\mathrm{char}}(I),\lfloor H/4\rfloor)$ pixels, with $\tau_1=10$ by default. We denote the smallest of the four vertical distances between the top and bottom edges of two boxes by $d_y$.

\begin{algorithm*}
\caption{Mining positive and negative crop pairs for training \(\mathcal{F}_\theta\).}
\label{alg:contrastive-training}
\begin{algorithmic}[1]
\scalefont{0.83}
\State \textbf{Input:} Document image \(I_0\), character-level OCR, thresholds \(\tau_0,\tau_1\), tolerance \(\varepsilon\), negatives per anchor \(N\), augmented hard negatives per anchor \(M_{\text{alt}}\), function \(\mathrm{ExtractLineSegments}\) (\cref{appendix:getlines}), function \(\mathrm{GetAlteredCrop}\) (\cref{appendix:metadataaug})
\State \textbf{Output:} contrastive pairs and corresponding negative crops

\Statex
\State \(\{(b_i,\ell_i)\}_{i=1}^n \gets \mathrm{ExtractLineSegments}(\mathrm{OCR}(I_0),\texttt{mode}{=}\texttt{contrastive})\)
\State For each \(b_i=(x_i,y_i,w_i,h_i)\), define its center \(c_i=(x_i+\tfrac12w_i,\;y_i+\tfrac12h_i)\)
\State Define \(\operatorname{CharMatch}(b_a,b_b)\) to be true if at least one box is blank, and otherwise to require equal character counts
\State \(\overline w \gets\) average character width in \(I_0\); \(h_{\mathrm{char}}(I_0) \gets\) average character height in \(I_0\)
\State Initialize collection \(\mathcal{S} \gets \emptyset\)

\For{each anchor index \(i\)}
    \State \(a_i \gets \mathrm{crop}(I_0, b_i)\)
    \State Initialize \(\mathcal{P}_i \gets \emptyset\), \(\mathcal{N}_i \gets \emptyset\)

    \State \textit{Step 1: Positive mining.}
    \For{each \(j \neq i\)}
        \If{\(\ell_j = \ell_i\), \(w_j = w_i\), \(h_j = h_i\), \(\operatorname{CharMatch}(b_i,b_j)\), and \(\lVert c_i-c_j\rVert_2 < \tau_0 \cdot \overline w\)}
            \State \(\mathcal{P}_i \gets \mathcal{P}_i \cup \{\mathrm{crop}(I_0, b_j)\}\)
        \EndIf
    \EndFor
    \If{\(|\mathcal{P}_i| = 0\)} 
        \State \textbf{continue}
    \Else
    \State $p \leftarrow \textsc{RandomChoice}(\mathcal{P}_i)$
    \State $\mathcal{P}_i \leftarrow \{p\}$
\EndIf

    \State \textit{Step 2: Anchor-based hard negatives.}
    \For{\(t = 1\) to \(M_{\text{alt}}\)}
        \State \(a_i^{\text{alt}} \gets \mathrm{GetAlteredCrop}(I_0, b_i)\)
        \State \(\mathcal{N}_i \gets \mathcal{N}_i \cup \{a_i^{\text{alt}}\}\)
    \EndFor

    \State \textit{Step 3: Intra-image negatives.}
\For{each \(j \neq i\), in increasing order of \(\left|\log \frac{w_j h_i}{h_j w_i}\right|\)}        \State \(a_j \gets \mathrm{crop}(I_0, b_j)\)
        \If{\(d_y(b_i,b_j)\ge g_{I_0}\), \(\operatorname{CharMatch}(b_i,b_j)\), and \(\frac{w_j/h_j}{w_i/h_i} \in [1-\varepsilon, 1+\varepsilon]\)}
            \State \(a_j' \gets \mathrm{resize}(a_j, (w_i, h_i))\)
            \State \(\mathcal{N}_i \gets \mathcal{N}_i \cup \{a_j'\}\)
            \If{\(|\mathcal{N}_i| \ge N\)} \State \textbf{break} \EndIf
        \EndIf
    \EndFor

    \State \textit{Step 4: Cross-document negatives.}
    \While{\(|\mathcal{N}_i| < N\)}
        \State Select a new document \(I'\)
        \State \(\{(b_k,\ell_k)\} \gets \mathrm{ExtractLineSegments}(\mathrm{OCR}(I'),\texttt{mode}{=}\texttt{contrastive})\)
        \For{each \((b_k,\ell_k)\) satisfying $d_y(b_i,b_k)\ge g_{I_0}$ when the pages have the same dimensions, in increasing order of \(\left|\log \frac{w_k h_i}{h_k w_i}\right|\)}
            \If{\(\operatorname{CharMatch}(b_i,b_k)\) and \(\frac{w_k/h_k}{w_i/h_i}\in[1-\varepsilon,1+\varepsilon]\)}
                \State \(a_k' \gets \mathrm{resize}(\mathrm{crop}(I', b_k), (w_i, h_i))\)
                \State \(\mathcal{N}_i \gets \mathcal{N}_i \cup \{a_k'\}\)
            \EndIf
            \If{\(|\mathcal{N}_i| = N\)} \State \textbf{break} \EndIf
        \EndFor
    \EndWhile

    \State Add tuple \((a_i, \mathcal{P}_i, \mathcal{N}_i)\) to \(\mathcal{S}\)
\EndFor

\State \Return \(\mathcal{S}\)
\end{algorithmic}
\end{algorithm*}
\section{Human Analysis}
\label{sec:human-review}

\subsection{Protocol}

Each case presents three crops: the \emph{anchor}, and two candidates shown as
\emph{Option 1} and \emph{Option 2}. One candidate is the mined positive and the
other the mined negative. The annotator is asked a single question: which option
is more visually similar to the anchor. Annotators are instructed to match
visual \emph{style} rather than content, and are given the following criteria as
guidance. The list is indicative rather than exhaustive, and annotators are free
to rely on any other cue they notice:
\begin{enumerate}
  \item \textbf{Letterforms.} Typeface and letter shape; letter size; stroke
        thickness; slant.
  \item \textbf{Layout.} Letter spacing; vertical placement of the text within
        the crop, i.e.\ whether the two crops placed side by side would show
        their text at the same line level.
  \item \textbf{Colour.} Text colour; background colour and texture.
  \item \textbf{Capture quality.} Brightness and contrast; sharpness or blur.
\end{enumerate}
Annotators must choose one of the two candidates.

\begin{figure}[!htbp]
\centering
\includegraphics[width=\columnwidth]{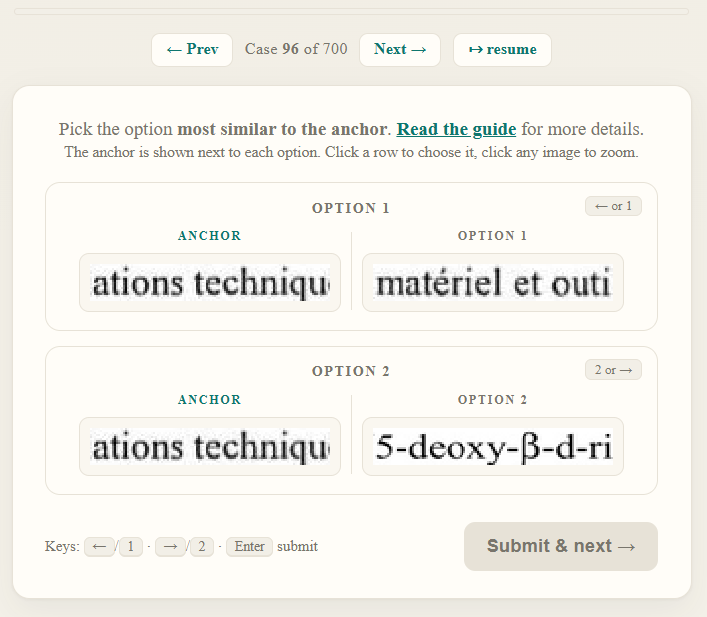}
\caption{The annotation interface. The anchor is repeated next to each candidate
so that both comparisons are made side by side, and the annotator selects one of
the two rows.}
\label{fig:interface}
\end{figure}

The study is blinded: the assignment of positive and negative to Option 1 and
Option 2 is randomised independently for each case, and annotators are never
shown which candidate was mined as the positive, nor any other annotator's
decision. Crops are presented at their original resolution, upscaled with
nearest-neighbour interpolation so that small crops remain legible without
introducing interpolation artefacts. \Cref{fig:interface} shows the annotation
interface.

\subsection{Design}

Five annotators each judge 700 cases: 600 assigned uniquely to that annotator
and 100 drawn from a shared block labelled by all five. This yields 3{,}500
judgments over 3{,}100 distinct cases. The shared block supports the
inter-annotator agreement analysis, while the unique blocks maximise coverage.
No annotator sees the same case twice.
\paragraph{Ethics and consent.}
Five volunteers participated in this minimal-risk,
non-invasive visual annotation study and provided
verbal informed consent. Participation was voluntary,
and individual responses are reported using anonymous
identifiers (R1--R5). Institutional ethics review was
not required for this type of study under the
applicable institutional guidelines.

\subsection{Results}

\begin{table}[!htbp]
\centering
\small
\caption{Per-annotator results over 700 cases each (3{,}500 judgments). Acc.\ is
the share of judgments in which the annotator chose the mined positive as the
more similar crop.}
\label{tab:reviewers}
\begin{tabular}{lrrr}
\toprule
Annotator & Acc.\ (\%) & Correct & Incorrect \\
\midrule
R1 & 97.7 & 684 & 16 \\
R2 & 98.0 & 686 & 14 \\
R3 & 97.4 & 682 & 18 \\
R4 & 96.6 & 676 & 24 \\
R5 & 96.0 & 672 & 28 \\
\midrule
All & 97.1 & 3400 & 100 \\
\bottomrule
\end{tabular}
\end{table}

\Cref{tab:reviewers} reports per-annotator outcomes. Annotators choose the mined
positive as the more similar crop in 97.1\% of judgments, ranging from 96.0\% to
98.0\% across annotators. The mined negative is preferred in 100 of 3{,}500
judgments (2.9\%).

\subsection{Agreement}

\begin{figure}[!htbp]
\centering
\includegraphics[width=\columnwidth]{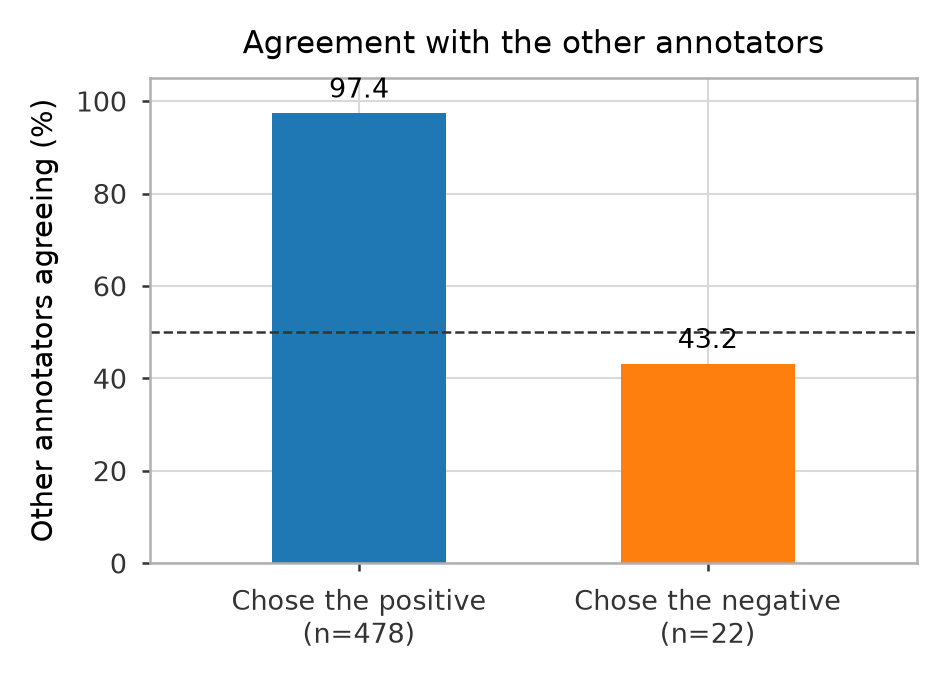}
\caption{For each of the 500 judgments on the 100 cases labelled by all five
annotators, the share of the other four who made the same choice, split by
whether the annotator chose the mined positive or the mined negative. The dashed
line marks the 50\% reference corresponding to a uniformly random binary choice.}
\label{fig:agreement}
\end{figure}

The 100 shared cases yield 500 judgments, for each of which we measure how many
of the other four annotators made the same choice. When an annotator selects the
mined positive, the other four agree 97.4\% of the time (478 judgments); when an
annotator selects the mined negative, they agree only 43.2\% of the time (22
judgments), below the 50\% reference corresponding to a uniformly random binary choice (\cref{fig:agreement}). Choosing
the positive is thus the consensus behaviour, whereas choosing the negative is an
isolated decision that the other annotators usually do not agree with. Mean
pairwise agreement across the shared block is 95.0\% and 89 of the 100 cases are
unanimous.

Aggregating the five annotators by majority vote suppresses most of this residual
disagreement. Of the 22 judgments that preferred the mined negative, 13 are
overturned by the majority (59.1\%) while a single one of the 478 judgments that
preferred the positive is lost (0.21\%), so 98 of the 100 shared cases end up
agreeing with the mined positive. Applying these two rates to the full corpus,
aggregation would move 59 of the 100 negative judgments to the positive and 7 of
the 3{,}400 positive judgments to the negative, leaving 48 of 3{,}500 in
disagreement, or 1.4\%. We take this as an estimate of the share of mined pairs
in which the negative is genuinely the better match, as opposed to those in which
an individual annotator simply erred.

\section{Generation algorithm details}
\label{sec:data-generation}
In \cref{main:generation}, we present our data generation pipeline. We further summarize the full procedure in \cref{alg:data-generation}. Given a nonempty set of candidates, $\mathrm{Select}$ samples among those above the similarity threshold, giving higher scores greater weight. If none qualifies, it returns the highest-scoring candidate. The edit is applied only if the selected score passes the threshold.

\begin{algorithm*}
\scalefont{0.7}
\caption{Tampering a Set of Document Images Using \(\mathcal{F}_\theta\) and \(\mathcal{G}_\theta\)}
\label{alg:data-generation}
\begin{algorithmic}[1]
\State \textbf{Input:} Document images \(\{I^{(d)}\}_{d=1}^D\); character-level OCR; threshold \(\tau_2\); aspect-ratio tolerance \(\varepsilon'\); probabilities \(p_{ins},p_{inp},p_{spl}\); maximum requested manipulations \(n_{\max}\); font set \(\{f_j\}\); functions \(\mathrm{ExtractLineSegments}\), \(\mathrm{Render}\), \(\mathrm{Inpaint}\), \(\mathcal{F}_\theta\), and \(\mathcal{G}_\theta\)
\State \textbf{Output:} Generated image and mask pairs

\Statex
\State \textit{Step 1: Preprocess documents and build the crop database}
\State \(\mathrm{DB}\gets\emptyset\)
\For{each document \(I^{(d)}\)}
  \State \(\{b_i^{(d)}\}_{i=1}^{k_d}\gets\mathrm{ExtractLineSegments}(\mathrm{OCR}(I^{(d)}),\texttt{mode}{=}\texttt{generation})\)
  \For{each segment \(b_i^{(d)}=(x_i,y_i,w_i,h_i)\)}
    \State \(c_i\gets\mathrm{crop}(I^{(d)},b_i^{(d)})\)
    \State Record \(d\), \(b_i^{(d)}\), \(\mathrm{text}_i\), \(\mathrm{isBlank}_i\), \(w_i\), \(h_i\), and \(\#\mathrm{chars}_i\)
    \If{\(\mathrm{isBlank}_i\)}
      \State Add the segment and its metadata to \(\mathrm{DB}\)
    \Else
      \State \(s_i\gets\mathrm{ExtractContextStripes}(I^{(d)},b_i^{(d)})\)
      \If{\(\mathcal{G}_\theta(c_i,s_i)>\tau_2\)}
        \State Add the segment and its metadata to \(\mathrm{DB}\)
      \EndIf
    \EndIf
  \EndFor
\EndFor

\Statex
\State \textit{Step 2: Generate images}
\For{each target document \(I\)}
  \State Initialize \(\mathrm{mask}\gets\mathbf{0}^{H\times W}\)
  \State Retrieve the retained target segments \(\mathcal{B}=\{b_i\}\) for \(I\)
  \State Sample \(n\sim\mathcal{U}(\{0,\ldots,n_{\max}\})\)
  \State Sample \(n\) distinct target segments \(\{r_1,\ldots,r_n\}\subset\mathcal{B}\) uniformly without replacement

  \For{each target region \(r_i=(x_i,y_i,w_i,h_i)\)}
    \If{\(r_i\) is blank and \texttt{Bernoulli}(\(p_{ins}\))}
      \State Find the nearest text segment \(c\in\mathcal{B}\) of size \((w_i,h_i)\); let \(t_c\) be its text
      \State \(v_c\gets\mathcal{F}_\theta(c)\)
      \State Estimate foreground color \(col_0\) using Sauvola thresholding
      \State Define \(col_k=col_0+(\Delta_R,\Delta_G,\Delta_B)\), with \(\Delta_X\in\{-2,-1,0,1,2\}\)
      \For{each font \(f_j\) and color \(col_k\)}
        \State \(P_{j,k}\gets\mathrm{Render}(t_c,f_j,col_k,\mathrm{crop}(I,r_i))\)
        \State \(v_{j,k}\gets\mathcal{F}_\theta(P_{j,k})\)
      \EndFor
      \State \((j^*,k^*)\gets\mathrm{Select}(\{s_{j,k}\})\) using the rendering scores. Skip this edit if none qualifies.
      \State Replace \(r_i\) by \(P_{j^*,k^*}\) and mark \(r_i\) in \(\mathrm{mask}\)

    \ElsIf{\(r_i\) is not blank and \texttt{Bernoulli}(\(p_{inp}\))}
      \State \((I,M_{\mathrm{inp}})\gets\mathrm{Inpaint}(I,r_i)\)
      \State \(\mathrm{mask}\gets\mathrm{mask}\lor M_{\mathrm{inp}}\)

    \Else
      \If{\texttt{Bernoulli}(\(p_{spl}\))}
        \State \(\mathcal{C}_{\mathrm{text}}\gets\) text crops in \(\mathrm{DB}\) from a different document, with matching size and character count, and text different from \(r_i\)
        \If{\(r_i\) is not blank}
          \State \(\mathcal{C}_{\mathrm{blank}}\gets\) blank crops in \(\mathrm{DB}\) from a different document, with matching size and character count
        \Else
          \State \(\mathcal{C}_{\mathrm{blank}}\gets\emptyset\)
        \EndIf
      \Else
        \State \(\mathcal{C}_{\mathrm{text}}\gets\) non-overlapping text crops from \(I\), excluding \(r_i\), with different text, matching character count, and aspect-ratio ratio in \([1/(1+\varepsilon'),1+\varepsilon']\)
        \If{\(r_i\) is not blank}
          \State \(\mathcal{C}_{\mathrm{blank}}\gets\) non-overlapping blank crops from \(I\), excluding \(r_i\), with matching character count and aspect-ratio ratio in \([1/(1+\varepsilon'),1+\varepsilon']\)
        \Else
          \State \(\mathcal{C}_{\mathrm{blank}}\gets\emptyset\)
        \EndIf
      \EndIf
      \State \(\mathcal{C}\gets\mathcal{C}_{\mathrm{text}}\cup\mathcal{C}_{\mathrm{blank}}\)
      \If{\(\mathcal{C}=\emptyset\)} \State \textbf{continue} \EndIf
      \State Resize every \(c_j\in\mathcal{C}\) to \((w_i,h_i)\)
      \State \(c_r\gets\mathrm{crop}(I,r_i)\), \(v_r\gets\mathcal{F}_\theta(c_r)\)
      \State \(j^*\gets\mathrm{Select}(\{s_j\})\) using the donor scores. Skip this edit if none qualifies.
      \State Replace \(r_i\) by \(c_{j^*}\) and mark \(r_i\) in \(\mathrm{mask}\)
    \EndIf
  \EndFor
  \State \textbf{yield} \((I,\mathrm{mask})\)
\EndFor
\end{algorithmic}
\end{algorithm*}

The valid source-target combinations are as follows. Insertion renders text in a blank target. Inpainting is applied only to a text target. Coverage replaces a text target with a blank source crop. Copy-move uses a text source from a non-overlapping region of the same document, and splicing uses a text source from a different document. Copy-move and splicing may target either text or blank regions. Blank-to-blank replacement is not allowed.

The number of manipulations is sampled uniformly as
\[
 n\sim\mathcal{U}(\{0,\ldots,n_{\max}\}).
\]
Each value therefore has probability \(1/(n_{\max}+1)\). With \(n_{\max}=5\), one sixth of the images are initially assigned zero requested manipulations. The number of requested manipulations is subsequently capped according to the available document content, and unsuccessful manipulation attempts may further reduce the final number of manipulations. Pristine images are retained as negative examples for image-level classification.
\section{Extracting Line Segments}
\label{appendix:getlines}

\noindent \texttt{ExtractLineSegments} converts raw OCR character boxes into a set of contiguous text (or blank) segments, each tagged with a line index.  These segment pairs \((b_i,\ell_i)\) serve as anchors for both our contrastive‐learning sampling and our tampering pipelines. The algorithm can be found in \cref{alg:extract-line-segments} and a detailed explanation is given below.

\medskip
\noindent\textbf{1. Line clustering.}\\
Given a character level OCR that returns a set of character‐level boxes \(\{b_i\}\), where
\(b_i = (x_i, y_i, w_i, h_i)\) denotes the top‐left corner \((x_i,y_i)\), width \(w_i\), and height \(h_i\), we first sort all \(b_i\) by their bottom edge \(y_i + h_i\).  Starting from the topmost box, we form a new line cluster \(\mathcal L_1\).  We then iterate through the remaining boxes in sorted order: if the vertical span of a box differs from the first box of the current cluster by at most a threshold \(\delta_y\), we add it to that cluster; otherwise we begin a new line cluster.  This produces
\(
\{\mathcal L_1,\mathcal L_2,\dots,\mathcal L_L\},
\)
where each \(\mathcal L_\ell\) is a vertically consistent set of boxes.

\medskip
\noindent\textbf{2. Segment generation.}\\
Within each line \(\mathcal L_\ell\), we sort its boxes by their horizontal midpoint \(x_i + \tfrac12w_i\).  We then enumerate every contiguous subsequence,
\[
\{\,b_{i_1},\,b_{i_2},\,\dots,\,b_{i_k}\}\subset\mathcal L_\ell,\quad k\ge1,
\]
and merge it into a single segment whose bounding box 
\[
b = \bigl(\min_j x_{i_j},\;\min_j y_{i_j},\;\max_j(x_{i_j}+w_{i_j})-\min_j x_{i_j},\]
\[\;\max_j(y_{i_j}+h_{i_j})-\min_j y_{i_j}\bigr)
\]

and which inherits the line index \(\ell\).  
Collecting all such \(\{(b_i,\ell_i)\}_{i=1}^n\) yields every possible contiguous run in each line. The text content of these runs is simply the concatenation of their character in their horizontal order
\medskip

\medskip
\noindent\textbf{3. Blank-segment construction.}\\
In contrastive-learning mode, we enrich the segment set with blank regions and search for a hard-negative partner for each horizontal blank. We iterate over an immutable copy of the original text segments, collect newly generated regions separately, and concatenate them after the original segments have been processed:
\begin{enumerate}
  \item Compute the average character‐width \(\bar w\) and height \(h_{\mathrm{char}}(I)\) from all single‐character boxes on page $I$.
  \item For each \((b_i,\ell_i)\) of size \((w_i,h_i)\), attempt to place a same‐sized box at horizontal offsets \(\pm k\,w_i\) (for \(1\le k\le \lfloor W/w_i\rfloor\)) that
    \begin{itemize}
    \item lies within the central third of the page width (applied only during contrastive‐learning sampling to avoid pairing text crops with blank regions near the page margins, where background color often shifts)
      \item does not overlap any existing box.
    \end{itemize}
    When found, add its bounding box with the same line index \(\ell_i\) and associate to it a string of “+” characters whose length matches the string in \((b_i)\). Note that using “+” here is arbitrary, any symbol not found naturally in documents will work. The only requirements are that these synthetic strings do not occur in real text (so they are not filtered out when selecting candidate crops with differing content) and that their length matches that of a nearby text segment, ensuring that, when rendered, the resulting font size is similar to nearby text regions.
    \item After placing a horizontal blank, search above and below $b_i$ for a non-overlapping box. Start at edge gap $g_I$ (\cref{sec:contrastive-training}) and increase the gap by one box height per trial.
\end{enumerate}
During tampering generation, blank candidates are instead obtained through a separate page-wide non-overlap search and are not subject to the central-third restriction.

\begin{algorithm}
\scalefont{0.76}
\caption{ExtractLineSegments}
\label{alg:extract-line-segments}
\begin{algorithmic}[1]
\Require Character boxes $\{b_i=(x_i,y_i,w_i,h_i)\}$, vertical threshold $\delta_y$ (set to 3), image width $W$, average char width $\bar w$, average char height $h_{\mathrm{char}}(I)$, processing mode $\texttt{mode}\in\{\texttt{contrastive},\texttt{generation}\}$
\Ensure Segment list $\{(b,\ell,\text{text})\}$
\State  1. Line clustering
\State Sort $\{b_i\}$ by $y_i + h_i$ ascending
\State $\mathcal L \gets []$, current cluster $\gets [\,]$
\For{each $b_i$ in sorted order}
  \If{current cluster is empty \textbf{or} $|y_i - y_{\text{first}}|\le\delta_y$ and $|y_i+h_i - (y_{\text{first}}+h_{\text{first}})|\le\delta_y$}
    \State add $b_i$ to current cluster
  \Else
    \State append current cluster to $\mathcal L$
    \State current cluster $\gets [\,b_i\,]$
  \EndIf
\EndFor
\State append current cluster to $\mathcal L$

\State  2. Segment generation
\State $S \gets []$
\For{each line $\mathcal L_\ell$ in $\mathcal L$}
  \State sort $\mathcal L_\ell$ by $x_i + \tfrac12 w_i$
  \For{every contiguous subsequence $(b_{i_1},\dots,b_{i_k})$, $k\ge1$}
    \State $x_{\min}\!\gets\!\min_j x_{i_j}$,\ 
          $y_{\min}\!\gets\!\min_j y_{i_j}$
    \State $w\!\gets\!\max_j(x_{i_j}{+}w_{i_j}) - x_{\min}$,\ 
          $h\!\gets\!\max_j(y_{i_j}{+}h_{i_j}) - y_{\min}$
    \State $\text{text}\gets$ concatenate chars of $b_{i_j}$ in order
    \State append $(\,(x_{\min},y_{\min},w,h),\,\ell,\,\text{text}\,)$ to $S$
  \EndFor
\EndFor

\State  3. Blank-segment construction
\State $S_{\mathrm{text}}\gets\mathrm{copy}(S)$, $S_{\mathrm{syn}}\gets[]$
\If{$\texttt{mode}=\texttt{contrastive}$}
  \For{each $(b_i,\ell_i,\text{text}_i)$ in $S_{\mathrm{text}}$}
    \State $(x,y,w,h)\gets b_i$
    \For{$k=1$ \textbf{to} $\lfloor W/w\rfloor$}
      \For{$d\in\{-1,+1\}$}
        \State $x'\gets x + d\cdot k\cdot w$
        \If{$x'\in[\tfrac W3,\tfrac{2W}3-w]$ \textbf{and} no overlap with any box}
          \State $\text{text} \gets \texttt{‘+’}^{\len(\text{text}_i)}$
          \State append $(\,(x',y,w,h),\,\ell_i,\,\text{text}\,)$ to $S_{\mathrm{syn}}$
          \State \textbf{break} both loops
        \EndIf
      \EndFor
    \EndFor
    \State skip this search if no horizontal blank was found
    \For{in-bounds $y\prime\in\{y-h-d_v,y+h+d_v\}$, $d_v=g_I,g_I+h,\ldots$}
      \For{$k=1$ \textbf{to} $\lfloor W/w\rfloor$}
        \For{$d\in\{-1,+1\}$}
          \State $x'\gets x + d\cdot k\cdot w$
          \If{inside page \textbf{and} no overlap}
            \State $\text{text} \gets \texttt{‘-’}^{\len(\text{text}_i)}$
            \State append $(\,(x',y',w,h),\,\ell_i,\,\text{text}\,)$ to $S_{\mathrm{syn}}$
            \State \textbf{end} this hard-negative search
          \EndIf
        \EndFor
      \EndFor
    \EndFor
  \EndFor
\Else
  \State $S_{\mathrm{syn}}\gets$ page-wide non-overlapping blank candidates obtained by the generation-time blank-region search
\EndIf

\State \Return $\mathrm{concat}(S_{\mathrm{text}},S_{\mathrm{syn}})$
\end{algorithmic}
\end{algorithm}

\section{More about the rendering and inpainting functions}

\subsection{Rendering}
\label{appendix:render}

In \cref{alg:data-generation}, the function
\[
  P_{j,k} = \mathrm{Render}\bigl(t_c, f_j, \mathrm{col}_k, \mathrm{crop}(I, r)\bigr)
\]
takes as input a text string \(t_c\), a font identifier \(f_j\), a foreground color
\(\mathrm{col}_k \in [0,255]^3\), and the target image patch
\(B = \mathrm{crop}(I,r) \in \mathbb{R}^{H \times W \times 3}\).
It outputs a new patch
\[
  P_{j,k} \in \mathbb{R}^{H \times W \times 3},
\]
with the same spatial size as \(B\), where the text \(t_c\) has been rasterized on top
of the original background using font \(f_j\) and color \(\mathrm{col}_k\).

Implementation-wise, the rendering function proceeds as follows:
\begin{itemize}
  \item A visible region inside the box is defined by subtracting fixed horizontal
        and vertical margins from the width and height of \(B\).
  \item A font scale and stroke thickness are chosen so that the rendered text
        fits entirely inside this region. This is done via a simple
        \emph{binary search} on the scale, using  OpenCV's 
        \texttt{getTextSize} as a feasibility check.
  \item Once a valid scale is found, the text is drawn onto a copy of \(B\),
         centered in the box, using
        OpenCV’s \texttt{putText} function.
\end{itemize}

The function therefore acts as a deterministic rasterizer:
\[
  (t_c, f_j, \mathrm{col}_k, B) \longmapsto P_{j,k},
\]
ensuring that the entire string lies within the target region while preserving
the original background outside the glyph strokes.

\subsection{Inpainting}
\label{appendix:inpaint}

In \cref{alg:data-generation}, the function
\[
  (I_{\mathrm{inp}},M_{\mathrm{inp}})
  =
  \mathrm{Inpaint}(I,r)
\]
takes the complete document image
\(I\in\mathbb{R}^{H\times W\times 3}\) and a target region
\(r=(x,y,w,h)\). It returns an inpainted image
\(I_{\mathrm{inp}}\) and a full-image binary mask
\(M_{\mathrm{inp}}\) indicating the pixels assigned to the inpainting operation.

We select one of the following two modes with equal probability:
\begin{itemize}
  \item \textbf{Full-box mode.}
  The complete target region is marked in the full-image mask:
  \[
    M_{\mathrm{inp}}[y:y+h,\;x:x+w]=1.
  \]
  Inpainting is applied to the complete document image, so the unmasked pixels surrounding the target region provide the context used to reconstruct its content.

  \item \textbf{Text-only mode.}
  The target region is converted to grayscale and a local threshold map is computed using Sauvola thresholding. Pixels darker than their local threshold are identified as foreground strokes and inserted into the corresponding location of \(M_{\mathrm{inp}}\). The remaining pixels of the target region stay unmasked.
\end{itemize}

In both modes, the selected pixels are reconstructed by applying OpenCV's Navier--Stokes inpainting method, \texttt{INPAINT\_NS}, to the complete image with an inpainting radius of \(3\):
\[
  I_{\mathrm{inp}}
  =
  \mathrm{OpenCVInpaint}
  \bigl(I,M_{\mathrm{inp}},3,\mathrm{INPAINT\_NS}\bigr).
\]
The full-box option uses context outside the target region, while the text-only option preserves the unmasked background pixels inside the region.

\section{More about $F_{\theta}$ and its training process}
\subsection{Data augmentation applied on anchors}
\label{appendix:metadataaug}

To generate hard negatives, we apply a series of randomized augmentations to the anchor crop itself. These augmented variants are then added to the anchor's negative set. The goal is to create visually inconsistent but structurally similar crops that challenge the contrastive model to focus on fine-grained visual differences.

Given a bounding box \(b_i = (x, y, w, h)\) defining the anchor crop in an image \(I\), we extract the crop and apply one of two augmentation strategies, as follows:

\paragraph{1. Geometric Perturbation.}
With a probability of 15\%, and only if the crop is not blank, we apply a vertical offset to the anchor crop. The offset is randomly sampled in proportion to the crop height. Depending on the sampled value, the crop is shifted either upward or downward. If the resulting crop dimensions differ from the original, it is resized back to the original width and height. This simulates misalignment.

\paragraph{2. Appearance-Based Transformations.}
Otherwise, we apply a composition of visual augmentations using Albumentations. The pool of available transformations includes RandomBrightnessContrast, HueSaturationValue, MotionBlur, RGBShift, ChannelShuffle, ColorJitter, and RandomTextColor. The number of applied augmentations \(k\) is drawn from a decreasing probability distribution \(P(k) \propto 1/k\). The augmentations are composed and applied in sequence.

\paragraph{3. Validity Check and Filtering.}
To ensure that each augmented crop is sufficiently distinct from the original anchor, we evaluate the following:
\begin{itemize}
    \item \textbf{Pixel difference ratio:} the percentage of differing pixels. We allow a minimal difference of 5 percent.
    \item \textbf{L2 distance:} the Euclidean distance between the raw RGB pixel vectors. We allow a minimal distance of 12.
\end{itemize}

If the augmented crop is too similar to the original anchor, new augmentations are sampled, until the criteria above are met or we exceed one hundred attempts.

\paragraph{4. Final Selection.}
If an augmented version passes the similarity threshold, it is added to the set of hard negatives \( \mathcal{N}_i \) corresponding to the anchor \( b_i \). Otherwise, the attempt is discarded.

\noindent This process is repeated \(M_{\text{alt}}\) times per anchor, resulting in multiple hard negatives that maintain spatial structure but differ in appearance, helping the contrastive model learn more discriminative features.

\subsection{Crop-Similarity Network  Architecture}

\label{appendix:metaarch}

The similarity network $\mathcal{F}_\theta$ is implemented using a modified ConvNeXt-style encoder. Given an input crop $x \in \mathbb{R}^{3 \times H \times W}$, it produces a 192-dimensional descriptor used with cosine similarity.

\paragraph{Backbone.}
We use a ConvNeXt-like architecture with stage depths $\{3,6,27\}$ and channel dimensions $\{32,96,192\}$, with two modifications: (i) the first downsampling layer uses stride $2$ (instead of the default $4$) to preserve higher spatial resolution in early features; and (ii) a fixed 2D sinusoidal positional encoding is injected into the feature map before the final stage to retain absolute spatial information. To avoid losing border pixels when $H$ or $W$ is odd, reflect padding is applied prior to stride-2 downsamples.

\paragraph{Pooling and Meta-head.}
After the final stage, the feature map $(B,192,H',W')$ is processed by three branches: (i) a cross-attention pooling layer with 8 learnable queries (each 128-D) attending over the spatial tokens (keys/values projected from the 192-D features to 128-D); (ii) global average pooling; and (iii) global max pooling. The branch outputs are concatenated to form a vector of size $192 + 192 + 8 \times 128$, which is passed through a small MLP ($\mathrm{Linear}\!\rightarrow\!\mathrm{GELU}\!\rightarrow\!\mathrm{Linear}$) to produce a 192-D descriptor.

\paragraph{Foreground/Background embeddings and usage.}
The 192-D descriptor is split into two L2-normalized 96-D vectors: a background embedding $f_{\mathrm{bg}}(x)$ (layout/texture) and a foreground embedding $f_{\mathrm{fg}}(x)$ (text cues such as font, color, alignment). For comparisons where both crops contain text, the final embedding for each crop is the 192-D concatenation $[f_{\mathrm{bg}}(x);\;f_{\mathrm{fg}}(x)]$, and the similarity is the average of background and foreground cosine scores:
\[
s(x,y)
=\tfrac{1}{2}\,\langle f_{\mathrm{bg}}(x),\,f_{\mathrm{bg}}(y)\rangle
+\tfrac{1}{2}\,\langle f_{\mathrm{fg}}(x),\,f_{\mathrm{fg}}(y)\rangle.
\]
If at least one crop is blank, we route through the background pathway only: the effective embedding reduces to $f_{\mathrm{bg}}(\cdot)$ (96-D), and similarity is computed solely in the background space,
\[
s(x,y)=\langle f_{\mathrm{bg}}(x),\,f_{\mathrm{bg}}(y)\rangle.
\]
\subsection{Sensitivity to visual characteristics}
\begin{figure}[!htbp]
    \centering
    \includegraphics[width=\linewidth]{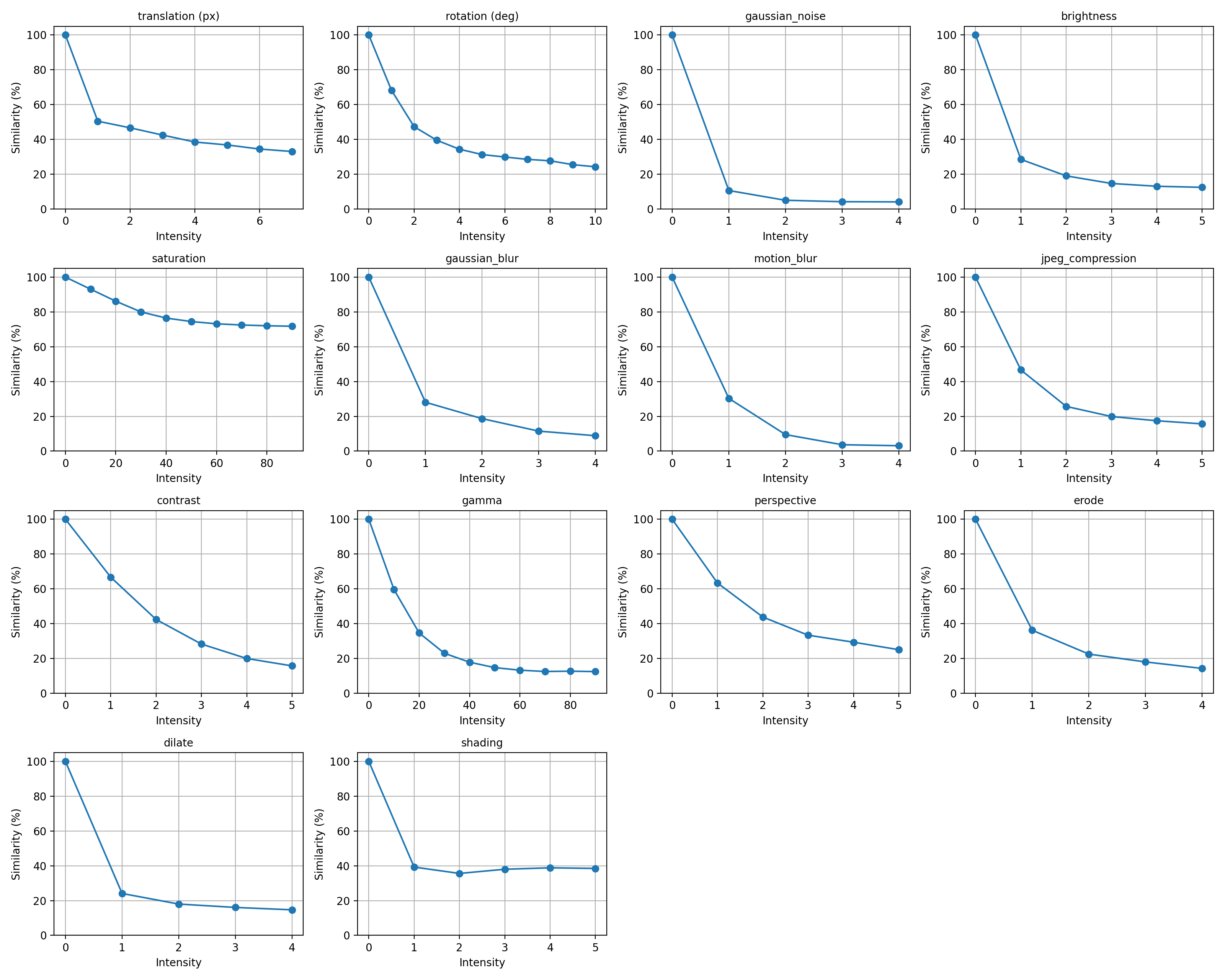}
    \caption{
        Similarity degradation curves for several visual transformations.
        Each curve shows the average percentage of retained similarity
        as the intensity of a single perturbation increases. }
    \label{fig:similarity-degradation}
\end{figure}

To verify that the crop-similarity network does not rely on a single trivial cue but instead exploits a broad range of visual characteristics, we test the sensitivity of the similarity score to local appearance changes. We first collect around 50,000 text-crop pairs for which the network assigns a high self-similarity score ($s > 0.9$). For each pair $(c, c_1)$, we generate perturbed versions $\tilde{c}$ of $c$ by applying one transformation at a time and measure the cosine similarity between the embeddings of $\tilde{c}$ and $c_1$ (and vice versa). For each transformation, we sweep an intensity parameter (e.g., from small to larger perturbations) and report the average percentage of remaining similarity across all crops. We consider the following transformations, each designed to probe a different type of invariance:
translation (horizontal and vertical shifts), which tests sensitivity to small misalignments of the text;
rotation;
Gaussian noise (increasing standard deviation) and JPEG compression (decreasing quality);
brightness and contrast adjustments;
gamma correction;
saturation changes;
Gaussian blur and motion blur;
erosion and dilation, which thin or thicken strokes and approximate the effects of binarization or resampling;
and shading, which applies a smooth illumination gradient across the crop to mimic non-uniform lighting. The results shown in \cref{fig:similarity-degradation} demonstrate that the network is sensitive to all these visual cues and does not rely on any single trivial characteristic.

\section{More About the Crop Quality Assessment Network Architecture and Data Pre-processing}
\subsection{Crop Quality Assessment Network Architecture}
\label{appendix:garch}

The bounding box quality network \(\mathcal{G}_\theta\) uses a multi-branch architecture that processes both the target crop and its surrounding context.

\paragraph{RGB Crop Encoder.}
The crop is processed by a lightweight three-stage 2D ConvNeXt-style convolutional encoder with channel dimensions \(\{32,96,192\}\) and three stride-2 downsampling layers. Reflect padding is applied before downsampling when a spatial dimension is odd, and a fixed 2D sinusoidal positional encoding is added before the final stage. Masked global average and maximum pooling are applied to the final 192-channel feature map, and their outputs are concatenated to produce a 384-dimensional crop representation.

\paragraph{Context Stripe Encoder.}
The context branch receives the top--bottom stripe \(\mathrm{tb}_k\) and the left--right stripe \(\mathrm{lr}_k\). Both are initially represented as 2D RGB regions with thickness \(t=9\).

Each stripe is first processed by a stripe-level 2D convolutional projection that aggregates the complete thickness at every position along the stripe. For the top--bottom stripe, this operation is equivalent to a 2D convolution with kernel size \((t,1)\):
\[
  \mathbb{R}^{3\times t\times W_{\mathrm{tb}}}
  \longrightarrow
  \mathbb{R}^{64\times 1\times W_{\mathrm{tb}}}.
\]
For the left--right stripe, it is equivalent to a 2D convolution with kernel size \((1,t)\):
\[
  \mathbb{R}^{3\times H_{\mathrm{lr}}\times t}
  \longrightarrow
  \mathbb{R}^{64\times H_{\mathrm{lr}}\times 1}.
\]

In the implementation, these 2D projections are computed by folding the RGB and thickness dimensions into \(3t=27\) channels and applying a pointwise 1D convolution:
\[
  (3,t,W_{\mathrm{tb}})
  \rightarrow
  (3t,W_{\mathrm{tb}})
  \xrightarrow{\mathrm{Conv1d}(1)}
  (64,W_{\mathrm{tb}}),
\]
\[
  (3,H_{\mathrm{lr}},t)
  \rightarrow
  (3t,H_{\mathrm{lr}})
  \xrightarrow{\mathrm{Conv1d}(1)}
  (64,H_{\mathrm{lr}}).
\]
This implementation is mathematically equivalent to the stripe-level 2D convolutions above. It combines the RGB values across the full stripe thickness while preserving the longer spatial dimension as a sequence.

The resulting top--bottom and left--right sequences are then processed separately by 1D CNNs. Each encoder contains 27 ConvNeXt-style 1D blocks with depthwise convolutions of kernel size 7, layer normalization, GELU activation, pointwise projections, residual connections, and layer scaling. Masked average and maximum pooling are applied to both sequences. The four resulting 64-dimensional vectors are concatenated to produce a 256-dimensional context representation.

\begin{table}[!htbp]
\centering
  \small
  \setlength{\tabcolsep}{1pt}
\caption{Detection performance on synthetic tampered datasets generated with and without \(\mathcal{F}_\theta\) and \(\mathcal{G}_\theta\). Lower F1 indicates more realistic and challenging tampering. Averages across different models are shown.}
\label{tab:tampering-quality}
\begin{tabular}{l|ccc}
\toprule
\textbf{Generation}  & \multicolumn{3}{c}{\textbf{Pixel-level}} \\
\textbf{Setting} &  \textbf{Prec} & \textbf{Rec} & \textbf{F1} \\
\midrule
ours & 67.6 & 29.3 & \textbf{40.1} \\
w/o \(\mathcal{G}_\theta\) & 68.4 & 39.8 & 49.7 \\
w/o \(\mathcal{F}_\theta\) & 69.9 & 71.5 & 68.5 \\
w/o \(\mathcal{F}_\theta\), \(\mathcal{G}_\theta\)  & 85.1 & 90.2 & 87.4 \\

\bottomrule
\end{tabular}
\end{table}

\paragraph{Classifier Head.}
The 384-dimensional crop representation and the 256-dimensional context representation are concatenated into a 640-dimensional vector. The classifier is

\[
\mathrm{Linear}(640,2560) \rightarrow\mathrm{GELU} \rightarrow\mathrm{Linear}(2560,2).
\]

The classifier produces two logits, \(z_0\) and \(z_1\), corresponding to well-defined and ill-defined crops, respectively. The quality score is computed as \(G_\theta=\sigma(z_0-z_1)\), which is equivalent to the softmax probability of the well-defined class. During training, an auxiliary classifier operating exclusively on the 384-dimensional crop representation is selected with probability 0.5. During standard inference, classification uses both the crop and its surrounding context.

We use a neural network instead of rules based on foreground and background estimation because those alternatives are slower. The speed comparison is reported below.
\begin{algorithm*}[t]
\caption{BorderIntegrity\textsc{Test}}
\label{alg:border-integrity-unified}
\begin{algorithmic}[1]
\Require Either an RGB image $I\in\mathbb{R}^{H\times W\times 3}$ and a box $(x_0,x_1,y_0,y_1)$ with $0\le x_0<x_1\le W$, $0\le y_0<y_1\le H$, or a grayscale crop $C\in\mathbb{R}^{h\times w}$; a boolean \texttt{global}; a boolean \texttt{invert\_mask}; a minimum component area \texttt{min\_component\_area} (set to $4$ pixels), below which a connected component is discarded as binarisation noise rather than treated as text. \textbf{Constant:} $\texttt{TOL}=1$.
\State \textbf{if} \texttt{global} \textbf{then}
\State \quad $h \gets y_1 - y_0$;\; $m \gets \max(\lfloor h/2 \rfloor, 8)$
\State \quad $x^{\mathrm{pad}}_0 \gets \max(0, x_0 - m)$,\; $x^{\mathrm{pad}}_1 \gets \min(W, x_1 + m)$
\State \quad $y^{\mathrm{pad}}_0 \gets \max(0, y_0 - m)$,\; $y^{\mathrm{pad}}_1 \gets \min(H, y_1 + m)$
\State \quad $\hat{I} \gets I[y^{\mathrm{pad}}_0:y^{\mathrm{pad}}_1,\;x^{\mathrm{pad}}_0:x^{\mathrm{pad}}_1]$;\; $\hat{G}\gets\textsc{Gray}(\hat{I})$
\State \textbf{else}
\State \quad $\hat{G}\gets C$;\; $x^{\mathrm{pad}}_0\gets 0$;\; $y^{\mathrm{pad}}_0\gets 0$
\State \textbf{end if}
\State \textbf{if} \texttt{invert\_mask} \textbf{then} $\hat{B}\gets\textsc{Otsu}(\hat{G},\textsc{BinaryInv})$ \textbf{else} $\hat{B}\gets\textsc{Otsu}(\hat{G},\textsc{Binary})$ \textbf{end if}
\State $\{(x_i,y_i,w_i,h_i,\mathrm{area}_i)\}_{i=1}^{N}\gets \textsc{CCStats}(\hat{B},8)$
\For{$i\gets 1$ \textbf{to} $N$}
  \If{$\mathrm{area}_i < \texttt{min\_component\_area}$} \textbf{continue} \EndIf
  \If{\texttt{global}}
    \State $X_1\gets x_i + x^{\mathrm{pad}}_0$,\; $Y_1\gets y_i + y^{\mathrm{pad}}_0$,\; $X_2\gets X_1 + w_i$,\; $Y_2\gets Y_1 + h_i$
    \State $\texttt{intersectsX}\gets (X_1 \le x_1)\land(X_2 \ge x_0)$,\; $\texttt{intersectsY}\gets (Y_1 \le y_1)\land(Y_2 \ge y_0)$
    \State $\texttt{fullyInside}\gets (X_1 \ge x_0{+}1)\land(X_2 \le x_1{-}1)\land(Y_1 \ge y_0{+}1)\land(Y_2 \le y_1{-}1)$
    \If{$\texttt{intersectsX}\land\texttt{intersectsY}\land \lnot \texttt{fullyInside}$} \Return \textbf{true} \EndIf
  \Else
    \State $X_1\gets x_i$,\; $Y_1\gets y_i$,\; $X_2\gets x_i{+}w_i$,\; $Y_2\gets y_i{+}h_i$
    \State $\texttt{touches}\gets (X_1 \le 1)\lor(Y_1 \le 1)\lor(X_2 \ge w{-}1)\lor(Y_2 \ge h{-}1)$
    \State $\texttt{inside}\gets (X_1 \ge 1)\land(Y_1 \ge 1)\land(X_2 \le w{-}1)\land(Y_2 \le h{-}1)$
    \If{$\texttt{touches}\land \lnot \texttt{inside}$} \Return \textbf{true} \EndIf
  \EndIf
\EndFor
\State \Return \textbf{false}
\end{algorithmic}
\end{algorithm*}

\begin{algorithm*}[t]
\caption{DataPreparation for $\mathcal{G}_\theta$}
\label{alg:data-prep}
\begin{algorithmic}[1]
\Require Image $I$, OCR-derived boxes $\mathcal{B}=\{(x_0^k,y_0^k,x_1^k,y_1^k)\}$ from \textsc{ExtractLineSegments}; per-bucket caps; target $N$ crops.
\State Initialize per-bucket counters for $y{=}1$ well-defined and $y{=}0$ ill-defined. Initialize counters for ill-defined normal and ill-defined manipulated.
\State Shuffle $\mathcal{B}$ and optionally subsample by width bins.
\For{\textbf{each} $b=(x_0,y_0,x_1,y_1)\in\mathcal{B}$}
  \State $g_{\mathrm{dark}} \gets \textsc{BorderIntegrityTest}(I,x_0,x_1,y_0,y_1,\texttt{global}{=}\textbf{true},\texttt{invert\_mask}{=}\textbf{true})$
  \State $g_{\mathrm{light}} \gets \textsc{BorderIntegrityTest}(I,x_0,x_1,y_0,y_1,\texttt{global}{=}\textbf{true},\texttt{invert\_mask}{=}\textbf{false})$
  \If{$g_{\mathrm{dark}}=\textbf{false}$ \textbf{or} $g_{\mathrm{light}}=\textbf{false}$}
     \State $y\gets 1$
     \State $\texttt{fg\_darker}\gets (g_{\mathrm{dark}}=\textbf{false})$
  \Else
     \If{ill-defined bucket not full and normal within balance} \State $y\gets 0$;\; increment ill-defined normal \Else \State \textbf{continue} \EndIf
  \EndIf

  \If{$y{=}1$ \textbf{and} ill-defined bucket not full}
     \State Choose a random subset of sides and small integer offsets $\Delta_s\ge 1$; choose per-side operation in \{pad, crop\}
     \State Build $\hat b=(\hat x_0,\hat y_0,\hat x_1,\hat y_1)$ by side-wise pad or crop; clip to image bounds; if degenerate, revert to $b$
     \State $g_{\mathrm{pert}} \gets \textsc{BorderIntegrityTest}(I,\hat x_0,\hat x_1,\hat y_0,\hat y_1,\texttt{global}{=}\textbf{true},\texttt{invert\_mask}{=}\texttt{fg\_darker})$
     \If{$g_{\mathrm{pert}}=\textbf{true}$}
       \State $y\gets 0$;\; $b\gets \hat b$
       \State $(x_0,y_0,x_1,y_1)\gets b$
       \State increment ill-defined manipulated
     \EndIf
  \EndIf

  \If{$y{=}0$}
    \State $C\gets I[y_0:y_1,x_0:x_1]$;\; $G\gets \textsc{Gray}(C)$
    \State $s_{\mathrm{dark}}\gets \textsc{BorderIntegrityTest}(G,\texttt{global}{=}\textbf{false},\texttt{invert\_mask}{=}\textbf{true})$
    \State $s_{\mathrm{light}}\gets \textsc{BorderIntegrityTest}(G,\texttt{global}{=}\textbf{false},\texttt{invert\_mask}{=}\textbf{false})$
    \If{\textbf{not} $(s_{\mathrm{dark}}=\textbf{true} \land s_{\mathrm{light}}=\textbf{true})$} \State \textbf{continue} \EndIf
  \EndIf

  \State If the per-bucket cap for $y$ is not exceeded, \textbf{store} $(b,y)$
  \If{stored crops reach target $N$} \textbf{break} \EndIf
\EndFor
\State \Return stored crops with labels
\end{algorithmic}
\end{algorithm*}

\subsection{Data Preparation}
\label{appendix:gprocess}

\subsubsection{Border Integrity Test}
\label{appendix:galgo}
\textsc{BorderIntegrityTest}, illustrated in \cref{alg:border-integrity-unified}, is a connected-components criterion on an Otsu-binarized image that flags border contact with a one-pixel tolerance when a foreground component touches the box boundary without being fully inside. The routine takes a boolean input \texttt{global}. When \texttt{global} is true the box is first expanded by a fixed margin and the same test is applied; when false the test is applied to the crop alone. For each original OCR-derived box, the global test is evaluated under both dark-text and light-text polarities. An original box is retained as naturally ill defined only when both global tests indicate border contact; otherwise, it is treated as a candidate well-defined box. To synthesize additional ill-defined examples, candidate well-defined boxes are perturbed and retested globally using the foreground polarity inferred from the original box. A perturbation is accepted when this test indicates border contact. Finally, every ill-defined example, whether naturally occurring or synthetically perturbed, is retained only when the local test indicates border contact under both polarities. Samples that fail these consistency checks are discarded.

\subsubsection{The Labeling Process}
\label{appendix:galll}
Given an image $I$, we obtain character boxes from OCR and group them into line segments using $\mathrm{ExtractLineSegments}$, which yields boxes $b=(x_0,y_0,x_1,y_1)$ with text metadata. Candidates are subsampled by width buckets to control size diversity. For each box we run \textsc{BorderIntegrityTest} with \texttt{global} true in both polarities. A box is labeled $y=1$ when at least one polarity indicates no border contact under the global test; otherwise it is labeled $y=0$, subject to per-bucket caps and simple balance constraints. To enrich ill-defined data when quotas are not met, some $y=1$ boxes are perturbed by padding or cropping their sides with small integer offsets, clipped to image bounds, and retested with the global setting under the inferred foreground polarity; if border contact is detected the perturbed box is accepted as $y=0$, otherwise it remains $y=1$ and the expansion or padding is discarded. As a final sanity check, every \(y=0\) crop is evaluated locally under both polarities and retained only when both tests indicate border contact. We note that the implementation stores the complementary class index \(1-y\), while the notation throughout the paper follows the quality-label convention defined in Sec. 3.2.

\subsubsection{Crop perturbation}
\label{appendix:gcroperturbation}

To enrich ill-defined examples, some $y=1$ crops are perturbed by padding or cropping the four sides by small integer offsets. Padding expands a side outward. Cropping moves it inward. For each active side $s\in\{\text{left},\text{right},\text{top},\text{bottom}\}$ we draw an integer offset $\Delta_s\in\{1,\dots,\Delta_{\max}\}$ with a distribution biased toward small values. Concretely:
\begin{itemize}
  \item \textbf{Range.} Let $h=y_1-y_0$ and $w=x_1-x_0$. We set $\Delta_{\max}=\min\!\bigl(20,\;\lfloor 0.3\cdot\max(h,w)\rfloor\bigr)$ to adapt the maximum step to the box scale while capping it in pixels.
  \item \textbf{Skew toward small moves.} We use a truncated geometric-like law $\Pr(\Delta_s=k)\propto \rho^{\,k-1}$ on $k\in\{1,\dots,\Delta_{\max}\}$ with $\rho\in(0,1)$, default $\rho=0.5$, which concentrates mass on $\{1,2,3\}$.
  \item \textbf{Coherent moves.} With probability $1/2$ we tie all active sides to the same offset value $\Delta$ and to the same operation, which produces more realistic, coherent expansions or contractions along multiple sides.
  \item \textbf{Nondegeneracy.} Before applying a crop operation on a side, we ensure the updated width and height remain positive by enforcing
  \[
  \hat w=\hat x_1-\hat x_0\ge 2,\qquad \hat h=\hat y_1-\hat y_0\ge 2.
  \]
  If a proposed $\Delta_s$ would violate these bounds, we reduce it to the largest value that preserves $\hat w\ge 2$ and $\hat h\ge 2$, or skip that side if no feasible value exists.
  \item \textbf{Clipping.} After the side-wise updates, the box is clipped to image bounds:
  \[
  \hat x_0\leftarrow\max(0,\hat x_0),\quad \hat y_0\leftarrow\max(0,\hat y_0),\] \[\quad
  \hat x_1\leftarrow\min(W,\hat x_1),\quad \hat y_1\leftarrow\min(H,\hat y_1).
  \]
\end{itemize}
These choices keep perturbations small and realistic, scale them with the content, and prevent degenerate boxes while allowing enough variability. The global test is rerun and only if border contact is detected the perturbed crop is accepted as ill defined with $y=0$. The process is illustrated in \cref{alg:data-prep}. 

\subsubsection{Augmentation.}
\label{appendix:gdataaug}
 With probability 0.5, we apply photometric augmentation to the crop and stripes. The transformation order is shared across the three inputs, while the application of individual transformations and their parameters are sampled independently for each input. The following operations are considered:
\begin{itemize}
    \item \textbf{Brightness/Contrast:} random brightness and contrast changes with limits \(0.4\) (applied with \(p=0.25\)).
    \item \textbf{Hue/Saturation/Value:} hue/saturation/value shifts up to \(\pm 40\) (applied with \(p=0.25\)).
    \item \textbf{RGB Channel Shifts:} per-channel shifts up to \(\pm 40\) (applied with \(p=0.25\)).
    \item \textbf{Color Jitter:} coupled brightness/contrast/saturation/hue jitter (same limits; \(p=0.25\)).
    \item \textbf{Channel Shuffle:} random permutation of RGB channels (\(p=0.2\)).
    \item \textbf{Text-Color Perturbation:} brighten low-luminance (foreground-like) pixels to vary ink color (threshold \(=40\); \(p=0.2\)).
    \item \textbf{Inversion:} full-color inversion (\(p=0.2\)).
    \item \textbf{Blur/Defocus (base):} Gaussian/median/box/motion blur with kernel limits \(3\!-\!7\) and defocus radius \(1\!-\!2\) (each \(p=0.15\)).
    \item \textbf{Blur/Defocus (for bad-defined):} a wider blur band with kernels up to \(31\) and defocus radius \(1\!-\!11\) to simulate stronger degradation.
    \item \textbf{Compression:} JPEG quality sampled in \([10,85]\) (\(p=0.2\)).
    \item \textbf{Sharpening:} \(\alpha\in[0.2,0.5]\), lightness \([0.5,1.0]\) (\(p=0.1\)).
    \item \textbf{Additive Noise:} Gaussian noise with \(\sigma\in[0.05,0.1]\) (\(p=0.1\)) and ISO noise with color shift \([0.01,0.02]\), intensity \([0.01,0.05]\) (\(p=0.1\)).
    \item \textbf{Multiplicative Noise:} per-channel multiplier in \([0.975,1.025]\) (\(p=0.1\)).
    \item \textbf{Occasional Grayscale:} convert to grayscale with small probability (\(p=0.01\)).
\end{itemize}

\subsection{More on the Crop Quality Assessment Network}
\label{appendix:evaluationcropqualityassesment}
\subsubsection{Comparison with algorithmic alternatives}
We compare the inference cost of our bounding-box quality assessment network against several algorithmic baselines that follow the procedure described in \cref{alg:border-integrity-unified}. The baselines rely on classical local-thresholding schemes, namely Niblack, adaptive mean, adaptive Gaussian, and Sauvola. As shown in \cref{tab:crop-quality-speed}, our network is substantially faster in both per-image latency and throughput.

\begin{table}[!htbp]
  \small
\caption{Inference speed comparison between the bounding-box quality assessment network and classical baseline algorithms. 
Lower is better for latency; higher is better for throughput. 
The slowdown ratio is defined as (classic / model) per-image time. The evaluation is done with a batch size of 1024.}
\label{tab:crop-quality-speed}
  \resizebox{\linewidth}{!}{%
  \setlength{\tabcolsep}{1pt}
\centering
\begin{tabular}{lccc}
\toprule
\textbf{Method} & \textbf{Per-image (ms)} &
\textbf{Throughput (img/s)} & \textbf{Slowdown ratio} \\
\midrule
bounding-box quality net (ours) 
& 0.159 $\pm$ 0.000 
& 6306.8 
& 1.00$\times$ \\
\midrule
Niblack
& 1.213 $\pm$ 0.018 
& 824.3 
& 7.65$\times$ \\
Adaptive mean
& 0.536 $\pm$ 0.007 
& 1866.4 
& 3.38$\times$ \\
Adaptive Gaussian 
& 0.638 $\pm$ 0.009 
& 1567.0 
& 4.02$\times$ \\
Sauvola
& 1.783 $\pm$ 0.021 
& 560.8 
& 11.25$\times$ \\
\bottomrule
\end{tabular}%
}
\end{table}

\subsubsection{Evaluating the Crop Quality Assessment Network}

We evaluate our bounding-box quality assessment network on all the DIBCO datasets, which provide human-annotated segmentation masks. These masks allow us to easily create both well-defined and ill-defined crops. For well-defined crops, we begin by selecting a random region and then iteratively shrink the bounding box in each direction until it touches a positive mask (or the box becomes empty and is discarded). For ill-defined crops, we simply choose a box randomly until we get one that cuts through a region labeled as positive. We do this 100 times per document. The results reported in \cref{tab:funsd-crop-quality} show that our bounding-box quality assessment network reliably distinguishes well-defined crops from corrupted ones. Combined with the improvements observed in our ablation experiments when using this model, these findings confirm that the network effectively fulfills its intended role.

\begin{table*}[!htbp]
  \small
  \setlength{\tabcolsep}{1pt}
\centering
\caption{bounding-box quality assessment classification results on data derived from the DIBCO datasets.}
\label{tab:funsd-crop-quality}
\begin{tabular}{lcc}
\toprule
\textbf{Method} & \textbf{Accuracy well-defined} & \textbf{Accuracy ill-defined} \\
\midrule
bounding-box quality assessment network 
& 98.6\% & 97.9 \%\\

\bottomrule
\end{tabular}
\end{table*}

\section{Previous data generation approaches}
\label{appendix:generationpreviouswork}
\subsection{FSTS synthesis pipeline.}

FSTS models text-image tampering using five representative manipulation types: copy-move, splicing, removal, insertion, and replacement. In the synthesis stage, the pipeline starts from a target image, extracts text-region coordinates using an OCR system, samples one tampering type together with its associated operation--parameter configuration and frequency weights from the modeled distribution, applies the corresponding scripted editing operations to the selected regions, merges the edited layers into a final tampered image, and finally generates the supervision mask by thresholding the difference between the tampered image and the original image. The synthesis process is decomposed into main processing and post-processing steps, and the reported frequencies guide operation selection during synthesis. The generation process follows a parent-to-child hierarchy: whenever several sub-operations share the same index, they are mutually exclusive alternatives whose frequencies indicate the preferred variant to select. In addition, all post-processing frequencies are scaled by a factor of $0.3$ before synthesis to avoid unrealistically complex samples.

For \emph{copy-move}, the synthesis process is organized into region sampling, geometric transformation, and visual trace concealment. It selects $1$--$12$ text regions, randomly chooses a source text region from the same image, retains $1$--$20$ characters from that source region, and chooses a target text region in the same image, with the generation rules additionally mentioning a nearby copy area in $9$-grid positions. The geometric transformation stage then uses one of two mutually exclusive extraction variants: either a Magic-Wand-like extraction with tolerance $1$--$50$, contiguous yes/no, and anti-alias yes/no, used with frequency $18.53\%$, or a channel-and-level adjustment for background removal and text-shape extraction using the red channel, input levels $130$--$237$, and output levels $0$--$255$, used with frequency $23.74\%$. This is followed by adaptive scaling to match the paste region ($73.50\%$) and region rotation of $0^\circ$--$5^\circ$ ($13.33\%$). The post-processing candidates are sharpen with amount $100$--$200\%$, radius $1$--$4$ pixels, and threshold $7$--$12$ levels ($8.90\%$); blur filter with default parameters ($5.71\%$); blur-more filter with default parameters ($3.74\%$); mean filter with default parameters ($5.50\%$); Gaussian blur with radius $0.1$--$3$ pixels ($12.70\%$); motion blur with angle $-15^\circ$--$15^\circ$ and radius $1$--$9$ pixels ($7.10\%$); radial blur with method spin/zoom and quality best/draft/good ($3.09\%$); smart blur with radius $0.1$--$10$ pixels, threshold $0.1$--$10$ levels, blur quality high/medium/low, and blur mode edge preservation/normal/stroke enhancement ($8.78\%$); custom convolution filtering with kernel values in $[-10,10]$, scale $1$--$20$, and offset $-5$--$5$ ($8.70\%$); color balance on midtones with slider values $-100$--$100$ ($4.18\%$); and color curves with raise-highlights or lower-shadows variants ($8.53\%$).

For \emph{splicing}, the process is again decomposed into region sampling, geometric transformation, and visual trace concealment. It selects $1$--$12$ text regions, randomly chooses a source text region from a different source image, retains $1$--$20$ characters, and pastes into a target text region in the target image. The geometric transformation stage again uses one of two mutually exclusive extraction variants: either tolerance-based text-shape extraction with tolerance $1$--$50$, contiguous yes/no, and anti-alias yes/no ($12.69\%$), or channel-and-level adjustment with red channel, input levels $130$--$237$, and output levels $0$--$255$ ($17.94\%$). This is followed by adaptive scaling to match the paste region ($78.00\%$) and region rotation of $0^\circ$--$5^\circ$ ($19.30\%$). The listed post-processing operations are sharpen with amount $100$--$200\%$, radius $1$--$4$ pixels, and threshold $7$--$12$ levels ($10.04\%$); Gaussian blur with radius $0.1$--$3$ pixels ($18.70\%$); lens blur with depth-map mode none, invert disabled, aperture shape, aperture radius $0$--$1$, blade curvature $0$--$1$, rotation angle $0^\circ$--$6^\circ$, brightness $100\%$, threshold $0$--$100\%$, amount $0$--$25\%$, and Gaussian/uniform distribution ($7.03\%$); motion blur with angle $-15^\circ$--$15^\circ$ and radius $1$--$9$ pixels ($5.40\%$); radial blur with method spin/zoom and quality best/draft/good ($3.11\%$); smart blur with radius $0.1$--$10$ pixels, threshold $0.1$--$10$ levels, blur quality high/medium/low, and blur mode edge preservation/stroke enhancement/normal ($3.90\%$); custom convolution filtering with kernel values in $[-10,10]$, scale $1$--$20$, and offset $-5$--$5$ ($2.70\%$); color curves with raise-highlights or lower-shadows variants ($17.45\%$); stroke with size $1$--$5$ pixels, position inside/center/outside, blend mode normal/multiply, opacity $50$--$100\%$, fill type color, and RGB values in $[0,255]^3$ ($8.75\%$); drop shadow with blend mode normal/multiply/darken, RGB values in $[0,255]^3$, opacity $5$--$23\%$, angle $-30^\circ$--$30^\circ$, distance $1$--$7$ pixels, spread $3$--$12\%$, size $1$--$17$ pixels, and noise $1$--$10\%$ ($6.99\%$); and hue/saturation adjustment with hue $-30$--$30$, saturation $-20$--$20$, and lightness $-30$--$30$ ($10.49\%$).

For \emph{removal}, the process is organized into region sampling, content removal, and geometric transformation. It selects $1$--$12$ text regions and a tampered text length of $1$--$20$ characters. The content-removal stage then uses one of several mutually exclusive operations: content-aware fill with $1$--$5$ iterations ($55.82\%$), solid-color fill with RGB values in $[0,255]^3$ ($9.76\%$), pure background cloning with normal blending mode ($11.52\%$), clone stamp with mode normal, opacity $100\%$, and flow $100\%$ ($10.45\%$), or healing brush with mode normal/replace and sampled source ($12.45\%$). The geometric-transformation stage then applies adaptive region scaling of $\pm5\%$ relative to the text region ($88.00\%$) and region rotation of $-5^\circ$--$5^\circ$ ($0.68\%$).

For \emph{insertion}, the synthesis process is organized into region sampling, text insertion, geometric transformation, and visual trace concealment. It selects $1$--$12$ non-text regions and a text length of $1$--$20$ characters. The insertion stage uses one of the listed fonts, namely Times New Roman, SimSun, KaiTi, Microsoft YaHei, or SimHei, together with anti-aliasing. For color adaptation, there are two alternatives: sampling the same color as the original text ($86.90\%$), or using a safety-color rule ($13.10\%$), where light backgrounds use RGB$(0$--$64,0$--$64,0$--$64)$ and dark backgrounds use RGB$(192$--$255,192$--$255,192$--$255)$. The geometric stage then applies adaptive scaling of $\pm5\%$ relative to the text region ($77.00\%$) and rotation of $-5^\circ$--$5^\circ$ ($12.03\%$). The post-processing candidates are sharpen with $1$--$5$ iterations, strength $400$--$500\%$, radius $50$--$60$ pixels, and threshold $2$--$3$ levels ($12.73\%$); Gaussian blur with radius $0.5$--$1.2$ pixels ($16.80\%$); outer glow with color RGB$(83,79,79)$, opacity $17\%$, noise $35$--$45\%$, and spread $5$--$8$ pixels ($7.51\%$); additive noise with amount $0.10\%$--$35\%$, Gaussian/uniform distribution, and monochromatic yes/no ($16.54\%$); stroke with size $1$--$5$ pixels, position inside/center/outside, blend mode normal/multiply, and RGB color in $[0,255]^3$ ($15.33\%$); and drop shadow with blend mode normal/multiply/darken, RGB color in $[0,255]^3$, opacity $5$--$23\%$, angle $-30^\circ$--$30^\circ$, distance $1$--$7$ pixels, spread $3$--$12\%$, size $1$--$17$ pixels, and noise $1$--$10\%$ ($5.26\%$).

For \emph{replacement}, the process is organized into region sampling, content removal, text insertion, geometric transformation, and visual trace concealment. The region-sampling stage uses $1$--$12$ text regions and a tampered text length of $1$--$20$ characters. The content-removal stage uses one of five mutually exclusive alternatives: content-aware fill with $1$--$5$ iterations ($61.70\%$), solid-color fill with RGB values in $[0,255]^3$ ($9.60\%$), pure background cloning with normal blending mode ($9.50\%$), clone stamp with mode normal, opacity $100\%$, and flow $100\%$ ($10.40\%$), or healing brush with mode normal/replace and sampled source ($8.80\%$). The text-insertion stage uses the same font set as insertion, namely Times New Roman, SimSun, KaiTi, Microsoft YaHei, and SimHei, together with anti-aliasing, and two color-selection alternatives: same as the original text color ($88.40\%$) or the same safety-color rule as insertion ($11.60\%$). The geometric stage then applies adaptive scaling of $\pm5\%$ relative to the text region ($43.50\%$) and rotation of $-5^\circ$--$5^\circ$ ($33.33\%$). The post-processing candidates are sharpen with $1$--$5$ iterations, strength $400$--$500\%$, radius $50$--$60$ pixels, and threshold $2$--$3$ levels ($12.69\%$); Gaussian blur with radius $0.5$--$1.2$ pixels ($11.91\%$); surface blur with radius $1$--$15$ pixels and threshold $5$--$25$ levels ($7.60\%$); motion blur with angle $-30^\circ$--$30^\circ$ and distance $1$--$20$ pixels ($7.63\%$); outer glow with color RGB$(83,79,79)$, opacity $17\%$, and noise $35$--$45\%$ ($13.68\%$); additive noise with amount $0.10\%$--$35\%$, Gaussian/uniform distribution, and monochromatic yes/no ($10.47\%$); stroke with size $1$--$5$ pixels, position inside/center/outside, blend mode normal/multiply, opacity $50$--$100\%$, fill type color, and RGB values in $[0,255]^3$ ($10.20\%$); and drop shadow with blend mode normal/multiply/darken, RGB values in $[0,255]^3$, opacity $5$--$23\%$, angle $-30^\circ$--$30^\circ$, distance $1$--$7$ pixels, spread $3$--$12\%$, size $1$--$17$ pixels, and noise $1$--$10\%$ ($8.81\%$).

We note that there is no open-source code for this generation approach, and we therefore implement it by strictly following the paper. Some implementation details remain unspecified, including the method used to detect non-text regions for insertion, the rule used to generate inserted or replacement text strings, and the exact heuristic used to match the original text color and to determine whether a background is light or dark. In our reproduction, we therefore follow exactly all details stated above, and for every remaining unspecified component, we use the same heuristics as in our own data generation pipeline for direct comparability. To determine whether a background is light or dark, we use the Sauvola algorithm.
\subsection{DocTamper generation pipeline.}

Starting from an authentic document image, \citet{Doctamper} first extracts OCR text boxes, including both word-level and character-level boxes. A Sauvola-based binarization is then applied to obtain a stroke-level foreground mask and to separate text foreground from background. For each text instance, a coarse font/style descriptor is recorded using its size and appearance, namely its height, width, foreground color, and background color. Concretely, the foreground descriptor is computed from statistics inside the stroke-level mask, while the background descriptor is computed from statistics inside the text box. Text instances with similar recorded font information are then grouped together.

Tampering is synthesized with two described branches. In the selective copy-paste branch, a source text region and a target text region are selected from the same compatible group, the source crop is resized to the target box when needed, pasted into the target location, and then post-processed. This branch operates by randomly swapping text positions within each group, followed by random post-processing. In the selective generation branch, the original text in the target region is erased, a similar TTF font is selected from the recorded font information, and new text is rendered into the erased region, followed by post-processing. The generation process is illustrated as selecting a digital font with the same height and foreground color as the target text, erasing the original content, printing the new text, and then post-processing the result. These operations are used to synthesize the tampered document image.

Some low-level implementation details can be partially recovered from the released toy script. In that simplified script, OCR boxes are stored as $(x,y,w,h)$, a binary text mask is loaded for each image, and foreground/background statistics are computed from masked and non-masked pixels inside each target box. Candidate matches are filtered using approximate width and height tolerances of $[0.9,1.1]\times$ together with foreground/background mean and standard-deviation tolerances of $\pm 20$ and $\pm 4$. Once a candidate is accepted, the source crop is resized to the target box and pasted into it.

However, since the released code is only a toy version, several implementation details still remain unavailable. These include the exact OCR configuration, the probabilities used to choose synthesis branches or source/target boxes, the number of edits inserted into each synthesized image, the exact erase and post-processing functions, the font library and font-selection rule and the replacement-text sampling rule. In our reproduction, we therefore follow exactly all generation details described above, and for every remaining missing detail we use the same values, rules, and functions as in our own data generation pipeline for comparability.

\subsection{Generation method for TextTamper}
The generation pipeline in \citet{newdataandopenopen} consists of two stages. First, a plain forgery is generated by selecting a source image $S$ and a target image $T$, extracting their binary text masks $M_s$ and $M_t$ with a text segmentation method, randomly cropping a textual region from the source image, and pasting it onto a non-text region of the target image. This is formulated as
\[
P = P(R(S,M_s),\,T,\,1-M_t),
\]
where $R(\cdot,\cdot)$ randomly crops a textual region from the source image and $P(\cdot,\cdot,\cdot)$ pastes it onto the target image. The resulting plain forgery $P$ is then passed through a distortion function
\[
I = D(P),
\]
which randomly applies either an unmalicious distortion or a malicious post-processing operation. The unmalicious distortions are JPEG compression, screen recapture, and blurring/noising. The malicious post-processing operation is image blending, for which the two methods used are Poisson blending and deep image blending. The final tampering mask is obtained from the difference between the distorted image $I$ and the target image $T$, followed by binarization and dilation, with
\[
M = B(T-I),
\]
where $B(x)=0$ if $x=0$ and $B(x)=1$ otherwise.

Some information is missing from the paper: the exact constraints used to select a valid non-text paste location in $T$, the sampling probabilities of the different distortion branches, the parameter ranges of JPEG compression, blurring, and noising. In addition, the data-generation code is not publicly released. In our reproduction, we follow exactly the details specified above, and for every unspecified component, we use the same values, rules, and functions as in our own data generation pipeline for comparability when possible.

\subsection{Generator-based datasets}
\label{appendix:generator-based-datasets}

To compare our training data against recent generative text-editing approaches, we construct a training set by combining samples from DanceText~\cite{qu2026detect}, ScaleEdit-12M~\cite{chen2026scaleedit}, AIForge-Doc~\cite{wu2026aiforgedoc,aiforgedoc2v2026}, FantasyID~\cite{Korshunov_IJCB-2_2025}, and VDE-Bench~\cite{yi2026vdebench}. After filtering, the combined dataset contains 972,139 images: 346,926 tampered and 125,978 authentic images from DanceText; 381,697 poster and 107,322 GUI images from ScaleEdit-12M; 4,061 and 3,066 images from AIForge-Doc v1 and v2, respectively; 2,201 images from FantasyID; and 888 images from VDE-Bench.

We retain full images containing localized text manipulations produced automatically by generative editing methods, including documents, posters, graphical user interfaces, and identification documents. We use the localization masks provided by the dataset authors when available; otherwise, masks are obtained from differences between aligned original and edited images. All samples are used at the full-image level rather than as isolated text crops.

\section{Examples of Generated Manipulations}
\label{appendix:exemplesgeenration}
\begin{figure*}[!tp]
  \centering
  \begin{tabular}{cccc}
    \includegraphics[width=0.203\textwidth]{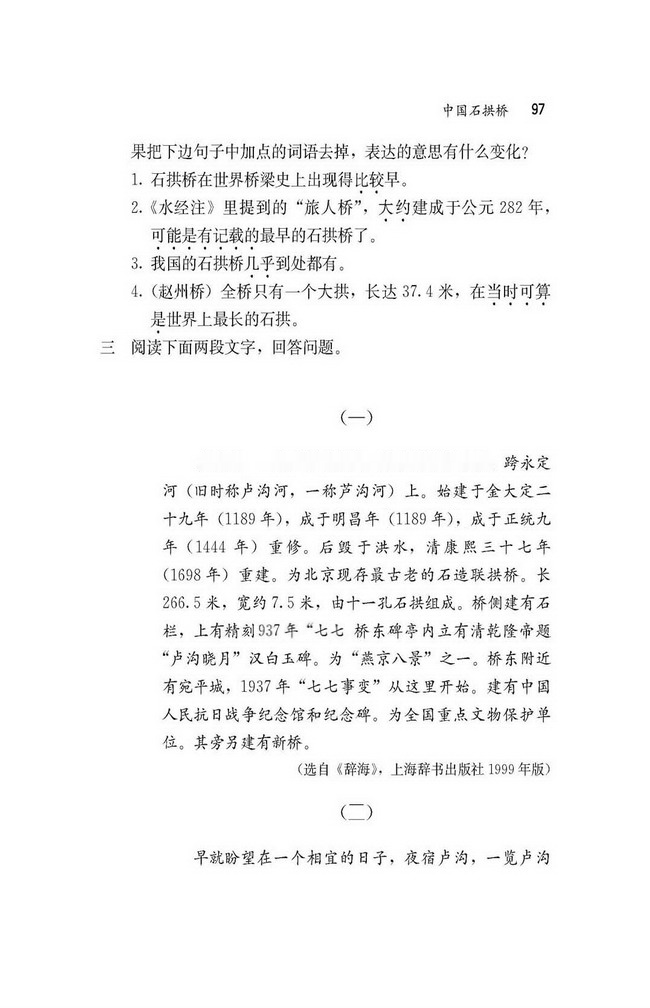} &
    \includegraphics[width=0.203\textwidth]{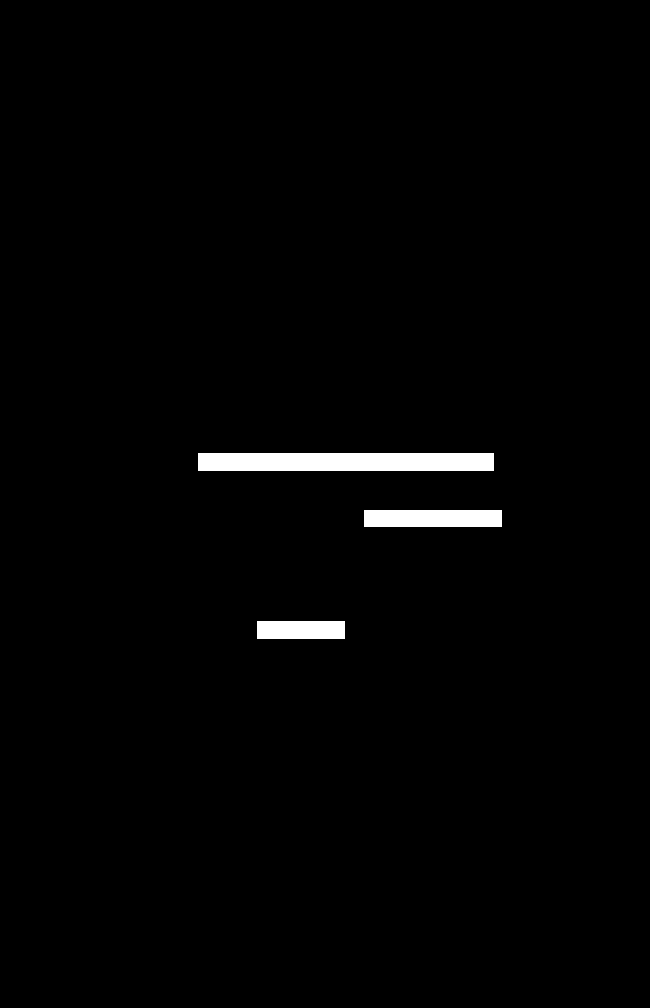} &
    \includegraphics[width=0.203\textwidth]{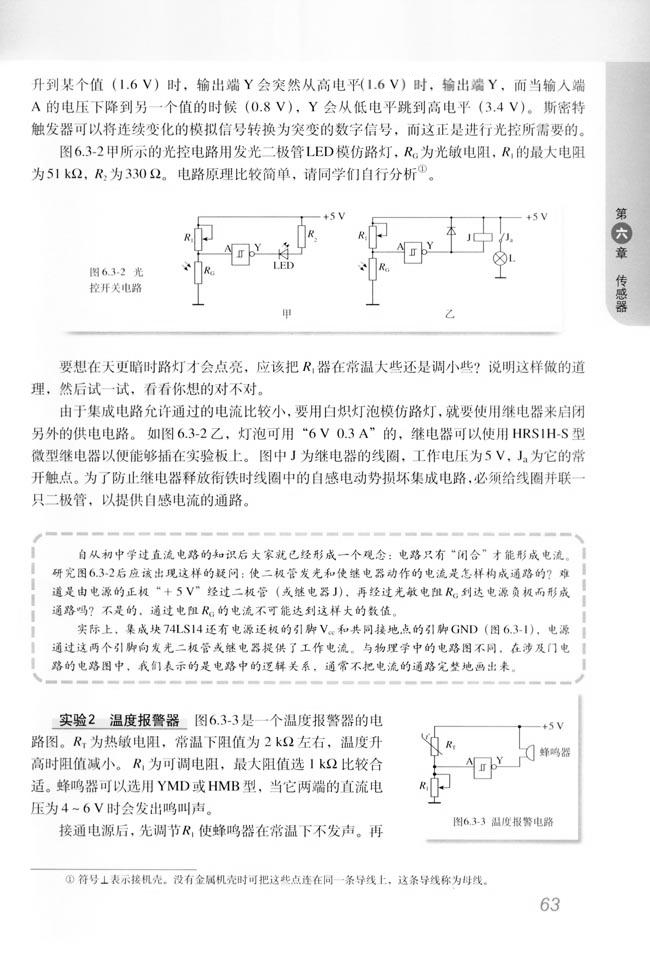} &
    \includegraphics[width=0.203\textwidth]{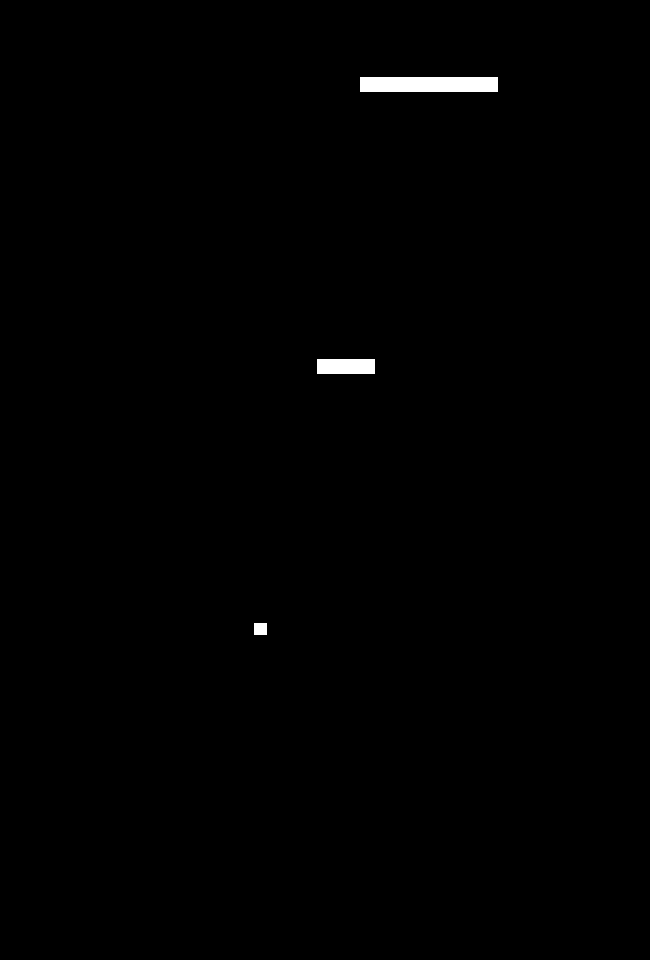} \\
    \includegraphics[width=0.203\textwidth]{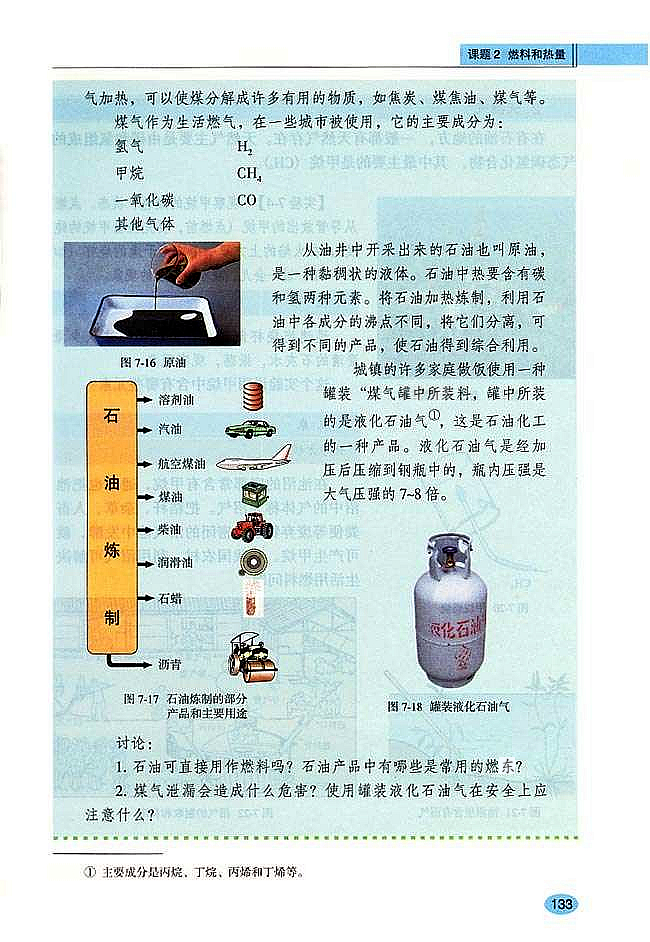} &
    \includegraphics[width=0.203\textwidth]{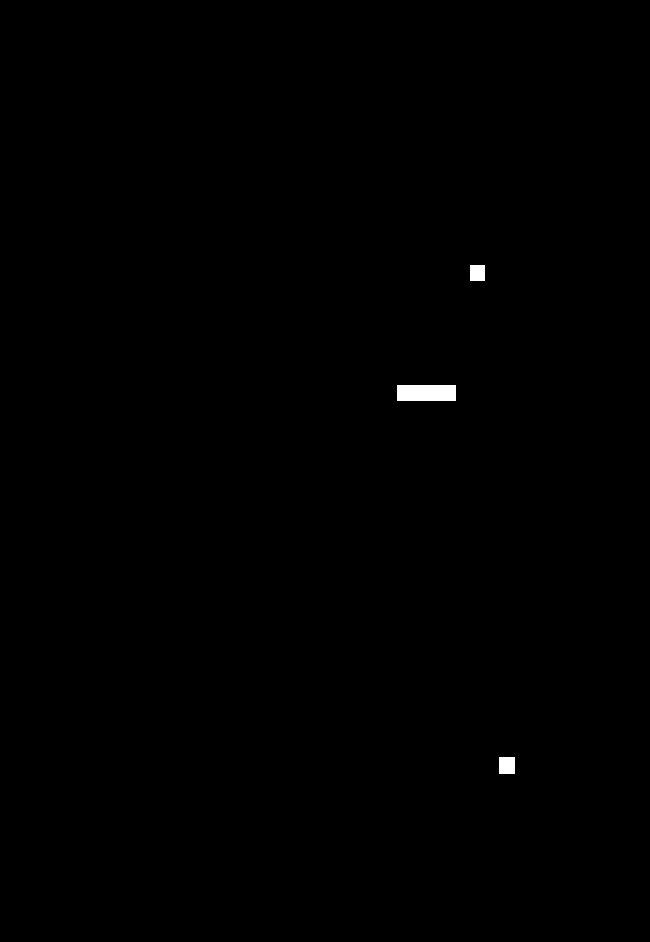} &
    \includegraphics[width=0.203\textwidth]{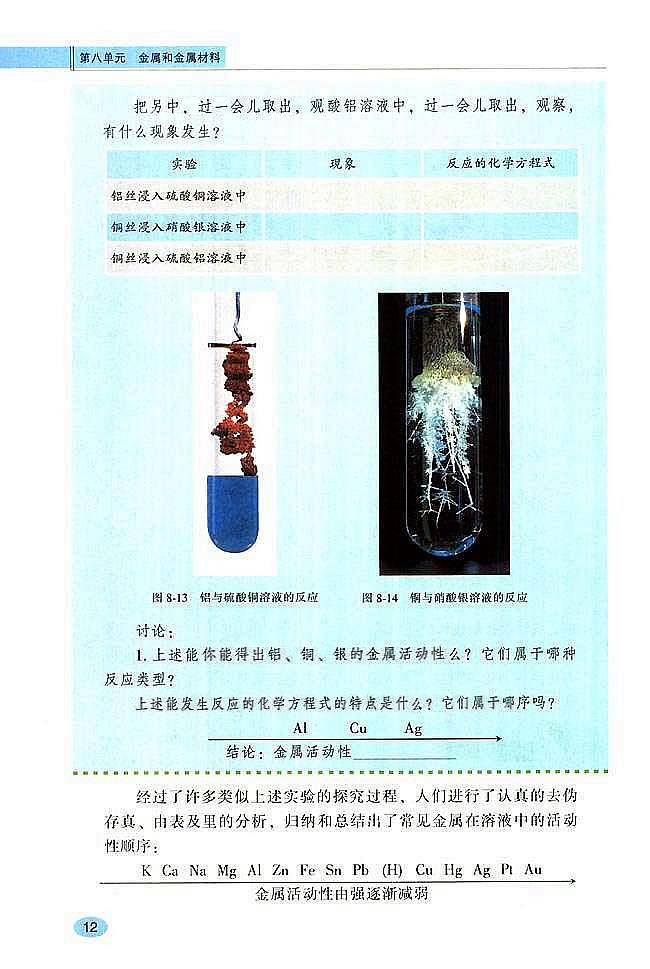} &
    \includegraphics[width=0.203\textwidth]{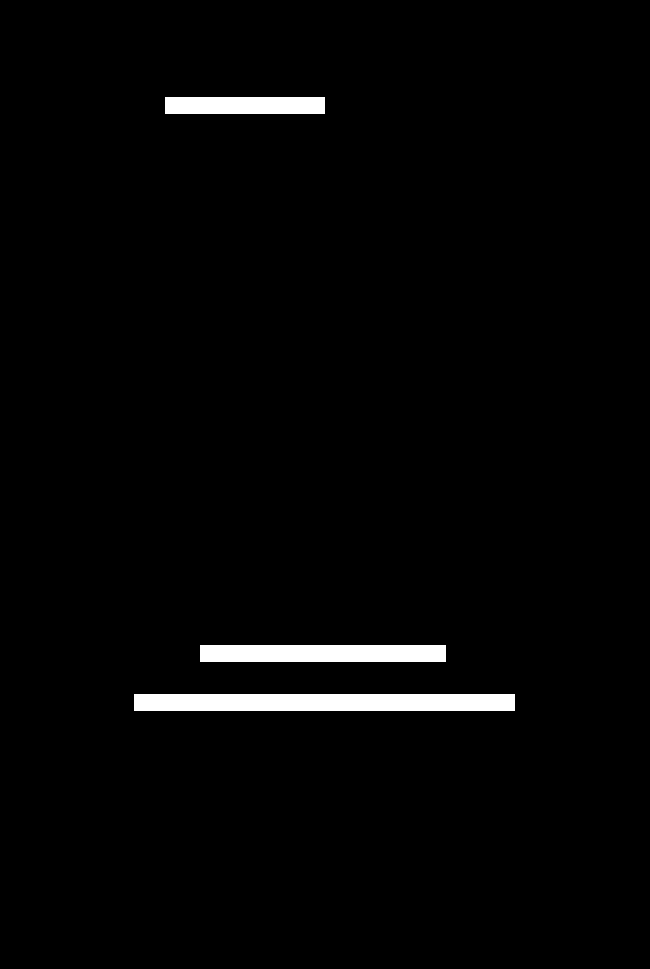} \\
    \includegraphics[width=0.203\textwidth]{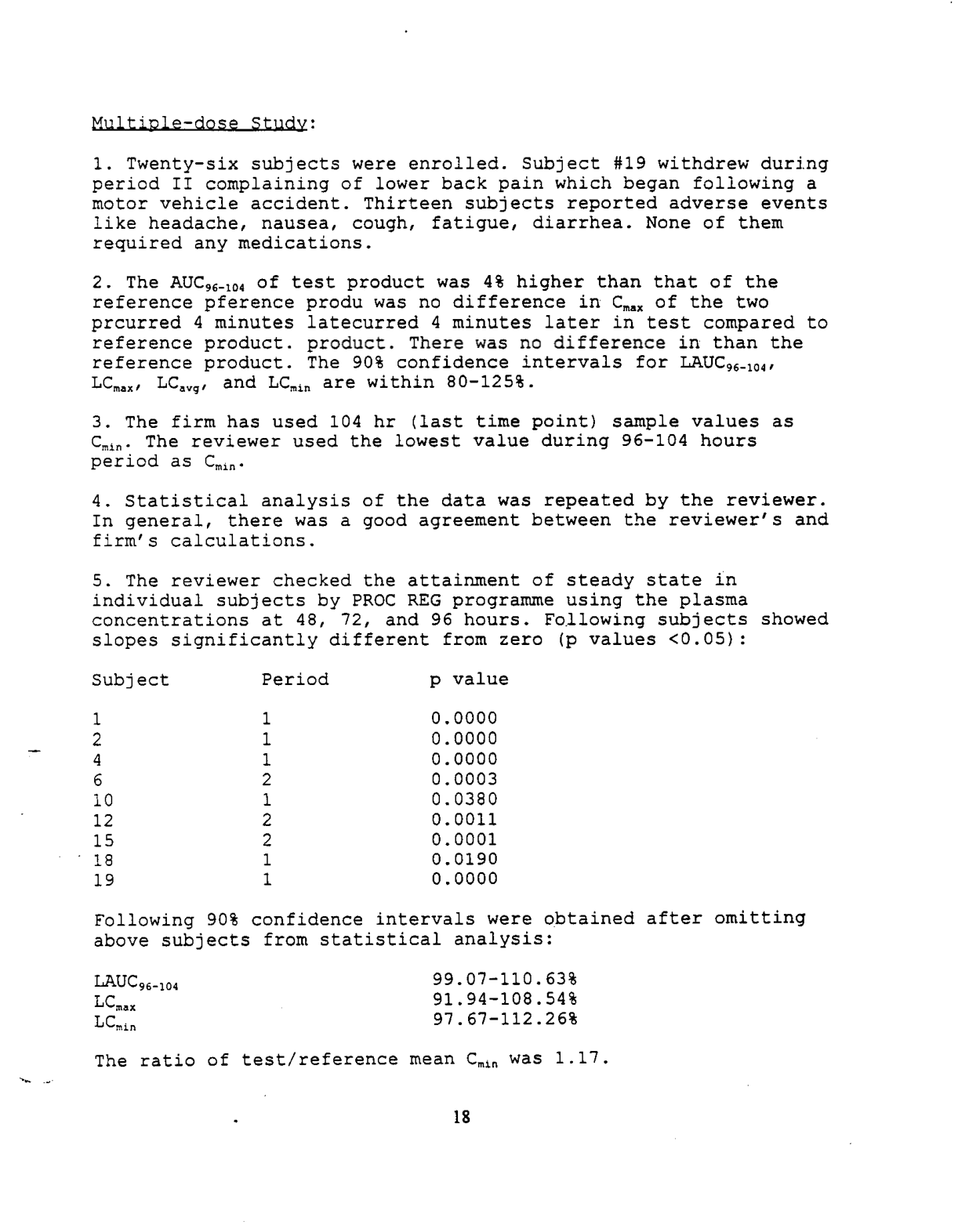} &
    \includegraphics[width=0.203\textwidth]{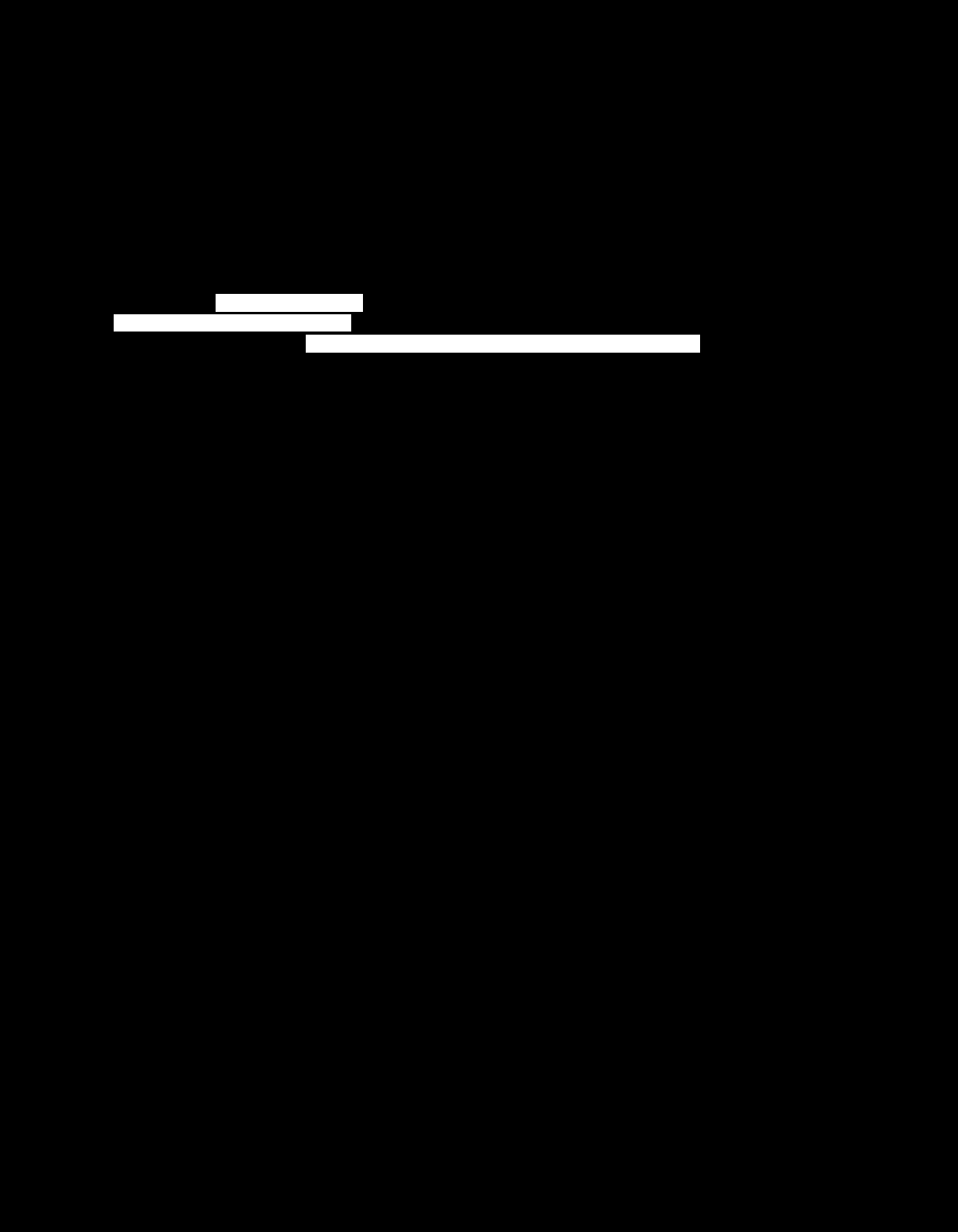} &
    \includegraphics[width=0.203\textwidth]{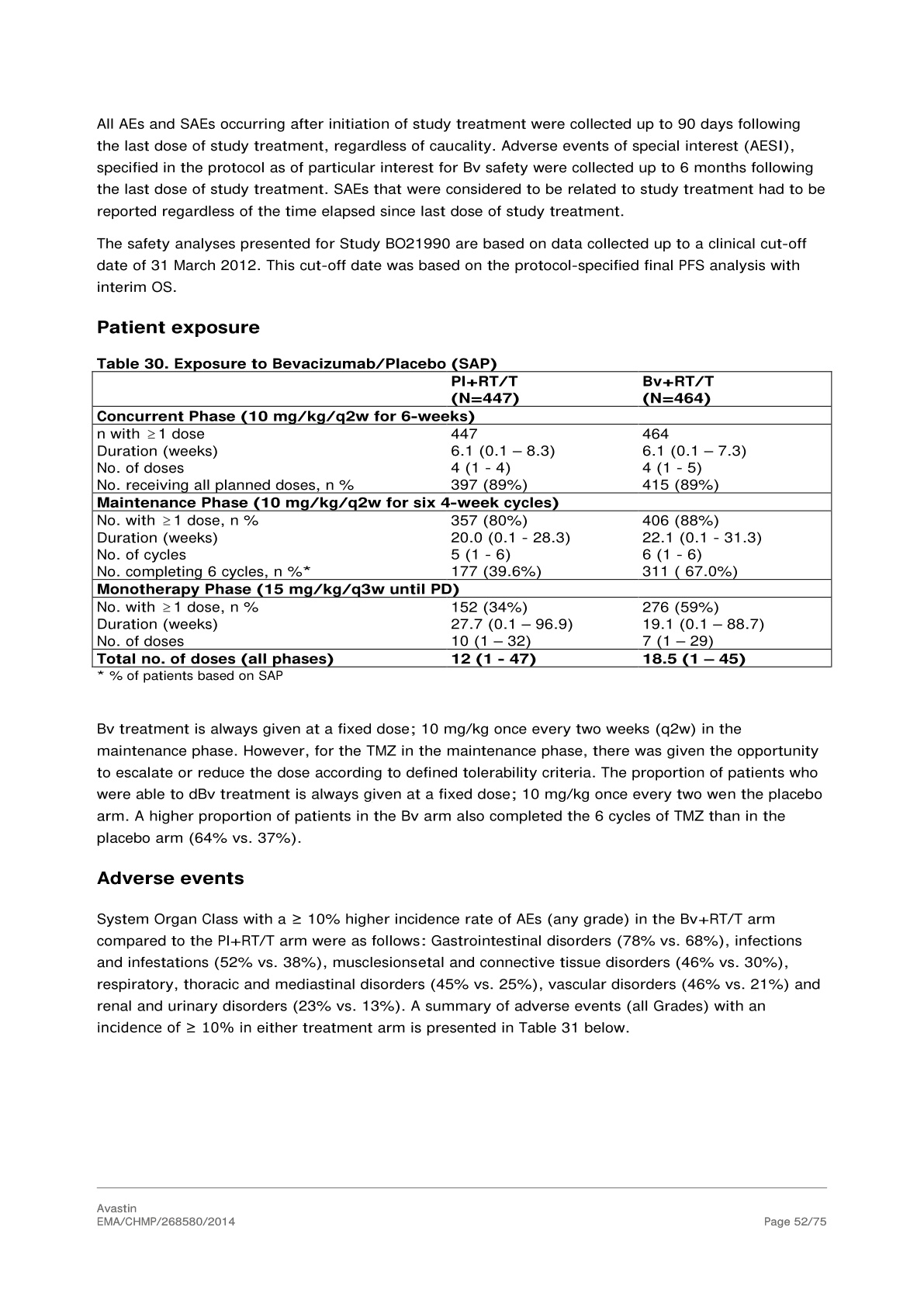} &
    \includegraphics[width=0.203\textwidth]{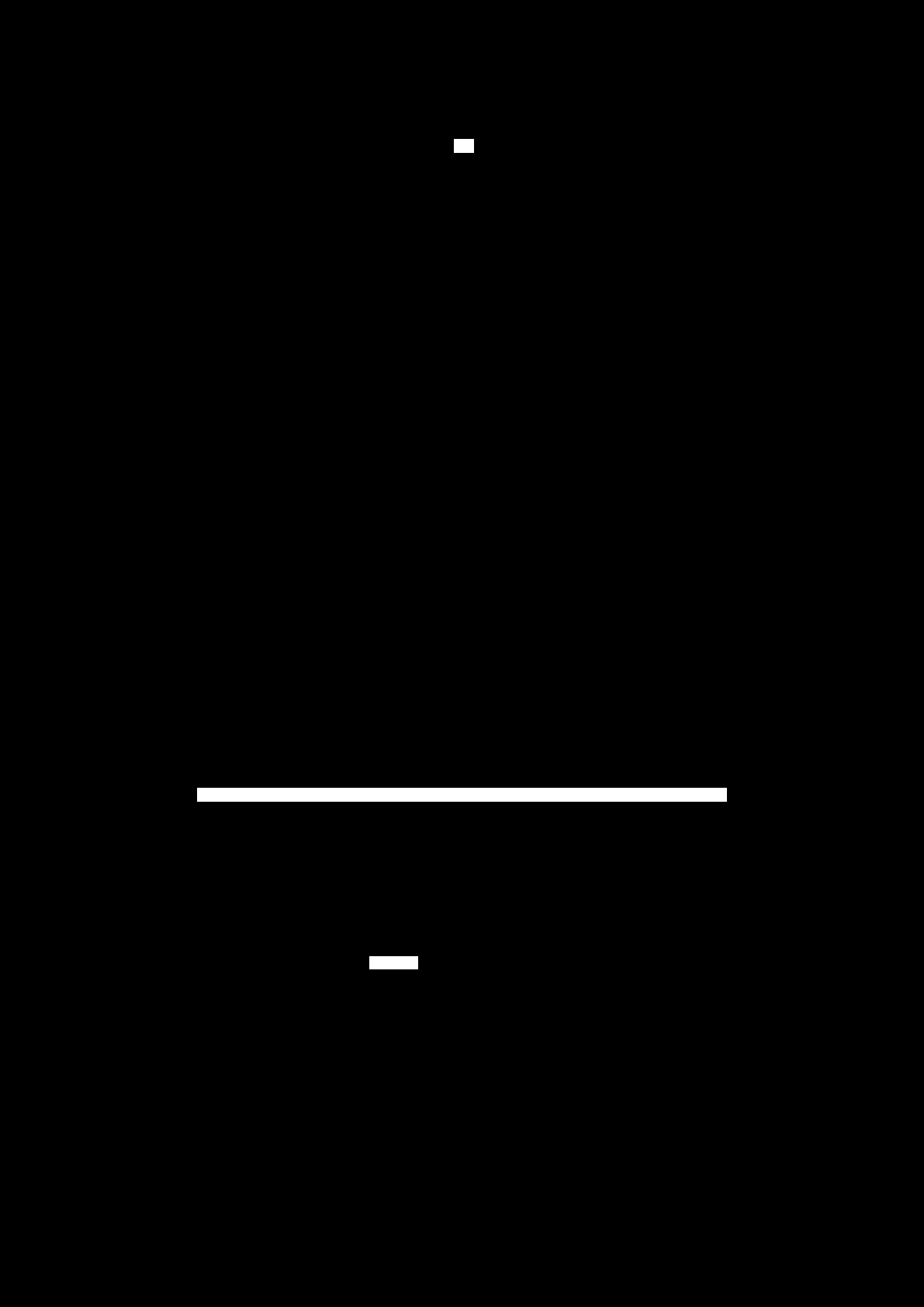} \\
    \includegraphics[width=0.203\textwidth]{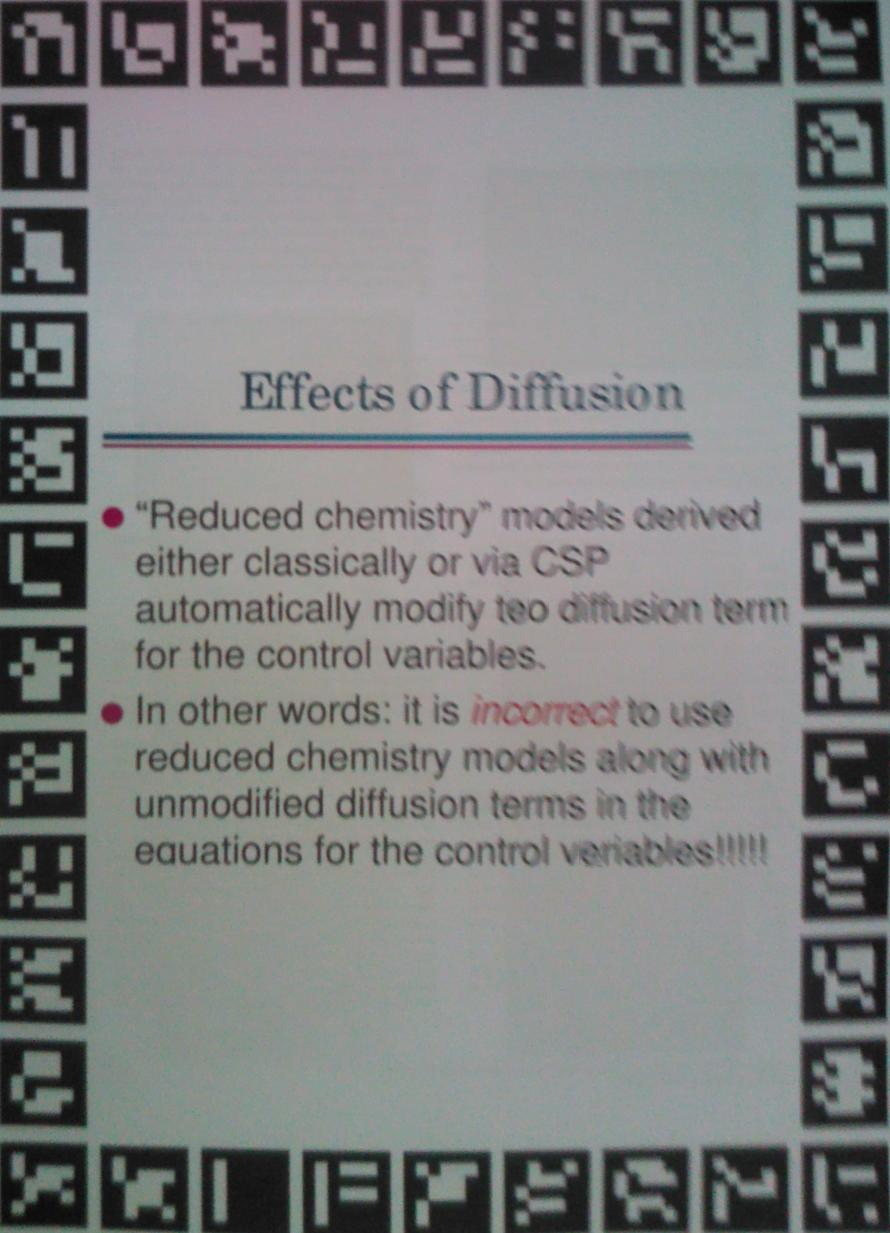} &
    \includegraphics[width=0.203\textwidth]{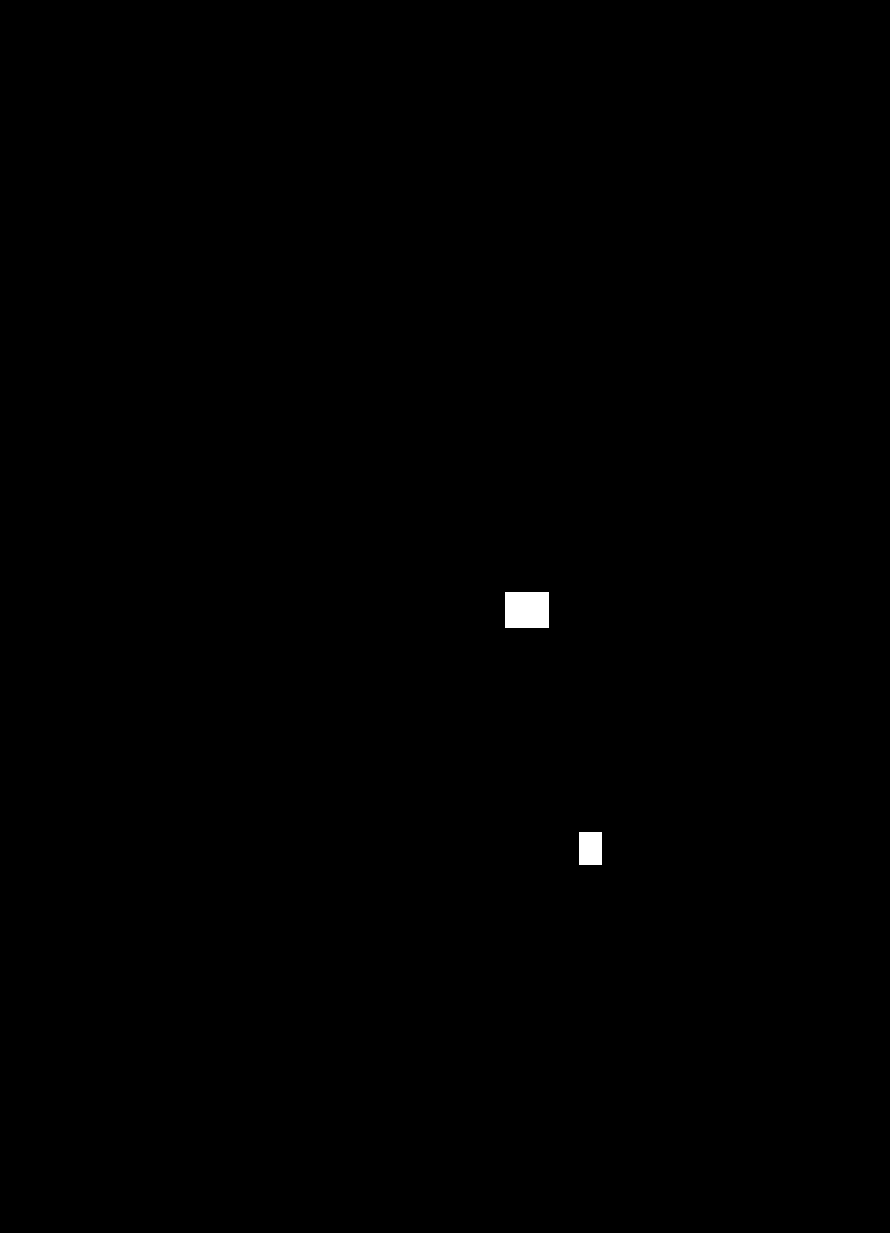} &
    \includegraphics[width=0.203\textwidth]{exemple_generated/image_7.png} &
    \includegraphics[width=0.203\textwidth]{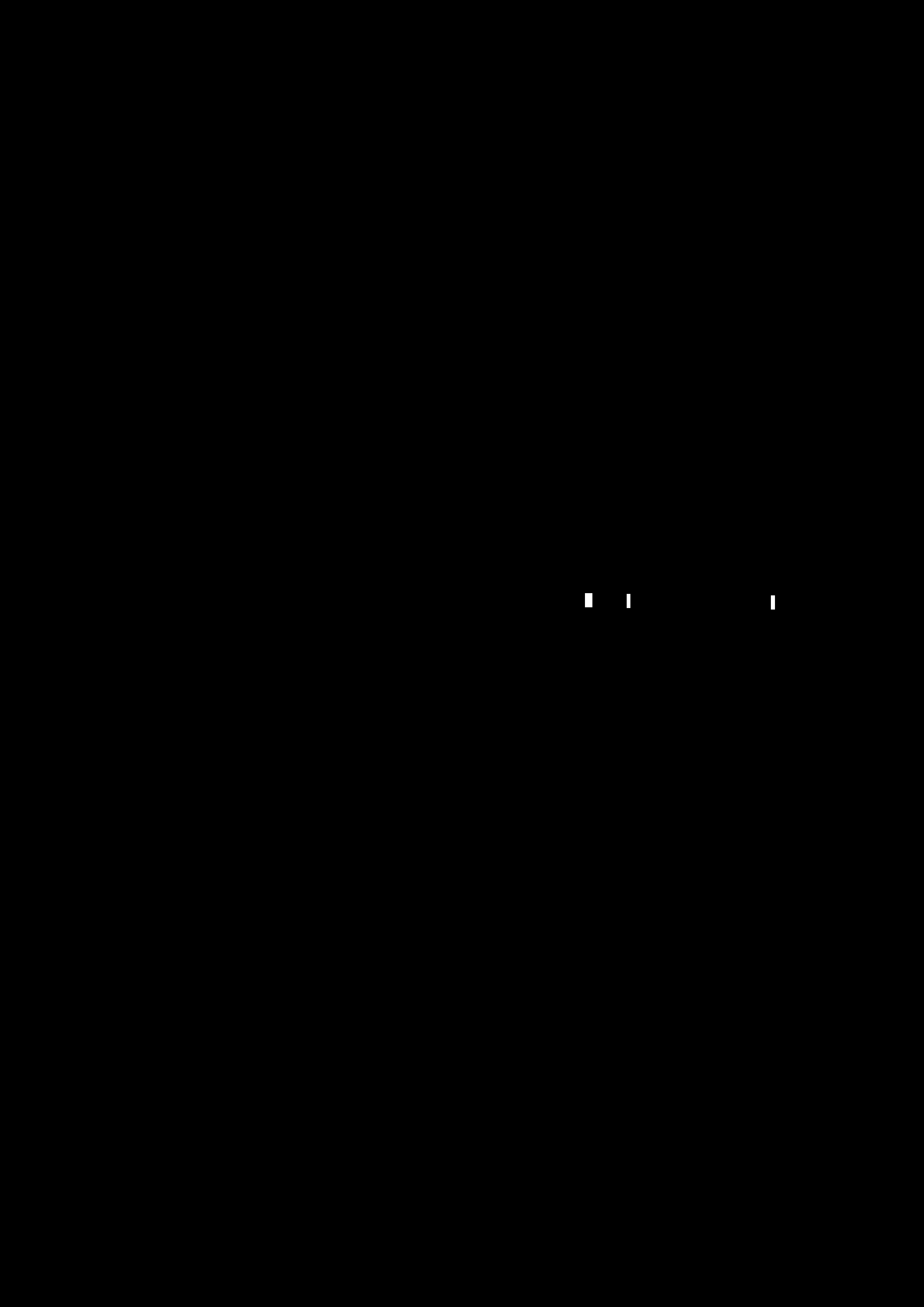} \\
  \end{tabular}
  \caption{Examples of generated tampered documents and their corresponding masks.}
  \label{fig:appendix:examples_0_7}
\end{figure*}

\begin{figure*}[!tp]
  \centering
  \begin{tabular}{cccc}
    \includegraphics[width=0.203\textwidth]{exemple_generated/image_8.png} &
    \includegraphics[width=0.203\textwidth]{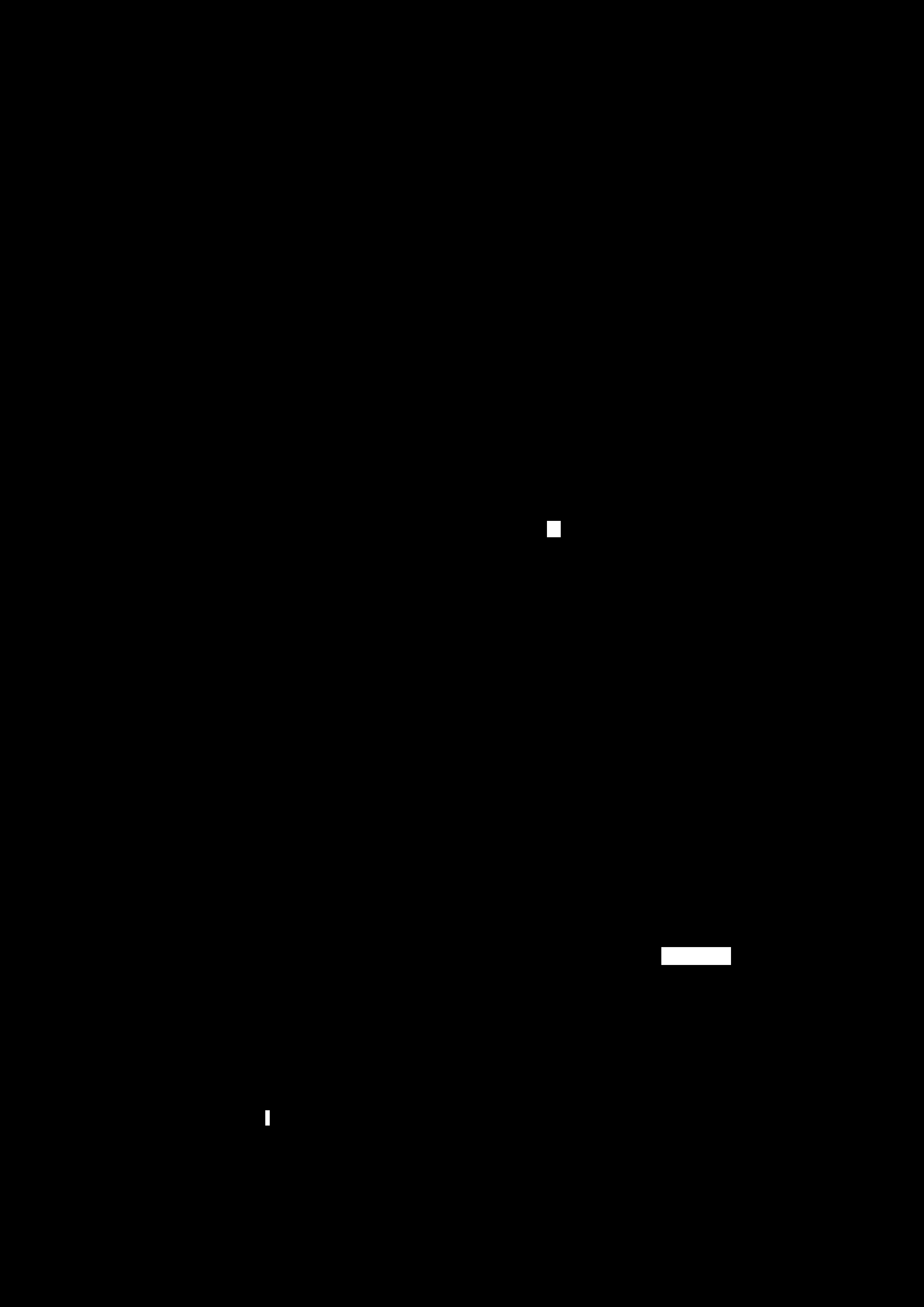} &
    \includegraphics[width=0.203\textwidth]{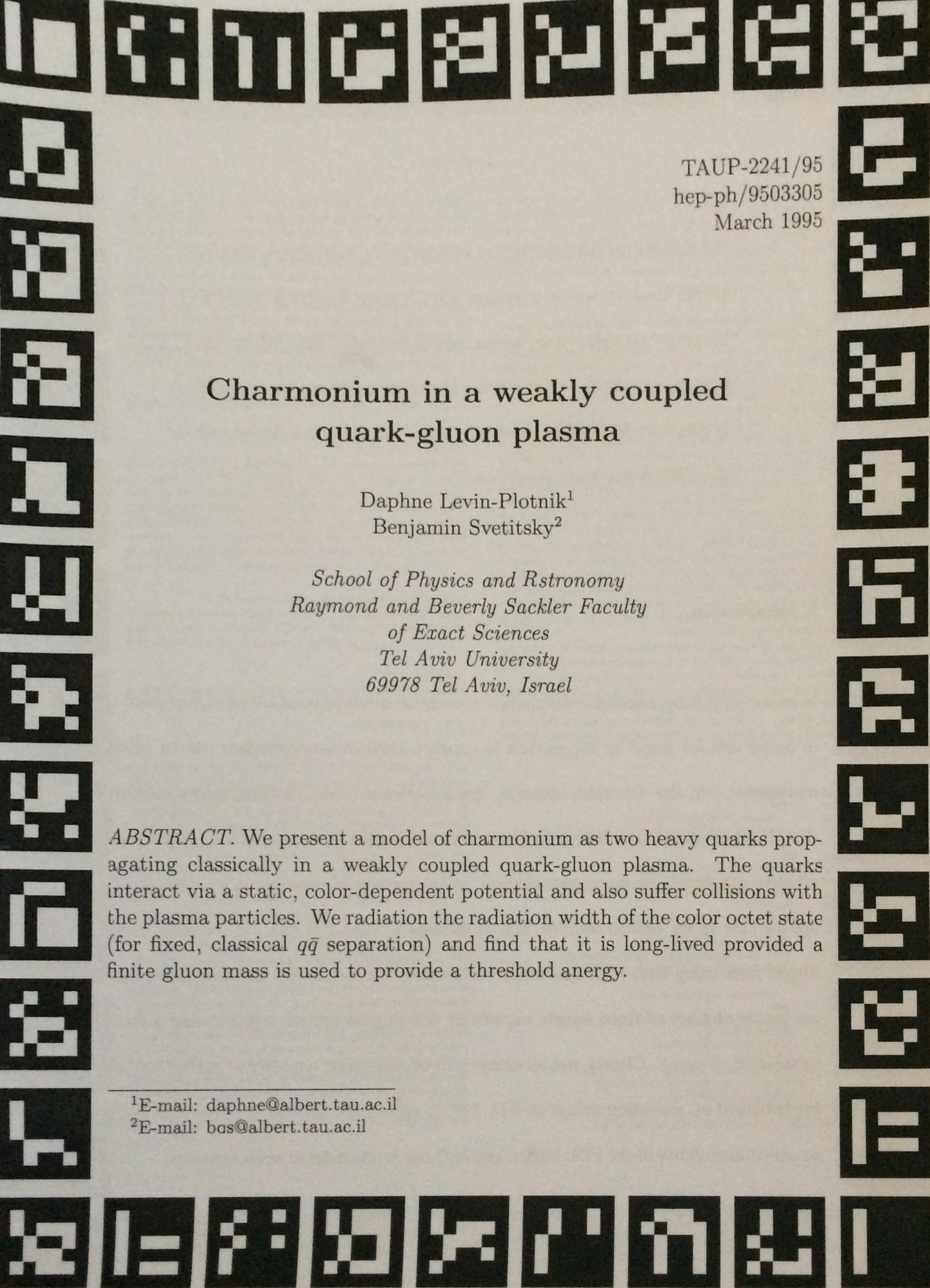} &
    \includegraphics[width=0.203\textwidth]{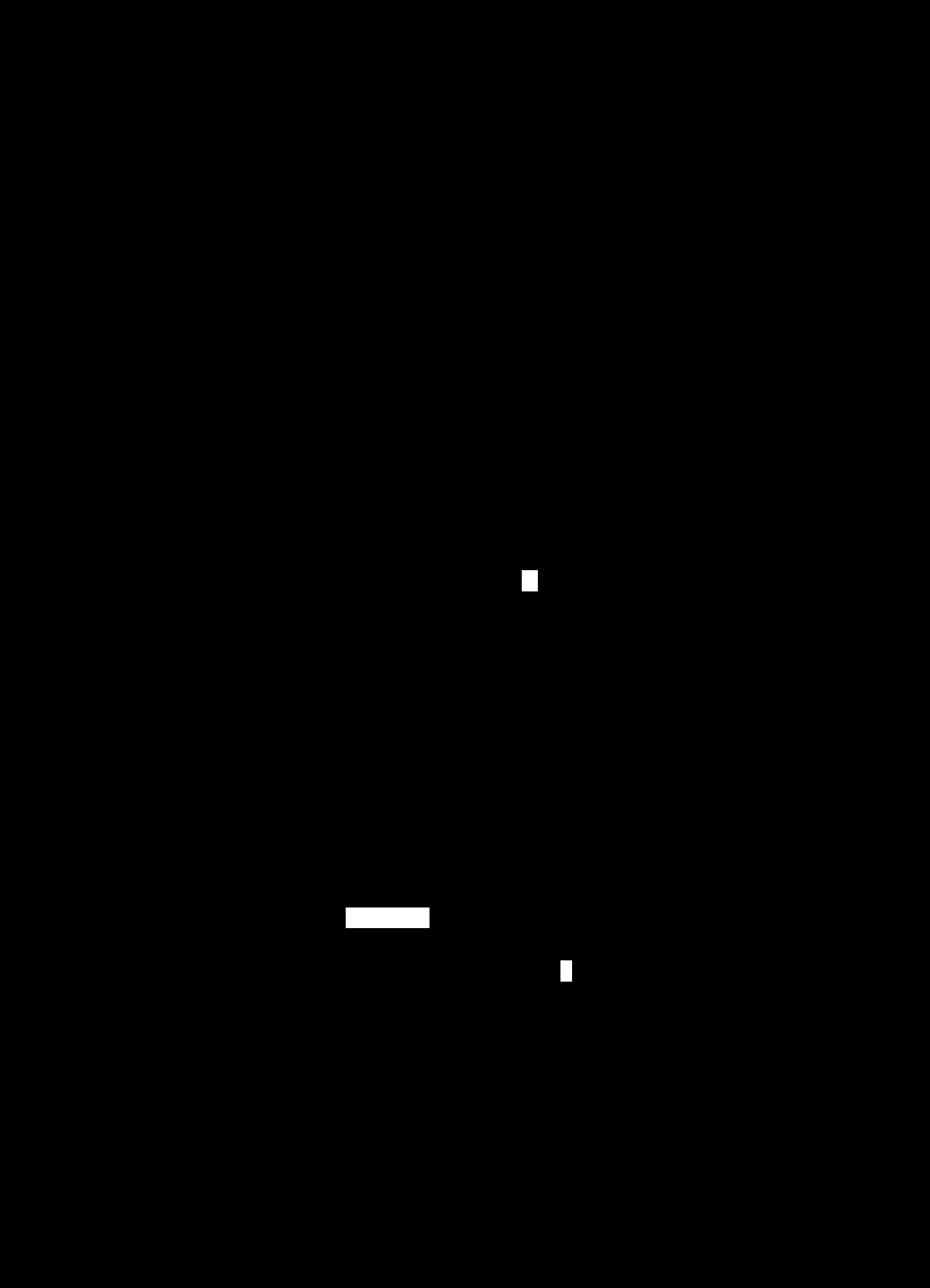} \\
    \includegraphics[width=0.203\textwidth]{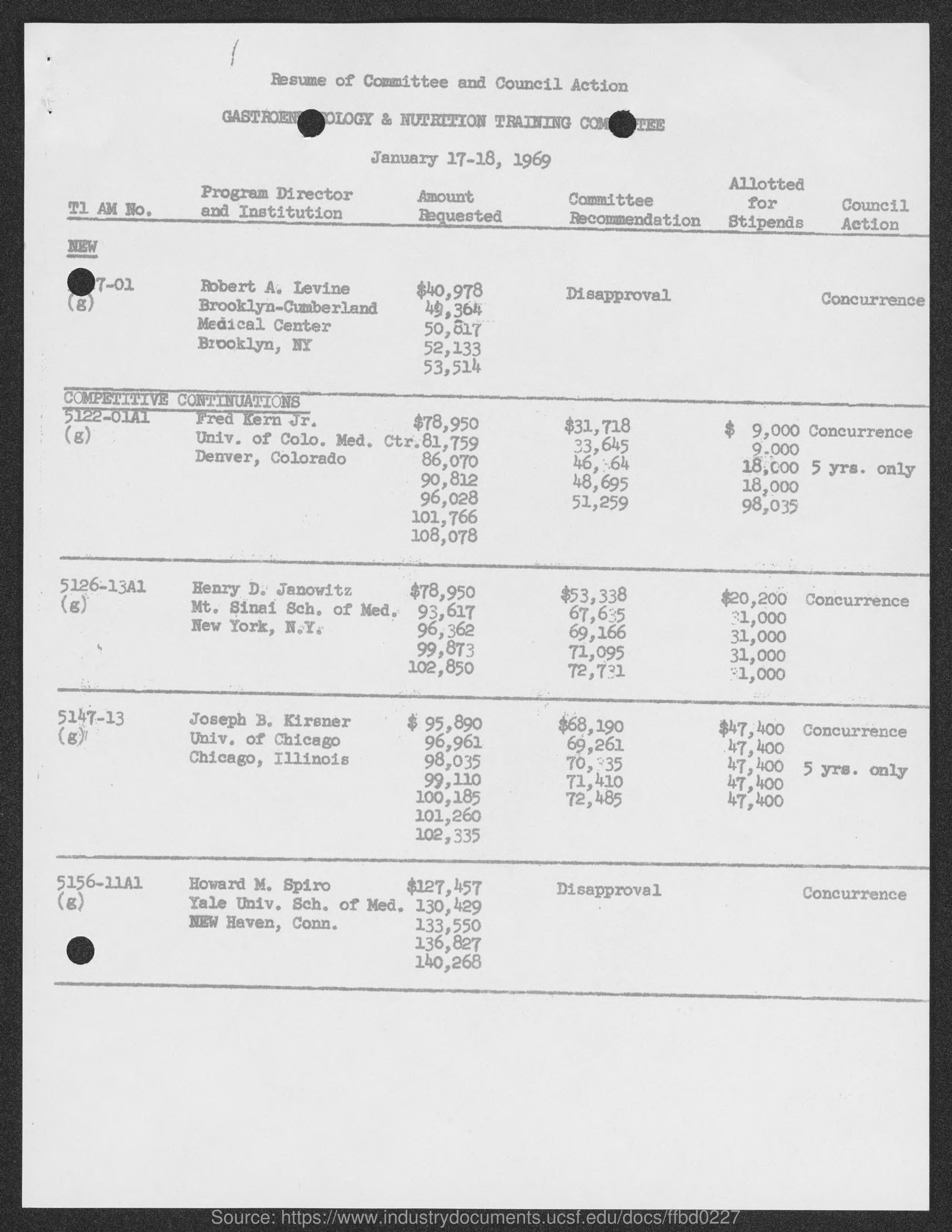} &
    \includegraphics[width=0.203\textwidth]{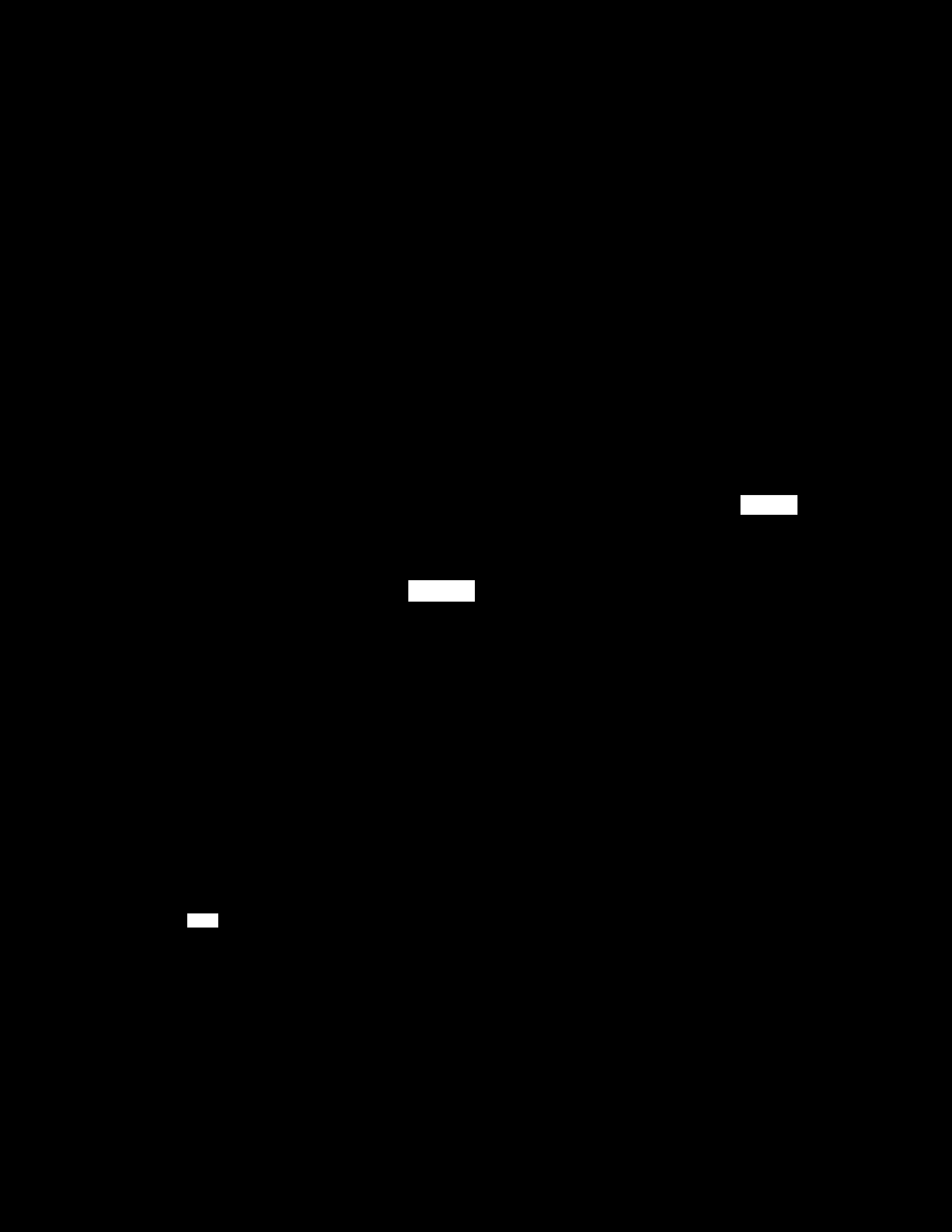} &
    \includegraphics[width=0.203\textwidth]{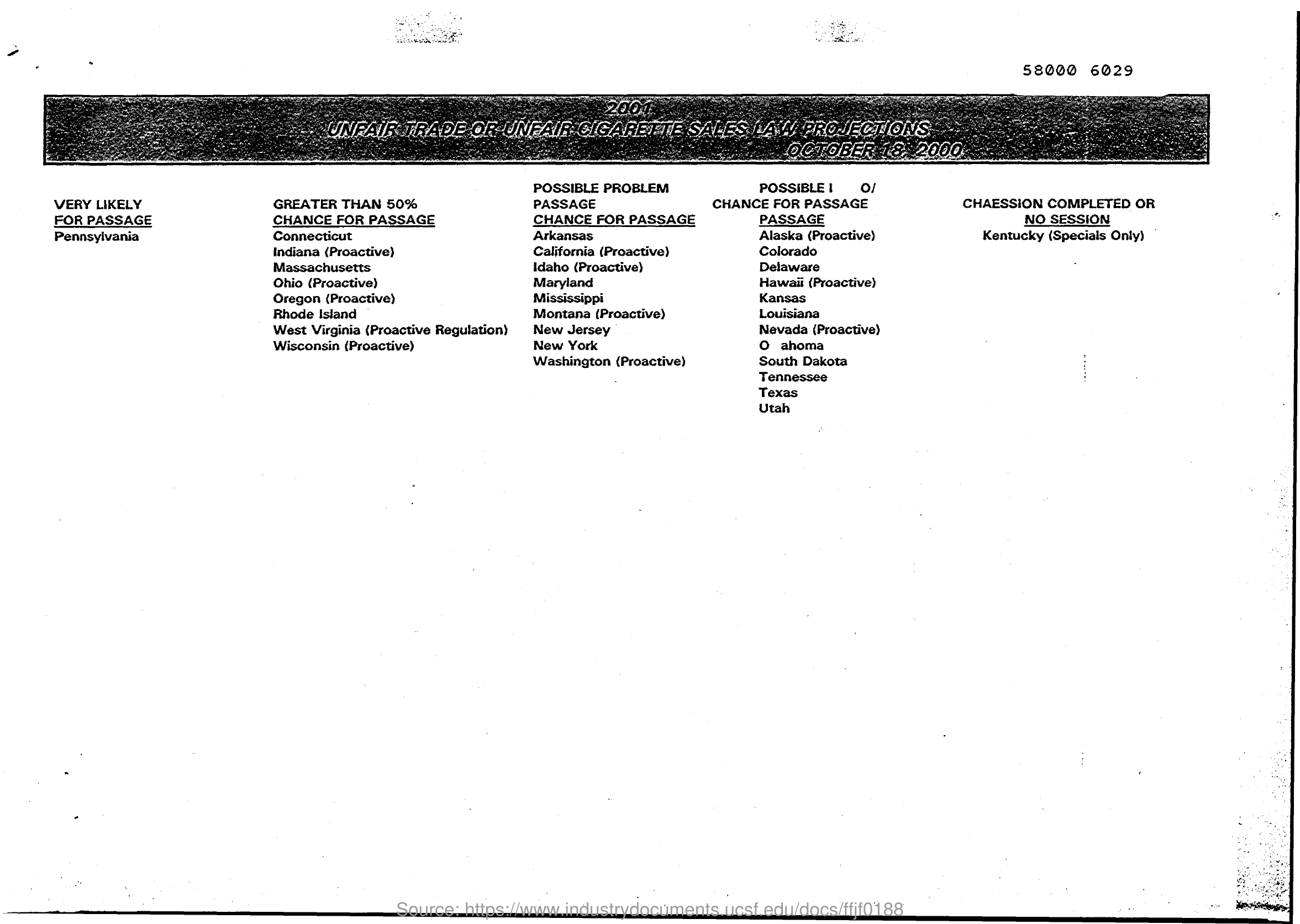} &
    \includegraphics[width=0.203\textwidth]{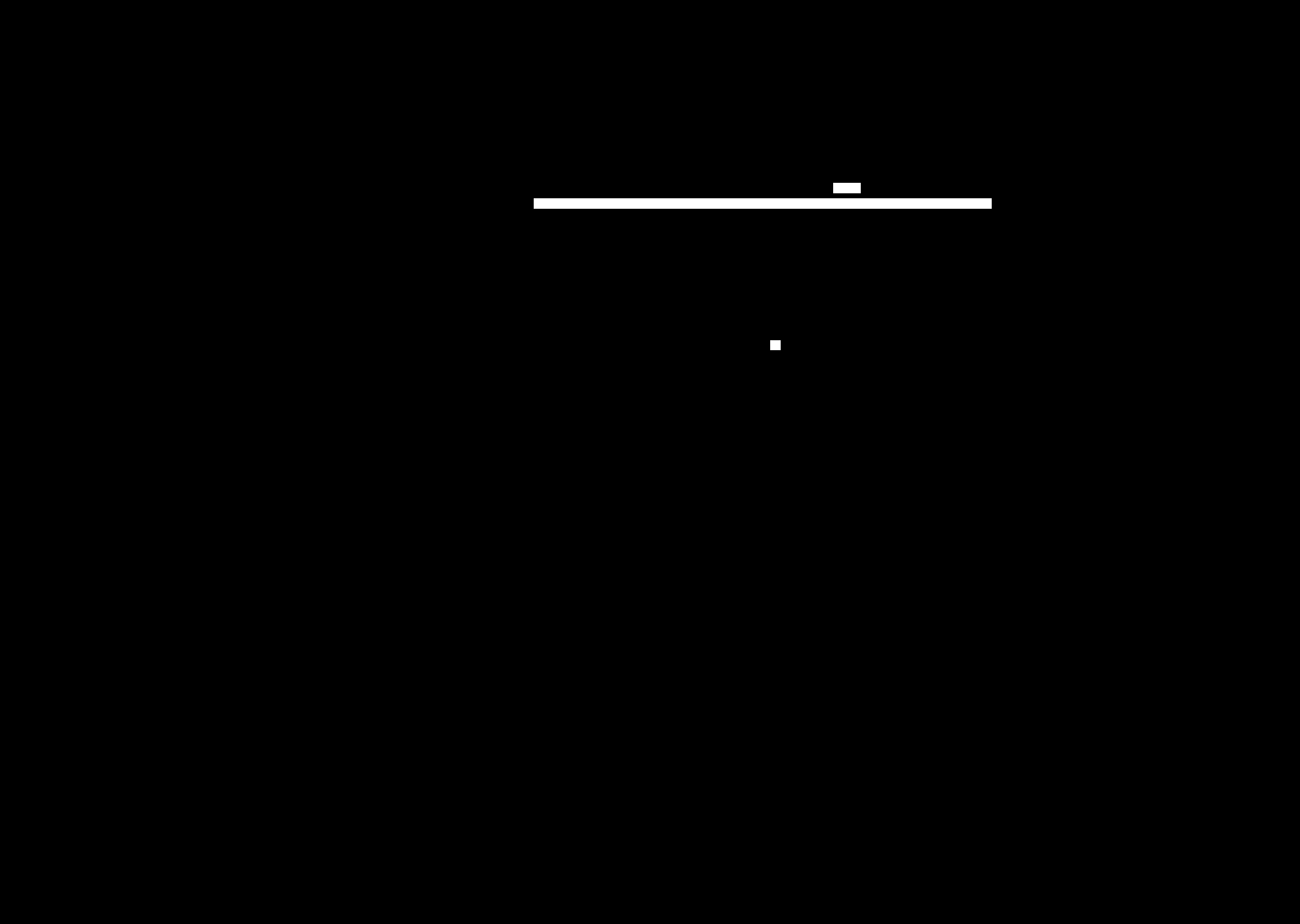} \\
    \includegraphics[width=0.203\textwidth]{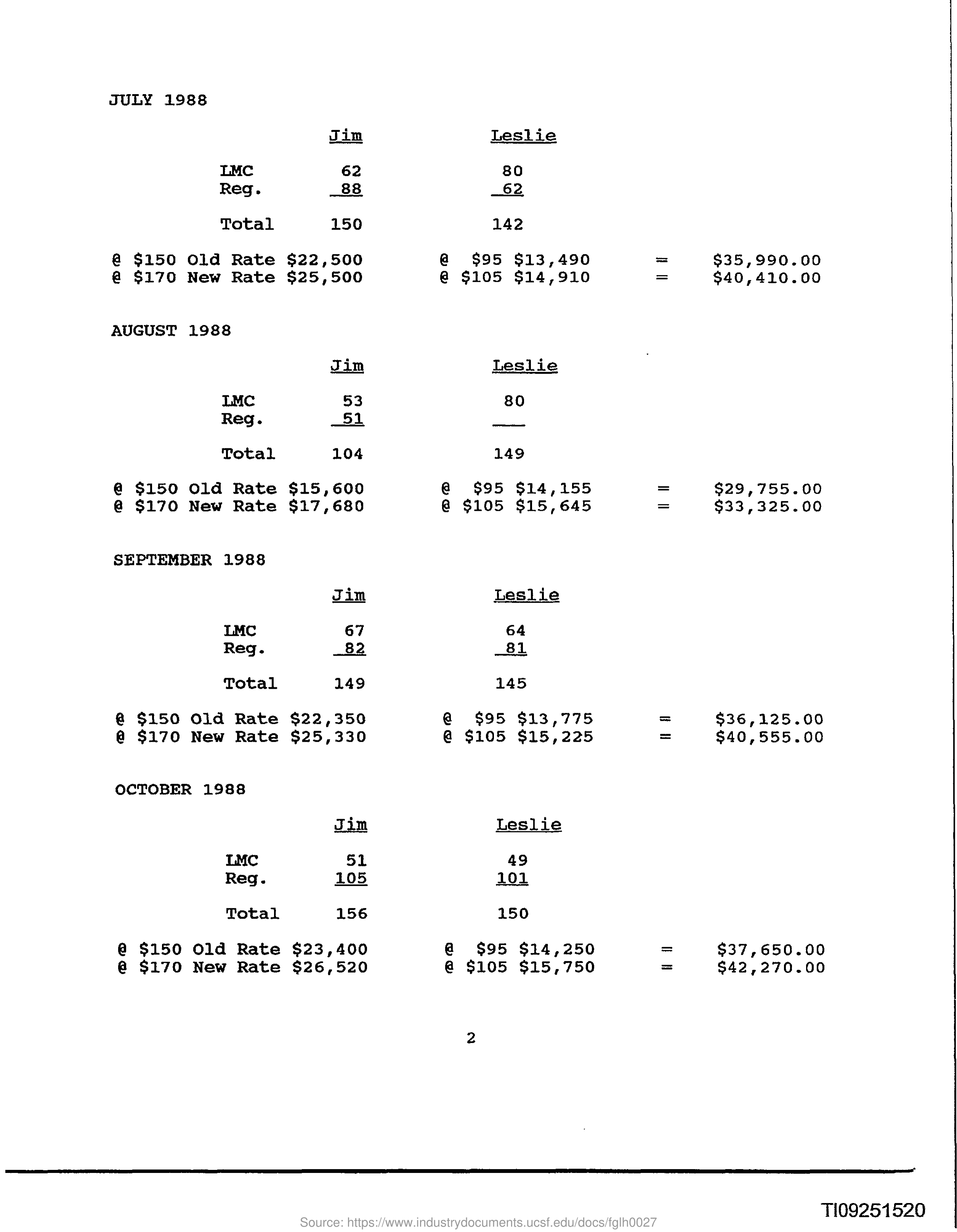} &
    \includegraphics[width=0.203\textwidth]{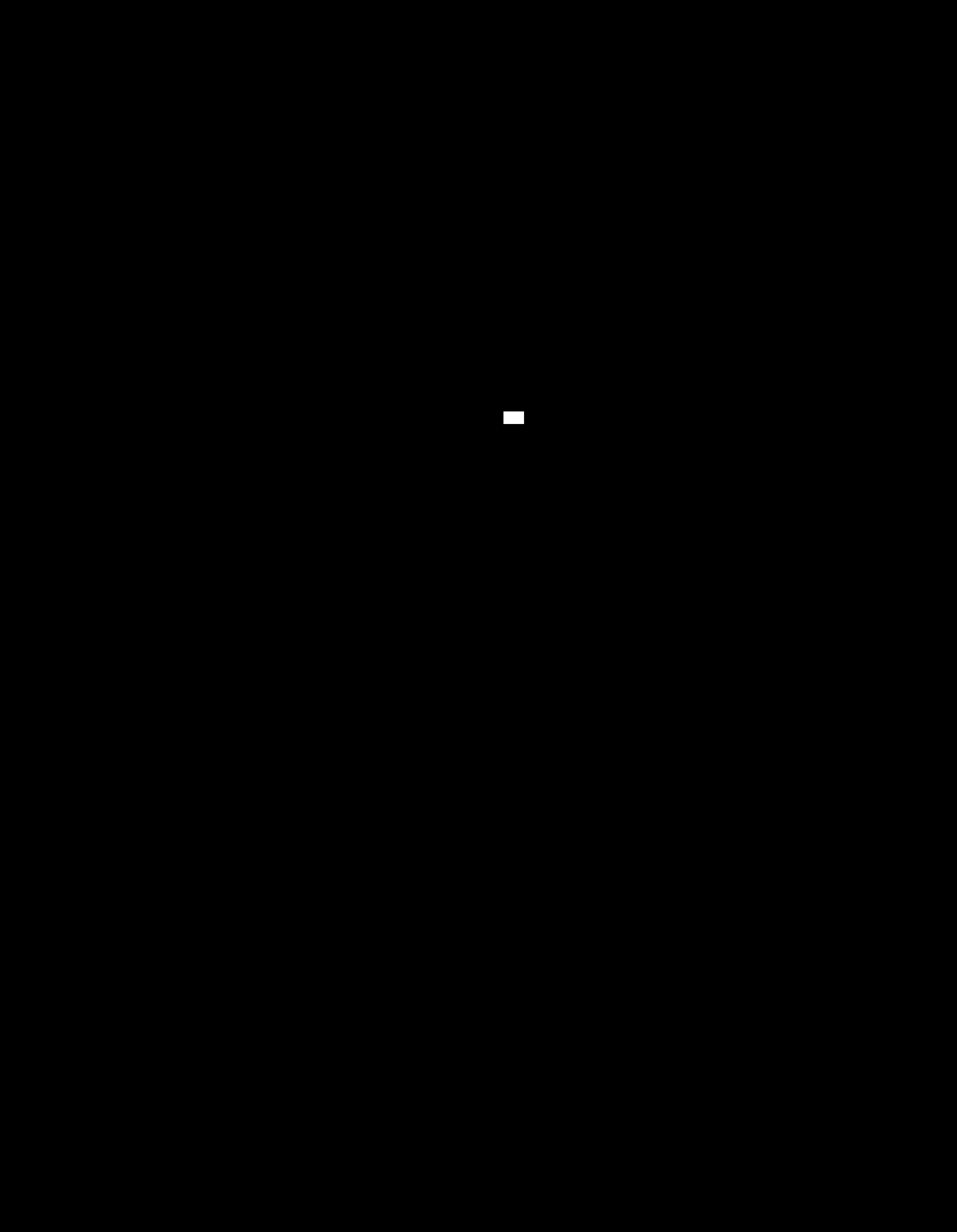} &
    \includegraphics[width=0.203\textwidth]{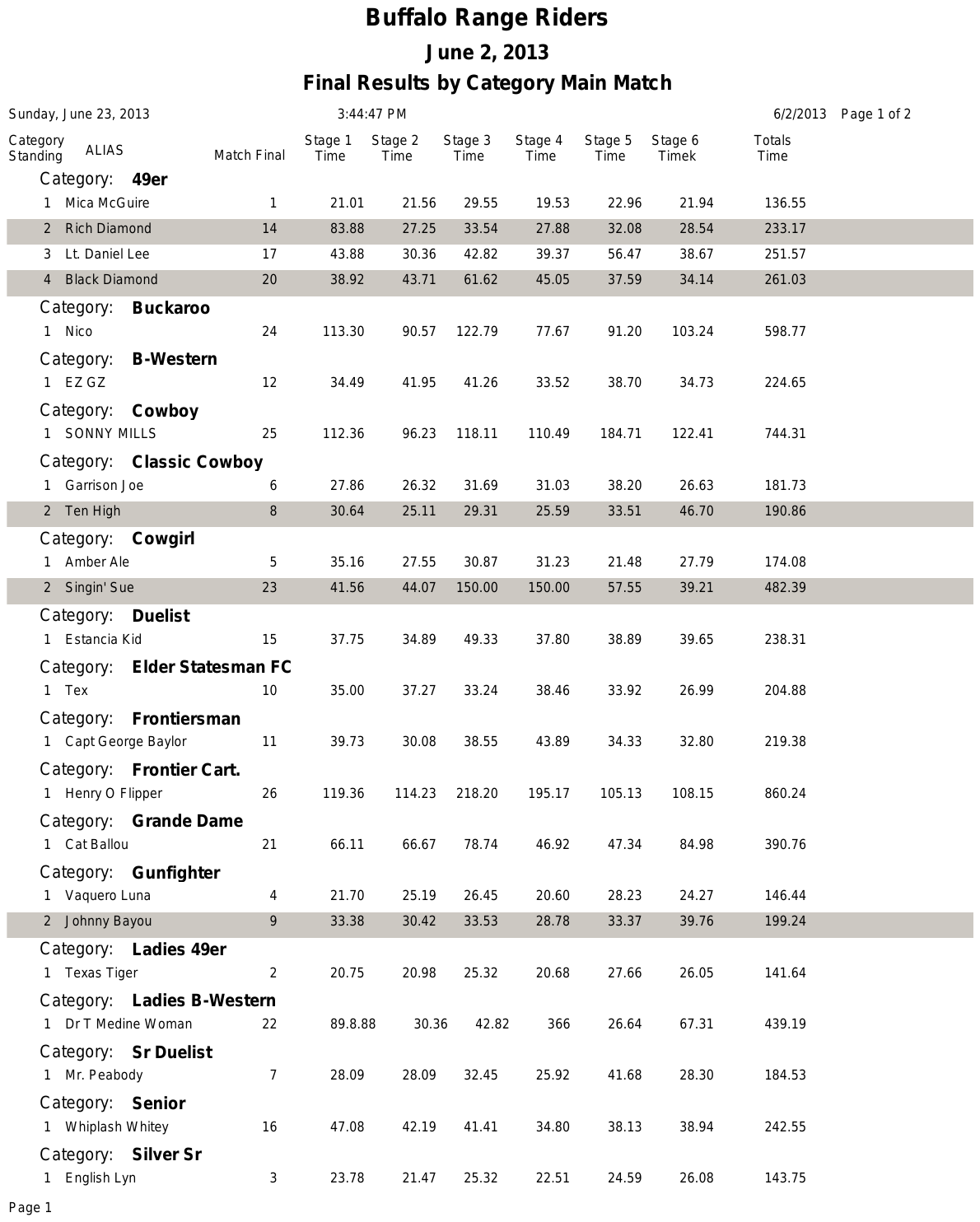} &
    \includegraphics[width=0.203\textwidth]{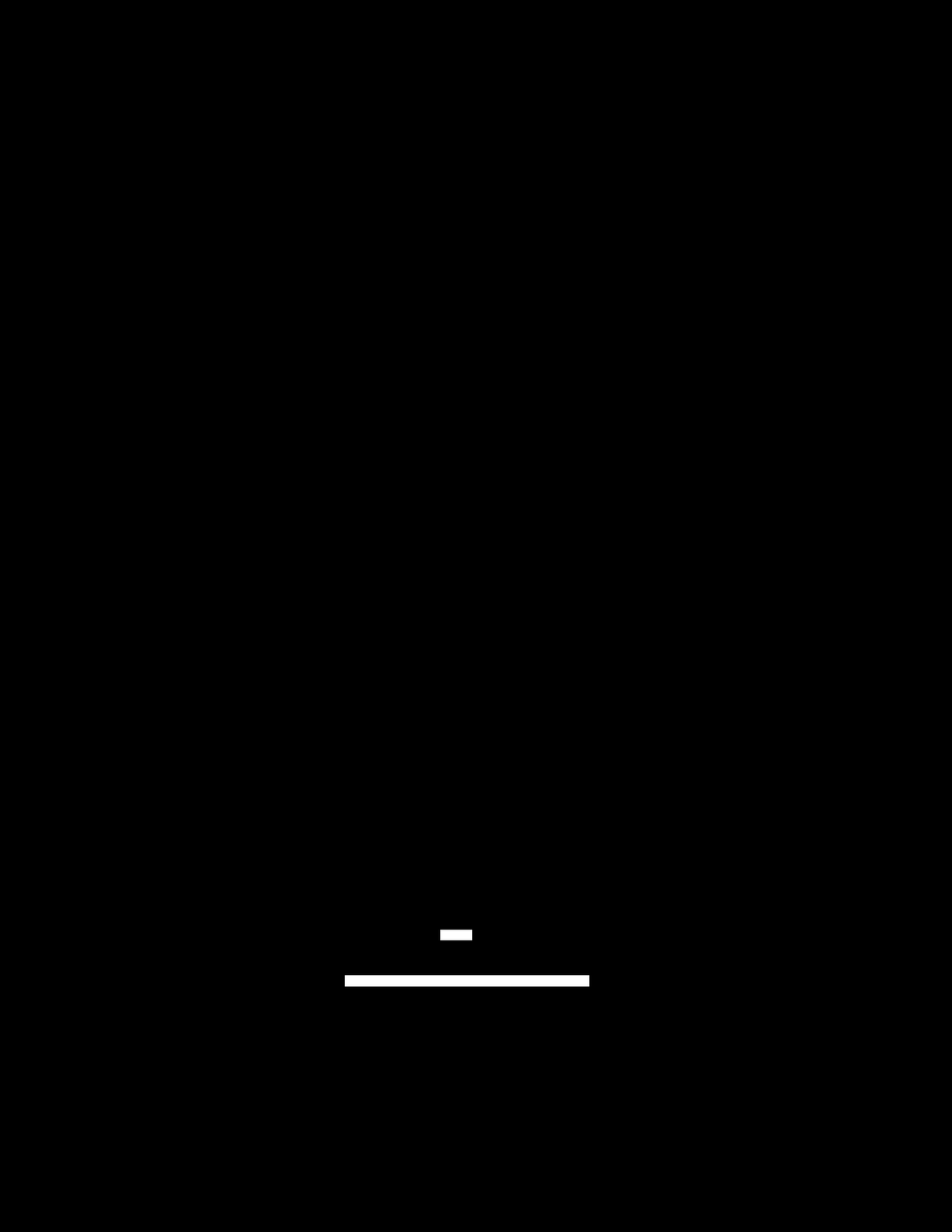} \\
    \includegraphics[width=0.203\textwidth]{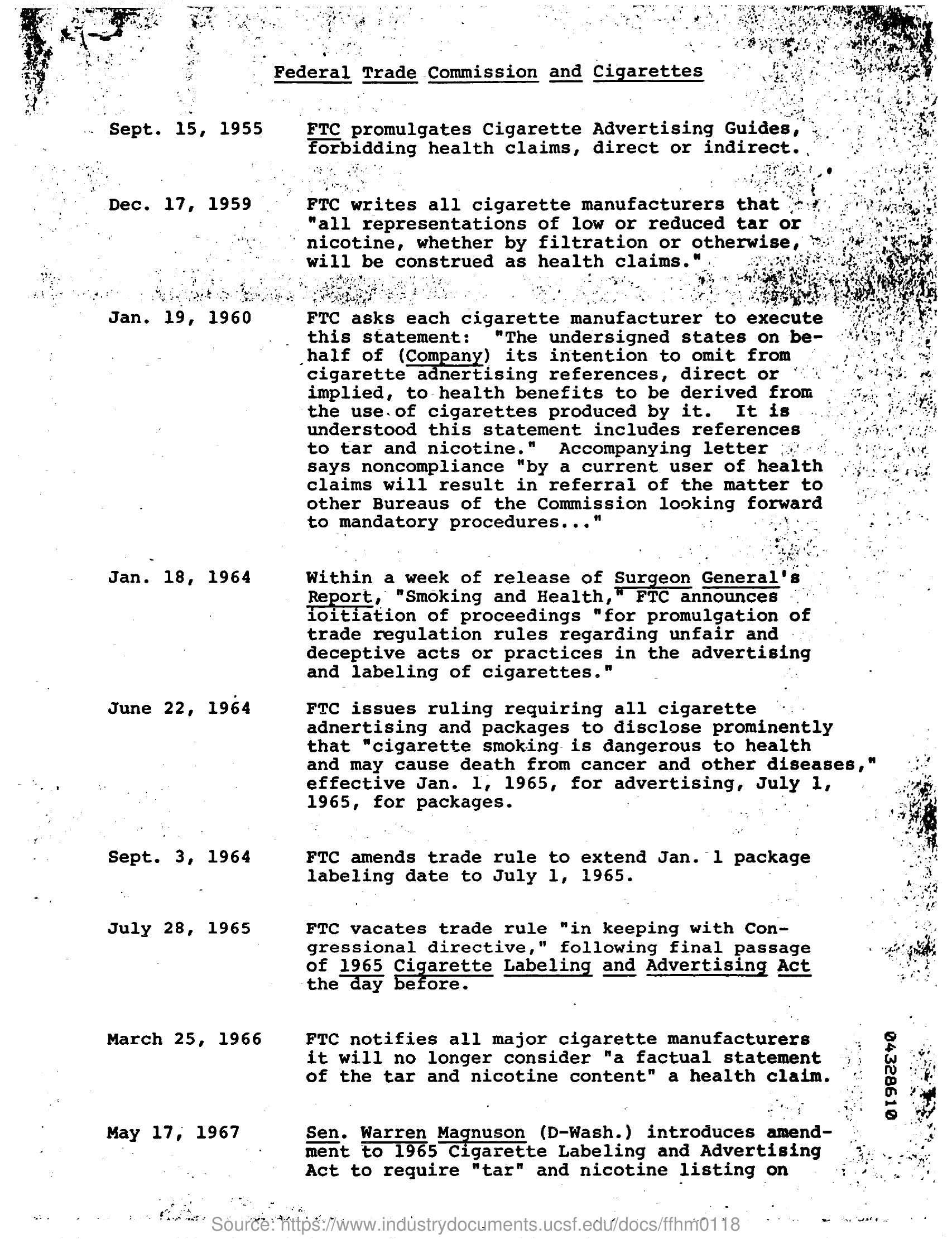} &
    \includegraphics[width=0.203\textwidth]{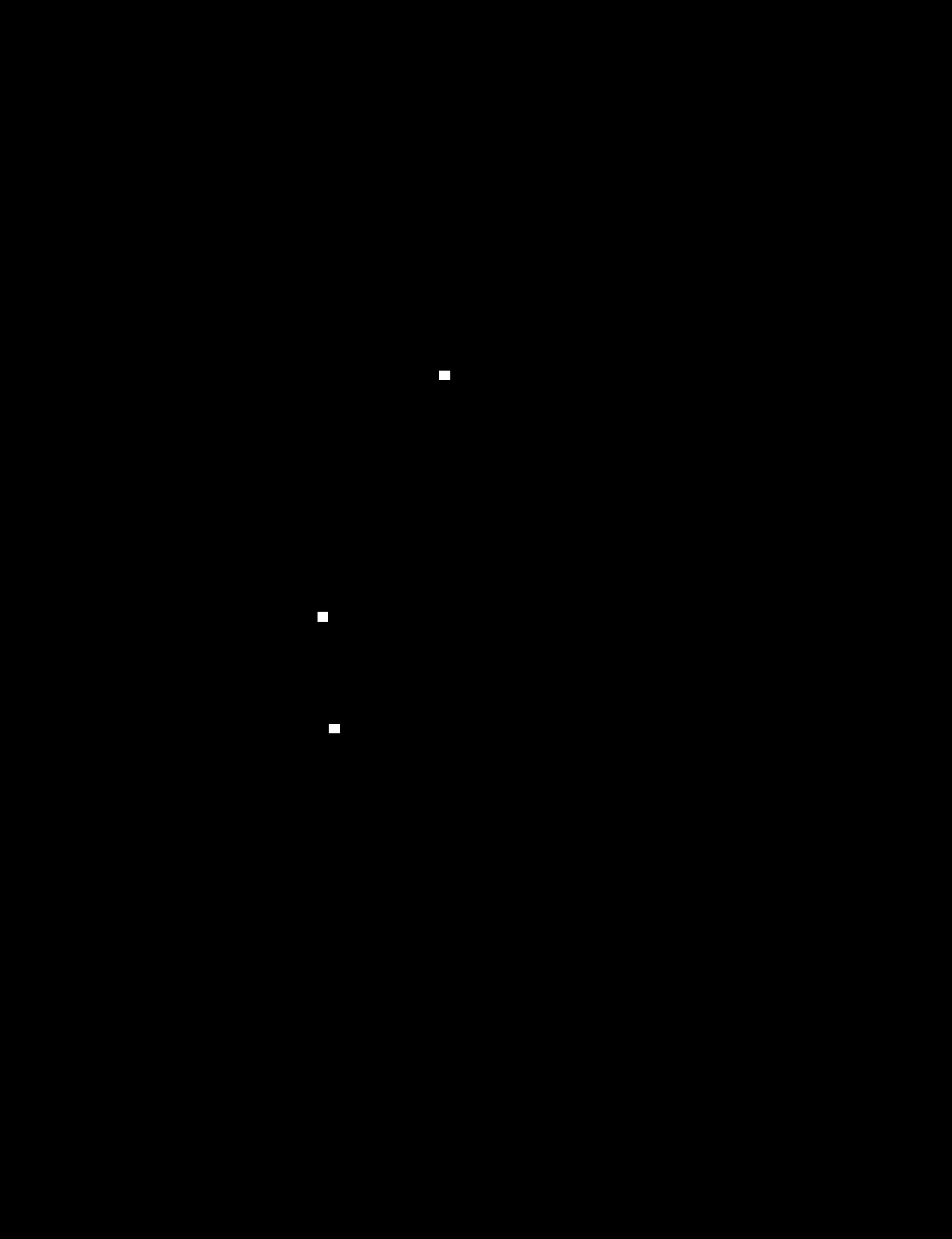} &
    \includegraphics[width=0.203\textwidth]{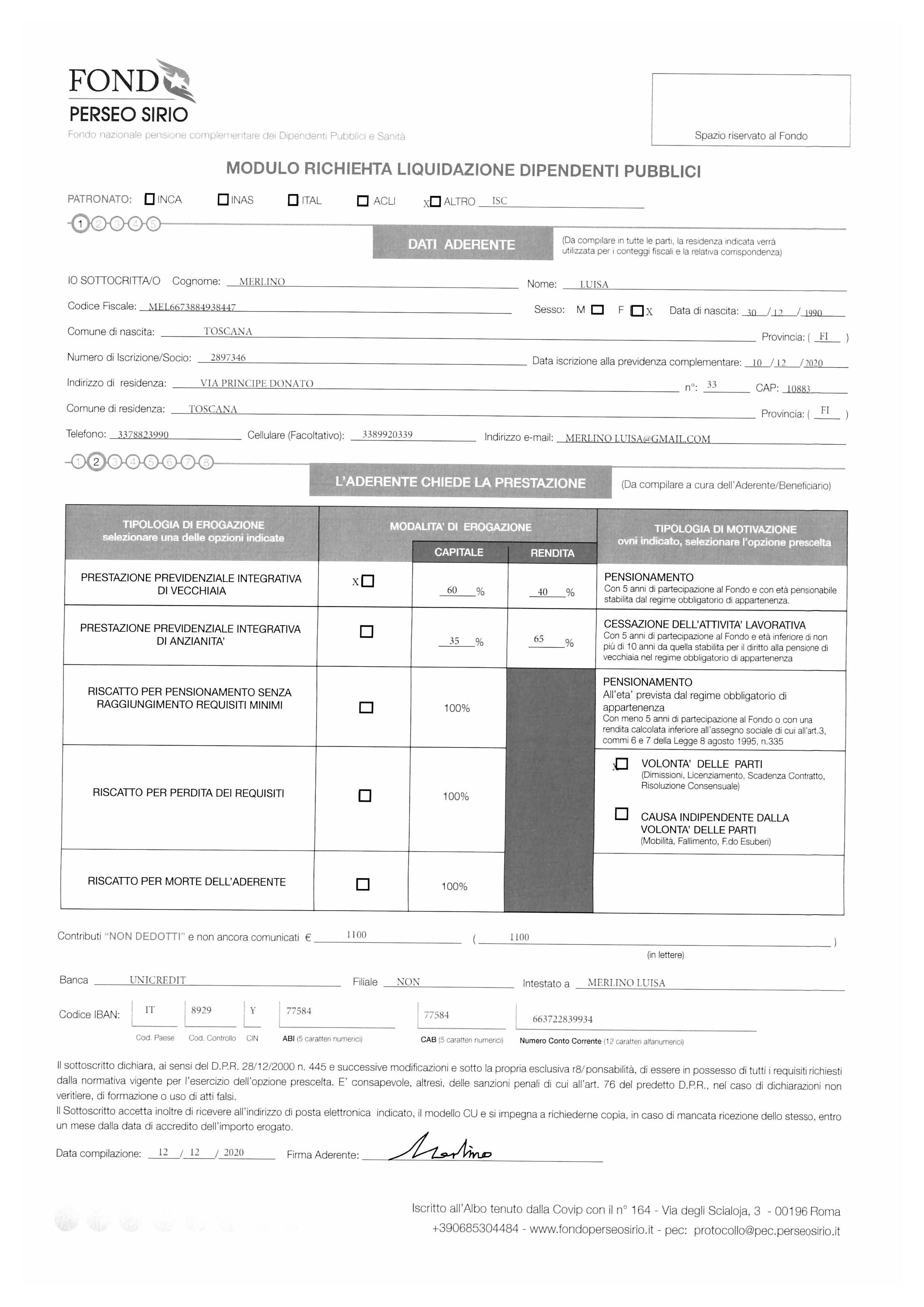} &
    \includegraphics[width=0.203\textwidth]{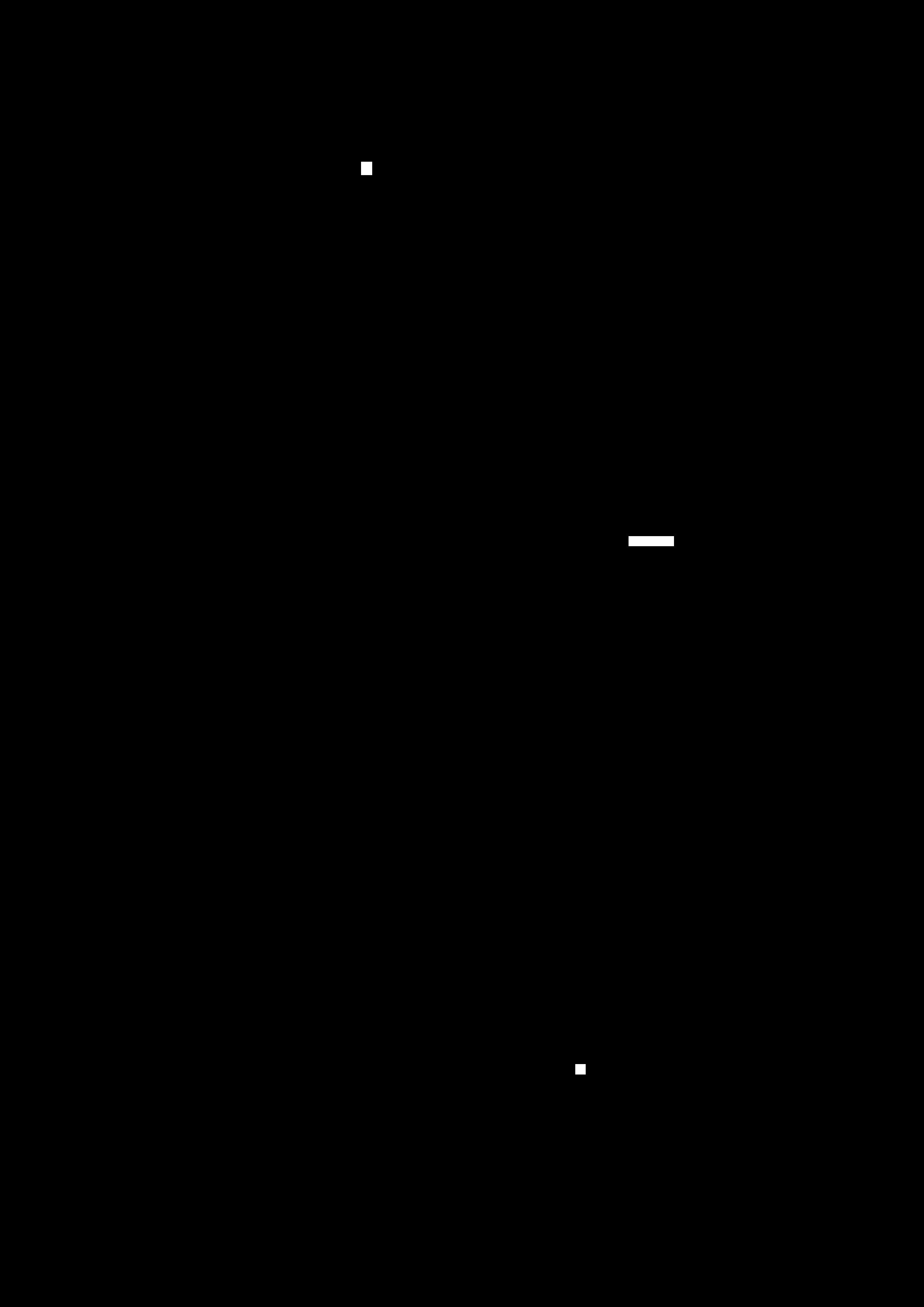} \\
  \end{tabular}
  \caption{Examples of generated tampered documents and their corresponding masks.}
  \label{fig:appendix:examples_8_15}
\end{figure*}

\begin{figure*}[!tp]
  \centering
  \begin{tabular}{cccc}
    \includegraphics[width=0.203\textwidth]{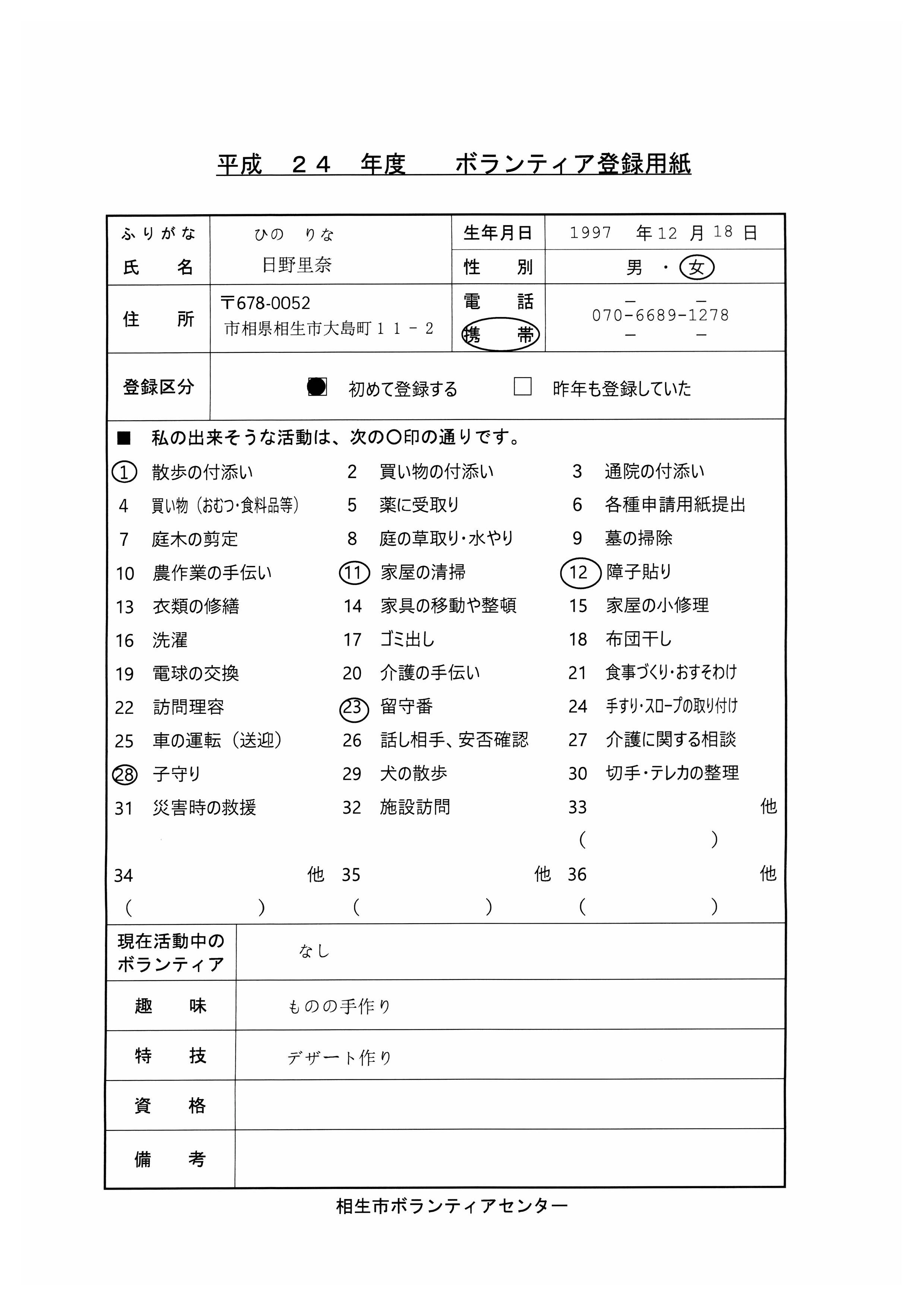} &
    \includegraphics[width=0.203\textwidth]{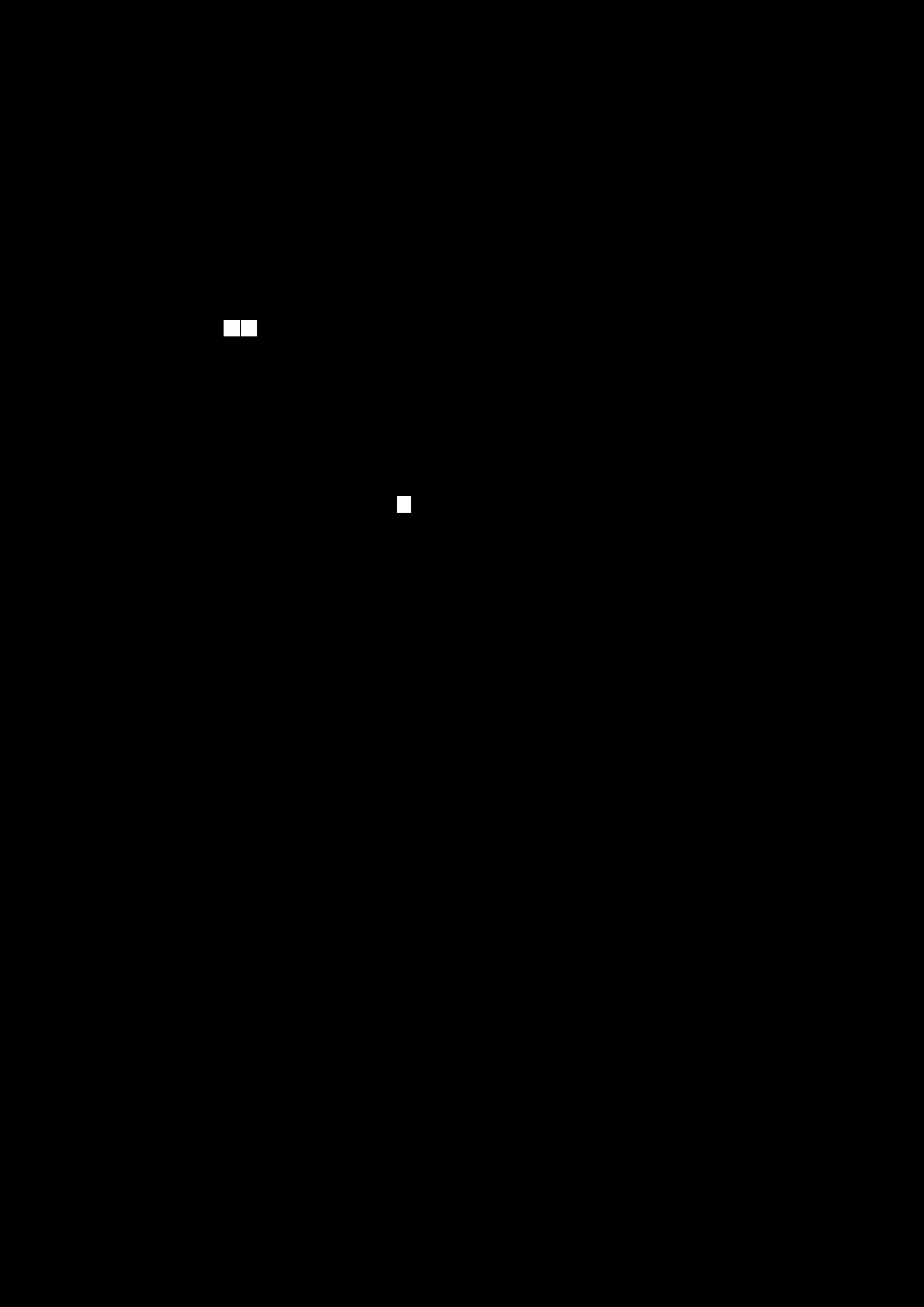} &
    \includegraphics[width=0.203\textwidth]{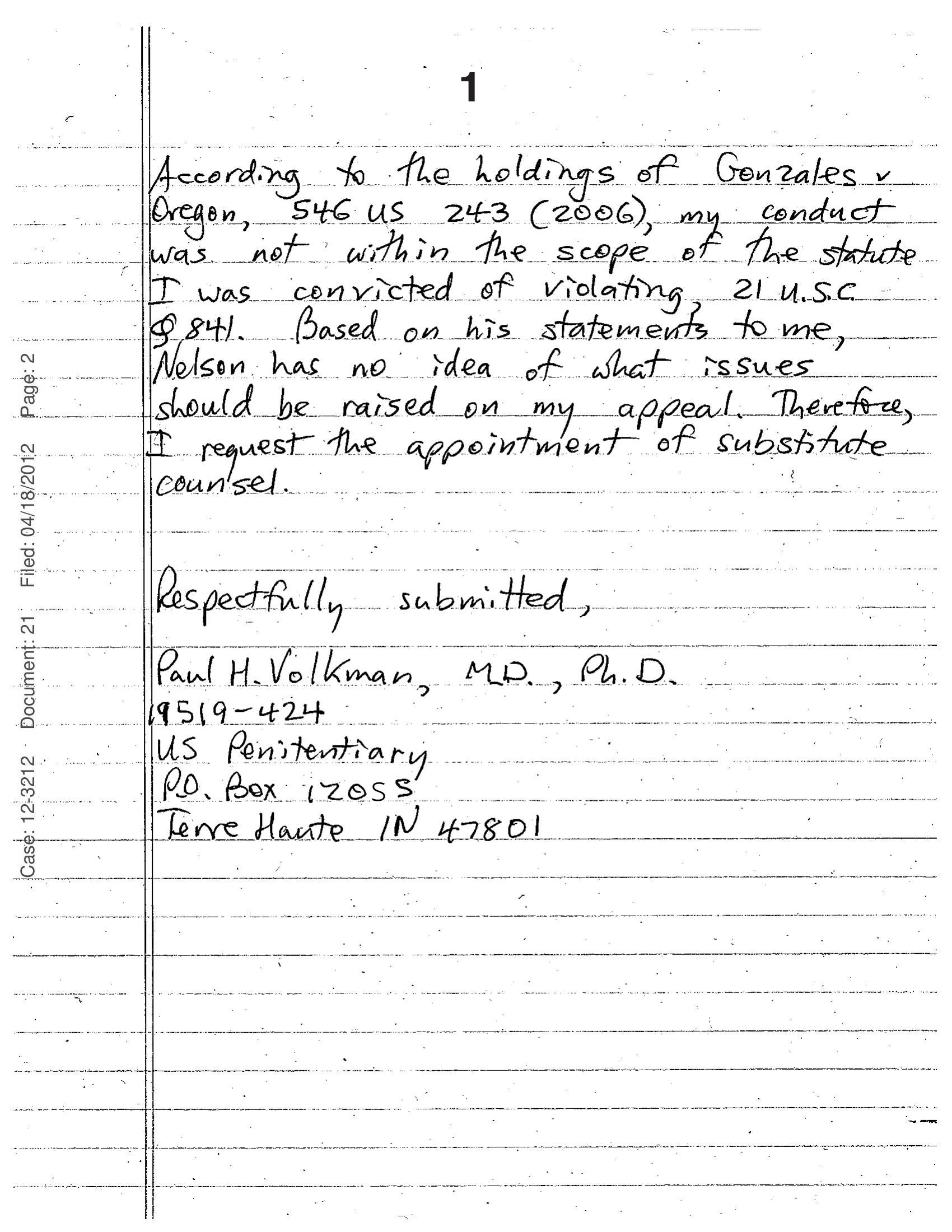} &
    \includegraphics[width=0.203\textwidth]{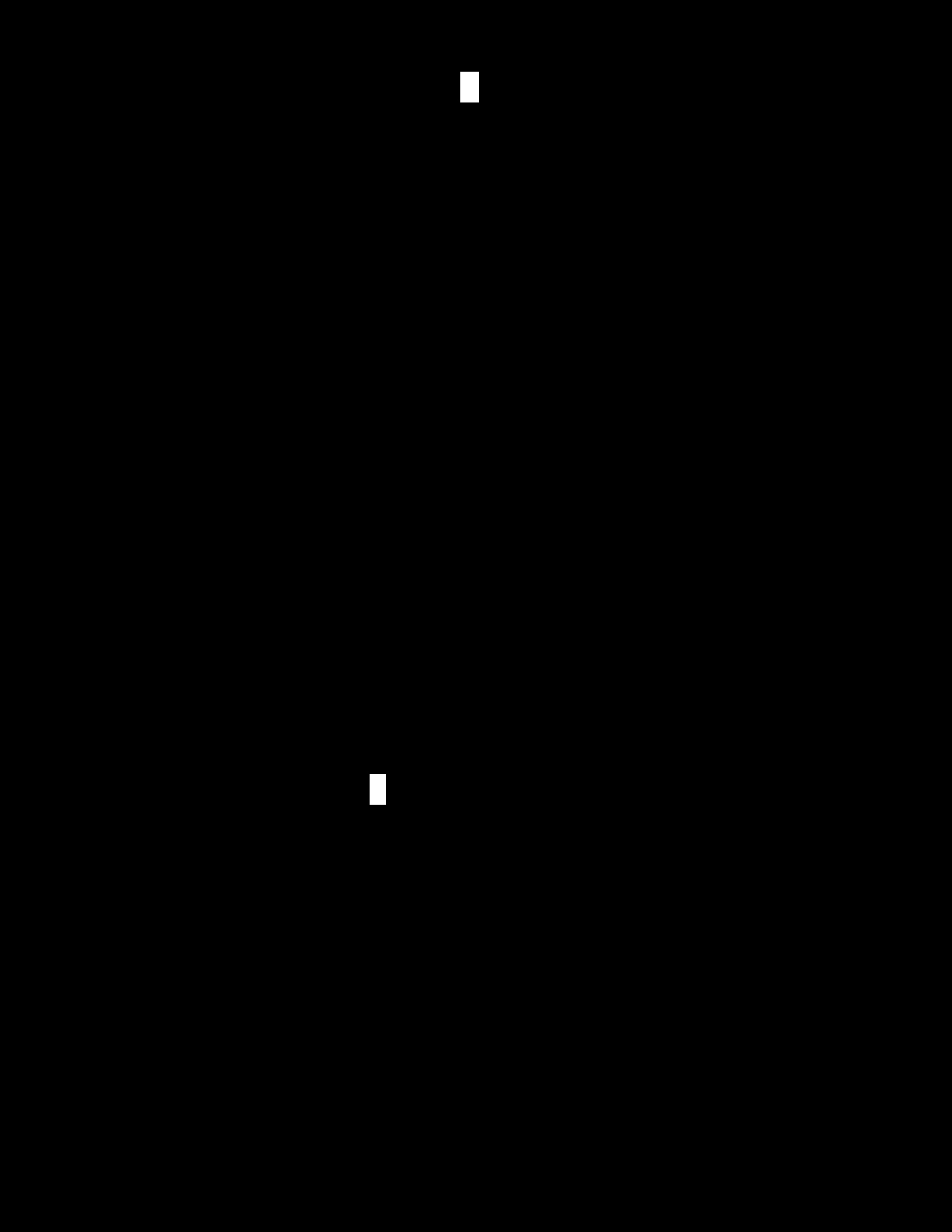} \\
    \includegraphics[width=0.203\textwidth]{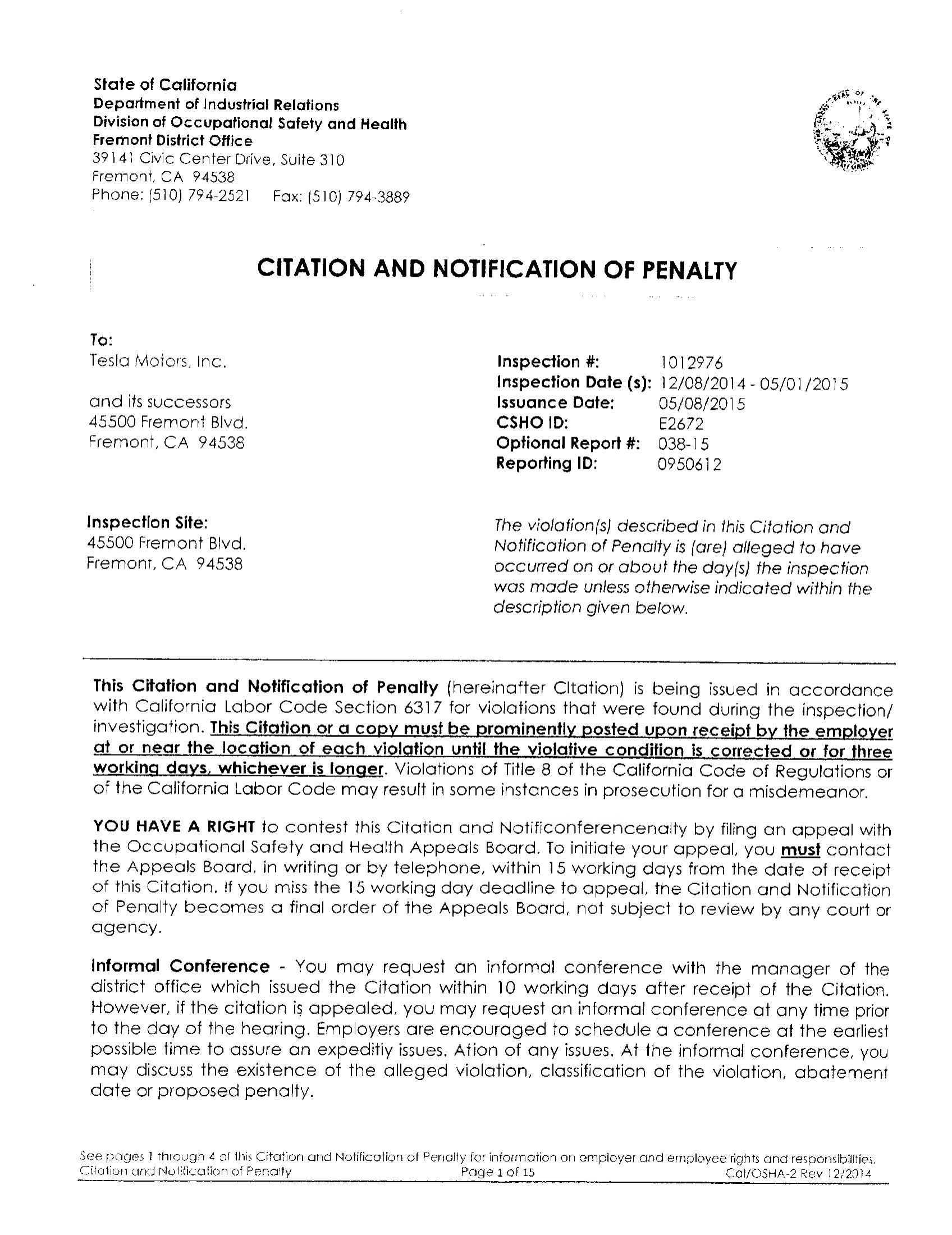} &
    \includegraphics[width=0.203\textwidth]{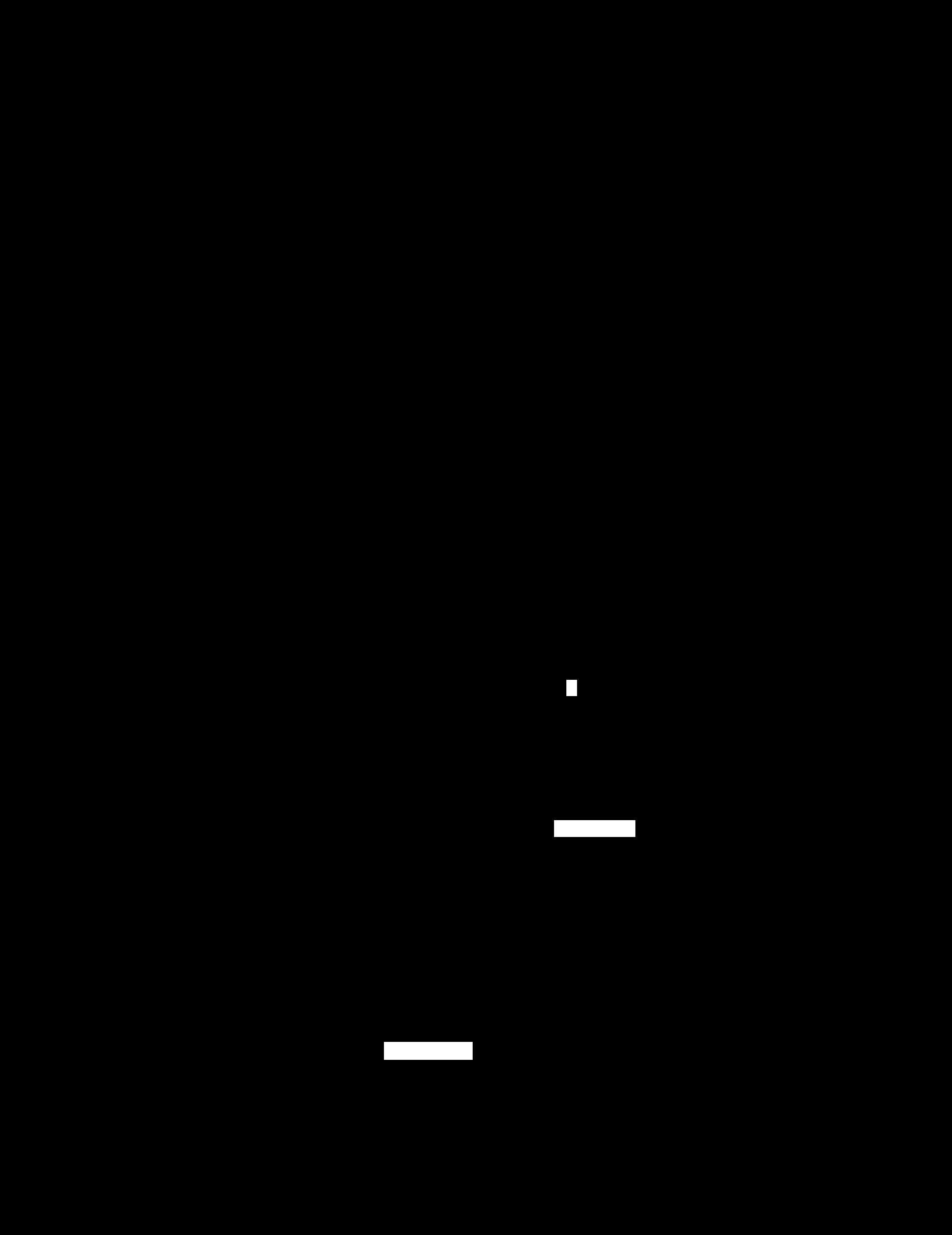} &
    \includegraphics[width=0.203\textwidth]{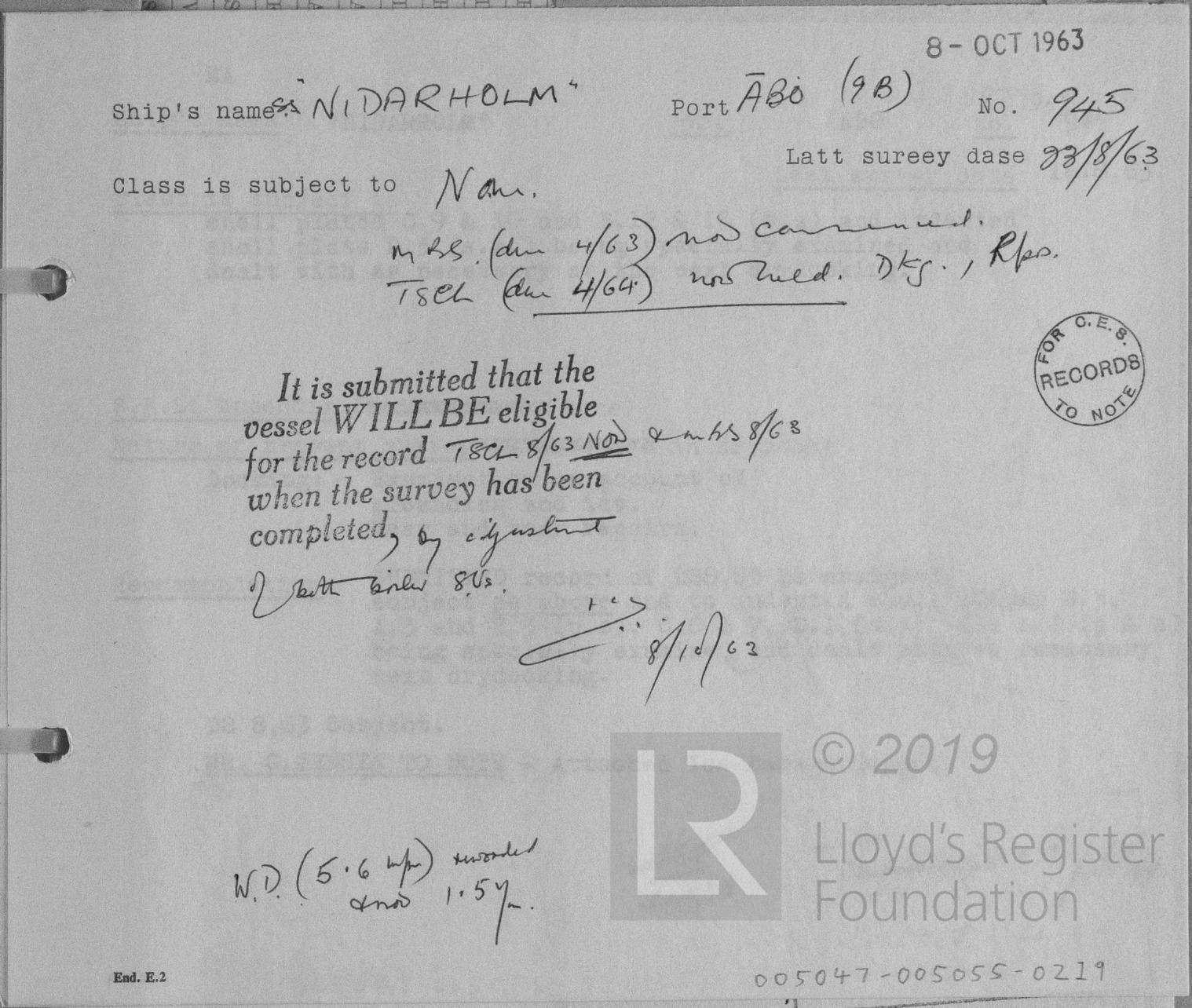} &
    \includegraphics[width=0.203\textwidth]{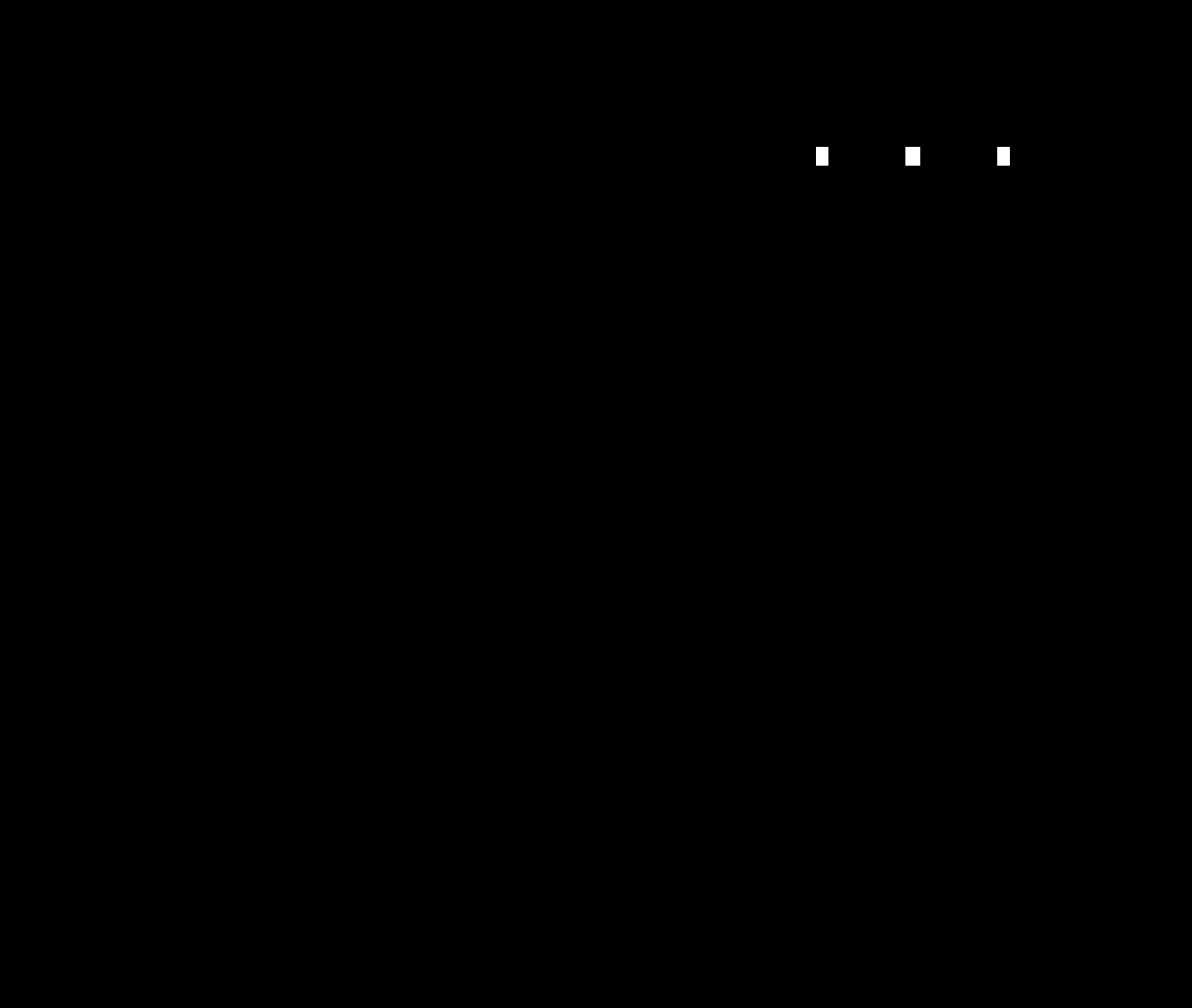} \\
    \includegraphics[width=0.203\textwidth]{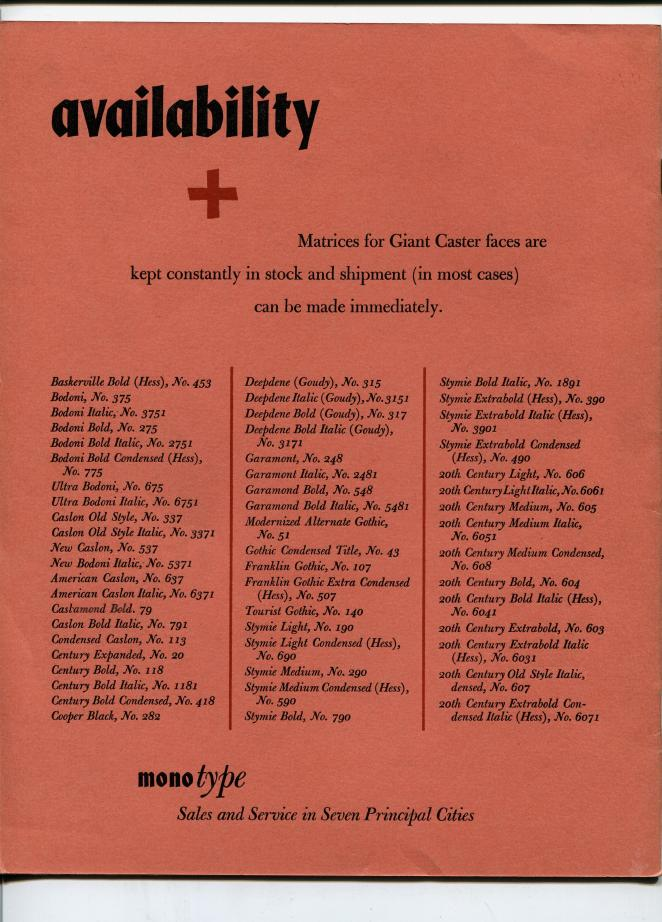} &
    \includegraphics[width=0.203\textwidth]{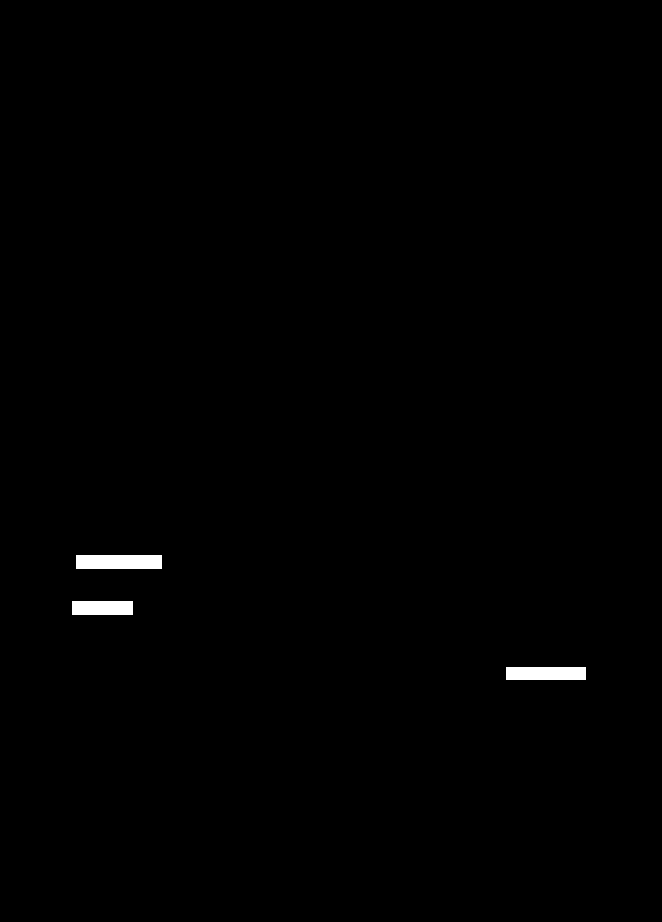} &
    \includegraphics[width=0.203\textwidth]{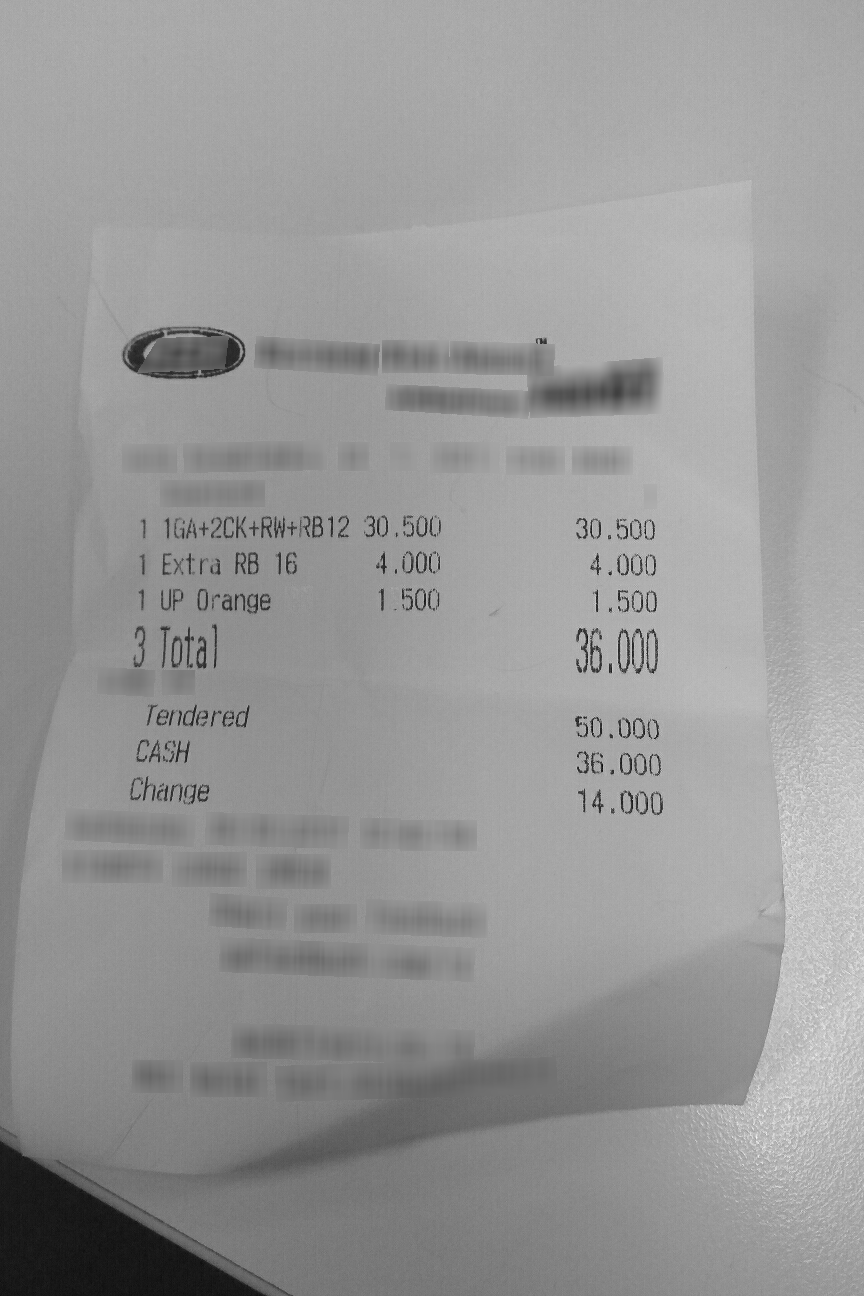} &
    \includegraphics[width=0.203\textwidth]{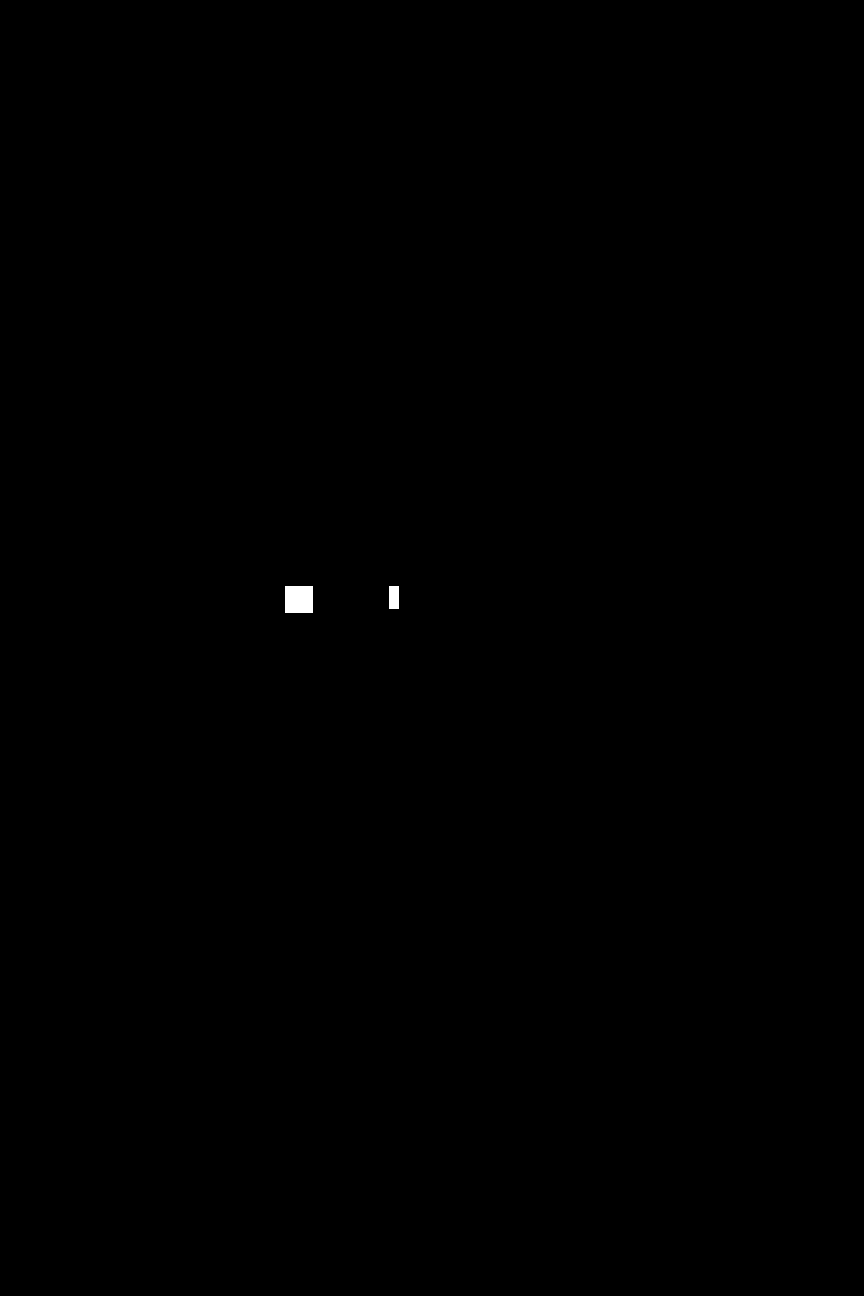} \\
    \includegraphics[width=0.203\textwidth]{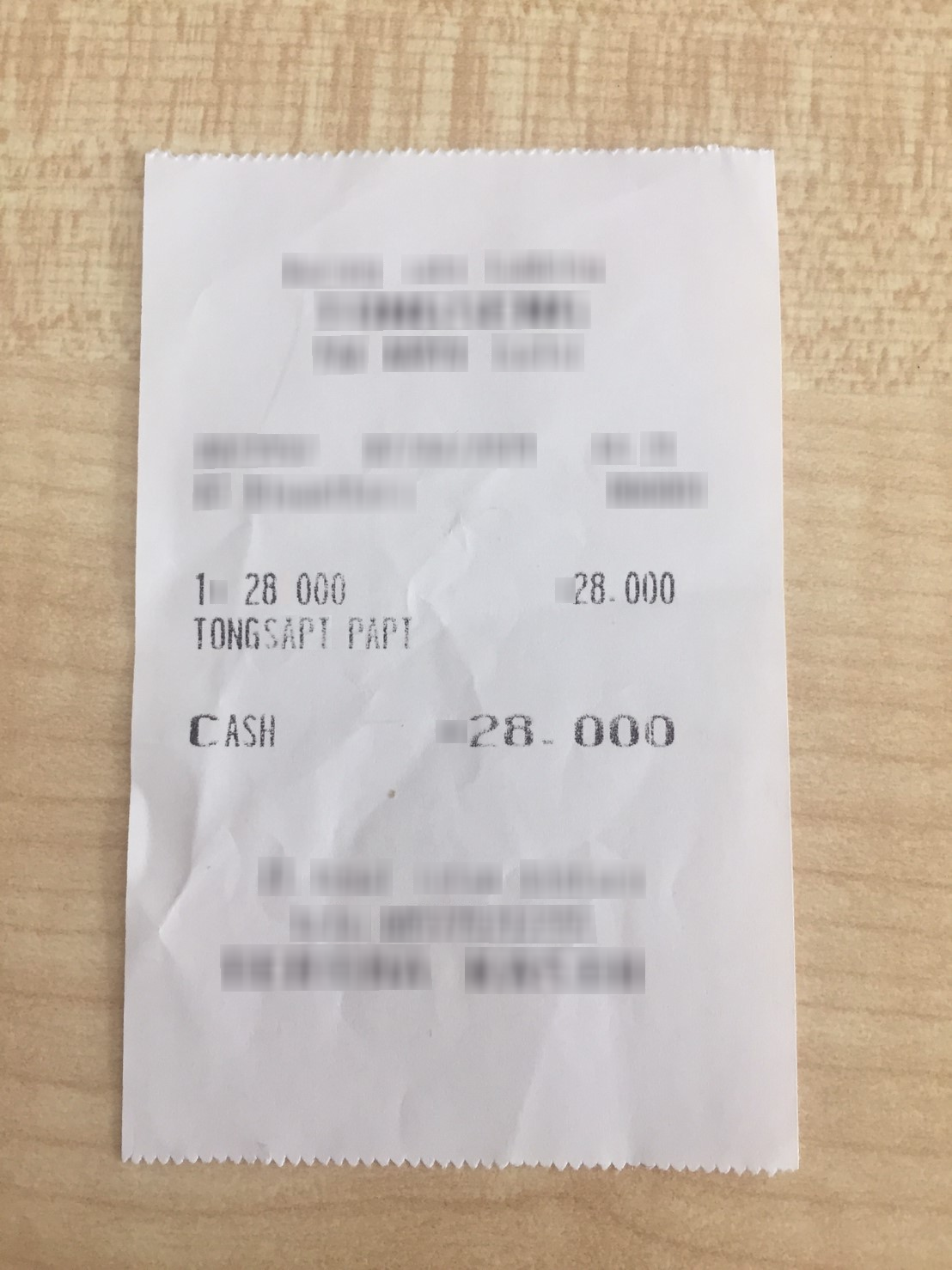} &
    \includegraphics[width=0.203\textwidth]{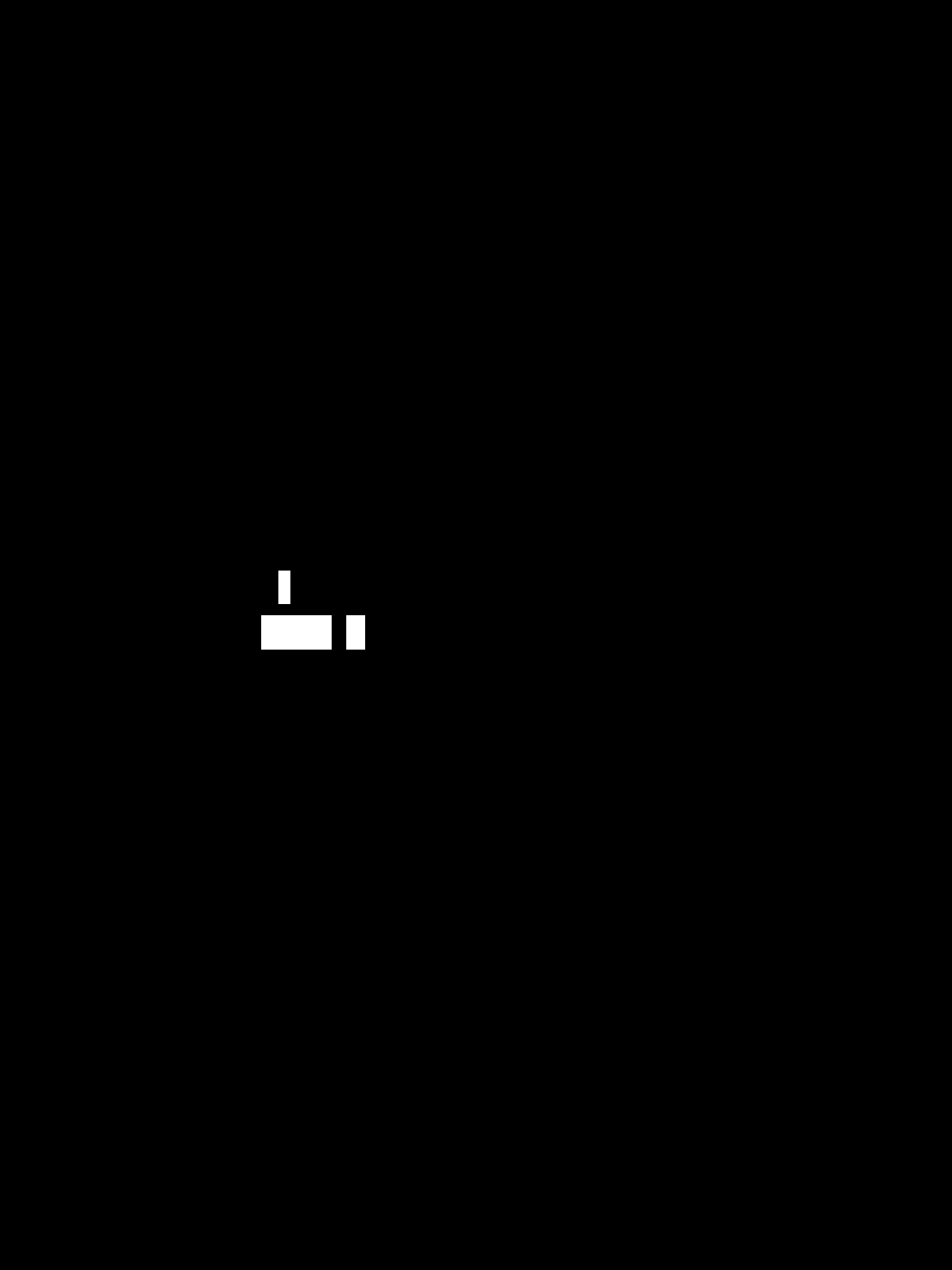} &
    \includegraphics[width=0.203\textwidth]{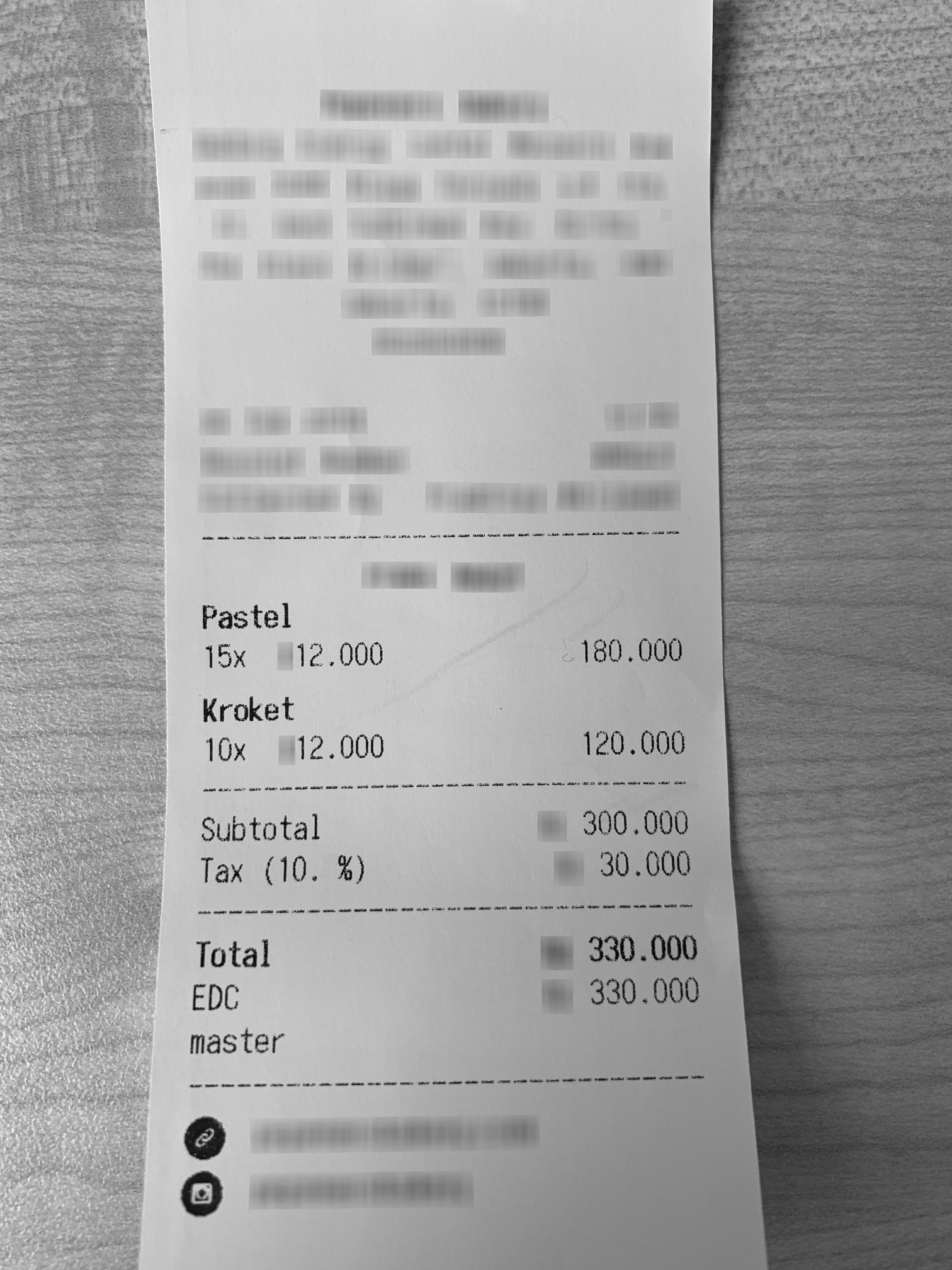} &
    \includegraphics[width=0.203\textwidth]{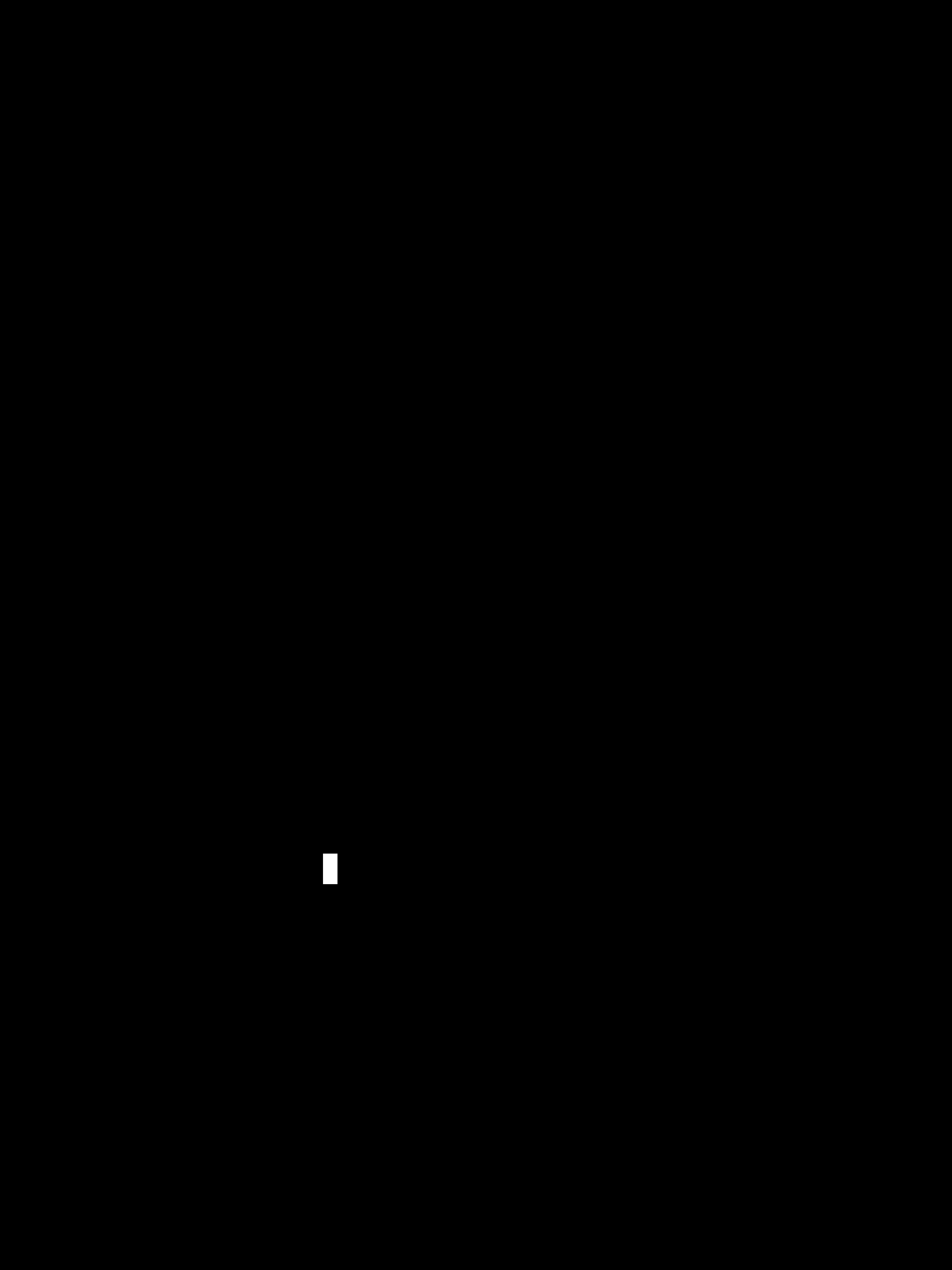} \\
  \end{tabular}
  \caption{Examples of generated tampered documents and their corresponding masks.}
  \label{fig:appendix:examples_16_23}
\end{figure*}

To illustrate the diversity and realism of our tampered document generation pipeline, we provide examples of tampered documents alongside their ground‐truth masks. \Cref{fig:appendix:examples_0_7}, \cref{fig:appendix:examples_8_15}, and \cref{fig:appendix:examples_16_23} present 24 samples covering all five types of tampering (copy-move, splicing, insertion, inpainting, and coverage) across a variety of document layouts, fonts, and background textures.

\section{Data Scraping}
\label{appendix:scraping}

To further diversify our training corpus, we scrape additional document images from publicly available sources. In particular, we collect 175k documents from the Official Journal of the French Republic (\url{https://www.journal-officiel.gouv.fr}), which provides scans of administrative texts, including laws, decrees, and official announcements. These documents offer complex layouts, multilingual content, and high visual diversity, making them valuable for our task. We note that we only scrape documents that are in the public domain. 

\section{Filtering for Out-of-Domain Evaluation}
\label{appendix:outofdomain}
To ensure out-of-domain evaluation and prevent data leakage, we remove any image that either fully or partially matches any image from the source data used to construct the evaluation datasets. Concretely, we slide a $64 \times 64$ patch over the images, compute a hash for each patch, and discard any training image whose patch hashes overlap with those of the evaluation datasets.

\section{Data augmentation applied during training the various tampering detection and localization networks}
\label{appendix:dataaug}
During training, we apply, for all models and datasets, the same data augmentation operations that consist of a series of augmentations applied on both the document images and their corresponding tampering masks. The augmentations are probabilistically applied and consist of both geometric and photometric transformations. Below, we detail each augmentation used.

\subsection{Photometric Augmentations (Image Only)}
\begin{itemize}
    \item \textbf{Brightness Adjustment:} Using \texttt{ColorJitter}, brightness is randomly modified in the range \([0.3, 2.0]\).
    \item \textbf{Contrast Adjustment:} Image contrast is randomly perturbed in the range \([0.3, 2.0]\).
    \item \textbf{Saturation Adjustment:} Saturation is altered within the same range \([0.3, 2.0]\).
    \item \textbf{Gaussian Blur:} Randomly applied with kernel sizes from \{3, 5, 7\}.
\end{itemize}

\subsection{Geometric Augmentations (Image + Mask)}
\begin{itemize}
    \item \textbf{Small Angle Rotation:} Images and masks are rotated by a random angle between \(-5^\circ\) and \(+5^\circ\).
    \item \textbf{90°-Based Rotations:} Random rotation of either \(+90^\circ\), \(-90^\circ\), or \(180^\circ\).
    \item \textbf{Horizontal and Vertical Flips:} Horizontal and vertical flip decisions are sampled independently of each other with probability 0.05, and every selected flip is applied jointly to the image and mask using the same parameters.
    \item \textbf{Random Shrink and Rescale:} The image and mask are jointly scaled by a factor from \([0.5,2.0]\) using identical geometric parameters. The RGB image uses one of \texttt{BILINEAR}, \texttt{BICUBIC}, or \texttt{LANCZOS}, while the mask uses nearest-neighbor interpolation.
    \item \textbf{Perspective Warp:} With small probability (0.1), homographic transformations are applied by perturbing image corners.
\end{itemize}

\subsection{Compression Artifacts (Image Only)}
To simulate JPEG compression:
\begin{itemize}
    \item A random number (0–3) of JPEG recompressions are applied with quality values randomly selected from a predefined range (75–100).
\end{itemize}

The probabilities are controlled via a configuration file, which is included in the codebase and contains the default values used in our training.

\section{Robustness across dataset sizes and random seeds}
\label{smaller_scale}

We further analyze how the benefit of our data-generation pipeline evolves when the amount of source data used for synthetic generation is reduced. Let the \emph{full} setting denote the main source dataset used in the previous experiments. The \emph{half}, \emph{quarter}, and \emph{eighth} settings correspond to using respectively $1/2$, $1/4$, and $1/8$ of the source images from the main dataset. For each fraction, we first sample the corresponding subset of source images and then regenerate the synthetic manipulations from scratch for each pipeline, using the same image pool for all generation methods at each iteration. For every scale, we compare our method against the strongest of the three previous pipelines \cite{Doctamper, dataset2,newdataandopenopen}.
As results at some scales are close, we report the best F1 score each time, avoiding the variance that a fixed threshold may introduce. \Cref{tab:size_scaling} summarizes the results under different data scales.
The results show that the advantage of our generation strategy remains consistent across all dataset sizes. For ASC-Former, our method improves the average, across RTM, FindItAgain, and FindIt, of the best pixel-level F1 obtained on each dataset from $14.14$ to $17.33$ at full scale, from $14.09$ to $16.96$ at half scale, from $13.85$ to $16.45$ at quarter scale, and from $13.77$ to $16.07$ at eighth scale. For DTD, the corresponding average, across RTM, FindItAgain, and FindIt, of the best pixel-level F1 obtained on each dataset increases from $12.35$ to $13.73$, from $11.86$ to $13.72$, from $11.39$ to $13.60$, and from $10.96$ to $13.55$. For FFDN, the gains are larger still, with the average, across RTM, FindItAgain, and FindIt, of the best pixel-level F1 obtained on each dataset increasing from $21.07$ to $28.80$ at full scale, from $20.85$ to $26.94$ at half scale, from $20.51$ to $25.58$ at quarter scale, and from $20.24$ to $24.57$ at eighth scale. These results indicate that the benefit of higher-quality synthetic tampering does not disappear in the low-data regime and suggest that the gap may grow further with more scaling.

\begin{figure}[!htbp]
    \centering
    \includegraphics[width=\linewidth]{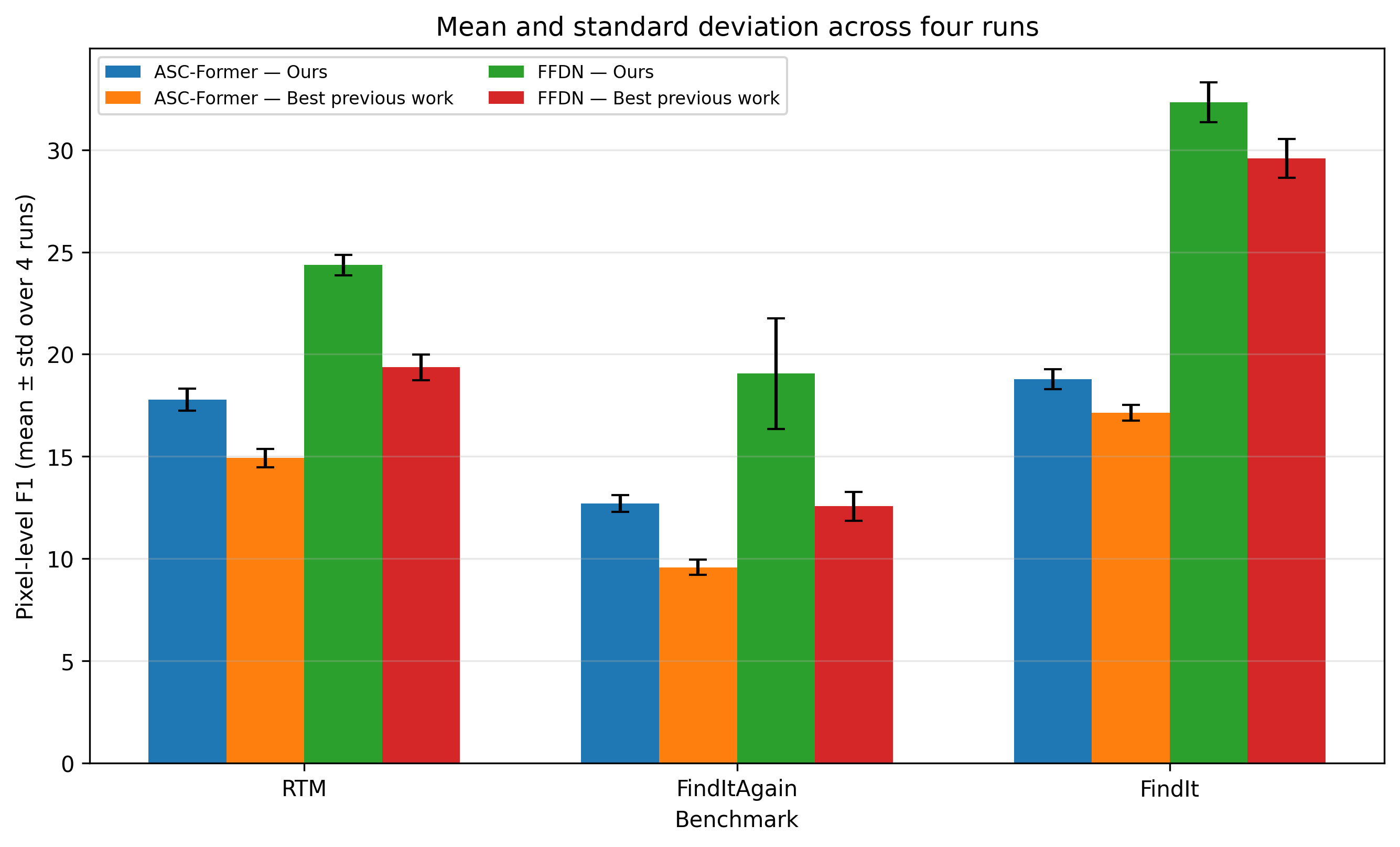}
    \caption{Performance stability across four independently generated training sets. We report the mean and standard deviation of the best pixel-level F1 obtained in each run for ASC-Former and FFDN on RTM, FindItAgain, and FindIt. For each dataset, we compare our method with the strongest previous pipeline, selected as the one with the highest mean among \cite{Doctamper,dataset2,newdataandopenopen}. We generate 700,000 images per run. Our data generation pipeline consistently achieves higher mean performance than the strongest prior pipeline, while the observed standard deviations remain small relative to the performance gap, indicating that the gains are robust to random regeneration.}
    \label{fig:quarter_stats}
\end{figure}

We also study the variability induced by the random generation process and present the results in \cref{fig:quarter_stats}. To this end, we focus on the quarter-size setting and regenerate the synthetic training set four times with different random seeds and different chosen source images (always same across generation pipelines), then retrain ASC-Former and FFDN on each version. The resulting mean and standard deviation confirm that the gains are stable. For ASC-Former, our method improves the pixel-level F1 from $14.92 \pm 0.45$ to $17.78 \pm 0.54$ on RTM, from $9.58 \pm 0.37$ to $12.69 \pm 0.41$ on FindItAgain, and from $17.14 \pm 0.39$ to $18.79 \pm 0.49$ on FindIt. For FFDN, the corresponding values increase from $19.36 \pm 0.63$ to $24.36 \pm 0.51$ on RTM, from $12.56 \pm 0.71$ to $19.05 \pm 2.70$ on FindItAgain, and from $29.59 \pm 0.95$ to $32.33 \pm 0.97$ on FindIt. In all cases, the margin between our method and the baseline is clearly larger than the observed run-to-run variation, which supports that the improvements are not due to a favorable random seed.

\begin{table}[!htbp]
    \centering
      \caption{Pixel-level F1 under different generated dataset sizes. For each run, we report the best F1 score. For each size, \emph{Best prior} denotes the strongest among \cite{Doctamper, dataset2,newdataandopenopen}.}
    \label{tab:size_scaling}
    \resizebox{\linewidth}{!}{%
    \setlength{\tabcolsep}{2pt}
    \footnotesize
    \begin{tabular}{llcccccccc}
        \toprule
        & & \multicolumn{4}{c}{Best prior} & \multicolumn{4}{c}{Ours} \\
        \cmidrule(lr){3-6} \cmidrule(lr){7-10}
        Benchmark & Model & 1/8 & 1/4 & 1/2 & Full & 1/8 & 1/4 & 1/2 & Full \\
        \midrule
        \multirow{3}{*}{RTM}
        & ASC-Former & 14.95 & 14.92 & 15.10 & 14.97 & \textbf{17.20} & \textbf{17.88} & \textbf{18.63} & \textbf{19.24} \\
        & DTD        &  9.88 & 10.09 & 10.46 & 10.82 & \textbf{12.71} & \textbf{12.79} & \textbf{12.98} & \textbf{13.11} \\
        & FFDN       & 19.29 & 19.38 & 19.47 & 19.51 & \textbf{23.49} & \textbf{24.58} & \textbf{24.46} & \textbf{24.87} \\
        \midrule
        \multirow{3}{*}{FindItAgain}
        & ASC-Former &  9.25 &  9.54 &  9.96 & 10.29 & \textbf{12.40} & \textbf{12.79} & \textbf{13.31} & \textbf{13.62} \\
        & DTD        & 11.10 & 11.35 & 11.63 & 11.83 & \textbf{13.17} & \textbf{13.18} & \textbf{13.27} & \textbf{13.21} \\
        & FFDN       & 12.52 & 12.57 & 12.68 & 12.72 & \textbf{16.53} & \textbf{18.99} & \textbf{22.90} & \textbf{26.53} \\
        \midrule
        \multirow{3}{*}{FindIt}
        & ASC-Former & 17.12 & 17.10 & 17.20 & 17.15 & \textbf{18.61} & \textbf{18.69} & \textbf{18.95} & \textbf{19.12} \\
        & DTD        & 11.91 & 12.73 & 13.49 & 14.40 & \textbf{14.78} & \textbf{14.82} & \textbf{14.90} & \textbf{14.88} \\
        & FFDN       & 28.92 & 29.57 & 30.41 & 30.97 & \textbf{33.68} & \textbf{33.18} & \textbf{33.46} & \textbf{35.00} \\
        \bottomrule
    \end{tabular}%
    }
\end{table}

\section{Per-Model and Per-Dataset Results}

\subsection{Complete comparison with previous data-generation methods}
\label{appendix:full-controlled-results}

\Cref{tab:out-of-domain-full-results} provides the complete controlled results for the data-generation methods of TextTamper~\cite{newdataandopenopen}, DocTamper~\cite{Doctamper}, FSTS~\cite{dataset2}, and ours. The condensed comparison in \cref{tab:out-of-domain-results} selects the strongest previous generator independently for image-level and pixel-level F1. We form the FindItAgain masks from the annotated modified rectangles and exclude the original copy-source regions.

\noindent\textbf{Controlled generation protocol.}
To ensure a fair comparison, we implemented the three previous data-generation approaches and our method within a shared tree-structured generation script. For a given source document and random seed, decisions common to multiple approaches are sampled once and reused across their corresponding branches. This ensures that shared stochastic choices lead to the same outcomes, while decisions and operations specific to each approach remain governed by its original generation rules. This design does not eliminate randomness, but controls the variability introduced by shared random decisions for a fixed seed, allowing differences in performance to be more directly attributed to the generation strategies rather than unrelated sampling choices. Consequently, approaches sharing similar generation operations can produce more comparable training data. 

\begin{table*}[!t]
\centering
\caption{Out-of-domain document tampering detection results on RTM, FindItAgain, and FindIt for several models trained on data generated using different data generation approaches. P, R, and F1 correspond to precision, recall, and F1 score, while Image and Pixel denote image-level classification and pixel-level localization performance. Note that the source images used to generate the training data are the same across models and data generation methods as described in \cref{sec:data}, and each detector uses the same training protocol across generators as described in \cref{sec:training}.
}

\label{tab:out-of-domain-full-results}
{\small
\setlength{\tabcolsep}{1mm}
\resizebox{\textwidth}{!}{%
\begin{tabular}{l|l|ccc|ccc|ccc|ccc|ccc|ccc}
\toprule
\multirow{2}{*}{\textbf{Model}} & \multirow{2}{*}{\textbf{Data Gen.}} &
\multicolumn{6}{c|}{\textbf{RTM}} &
\multicolumn{6}{c|}{\textbf{FindItAgain}} &
\multicolumn{6}{c}{\textbf{FindIt}} \\
\cmidrule(lr){3-8}\cmidrule(lr){9-14}\cmidrule(lr){15-20}
 & & \multicolumn{3}{c|}{Image} & \multicolumn{3}{c|}{Pixel} &
 \multicolumn{3}{c|}{Image} & \multicolumn{3}{c|}{Pixel} &
 \multicolumn{3}{c|}{Image} & \multicolumn{3}{c}{Pixel} \\
 & & P & R & F1 & P & R & F1 &
 P & R & F1 & P & R & F1 &
 P & R & F1 & P & R & F1 \\
\midrule
\multirow{4}{*}{PSCC-Net}
& \cite{dataset2} & 37.9 & 29.1 & 32.9 & 7.4 & 9.8 & 8.4 & 47.8 & 31.4 & 37.9 & 8.6 & 9.1 & 8.8 & 60.0 & 52.5 & 56.0 & 11.0 & 11.7 & 11.3 \\

& \cite{Doctamper} & 38.5 & 27.9 & 32.3 & 7.5 & 10.0 & 8.6 & 47.8 & 31.4 & 37.9 & 8.7 & 9.2 & 8.9 & 61.4 & 53.8 & 57.3 & 11.2 & 11.9 & 11.5 \\

& \cite{newdataandopenopen} & 33.8 & 34.1 & 33.9 & 7.9 & 8.1 & 8.0 & 52.0 & 37.1 & 43.3 & 7.5 & 8.7 & 8.1 & 60.9 & 52.5 & 56.4 & 8.9 & 9.7 & 9.3 \\

& Ours & \textbf{39.3} & \textbf{40.2} & \textbf{39.7} & \textbf{10.5} & \textbf{10.8} & \textbf{10.6} & \textbf{56.8} & \textbf{60.0} & \textbf{58.3} & \textbf{10.6} & \textbf{12.8} & \textbf{11.6} & \textbf{62.9} & \textbf{55.0} & \textbf{58.7} & \textbf{12.0} & \textbf{12.1} & \textbf{12.0} \\
\midrule
\multirow{4}{*}{CAT-Net}
& \cite{dataset2} & 36.0 & 32.2 & 34.0 & 8.6 & 9.9 & 9.2 & 37.0 & 28.6 & 32.3 & 10.4 & 5.2 & 6.9 & 47.1 & 40.0 & 43.2 & 10.1 & 9.9 & 10.0 \\

& \cite{Doctamper} & 36.5 & 32.7 & 34.5 & 8.7 & 10.1 & 9.3 & 37.0 & 28.6 & 32.3 & 9.5 & 4.9 & 6.5 & 47.8 & 41.2 & 44.3 & 10.3 & 10.1 & 10.2 \\
& \cite{newdataandopenopen} & 36.5 & 31.5 & 33.8 & 8.9 & 9.4 & 9.1 & 34.4 & 31.4 & 32.8 & 10.7 & 5.2 & 7.0 & 42.1 & 40.0 & 41.0 & 11.8 & 8.3 & 9.7 \\
& Ours & \textbf{40.0} & \textbf{35.0} & \textbf{37.3} & \textbf{9.2} & \textbf{11.0} & \textbf{10.0} & \textbf{48.0} & \textbf{34.3} & \textbf{40.0} & \textbf{11.0} & \textbf{5.6} & \textbf{7.4} & \textbf{51.5} & \textbf{42.5} & \textbf{46.6} & \textbf{13.0} & \textbf{10.5} & \textbf{11.7} \\
\midrule
\multirow{4}{*}{DTD}
& \cite{dataset2} & 36.5 & 52.1 & 42.9 & 10.8 & 10.0 & 10.4 & 48.1 & 37.1 & 41.9 & 9.8 & 9.9 & 9.9 & 56.4 & 55.0 & 55.7 & 13.6 & 13.5 & 13.5 \\
& \cite{Doctamper} & 36.4 & 52.0 & 42.8 & 9.8 & 10.2 & 10.0 & 48.1 & 37.1 & 41.9 & 8.8 & 10.2 & 9.4 & 55.6 & 56.2 & 55.9 & 13.6 & 13.9 & 13.7 \\
& \cite{newdataandopenopen} & 36.5 & 50.8 & 42.5 & 10.8 & 7.6 & 8.9 & 42.9 & 34.3 & 38.1 & 9.8 & 7.0 & 8.2 & 46.9 & 37.5 & 41.7 & 11.9 & 7.4 & 9.1 \\
& Ours &
\textbf{36.6} & \textbf{73.3} & \textbf{48.9} & \textbf{14.9} & \textbf{11.4} & \textbf{12.9} & \textbf{50.0} & \textbf{40.0} & \textbf{44.4} & \textbf{12.5} & \textbf{12.4} & \textbf{12.5} & \textbf{62.3} & \textbf{66.2} & \textbf{64.2} & \textbf{14.0} & \textbf{14.4} & \textbf{14.2} \\
\midrule
\multirow{4}{*}{ADCD-Net}
& \cite{dataset2} & 41.2 & 47.9 & 44.3 & 13.0 & 14.1 & 13.5 & 42.9 & 51.4 & 46.8 & 10.8 & 11.3 & 11.0 & 76.1 & 63.8 & 69.4 & 17.5 & 17.0 & 17.2 \\
& \cite{Doctamper} & 41.8 & 48.6 & 45.0 & 12.5 & 14.3 & 13.3 & 43.9 & 51.4 & 47.4 & 11.0 & 11.5 & 11.2 & 77.6 & 65.0 & 70.8 & 17.8 & 17.3 & 17.6 \\
& \cite{newdataandopenopen} & 39.3 & 47.1 & 42.8 & 12.2 & 10.9 & 11.5 & 39.1 & 51.4 & 44.4 & 10.7 & 11.2 & 10.9 & 77.3 & 63.8 & 69.9 & 17.5 & 17.0 & 17.2 \\
& Ours & \textbf{47.0} & \textbf{51.7} & \textbf{49.2} & \textbf{24.9} & \textbf{15.7} & \textbf{19.3} & \textbf{52.3} & \textbf{65.7} & \textbf{58.2} & \textbf{13.6} & \textbf{14.9} & \textbf{14.2} & \textbf{82.8} & \textbf{66.2} & \textbf{73.6} & \textbf{19.1} & \textbf{17.9} & \textbf{18.5} \\
\midrule
\multirow{4}{*}{ASC-Former}
& \cite{dataset2} & 40.1 & 29.6 & 34.1 & 15.2 & 13.4 & 14.2 & 36.2 & 60.0 & 45.2 & 10.6 & 9.7 & 10.1 & 53.7 & 63.8 & 58.3 & 24.5 & 12.0 & 16.1 \\
& \cite{Doctamper} & 40.7 & 30.0 & 34.5 & 15.4 & 13.6 & 14.4 & 36.8 & 60.0 & 45.7 & 10.8 & 9.8 & 10.2 & 54.7 & 65.0 & 59.4 & 24.9 & 12.2 & 16.4 \\
& \cite{newdataandopenopen} & 35.8 & 26.1 & 30.2 & 13.9 & 12.0 & 12.9 & 31.1 & 54.3 & 39.6 & 9.4 & 8.8 & 9.1 & 46.9 & 57.5 & 51.7 & 22.4 & 10.4 & 14.2 \\
& Ours & \textbf{44.3} & \textbf{32.3} & \textbf{37.4} & \textbf{23.4} & \textbf{16.1} & \textbf{19.1} & \textbf{38.6} & \textbf{62.9} & \textbf{47.8} & \textbf{15.3} & \textbf{12.3} & \textbf{13.6} & \textbf{69.6} & \textbf{68.8} & \textbf{69.2} & \textbf{30.1} & \textbf{13.9} & \textbf{19.0} \\
\midrule
\multirow{4}{*}{FFDN}
& \cite{dataset2} & 45.4 & 46.2 & 45.8 & 19.5 & 16.8 & 18.0 & 51.4 & 51.4 & 51.4 & 13.6 & 10.6 & 11.9 & 76.2 & 60.0 & 67.1 & 28.2 & 25.1 & 26.6 \\
& \cite{Doctamper}
& 46.1 & 46.9 & 46.5 & 19.8 & 17.1 & 18.4 & 47.5 & 54.3 & 50.7 & 13.0 & 8.7 & 10.4 & 77.8 & 61.3 & 68.5 & 28.6 & 25.5 & 27.0 \\
& \cite{newdataandopenopen} & 42.7 & 45.4 & 44.0 & 18.1 & 14.6 & 16.2 & 52.0 & 37.1 & 43.3 & 12.0 & 10.6 & 11.3 & 74.5 & 47.5 & 58.0 & 30.2 & 23.9 & 26.7 \\
& Ours & \textbf{46.4} & \textbf{51.0} & \textbf{48.6} & \textbf{28.2} & \textbf{20.6} & \textbf{23.8} & \textbf{69.0} & \textbf{57.1} & \textbf{62.5} & \textbf{31.6} & \textbf{21.4} & \textbf{25.5} & \textbf{82.3} & \textbf{70.0} & \textbf{75.7} & \textbf{36.3} & \textbf{28.3} & \textbf{31.8} \\
\midrule

\multirow{4}{*}{DITL$^2$}
& \cite{dataset2}
& 57.4 & 15.9 & 24.9 & 13.9 & 12.1 & 12.9
& 40.0 & 40.0 & 40.0 & 12.1 & 9.7 & 10.8
& 61.1 & 68.8 & 64.7 & 18.9 & 13.9 & 16.0 \\
& \cite{Doctamper}
& 54.7 & 12.2 & 20.0 & 12.1 & \textbf{12.9} & 12.5
& 35.0 & 40.0 & 37.3 & 12.9 & 6.3 & 8.5
& 65.9 & 72.5 & 69.0 & 19.3 & 14.4 & 16.5 \\
& \cite{newdataandopenopen}
& 53.7 & 13.7 & 21.8 & 11.5 & 12.0 & 11.7
& 33.3 & 37.1 & 35.1 & 11.6 & 6.9 & 8.7
& 65.9 & 70.0 & 67.9 & 18.4 & 14.0 & 15.9 \\
& Ours
& \textbf{61.0} & \textbf{16.5} & \textbf{26.0} & \textbf{18.3} & 12.6 & \textbf{14.9}
& \textbf{44.4} & \textbf{45.7} & \textbf{45.1} & \textbf{13.2} & \textbf{10.4} & \textbf{11.6}
& \textbf{67.0} & \textbf{73.8} & \textbf{70.2} & \textbf{20.7} & \textbf{14.7} & \textbf{17.1} \\

\bottomrule
\end{tabular}}}

\end{table*}

\subsection{Image-Level Comparison with Generator-Based Datasets}
\label{appendix:generator-image-results}

\Cref{tab:generator-image-f1} reports the image-level
F1 scores obtained by four architectures pretrained on
our generated dataset or the generator-based datasets
described in \cref{appendix:generator-based-datasets}.
Our approach achieves higher image-level F1 across all
four architectures and all three human-made benchmarks.
These findings complement the pixel-level results in
\cref{tab:doctamper-full}, showing that the benefits
of our generated training data extend to document-level
tampering detection.

\begin{table}[!t]
\centering
\small
\setlength{\tabcolsep}{4pt}
\caption{Image-level F1 (\%) when pretrained on our generated dataset or the generator-based datasets described in \cref{appendix:generator-based-datasets}.}
\label{tab:generator-image-f1}
\begin{tabular}{llccc}
\toprule
\textbf{Model} & \textbf{Training data}
& \textbf{RTM} & \textbf{FindItAgain} & \textbf{FindIt} \\
\midrule
\multirow{2}{*}{ASC-Former}
& Generator-based & 32.8 & 44.9 & 37.9 \\
& Ours & 37.4 & 47.8 & 69.2 \\
\midrule
\multirow{2}{*}{FFDN}
& Generator-based & 35.7 & 26.7 & 75.0 \\
& Ours & 48.6 & 62.5 & 75.7 \\
\midrule
\multirow{2}{*}{ADCD-Net}
& Generator-based & 44.8 & 51.9 & 48.1 \\
& Ours & 49.2 & 58.2 & 73.6 \\
\midrule
\multirow{2}{*}{DITL$^2$}
& Generator-based & 13.1 & 9.1 & 48.7 \\
& Ours & 26.0 & 45.1 & 70.2 \\
\bottomrule
\end{tabular}
\end{table}

\subsection{Fine-Tuning Results}
\label{appendix:finetuningresultsdetails}
We report the average fine-tuning performance across all datasets and models in \cref{tab:finetuningresults}, and provide detailed per-dataset results in \cref{tab:finetune-per-dataset}. These results confirm that our data generation method provides the strongest average performance across the evaluated datasets and models.

\begin{table*}[!tp]
\centering
\caption{Document tampering detection and localization results after fine-tuning.
Models are pretrained on synthetic data generated by different approaches and then fine-tuned on each target dataset’s training split. Results are reported on RTM, FindItAgain, and FindIt. P, R, and F1 correspond to precision, recall, and F1 score, while Image and Pixel denote image-level classification and pixel-level localization performance.}
\label{tab:finetune-per-dataset}
\resizebox{\linewidth}{!}{%
\setlength{\tabcolsep}{0.8pt}
\tiny
\begin{tabular}{l|l|ccc|ccc|ccc|ccc|ccc|ccc}
\toprule
\multirow{2}{*}{\textbf{Model}} & \multirow{2}{*}{\textbf{Data Gen.}} &
\multicolumn{6}{c|}{\textbf{RTM}} & 
\multicolumn{6}{c|}{\textbf{FindItAgain}} & 
\multicolumn{6}{c}{\textbf{FindIt}} \\
\cmidrule(lr){3-8}\cmidrule(lr){9-14}\cmidrule(lr){15-20}
 & & \multicolumn{3}{c|}{Image} & \multicolumn{3}{c|}{Pixel} &
 \multicolumn{3}{c|}{Image} & \multicolumn{3}{c|}{Pixel} &
 \multicolumn{3}{c|}{Image} & \multicolumn{3}{c}{Pixel} \\
 & & P & R & F1 & P & R & F1 &
 P & R & F1 & P & R & F1 &
 P & R & F1 & P & R & F1 \\
\midrule
\multirow{4}{*}{PSCC-Net}
& FSTS &
50.3 & 90.1 & 64.6 & 15.5 & 18.2 & 16.7 &
61.3 & 54.3 & 57.6 & 14.4 & 31.0 & 19.7 &
74.4 & 72.5 & 73.4 & 34.9 & 34.1 & 34.5 \\
& DocTamper 
& 54.8 & 88.5 & 67.7 & 15.2 & 17.8 & 16.4
& 59.4 & 54.3 & 56.7 & 14.1 & 31.3 & 19.4
& 74.7 & 73.7 & 74.2 & 35.3 & 34.4 & 34.8 \\
& TextTamper 
& 52.1 & 88.2 & 65.5 & 14.3 & 16.7 & 15.4
& 59.4 & 54.3 & 56.7 & 13.3 & 28.7 & 18.2
& 71.4 & 68.8 & 70.1 & 32.4 & 34.2 & 33.3 \\
& Ours 
& 53.3 & 90.2 & 67.0 & 15.9 & 20.4 & 17.9
& 73.1 & 54.3 & 62.3 & 30.1 & 22.6 & 25.8
& 71.6 & 72.5 & 72.1 & 34.3 & 32.7 & 33.5 \\
\midrule
\multirow{4}{*}{CAT-Net}
 & FSTS &
63.1 & 76.5 & 69.2 & 24.1 & 18.1 & 20.7 &
36.1 & 37.1 & 36.6 & 6.4 & 5.7 & 6.0 &
61.6 & 76.2 & 68.2 & 48.8 & 27.8 & 35.5 \\
& DocTamper 
& 63.7 & 75.5 & 69.1 & 23.6 & 18.3 & 20.6
& 35.1 & 37.1 & 36.1 & 6.3 & 5.6 & 5.9
& 60.2 & 77.5 & 67.8 & 47.8 & 27.3 & 34.8 \\
& TextTamper 
& 57.1 & 72.2 & 63.8 & 24.1 & 18.2 & 20.7
& 33.3 & 34.3 & 33.8 & 6.7 & 5.8 & 6.2
& 75.9 & 75.0 & 75.5 & 48.9 & 27.6 & 35.3 \\
& Ours 
& 59.6 & 77.6 & 67.4 & 26.1 & 18.1 & 21.4
& 77.8 & 40.0 & 52.8 & 9.4 & 7.2 & 8.1
& 78.6 & 68.7 & 73.3 & 52.0 & 28.9 & 37.2 \\
\midrule
\multirow{4}{*}{DTD}
&  FSTS &
65.4 & 76.3 & 70.5 & 21.4 & 18.7 & 20.0 &
75.0 & 25.7 & 38.3 & 16.6 & 13.1 & 14.6 &
79.2 & 76.2 & 77.7 & 40.4 & 46.5 & 43.2 \\
& DocTamper 
& 64.1 & 77.1 & 70.0 & 21.0 & 18.3 & 19.6
& 75.0 & 25.7 & 38.3 & 16.3 & 13.2 & 14.6
& 77.9 & 75.0 & 76.4 & 40.1 & 46.4 & 43.0 \\
& TextTamper 
& 64.8 & 72.1 & 68.3 & 21.7 & 19.5 & 20.5
& 61.1 & 31.4 & 41.5 & 15.4 & 12.9 & 14.0
& 68.9 & 77.5 & 72.9 & 39.2 & 45.1 & 41.9 \\
& Ours 
& 67.5 & 73.9 & 70.6 & 23.0 & 22.0 & 22.5 
& 80.0 & 34.3 & 48.0 & 16.9 & 13.0 & 14.7
& 83.8 & 77.5 & 80.5 & 40.5 & 46.6 & 43.3 \\
\midrule
\multirow{4}{*}{ADCD-Net}
&  FSTS &
72.0 & 53.2 & 61.2 & 28.0 & 29.2 & 28.5 &
58.1 & 51.4 & 54.5 & 16.9 & 14.5 & 15.6 &
88.3 & 85.0 & 86.6 & 35.4 & 39.3 & 37.2 \\
& DocTamper 
& 72.7 & 52.2 & 60.8 & 28.3 & 28.6 & 28.4 &
56.3 & 51.4 & 53.7 & 11.7 & 14.2 & 12.8
& 88.2 & 83.8 & 85.9 & 34.7 & 38.5 & 36.5  \\
& TextTamper 
& 62.7 & 57.2 & 59.8 & 27.3 & 25.6 & 26.4 &
60.0 & 51.4 & 55.4 & 15.7 & 14.3 & 15.0
& 77.4 & 81.2 & 79.3 & 33.1 & 35.0 & 34.0  \\
& Ours 
& 60.0 & 67.4 & 63.5 & 27.9 & 29.5 & 28.6 
& 64.9 & 68.6 & 66.7 & 27.1 & 21.2 & 23.8
& 88.3 & 85.0 & 86.6 & 38.8 & 41.3 & 40.0 \\
\midrule
\multirow{4}{*}{ASC-Former}
& FSTS
& 51.0 & 46.0 & 48.4 & 26.0 & 27.0 & 26.6
& 63.6 & 40.0 & 49.1 & 12.5 & 14.9 & 13.6
& 84.0 & 78.8 & 81.3 & 55.0 & 39.0 & 45.6 \\
& DocTamper 
& 52.6 & 47.4 & 49.9 & 27.6 & 25.5 & 26.5 &
63.6 & 40.0 & 49.1 & 10.4 & 14.6 & 12.1
& 81.2 & 81.3 & 81.2 & 56.6 & 40.8 & 47.4  \\
& TextTamper 
& 43.0 & 33.1 & 37.4 & 20.5 & 24.9 & 22.5
& 61.9 & 37.1 & 46.4 & 10.2 & 13.2 & 11.5
& 75.0 & 75.0 & 75.0 & 48.5 & 44.3 & 46.3 \\
& Ours 
& 51.7 & 46.3 & 48.9 & 25.7 & 27.7 & 26.7 
& 63.6 & 40.0 & 49.1 & 34.4 & 15.1 & 21.0
& 89.7 & 76.3 & 82.4 & 56.1 & 39.6 & 46.4 \\
\midrule
\multirow{4}{*}{FFDN}
& FSTS
& 78.0 & 70.5 & 74.1 & 35.0 & 37.5 & 36.2
& 60.0 & 60.0 & 60.0 & 35.5 & 22.5 & 27.5
& 88.6 & 87.5 & 88.1 & 62.0 & 46.0 & 52.8 \\
& DocTamper 
& 76.6 & 70.6 & 73.5 & 33.9 & 37.1 & 35.4
& 58.3 & 60.0 & 59.2 & 33.4 & 20.6 & 25.5
& 86.1 & 85.0 & 85.5 & 59.6 & 43.8 & 50.5 \\
& TextTamper 
& 70.1 & 65.2 & 67.6 & 31.2 & 36.4 & 33.6
& 61.8 & 60.0 & 60.9 & 34.4 & 22.6 & 27.3
& 83.8 & 83.8 & 83.8 & 57.5 & 44.9 & 50.4 \\
& Ours 
& 78.6 & 75.1 & 76.8 & 36.1 & 39.0 & 37.5
& 80.8 & 60.0 & 68.9 & 46.4 & 29.8 & 36.3
& 93.5 & 90.0 & 91.7 & 71.9 & 46.9 & 56.8 \\

\bottomrule
\end{tabular}%
}

\end{table*}

\noindent\textbf{DITL$^2$ and fine-tuning.}
 We exclude DITL$^2$~\cite{li2025ditl2} from the target-dataset fine-tuning experiment because it is specifically designed for cross-domain generalization. Its original framework already incorporates a dedicated parameter adaptation stage, FDPA, to transfer learned tampering features while mitigating overfitting. We retain this stage when training DITL$^2$ on the synthetic datasets and evaluate its generalization to unseen benchmarks without additional target-dataset fine-tuning. Adapting it further to the evaluation datasets would assess a different, target-domain adaptation setting. We therefore restrict the separate target-dataset fine-tuning comparison to the other six architectures.

\subsection{Image-Level Evaluation of DITL$^2$}
\label{appendix:ditl-image-eval}

 The official DITL$^2$~\cite{li2025ditl2} implementation, which we used in our study, does not have an image-level classification head and only predicts pixel-level tampering probabilities. We therefore derive image-level predictions from its pixel probability maps. An image is considered tampered if its ground-truth mask contains at least one positive pixel.

\subsection{Ablation Results}
\label{appendix:ablation}
We report the average ablation results across all datasets and models in \cref{tab:ablation-f1-permodel}, and provide detailed per-dataset results in \cref{tab:ablation-per-dataset-results}. Removing $\mathcal{F}_\theta$ permits more visually mismatched source-target pairs, while removing $\mathcal{G}_\theta$ permits more imprecise crops. Both changes reduce downstream localization performance, and removing both components produces the largest degradation. These results support that the two networks address distinct and complementary failure modes.

\section{Threshold-Independent Evaluation}
\label{appendix:threshold-metrics}

The main evaluation reports pixel-level precision, recall, and F1. To compare the two supervision sources independently of the decision threshold, we additionally evaluate six of the seven detectors, excluding DITL$^2$, using pixel AUROC and pixel AUPRC on RTM, FindItAgain, and FindIt.

\begin{table}[!tp]
    \centering
    \caption{Pixel AUROC across the six localization models and three human-made benchmarks. Each model is trained separately using data generated by the indicated approach. \emph{Best prior} denotes the strongest among \cite{Doctamper,dataset2,newdataandopenopen}.}
    \label{tab:pixel-auroc}
    \resizebox{\columnwidth}{!}{
    \begin{tabular}{l|cc|cc|cc}
        \toprule
        \multirow{2}{*}{\textbf{Model}} & \multicolumn{2}{c|}{\textbf{RTM}} &
        \multicolumn{2}{c|}{\textbf{FindItAgain}} & \multicolumn{2}{c}{\textbf{FindIt}} \\
        & \textbf{Best prior} & \textbf{Ours} & \textbf{Best prior} & \textbf{Ours}
          & \textbf{Best prior} & \textbf{Ours} \\
        \midrule
        CAT-Net & 60.0 & 60.7 & 58.4 & 58.8 & 60.2 & 61.5 \\
        PSCC-Net & 61.6 & 63.8 & 63.4 & 66.8 & 63.9 & 64.4 \\
        DTD & 68.9 & 72.7 & 67.0 & 70.8 & 72.8 & 73.5 \\
        ASC-Former & 63.9 & 67.8 & 63.9 & 67.7 & 62.5 & 64.1 \\
        FFDN & 70.8 & 75.9 & 66.7 & 82.2 & 72.7 & 76.1 \\
        ADCD-Net & 69.1 & 75.9 & 78.1 & 84.5 & 74.9 & 76.0 \\
        \midrule
        Average & 65.7 & 69.5 & 66.2 & 71.8 & 67.8 & 69.3 \\
        \bottomrule
    \end{tabular}}
\end{table}

\begin{table}[!tp]
    \centering
    \caption{Pixel AUPRC across the six localization models and three human-made benchmarks. Each model is trained separately using data generated by the indicated approach. \emph{Best prior} denotes the strongest among \cite{Doctamper,dataset2,newdataandopenopen}. The positive-pixel rate, which is the pixel AUPRC of a random predictor, is $0.36\%$ on RTM, $0.05\%$ on FindItAgain, and $0.43\%$ on FindIt.}
    \label{tab:pixel-auprc}
    \resizebox{\columnwidth}{!}{
    \begin{tabular}{l|cc|cc|cc}
        \toprule
        \multirow{2}{*}{\textbf{Model}} & \multicolumn{2}{c|}{\textbf{RTM}} &
        \multicolumn{2}{c|}{\textbf{FindItAgain}} & \multicolumn{2}{c}{\textbf{FindIt}} \\
        & \textbf{Best prior} & \textbf{Ours} & \textbf{Best prior} & \textbf{Ours}
          & \textbf{Best prior} & \textbf{Ours} \\
        \midrule
        CAT-Net & 4.3 & 4.8 & 1.6 & 1.7 & 5.0 & 6.1 \\
        PSCC-Net & 3.3 & 4.5 & 2.6 & 3.9 & 5.4 & 5.8 \\
        DTD & 5.0 & 7.0 & 3.6 & 5.2 & 8.5 & 9.0 \\
        ASC-Former & 6.1 & 9.5 & 2.2 & 3.4 & 7.1 & 8.9 \\
        FFDN & 11.8 & 17.5 & 4.3 & 13.9 & 19.4 & 25.0 \\
        ADCD-Net & 6.9 & 12.1 & 6.2 & 9.0 & 12.5 & 13.5 \\
        \midrule
        Average & 6.2 & 9.2 & 3.4 & 6.2 & 9.7 & 11.4 \\
        \bottomrule
    \end{tabular}}
\end{table}

\Cref{tab:pixel-auroc,tab:pixel-auprc} show that the improvement is not an artefact of the operating point. Averaged over all model--benchmark combinations, models trained on our data reach 70.2 pixel AUROC and 8.9 pixel AUPRC, against 66.6 and 6.4 for the best prior generation approach, and they are ahead in every one of the eighteen model--benchmark combinations under both metrics. The relative margin is larger under pixel AUPRC, at 39\% compared with 5\% for pixel AUROC. Fewer than one percent of the pixels are tampered on all three benchmarks, so pixel AUPRC places greater emphasis on the high-precision operating region. FFDN and ADCD-Net are the two strongest architectures under both threshold-independent metrics.

\section{Generalization to AI-Generated Tampering}
\label{appendix:ai-tampering}

\paragraph{Evaluation setup.}
To assess whether localization models trained on synthetic data generalize to
manipulation mechanisms absent from training, we construct three test sets from
held-out pristine pages of our source corpus using FLUX-Text~\cite{Flux-text},
AnyText~\cite{Anytext}, and Qwen-Image-Edit~\cite{wu2025qwen}. Each test set
contains one manipulated crop per source document together with the
corresponding unmodified crop, giving 500 manipulated and 500 pristine crops for
FLUX-Text, 500 and 500 for AnyText, and 500 and 500 for Qwen-Image-Edit. For each page, an OCR pass proposes candidate word boxes, from which one box is
sampled. The editor is prompted to replace the text in that box with a string of
the same length in which every alphanumeric character is replaced by a different
character. The edited output is composited into the page through the sampled box
dilated by three pixels, and that entire compositing region defines the
ground-truth mask; pixels outside it retain their original values. A second OCR pass checks whether the transcription of the edited region changes. Cases with an unchanged transcription are regenerated once. FLUX-Text is run for 28 denoising steps and AnyText for 20 steps with a guidance
scale of 9.0. For Qwen-Image-Edit we use the Qwen-Image-Edit-2511 checkpoint,
run for 40 denoising steps with a classifier-free guidance scale
(\texttt{true\_cfg\_scale}) of 4.0 and a 4-bit quantized transformer. All crops are evaluated in a window of
1024 pixels, at native resolution and without rescaling. We report pixel-level precision,
recall, and F1 for the three strongest architectures of
\cref{tab:out-of-domain-results}, namely ASC-Former, FFDN, and ADCD-Net,
pretrained either on data generated by our approach or by the best prior
generation method, and evaluated without fine-tuning.
\begin{table}[!tp]
    \centering
    \caption{Generalization to unseen AI-generated text manipulations.
    ASC-Former, FFDN, and ADCD-Net are pretrained on data generated by our
    approach or by the best prior generation method, and evaluated without
    fine-tuning. \emph{Best prior} denotes the strongest among
    \cite{Doctamper,dataset2,newdataandopenopen}. We report
    pixel-level precision, recall, and F1; the higher F1 of each pair of rows is
    shown in bold. The best prior generation method is selected separately for each generator--architecture pair.}
    \label{tab:ai-tampering}
    \resizebox{\columnwidth}{!}{
    \begin{tabular}{ll|ccc|ccc|ccc}
        \toprule
        \multirow{2}{*}{\textbf{Generator}} & \multirow{2}{*}{\textbf{Data Gen.}} &
        \multicolumn{3}{c|}{\textbf{ASC-Former}} &
        \multicolumn{3}{c|}{\textbf{FFDN}} &
        \multicolumn{3}{c}{\textbf{ADCD-Net}} \\
        & & \textbf{P} & \textbf{R} & \textbf{F1}
          & \textbf{P} & \textbf{R} & \textbf{F1}
          & \textbf{P} & \textbf{R} & \textbf{F1} \\
        \midrule
        \multirow{2}{*}{FLUX-Text}
        & Best prior & 95.9 & 51.4 & 66.9 & 87.2 & 72.1 & 78.9 & 83.9 & 71.8 & 77.4 \\
        & Ours & 82.4 & 63.6 & \textbf{71.8} & 90.1 & 75.7 & \textbf{82.3} & 86.3 & 72.8 & \textbf{79.0} \\
        \midrule
        \multirow{2}{*}{AnyText}
        & Best prior & 95.7 & 54.9 & 69.8 & 90.5 & 74.4 & 81.7 & 81.7 & 69.2 & \textbf{74.9} \\
        & Ours & 81.7 & 63.7 & \textbf{71.6} & 91.9 & 76.0 & \textbf{83.2} & 83.3 & 67.9 & 74.8 \\
        \midrule
        \multirow{2}{*}{Qwen-Image-Edit}
        & Best prior & 96.4 & 59.9 & 73.9 & 87.8 & 82.3 & 85.0 & 89.0 & 81.7 & 85.2 \\
        & Ours & 84.6 & 66.7 & \textbf{74.6} & 91.0 & 83.1 & \textbf{86.9} & 92.7 & 80.1 & \textbf{85.9} \\
        \bottomrule
    \end{tabular}}
\end{table}

On these AI-generated test sets, averaging the three generator-specific scores gives 84.1
for FFDN, 79.9 for ADCD-Net, and 72.7 for ASC-Former, against 81.9, 79.2, and
70.2 respectively for the best prior generation method. Our supervision leads in eight of the nine
generator--architecture pairs, the only exception being ADCD-Net on AnyText,
where it trails by 0.1 F1.

The margins are narrower than those observed on the human-made benchmarks of
\cref{tab:out-of-domain-results}. This is mainly because these cases are easier. For models pretrained on our data, Qwen-Image-Edit yields the highest F1
for every architecture, while the lowest-F1 editor is FLUX-Text for FFDN and
AnyText for ASC-Former and ADCD-Net. This ordering does not follow visual
quality: Qwen-Image-Edit produces the most realistic edits of the three,
preserving the sharpness and the local contrast of the original text more
closely, which suggests that its regeneration leaves traces that are more
visible to the detectors than to a human observer. Overall,
supervision generated by our approach transfers to manipulation mechanisms that
are absent from training, and transfers better than supervision from prior
generation pipelines.

\begin{figure*}[!tp]
  \centering
  \includegraphics[height=0.86\textheight]{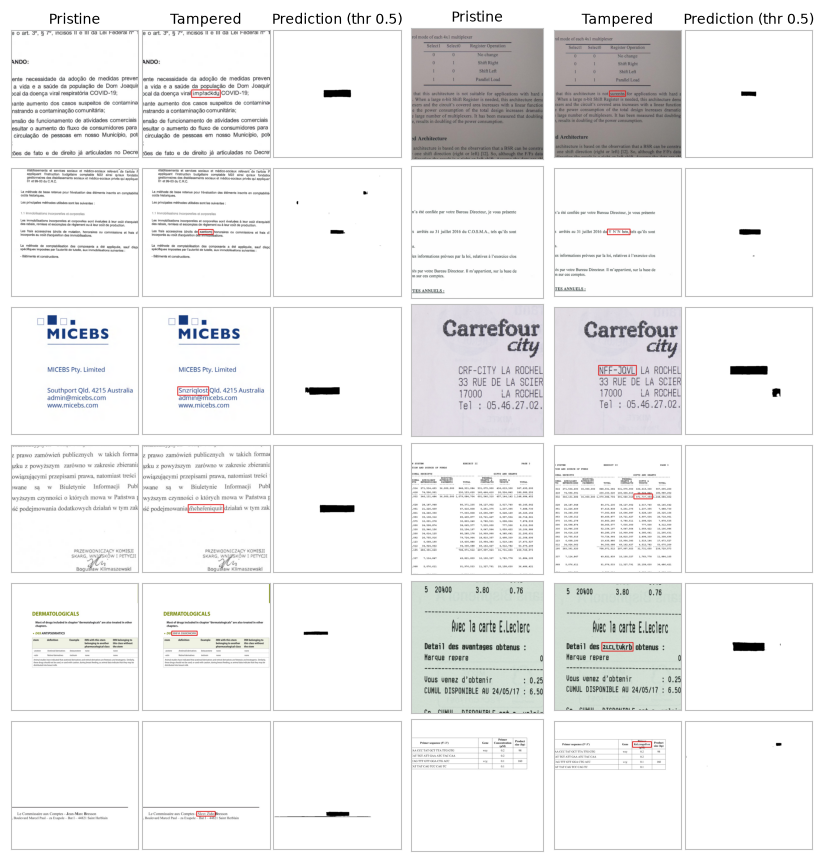}
  \caption{Twelve manipulations from the FLUX-Text test set, with the
  prediction of FFDN pretrained on our data. Columns: pristine crop,
  tampered crop with the ground-truth region outlined, predicted tampered region at a $0.5$ threshold.}
  \label{fig:qual-fluxtext-ffdn}
\end{figure*}

\begin{figure*}[!tp]
  \centering
  \includegraphics[height=0.86\textheight]{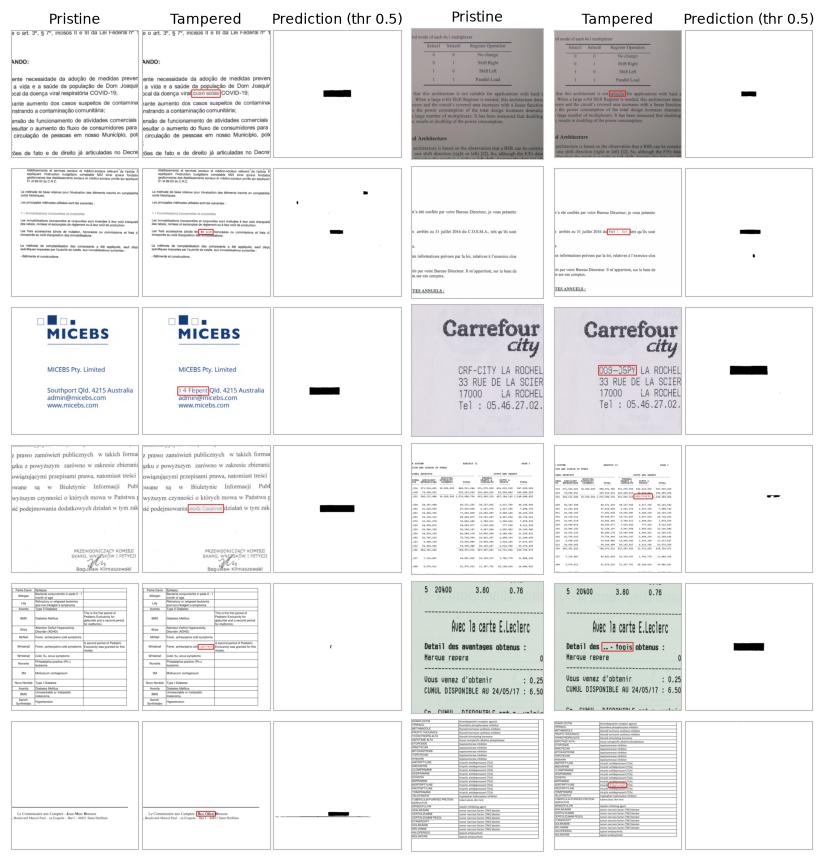}
  \caption{Twelve manipulations from the AnyText test set, with the
  prediction of FFDN pretrained on our data. Columns: pristine crop,
  tampered crop with the ground-truth region outlined, predicted tampered region at a $0.5$ threshold.}
  \label{fig:qual-anytext-ffdn}
\end{figure*}

\begin{figure*}[!tp]
  \centering
  \includegraphics[height=0.86\textheight]{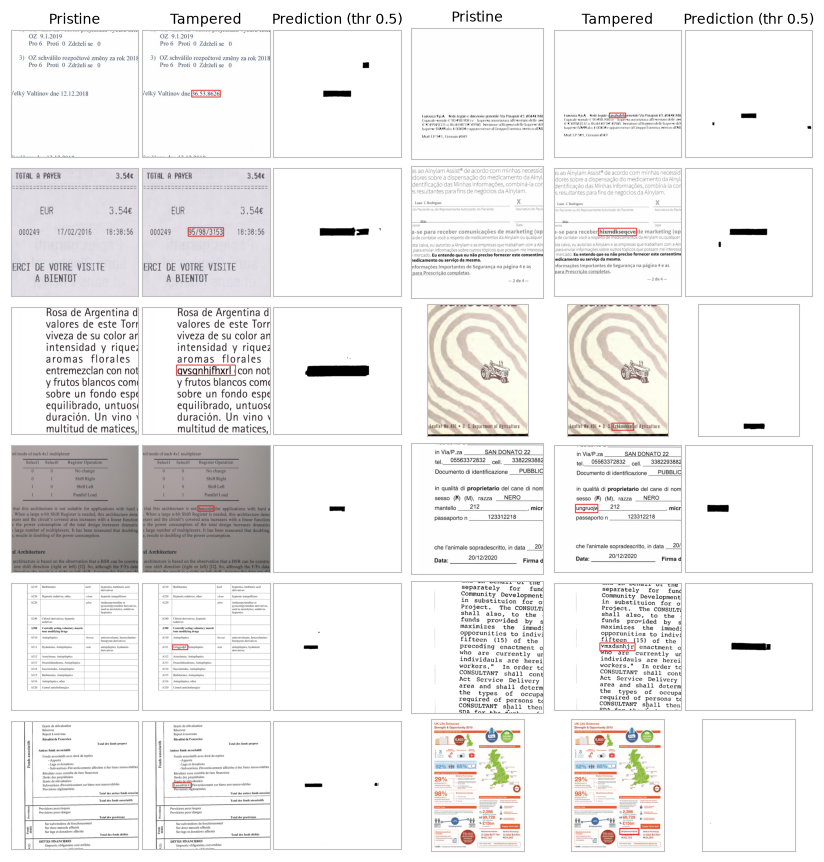}
  \caption{Twelve manipulations from the Qwen-Image-Edit test set, with the
  prediction of FFDN pretrained on our data. Columns: pristine crop,
  tampered crop with the ground-truth region outlined, predicted tampered region at a $0.5$ threshold.}
  \label{fig:qual-qwen-ffdn}
\end{figure*}

\paragraph{Qualitative examples.}
\Cref{fig:qual-fluxtext-ffdn,fig:qual-anytext-ffdn,fig:qual-qwen-ffdn} show, for
each generator, Twelve manipulations from the corresponding test
set, together with the prediction of FFDN pretrained on our data. Each row shows
the pristine crop, the tampered crop with the manipulated region outlined, and
the pixels predicted as tampered at a $0.5$ decision threshold. The sample is
restricted to scanned documents for comparability, excluding camera-captured and
warped sources, and to manipulations of at least eight characters.

\paragraph{Relation to AI-generated text tampering.}
\citet{qu2025revisiting} study open-set tampered scene-text detection across seen and unseen text-editing models. More recently, \citet{qu2026detect} investigate AI-counterfeited text images produced through full-image generation, text removal, and text editing. These works primarily study detector and dataset design for AI-generated counterfeits. Our work addresses a complementary problem by generating realistic and scalable supervision for document-tampering localization.

Our main evaluation focuses on human-made document manipulations, where source regions and editing operations can be deliberately selected to match the surrounding typography, background, alignment, and layout. Nevertheless, the results above show that models trained exclusively on data generated by our approach generalize strongly to unseen AI-based editors.

We do not include generative text-editing models in our data-generation pipeline because our objective is to evaluate whether supervision produced by our generator transfers to external manipulation mechanisms.
 Incorporating the evaluated AI editors during training would weaken this test and tie the generated dataset to particular, rapidly evolving model families. AI editors could nevertheless complement our framework as an additional source of manipulation diversity in future work.

\begin{table*}[!tp]
\centering
\caption{Ablation results on RTM, FindItAgain, and FindIt.}
\label{tab:ablation-per-dataset-results}
\resizebox{\linewidth}{!}{%
\setlength{\tabcolsep}{0.5pt}
\fontsize{6.5pt}{7.8pt}\selectfont
\begin{tabular}{l|l|ccc|ccc|ccc|ccc|ccc|ccc}
\toprule
\multirow{2}{*}{\textbf{Model}} & \multirow{2}{*}{\textbf{Data Gen.}} &
\multicolumn{6}{c|}{\textbf{RTM}} & 
\multicolumn{6}{c|}{\textbf{FindItAgain}} & 
\multicolumn{6}{c}{\textbf{FindIt}} \\
\cmidrule(lr){3-8}\cmidrule(lr){9-14}\cmidrule(lr){15-20}
 & & \multicolumn{3}{c|}{Image} & \multicolumn{3}{c|}{Pixel} &
 \multicolumn{3}{c|}{Image} & \multicolumn{3}{c|}{Pixel} &
 \multicolumn{3}{c|}{Image} & \multicolumn{3}{c}{Pixel} \\
 & & P & R & F1 & P & R & F1 &
 P & R & F1 & P & R & F1 &
 P & R & F1 & P & R & F1 \\
\midrule

\multirow{2}{*}{PSCC-Net}
& Ours & 39.3 & 40.2 & 39.7 & 10.5 & 10.8 & 10.6 & 56.8 & 60.0 & 58.3 & 10.6 & 12.8 & 11.6 & 62.9 & 55.0 & 58.7 & 12.0 & 12.1 & 12.0 \\

& w/o $F_{\theta},G_{\theta}$ & 38.8 & 39.7 & 39.2 & 6.8 & 9.9 & 8.1 & 42.4 & 40.0 & 41.2 & 8.3 & 10.7 & 9.4 & 46.0 & 57.5 & 51.1 & 11.9 & 11.8 & 11.8 \\
\midrule

\multirow{2}{*}{CAT-Net}
& Ours & 40.0 & 35.0 & 37.3 & 9.2 & 11.0 & 10.0 & 48.0 & 34.3 & 40.0 & 11.0 & 5.6 & 7.4 & 51.5 & 42.5 & 46.6 & 13.0 & 10.5 & 11.7 \\
& w/o $F_{\theta},G_{\theta}$ & 38.6 & 34.8 & 36.6 & 9.7 & 10.1 & 9.9 & 48.0 & 34.3 & 40.0 & 10.5 & 5.6 & 7.3 & 51.6 & 40.0 & 45.1 & 12.7 & 10.1 & 11.2 \\
\midrule

\multirow{2}{*}{DTD}& Ours & 36.6 & 73.3 & 48.9 & 14.9 & 11.4 & 12.9 & 50.0 & 40.0 & 44.4 & 12.5 & 12.4 & 12.5 & 62.3 & 66.2 & 64.2 & 14.0 & 14.4 & 14.2  \\
& w/o $F_{\theta},G_{\theta}$ & 35.7 & 65.7 & 46.2 & 12.9 & 10.9 & 11.8
& 46.4 & 37.1 & 41.3 & 10.7 & 9.9 & 10.3
& 60.3 & 55.0 & 57.5 & 14.7 & 13.5 & 14.0 \\
\midrule

\multirow{2}{*}{ADCD-Net}
& Ours & 47.0 & 51.7 & 49.2 & 24.9 & 15.7 & 19.3 & 52.3 & 65.7 & 58.2 & 13.6 & 14.9 & 14.2 & 82.8 & 66.2 & 73.6 & 19.1 & 17.9 & 18.5 \\
& w/o $F_{\theta},G_{\theta}$ & 44.3 & 51.3 & 47.5 & 16.4 & 16.7 & 16.6 & 43.5 & 57.1 & 49.4 & 12.6 & 11.9 & 12.2 & 69.6 & 68.7 & 69.2 & 17.5 & 19.4 & 18.4 \\
\midrule

\multirow{2}{*}{ASC-Former}
& Ours & 44.3 & 32.3 & 37.4 & 23.4 & 16.1 & 19.1 & 38.6 & 62.9 & 47.8 & 15.3 & 12.3 & 13.6 & 69.6 & 68.8 & 69.2 & 30.1 & 13.9 & 19.0 \\
& w/o $F_{\theta},G_{\theta}$ & 44.3 & 32.3 & 37.4 & 12.5 & 20.4 & 15.5 & 35.8 & 54.3 & 43.2 & 13.4 & 8.4 & 10.3 & 66.7 & 62.5 & 64.5 & 24.9 & 12.2 & 16.4 \\
\midrule

\multirow{2}{*}{FFDN}
& Ours & 46.4 & 51.0 & 48.6 & 28.2 & 20.6 & 23.8 & 69.0 & 57.1 & 62.5 & 31.6 & 21.4 & 25.5 & 82.3 & 70.0 & 75.7 & 36.3 & 28.3 & 31.8 \\
& w/o $F_{\theta},G_{\theta}$ & 50.2 & 42.5 & 46.0 & 22.1 & 20.4 & 21.2 & 60.9 & 40.0 & 48.3 & 18.5 & 19.3 & 18.9 & 70.4 & 71.3 & 70.8 & 34.1 & 27.3 & 30.3 \\
\bottomrule
\end{tabular}%
}

\end{table*}

\section{Assessing the impact of F and G on tampering quality}
\label{appendix:generateddataquality}
\subsection{Quantitative assessment}
\subsubsection{Localization metrics as a proxy of Tampering Quality} Assessing the quality of tampered documents via human annotation is time-consuming and costly. To obtain a proxy for tampering quality, we evaluate fixed pretrained detectors on four test sets generated from the same base images using the full pipeline or ablations of \(\mathcal{F}_\theta\) and \(\mathcal{G}_\theta\). The images used come from the CC-MAIN-2021-31-PDF-UNTRUNCATED corpus, excluding any documents that were used during detector pretraining. 

Specifically, the fixed pretrained detectors are evaluated on:
\begin{itemize}
    \item 10,000 tampered documents generated using our full pipeline (including \(\mathcal{F}_\theta\) and \(\mathcal{G}_\theta\));
    \item 10,000 tampered documents generated using only \(\mathcal{G}_\theta\) (ablating $F_{\theta}$);
    \item 10,000 tampered documents generated using only \(\mathcal{F}_\theta\) (ablating $G_{\theta}$);
    \item 10,000 tampered documents generated without using \(\mathcal{F}_\theta\) or \(\mathcal{G}_\theta\).
\end{itemize}

Each image then undergoes a compression with a ratio of fifty percent. We then compare the localization results as a measure of the tampering signal present in each set. Since the same pretrained models and base images are used across all settings, a lower F1 score indicates that the manipulations are more difficult for the detectors to localize.
As shown in \cref{tab:tampering-quality}, documents generated using our full pipeline yield the lowest pixel-level F1 scores, supporting the effectiveness of the auxiliary networks in producing more realistic manipulations.
\subsection{Qualitative assessment}

\begin{figure}[!htbp]
    \centering
    \setlength{\tabcolsep}{2pt}
    \begin{tabular}{cc}
        \includegraphics[width=0.48\linewidth]{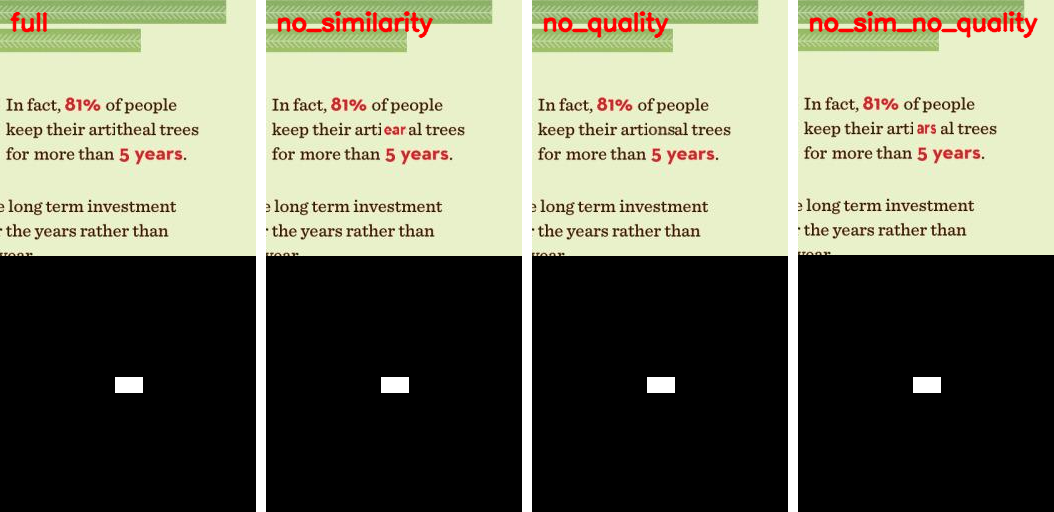} &
        \includegraphics[width=0.48\linewidth]{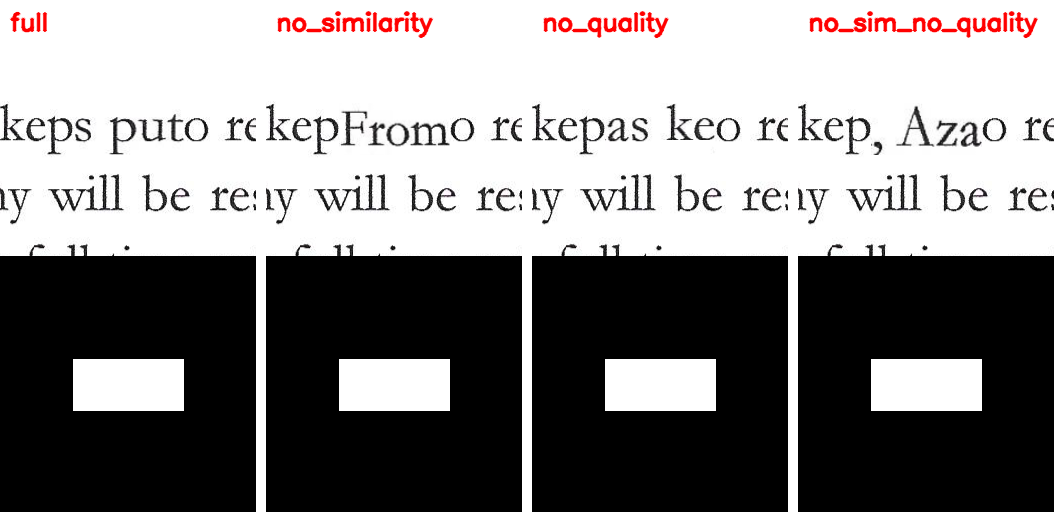} \\
        \includegraphics[width=0.48\linewidth]{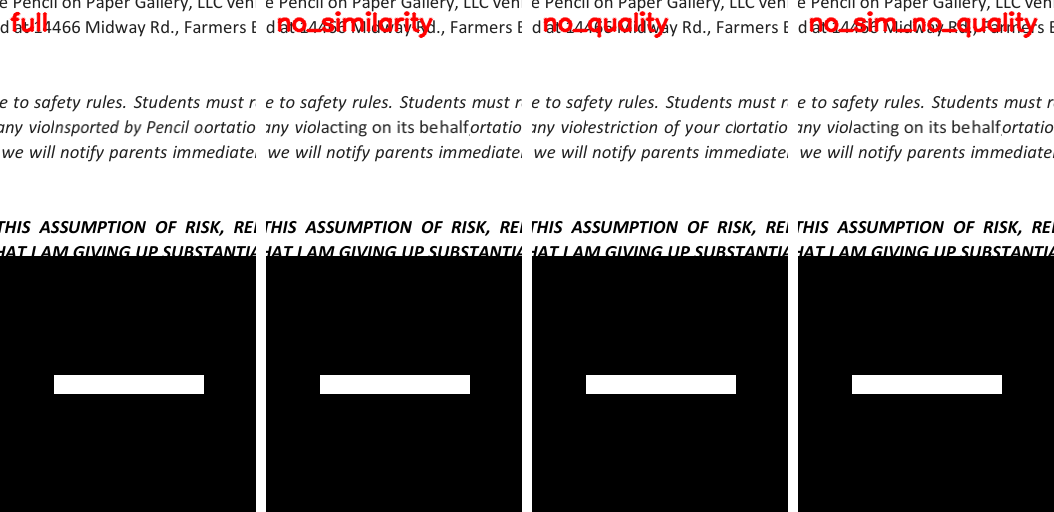} &
        \includegraphics[width=0.48\linewidth]{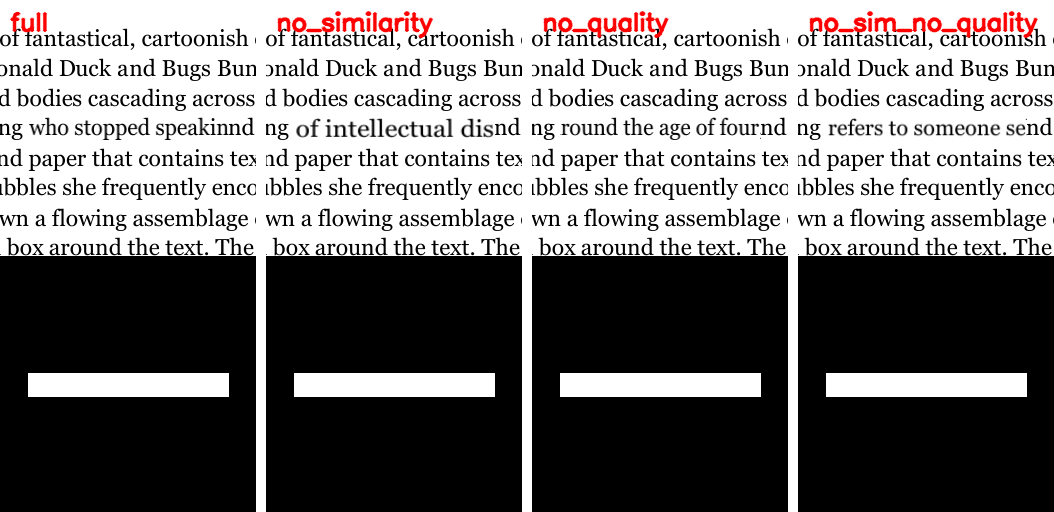} \\
    \end{tabular}
    \caption{
    Visual comparison of tampered regions generated under four settings: 
    with both $\mathcal{F}_\theta$ and $\mathcal{G}_\theta$ enabled (\textbf{full}), 
    with the similarity network $\mathcal{F}_\theta$ disabled (\textbf{no\_similarity}), 
    with the bounding-box quality assessment network $\mathcal{G}_\theta$ disabled (\textbf{no\_quality}), 
    and with both disabled (\textbf{no\_sim\_no\_quality}).  
    These examples illustrate the impact of each auxiliary network on tampering realism.
    }
    \label{fig:impact-aux-networks}
\end{figure}

Figure~\ref{fig:impact-aux-networks} provides a qualitative evaluation of the impact of the 
similarity network $\mathcal{F}_\theta$ and the bounding-box quality assessment network $\mathcal{G}_\theta$ on the realism of the generated tampering.  
Across a variety of document layouts and font styles, the benefits of the full pipeline are clear. When both networks are active (left column), the replaced regions match the surrounding text, producing visually coherent manipulations with no artifact left by ill-defined crops. Disabling $\mathcal{F}_\theta$ while keeping $\mathcal{G}_\theta$ (\textit{no\_similarity}) often leads to noticeable inconsistencies: characters become slightly misaligned, spacing drifts, and the texture or color among other visual characteristics of the pasted crop differ from the surrounding line. These artifacts appear because the system can no longer ensure that the selected source crop is visually compatible with the target region. Conversely, disabling $\mathcal{G}_\theta$ while keeping $\mathcal{F}_\theta$ (\textit{no\_quality}) results in bounding boxes that occasionally cut into neighboring characters or include traces of surrounding text. Although the chosen crop may be visually similar, the poor geometric fit introduces clear edge artifacts that would not appear in a realistic forgery. Finally, disabling both auxiliary networks (\textit{no\_sim\_no\_quality}) leads to the most degraded results. The pasted regions frequently exhibit mismatched fonts or colors and often overlap the original text in implausible ways.  
These examples indicate that the two components provide complementary benefits for producing visually coherent tampering. Overall, the qualitative comparison supports the use of both $\mathcal{F}_\theta$ and $\mathcal{G}_\theta$ in the generation pipeline.

\section{Further comparison of models trained using our data generation approach and previous work}
\subsection{Qualitative comparison of localization performance}
\label{appendix:qualitative}

\begin{figure}[!htbp]
    \centering
    \includegraphics[width=\linewidth]{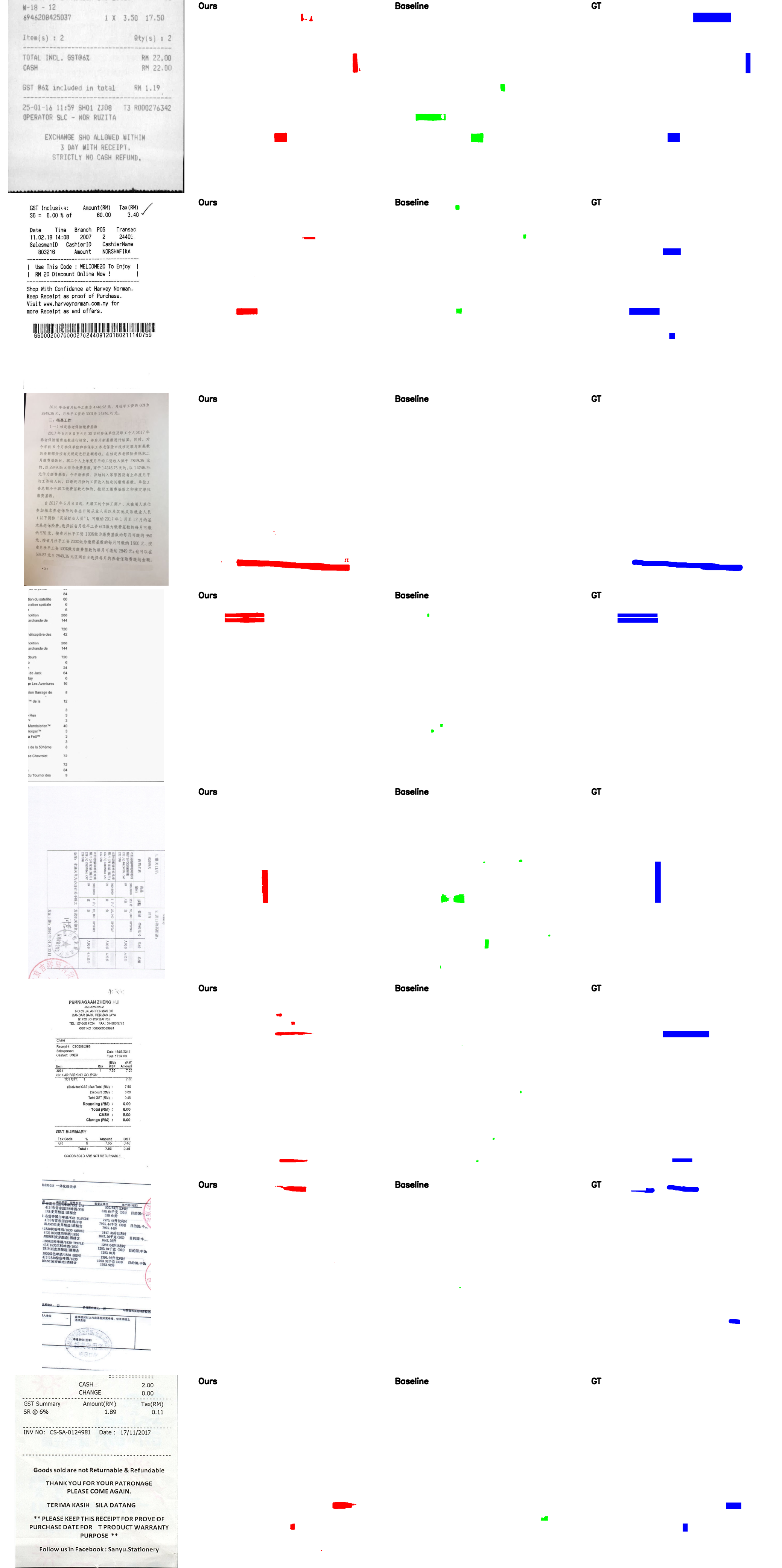}
    \caption{
    Qualitative comparison of the FFDN segmentation maps when the model is trained on data generated by our approach (\textbf{Ours}) versus the best of models trained on data generated by \cite{Doctamper,dataset2,newdataandopenopen}, which we denote collectively as \textbf{Baseline} in the figure. 
    Each block shows the input image (left), the predicted tampering mask of our model (red), the baseline prediction (green), and the ground-truth mask (blue). 
    Images and ground-truth masks are from the RTM dataset.
    }
    \label{fig:comparison-ours-vs-baseline}
\end{figure}

Figure~\ref{fig:comparison-ours-vs-baseline} compares the localization maps produced by FFDN when trained on data generated by our approach versus the best of models trained on data generated by \cite{Doctamper,dataset2,newdataandopenopen}. Overall, the examples illustrate that models trained on data generated by our approach localize tampered regions more accurately, especially in the harder cases. In these examples, both variants produce false positives. The model trained on our data produces fewer false positives and more frequently localizes difficult tampered regions.
These qualitative trends confirm that the proposed generation pipeline provides more informative supervision for challenging document forgeries, leading to sharper and more reliable localization.

\subsection{Precision-Recall curves}
\label{appendix:precrecallcurves}
\begin{figure}[!htbp]
    \centering
    \begin{subfigure}[b]{0.48\linewidth}
        \centering
        \includegraphics[width=\linewidth]{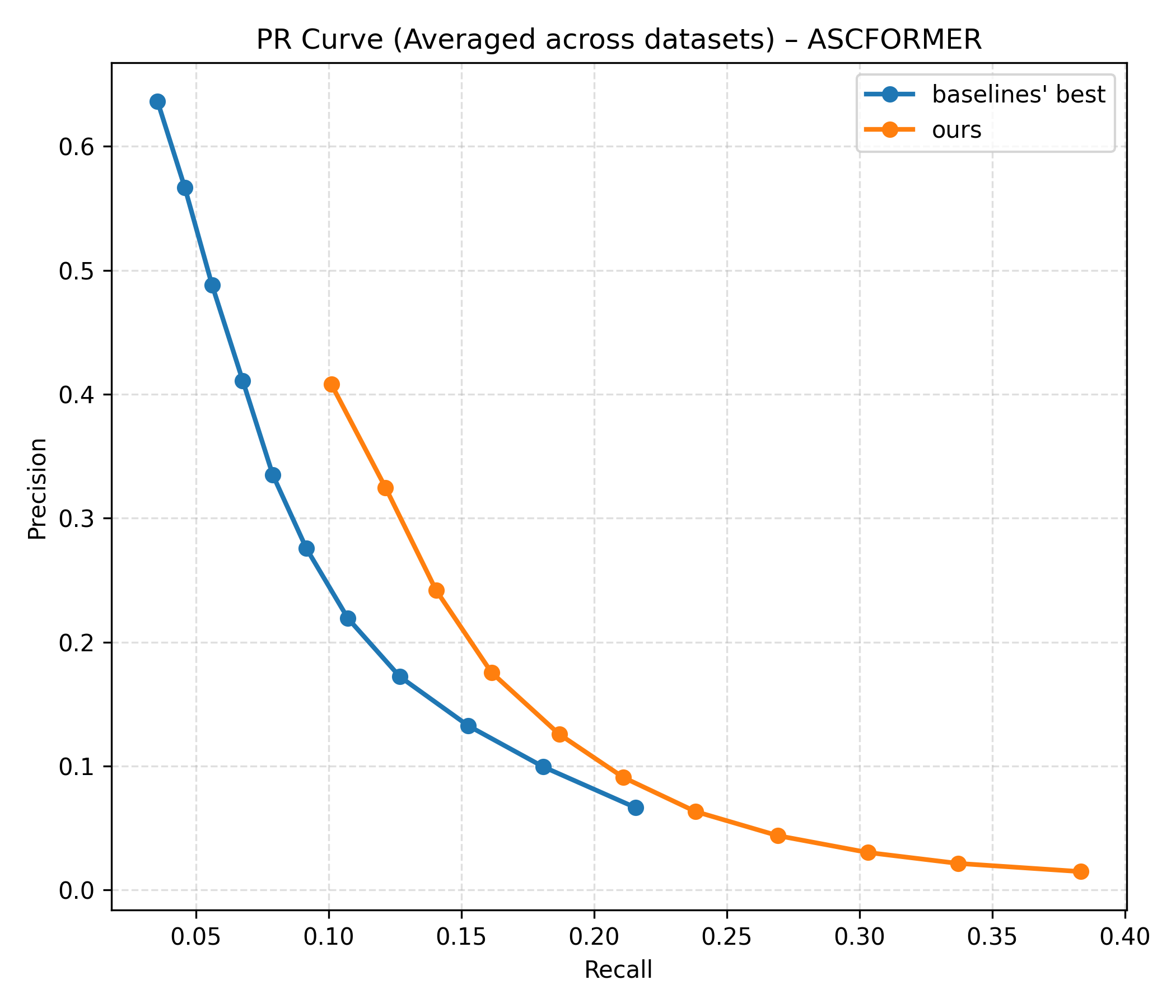}
        \caption{ASC-Former}
    \end{subfigure}
    \hfill
    \begin{subfigure}[b]{0.48\linewidth}
        \centering
        \includegraphics[width=\linewidth]{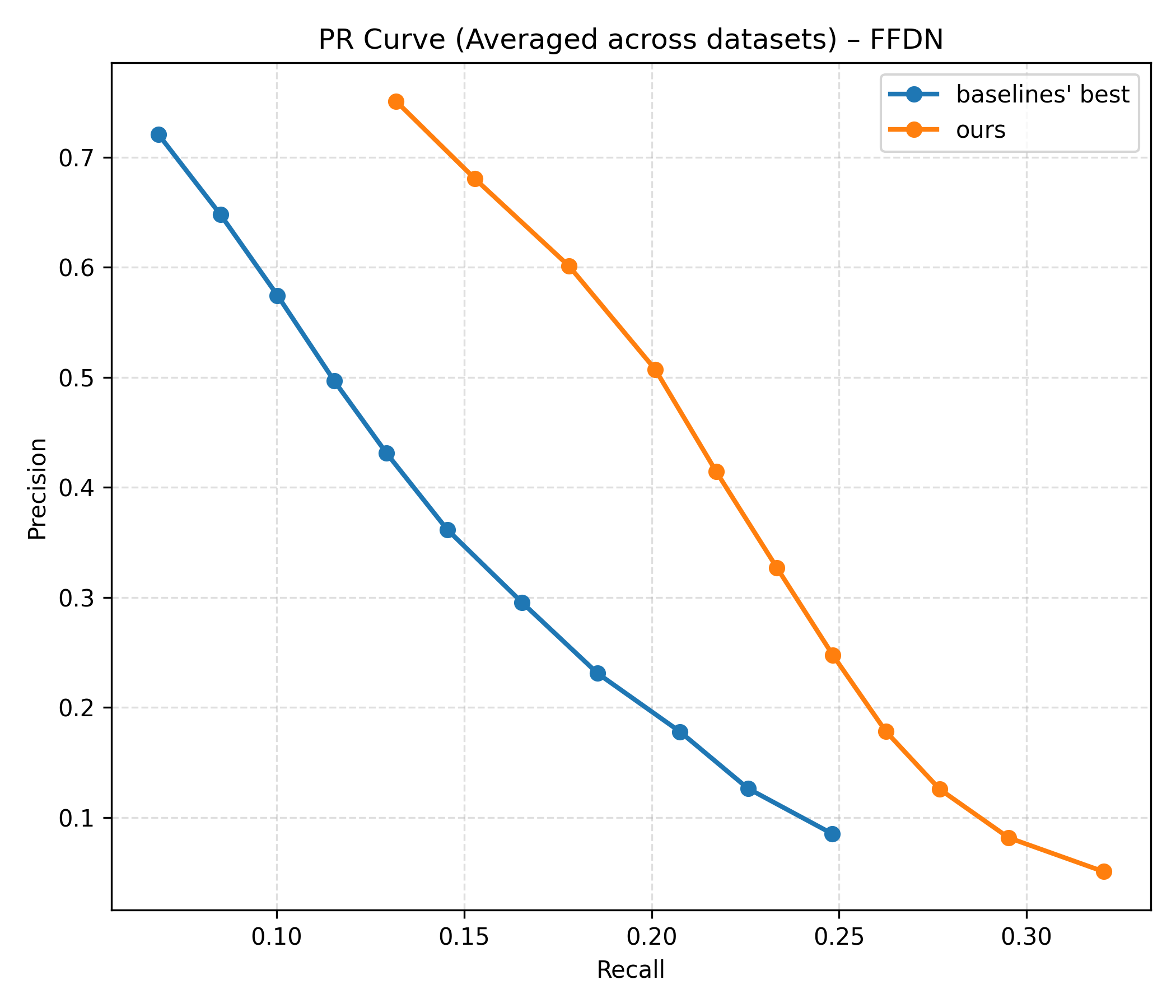}
        \caption{FFDN}
    \end{subfigure}
    \caption{Precision–recall curves averaged across datasets for ASC-Former and FFDN.}
    \label{fig:pr_curves}
\end{figure}

To further demonstrate that our data generation approach surpasses the baselines, we provide in \cref{fig:pr_curves} the precision–recall curves for FFDN and ASC-Former. These curves show that, at equal precision, the recall of the final models trained using our data generation approach is higher. The same also holds at equal recall.
\subsection{Per tampering type out-of-domain results}
\label{appendix:per-type}
RTM~\cite{RTM} provides tampering-type annotations, enabling us to evaluate models trained on data generated using our approach against those trained on the best baseline generator in the out-of-domain setting. We report the results in \cref{tab:data_gen_vs_xx}. We report IoU separately for the five tampering types: Copy-Move (CM), Splicing (SP), Insertion (IN), Inpainting (IP) and Coverage (CV). Training on data generated with our approach yields consistent gains for both architectures. The largest improvements are observed for CM, SP and CV: for example, FFDN gains nearly $4$ IoU points on both CM and SP (from $11.8\%$ to $15.7\%$ and from $11.3\%$ to $15.2\%$), and more than $4$ points on CV (from $11.3\%$ to $15.9\%$). ASC-Former exhibits the same pattern, with gains of more than $3$ IoU points on CM and SP and over $4$ points on CV.

\begin{table}[!htbp]
  \small
  \setlength{\tabcolsep}{1pt}
  \centering
  \caption{IoU performance (\%) of FFDN and ASC-Former trained on data generated with our method compared to the best baseline generator from~\cite{Doctamper,dataset2,newdataandopenopen}. CM, SP, IN, CV and IP denote Copy-Move, Splicing, Insertion, Coverage and Inpainting, respectively. The best performance in each column is highlighted in bold.}
  \label{tab:data_gen_vs_xx}
  \resizebox{\linewidth}{!}{\begin{tabular}{lcccccc}
    \toprule
    Method & CM & SP & IN & CV & IP & Mean (5 types) \\
    \midrule
    FFDN (ours, data-gen)            & \textbf{15.7} & \textbf{15.2} & \textbf{11.0} & \textbf{15.9} & \textbf{11.3} & \textbf{13.8} \\
    FFDN (best of~\cite{Doctamper,dataset2,newdataandopenopen})         & 11.8          & 11.3          & 10.6          & 11.3          & 10.5          & 11.1          \\
    ASC-Former (ours, data-gen)      & \textbf{12.8}          & \textbf{11.5}          & \textbf{7.8}          & \textbf{12.1}          & \textbf{9.3}           & \textbf{10.7}          \\
    ASC-Former (best of~\cite{Doctamper,dataset2,newdataandopenopen})   & 9.5           & 8.5           & 7.3           & 7.8           & 8.9           & 8.4           \\
    \bottomrule
  \end{tabular}}

\end{table}

\section{On the time cost and robustness of our data generation approach}
\subsection{Generation time analysis}
\label{appendix:datagentimeanalysis} 

The generation-time experiment was conducted using one NVIDIA A100 SXM4 GPU with 80 GB of memory and 16 CPU cores from an AMD Milan EPYC 7543 compute node. On this hardware, the average cost
of generating a single tampered region is approximately
$0.15$ seconds. This includes all operations. Within this total time cost, the two auxiliary networks used during generation,
$\mathcal{F}_\theta$ (crop similarity) and $\mathcal{G}_\theta$ (crop quality),
contribute about $19.3\%$ of the total per-region cost, while the remainder
is dominated by classical disk and CPU bound operations (image I/O, cropping, resizing),
text rendering, and inpainting. Generating the entire dataset of $2.8$ million generated images with
corresponding pixel-level masks required about $290$ hours on a single A100
with 16 CPU cores. In practice, the process
 can be parallelized and can be further sped up by distributing different
shards of the CSV list across multiple machines or GPUs, but the generation times reported here were measured on a single node.

Both networks are used only during offline dataset generation and introduce no additional computation into the downstream localization models.

\subsection{OCR quality robustness}
\label{appendix:ocrqualityrobustness} 

Our generation pipeline relies on character-level OCR to obtain the initial
bounding boxes used by \texttt{ExtractLineSegments} and by the subsequent
tampering stages. A natural concern is therefore how sensitive the overall
detection performance of models trained on this data is to the choice and accuracy of the OCR engine. We study this in two ways: (i) replacing the default OCR system with
a different engine (we choose Tesseract), and (ii) explicitly perturbing the OCR
bounding boxes before data generation.

The full results are reported in \cref{tab:ocr-full}.

\begin{table*}[!tp]
\centering
\caption{Full OCR robustness results with precision, recall, and F1 for both image-level and pixel-level evaluation across all three benchmarks. Scores are averaged across six of the seven detectors, excluding DITL$^2$.}
\label{tab:ocr-full}
\resizebox{\textwidth}{!}{%
\begin{tabular}{l|ccc|ccc|ccc|ccc|ccc|ccc}
\toprule
\multirow{2}{*}{\textbf{Setting}}
& \multicolumn{6}{c|}{\textbf{RTM}}
& \multicolumn{6}{c|}{\textbf{FindItAgain}}
& \multicolumn{6}{c}{\textbf{FindIt}} \\
\cmidrule(lr){2-7}\cmidrule(lr){8-13}\cmidrule(lr){14-19}
& \multicolumn{3}{c|}{\textbf{Image}} & \multicolumn{3}{c|}{\textbf{Pixel}}
& \multicolumn{3}{c|}{\textbf{Image}} & \multicolumn{3}{c|}{\textbf{Pixel}}
& \multicolumn{3}{c|}{\textbf{Image}} & \multicolumn{3}{c}{\textbf{Pixel}} \\
& P & R & F1 & P & R & F1
& P & R & F1 & P & R & F1
& P & R & F1 & P & R & F1 \\
\midrule
Google OCR  & 42.3 & 47.2 & 43.5 & 18.5 & 14.3 & 16.0 & 52.4 & 53.3 & 51.9 & 15.8 & 13.2 & 14.1 & 68.6 & 61.4 & 64.7 & 20.8 & 16.2 & 17.9 \\
Tesseract OCR       & 41.7 & 46.8 & 43.0 & 18.1 & 13.9 & 15.6 & 51.8 & 52.9 & 51.5 & 15.5 & 13.0 & 13.7 & 68.0 & 61.0 & 64.3 & 20.5 & 15.9 & 17.5 \\
Perturbed OCR boxes & 41.9 & 46.3 & 42.8 & 17.7 & 13.6 & 15.3 & 52.0 & 52.3 & 51.4 & 15.1 & 12.8 & 13.3 & 68.2 & 60.6 & 64.1 & 20.2 & 15.6 & 17.2 \\
\bottomrule
\end{tabular}%
}
\end{table*}

\paragraph{Replacing the OCR engine.}
In the first experiment, we replace the default OCR back-end, Google OCR, with
Tesseract, keeping all other components of the generation and training
pipeline unchanged. We then regenerate the full tampered training set and
retrain the downstream detection models on this new data. When evaluated in
the same out-of-domain setting as our main experiments, the average Pixel-level
and Image-level F1 score decreases by only $0.4$ points compared to the original
pipeline, with similarly small changes in precision and recall as presented in \cref{tab:ocr-robustness}.

\paragraph{Perturbing OCR bounding boxes.}
In the second experiment, we keep the default OCR engine but artificially
degrade its output before generation. For each OCR bounding box, we randomly
expand or contract its boundaries by an integer offset sampled uniformly from
$[-3, 3]$ pixels on each side. If the resulting width or height becomes
smaller than $3$ pixels, the box is discarded. Despite this additional noise, the final detection performance remains stable:
the average Pixel-level F1 score drops by only $0.7$ points
relative to the baseline, and Image-level results show a comparable small
degradation.

\section{Hyperparameters analysis}
\label{appendix:hyper}

\begin{table*}[!tp]
\centering

\setlength{\tabcolsep}{4pt}
\caption{Analysis of the effect of the hyperparameter choice on detection and localization performance. Results are averaged across ADCD-Net, ASC-Former, and FFDN, and across evaluation datasets. }
\label{tab:ablation-hyperparam}
\begin{tabular}{l|ccc|ccc}
\toprule
\textbf{Setting} & \multicolumn{3}{c|}{\textbf{Image-level}} & \multicolumn{3}{c}{\textbf{Pixel-level}} \\
 & \textbf{Prec} & \textbf{Rec} & \textbf{F1} & \textbf{Prec} & \textbf{Rec} & \textbf{F1} \\
\midrule
Ours & 59.1 & 58.4 & 58.0 & 24.7 & 17.9 & 20.5 \\
\midrule
\multicolumn{7}{l}{\textit{Contrastive}} \\
hard negatives $M_{\text{alt}}=30$ & 59.0 & 58.7 & 58.1 & 25.2 & 18.1 & 20.7 \\
hard negatives $M_{\text{alt}}=0$ (ablation)  & 58.0 & 56.6 & 56.8 & 23.9 & 16.8 & 19.5 \\
positive-pair distance threshold: $1.25\times\tau_0$ & 59.3 & 58.0 & 57.9 & 24.5 & 17.7 & 20.2 \\
positive-pair distance threshold: $0.75\times\tau_0$ & 58.7 & 58.5 & 57.8 & 24.4 & 17.9 & 20.3 \\
positive-pair distance threshold: $2.0\times\tau_0$ & 58.4 & 57.1 & 56.8 & 24.2 & 16.7 & 19.4 \\
negative-pair vertical distance threshold: $1.25\times\tau_1$ & 59.6 & 57.9 & 57.9 & 24.5 & 18.0 & 20.5 \\
negative-pair vertical distance threshold: $0.75\times\tau_1$ & 58.9 & 58.1 & 57.7 & 24.4 & 17.8 & 20.3 \\
negative-pair vertical distance threshold: $0.33\times\tau_1$ & 57.3 & 55.8 & 55.6 & 22.8 & 15.6 & 18.1 \\
aspect-ratio tolerance for negative mining: $1.25\times\epsilon$ & 59.2 & 58.3 & 58.0 & 24.6 & 18.0 & 20.4 \\
aspect-ratio tolerance for negative mining: $0.75\times\epsilon$ & 59.2 & 58.4 & 57.9 & 24.9 & 17.9 & 20.5 \\
aspect-ratio tolerance for negative mining: $\epsilon=0.0$ & 58.0 & 57.2 & 56.5 & 23.7 & 16.8 & 19.1 \\
\midrule
\multicolumn{7}{l}{\textit{Generation}} \\
aspect-ratio tolerance $\epsilon'=\infty$ (ablation) & 58.9 & 57.7 & 57.5 & 24.4 & 17.3 & 20.0 \\
aspect-ratio tolerance $\epsilon'=0.0$ & 58.4 & 57.9 & 57.0 & 24.0 & 17.2 & 19.6 \\
aspect-ratio tolerance $1.25\times\epsilon'$ & 59.2 & 58.5 & 58.1 & 24.8 & 17.9 & 20.4 \\
aspect-ratio tolerance $0.75\times\epsilon'$ & 59.0 & 58.1 & 57.8 & 24.5 & 17.9 & 20.3 \\
$p_{\text{spl}}=1.0$ (no copy-move) & 54.1 & 55.0 & 53.6 & 19.7 & 12.1 & 14.1 \\
$p_{\text{spl}}=0.0$ (no splicing) & 58.7 & 57.3 & 57.0 & 24.0 & 17.0 & 19.5 \\
\bottomrule
\end{tabular}

\end{table*}
 The default hyperparameter values were not chosen by tuning on the evaluation benchmarks, but from simple design considerations about the role of each component. For the contrastive module, we use standard contrastive settings for the temperature and number of negatives, and we set the remaining mining rules to obtain enough informative positives and negatives without making the supervision too noisy: $\tau_0$ balances positive coverage with visual similarity, $\tau_1$ makes negatives visually different from the anchor on average while preserving enough within-document negatives, $M_{\text{alt}}$ guarantees enough hard negatives, and $\epsilon$ avoids excessive stretching while keeping a sufficiently large negative pool. For the generation module, the tampering-type probabilities are heuristic design choices intended to maintain a diverse mixture of manipulation categories; $n_{\max}=5$ is used to avoid generating pages containing an excessive number of modified regions; and the aspect-ratio tolerance $\epsilon'$ is set to provide enough candidate crops for $\mathcal{F}_\theta$ while avoiding visibly unrealistic distortions. Overall, these choices are intended to place the method in a sensible operating regime rather than to optimize for a specific test set. To validate these choices and analyse the effect of each hyperparameter, we provide a sensitivity analysis for the main hyperparameters of the contrastive and generation modules. A full grid search would require regenerating large synthetic corpora and retraining several localization models, which is computationally expensive. We therefore vary one parameter at a time around the default configuration to assess local sensitivity and choose extreme values to test how much degradation they cause. For the contrastive module, the selected values test whether the mining procedure remains effective when positives are made slightly more or less permissive, when negatives are made slightly easier or harder, and when the amount of hard-negative augmentation is increased or removed. For the generation module, the selected values test whether the candidate pool remains realistic and sufficiently diverse when the aspect-ratio tolerance is moderately relaxed or tightened, and whether manipulation diversity remains important when splicing is overused or removed.

A practical difficulty in this analysis is that the final performance depends not only on the hyperparameter itself, but also on the randomness of data generation and contrastive pair construction. To make the comparisons fairer, we use paired randomization across compared settings while keeping the original algorithms unchanged. We note that the same strategy was used for the ablation study.

For the generation pipeline, we first fix the same source-image subset for all compared settings (which we did also to compare generation pipelines and which makes sense). Then, for each target image, we reuse the same random seed for the stochastic decisions that are already present in the generation algorithm: sampling the number of manipulated regions \(n\), sampling the manipulated regions themselves, and the Bernoulli decisions associated with insertion, inpainting, and splicing. When inpainting is selected, we also reuse the same seed for the choice between the two inpainting modes. The generator samples among qualifying donors or rendered candidates according to their similarity scores. Rendering itself is deterministic for a fixed text, font, color, and background.

For the contrastive pipeline, we apply the same principle only to the stochastic parts of the mining procedure. In the contrastive algorithm, positive mining uses random choice when several valid positives are available, and hard negatives are created through the randomized altered-crop procedure. We therefore reuse the same random seeds for these two operations across compared settings. By contrast, the intra-image negatives are traversed in a deterministic order, so no additional treatment is needed there. When cross-document negatives are required, we again keep the same seed across compared settings so that the sampled external negatives remain as aligned as possible. Finally, when retraining \(\mathcal{F}_\theta\) or the downstream localization models for the sensitivity study, we also keep the same seed across compared settings.

This paired protocol reduces variance without changing the underlying method. In particular, it avoids attributing differences to a hyperparameter when they are in fact caused by unrelated changes in sampled target regions, branch outcomes, or hard-negative augmentations. At the same time, it preserves the intended behavior of the method: candidate crops in generation are still chosen by \(\mathcal{F}_\theta\), and the contrastive pipeline still uses the same mining rules and the same randomized hard-negative construction. Thus, when two nearby settings produce different results, the difference is much more likely to come from the hyperparameter itself rather than from uncontrolled stochastic drift.

The results are reported in \cref{tab:ablation-hyperparam}. Overall, the method remains stable around the default configuration: moderate changes to \(\tau_0\), \(\tau_1\), \(\epsilon\), and \(\epsilon'\) lead to only small variations in performance, whereas clearly extreme settings consistently degrade both image-level and pixel-level metrics. This behavior matches the intended role of the default values. The contrastive thresholds are meant to provide enough informative pairs without introducing excessive ambiguity, and the aspect-ratio tolerances are meant to keep enough candidates while avoiding visibly unrealistic distortions. Likewise, the default splicing probability is chosen to preserve manipulation diversity rather than collapsing the generator to a single dominant tampering type.

A few nearby settings can match or slightly exceed the default on the averaged metrics. This is not surprising, since the reported numbers are averaged over several localization models and evaluation datasets, and the corresponding differences are very small. We nevertheless keep the original default values, because they were chosen from design considerations rather than selected after observing test-set performance. Retuning them to maximize performance on the evaluation benchmarks would amount to indirect test-set leakage. Our goal here is therefore not to claim that the default values are uniquely optimal, but rather to show that they lie in a sensible and stable operating regime.

The extreme cases are also informative. Removing hard negatives (\(M_{\text{alt}}=0\)) noticeably reduces performance, confirming that augmented difficult negatives help learn a more discriminative similarity function. Likewise, using an overly large positive-pair neighborhood (\(2.0\times\tau_0\)), a very small vertical separation for negatives (\(0.33\times\tau_1\)), or setting \(\epsilon=0\) reduces the usefulness or diversity of the mined contrastive pairs. On the generation side, both \(\epsilon'=\infty\) and \(\epsilon'=0\) are worse than the default for complementary reasons: the former allows unrealistic stretching, while the latter over-restricts the candidate pool. Finally, using only splicing (\(p_{\text{spl}}=1.0\)) or removing splicing entirely (\(p_{\text{spl}}=0.0\)) is also detrimental, supporting the use of a mixed generation strategy. The largest degradation occurs when copy-move is removed, suggesting that maintaining copy-move examples is particularly important for the evaluated benchmarks.

\section{Reproducibility Details}
\label{appendix:reproducibility}

\paragraph{Computational environment.}
 Detector pretraining used eight NVIDIA A100 GPUs per model except for ADCD-Net, which used 16, using PyTorch Lightning with distributed data parallelism. The generation-time experiment used one NVIDIA A100 GPU and 16 CPU cores. We train $\mathcal{F}_\theta$ with one document per GPU and accumulate gradients over eight steps. For $\mathcal{G}_\theta$, each GPU processes 32 prepared files per batch.

\paragraph{Software environment.}
The operating system used is
Red Hat Enterprise Linux 9 (RHEL 9).
Experiments were conducted using Python 3.11.12, PyTorch 2.7.0 with CUDA 12.8, torchvision 0.22.0, PyTorch Lightning 2.6.0, transformers 4.52.4, Albumentations 2.0.8, OpenCV 4.11.0.86, NumPy 2.2.6, scikit-image 0.25.2, and Google Cloud Vision 3.10.1. The reported experiments use \texttt{jpegio} 0.2.4 for JPEG quantization-table extraction. Environment and installation instructions are available in the \href{https://github.com/Mohamed-Dhouib/TDOC}{project repository}.

\paragraph{Optimization and input processing.}
Trainable parameters are optimized with AdamW using weight decay \(0.01\). The learning rate uses linear warmup followed by cosine decay. No gradient clipping is applied. Training uses bfloat16 mixed precision. RGB inputs are normalized using channel means \((0.485,0.455,0.406)\) and standard deviations \((0.229,0.224,0.225)\).

 We cap the training resolution at $1024 \times 1024$ and select the maximum resolution that fits in GPU memory with eight images per GPU, except for ADCD-Net, which uses four per GPU. We use $1024 \times 1024$ for DTD and ASC-Former, $768 \times 768$ for FFDN and CAT-Net, $512 \times 512$ for PSCC-Net and DITL$^2$, and $384 \times 384$ for ADCD-Net. Images smaller than the corresponding resolution are padded with white image pixels and zero-valued mask pixels, while larger images are cropped to that resolution. During evaluation, smaller pages are centrally padded and then restored to their original extent. For the document-level benchmarks, larger pages are processed using windows matching the architecture-specific input resolution, and overlapping probabilities are merged by pixelwise maximum. The final prediction is cropped to the original page dimensions rather than geometrically resized.

\paragraph{Randomness and experiment configurations.}
We use fixed random seeds for Python, NumPy, PyTorch, data-loader workers, and distributed training. For each controlled comparison, the same random-seed configuration is used across data-generation approaches for a given localization architecture. The \href{https://github.com/Mohamed-Dhouib/TDOC#resources}{project repository} provides training configurations and links to the model weights and to the fine-tuning and evaluation splits.

\paragraph{Number of runs.}
Unless explicitly stated otherwise, each reported result is obtained from one training run under the fixed configuration described above. The experiment in \cref{smaller_scale} uses four independently generated training datasets and four corresponding training runs for each reported configuration.

\paragraph{Evaluation metrics.}
We evaluate image-level detection and pixel-level localization using precision, recall, and F1 score for the tampered class:
\begin{equation}
\operatorname{Precision}
=
\frac{\mathrm{TP}}{\mathrm{TP}+\mathrm{FP}},
\qquad
\operatorname{Recall}
=
\frac{\mathrm{TP}}{\mathrm{TP}+\mathrm{FN}},
\end{equation}
\begin{equation}
\operatorname{F1}
=
\frac{2\operatorname{Precision}\operatorname{Recall}}
{\operatorname{Precision}+\operatorname{Recall}}.
\end{equation}

For pixel-level evaluation, TP, FP, and FN are accumulated globally over all pixels in the evaluation set before computing the metrics.  The six architectures other than DITL$^2$ include a two-class image-level classification head, whereas image-level predictions for DITL$^2$ are derived from its pixel probability maps, as described in \cref{appendix:ditl-image-eval}. The image-level target is positive if and only if the corresponding pixel mask contains at least one positive pixel. For architectures with a classification head, the training objective is
\[
  \mathcal{L}
  =
  \mathcal{L}_{\mathrm{seg}}
  +
  \lambda_{\mathrm{class}}\mathcal{L}_{\mathrm{class}},
  \qquad
  \lambda_{\mathrm{class}}=0.1,
\]
where both terms use focal loss with \(\gamma=2\). Image-level TP, FP, and FN counts are accumulated over the evaluation set.

We additionally report threshold-independent pixel AUROC and AUPRC in \cref{appendix:threshold-metrics}. ROC points are ordered by increasing false-positive rate, and AUROC is computed using trapezoidal integration:
\begin{equation}
\operatorname{AUROC}
=
\sum_{k=1}^{K}
\left(\operatorname{FPR}_{k}-\operatorname{FPR}_{k-1}\right)
\frac{\operatorname{TPR}_{k}+\operatorname{TPR}_{k-1}}{2}.
\end{equation}
AUPRC is computed as average precision with operating points ordered by increasing recall:
\begin{equation}
\operatorname{AUPRC}
=
\sum_{k=1}^{K}
\left(\operatorname{Recall}_{k}-\operatorname{Recall}_{k-1}\right)
\operatorname{Precision}_{k}.
\end{equation}
Here, \(\operatorname{TPR}=\operatorname{Recall}\) and
\[
\operatorname{FPR}
=
\frac{\mathrm{FP}}{\mathrm{FP}+\mathrm{TN}}.
\]

F1 is used as the primary threshold-dependent metric because both document-level detection and pixel-level localization are class-imbalanced and require balancing false positives and false negatives. AUPRC is particularly informative under pixel-level class imbalance, while AUROC summarizes the trade-off between the true-positive and false-positive rates across thresholds.

\section{Further Discussion}
\label{appendix:discussion}
Here we discuss points that were not discussed in the main text due to space limit, including limitations of our approach, evaluation challenges, and future directions.

\paragraph{Performance limitations and evaluation challenges.}
Although our proposed generation strategy provides a substantial performance boost for document tampering detection models, absolute performance remains low. We attribute this to several factors. First, the evaluation datasets contain many images with no tampered regions, which makes achieving high F1 challenging because the model must correctly classify images with no tampering while also detecting and localizing tampering in images that do contain it. Second, human-curated datasets used for evaluation often involve sophisticated, case-specific concealment techniques that are carefully designed to be visually convincing, making them inherently difficult to detect. These sophisticated techniques often include context-aware forgeries that blend seamlessly with the surrounding document content. Third, document tampering evaluation has long been dominated by synthetic datasets, and this reliance has hindered architectural progress toward robust generalization to real-world cases. The gap between synthetic and real-world tampering remains substantial, and closing this gap will require additional research into both data generation and detection model design.

\paragraph{Learning-based approaches for advanced concealment.}
We experimented with simple rule-based local post-processing to improve detection performance; however, we observed performance degradation mainly because these techniques, if not used correctly for each case, can end up revealing the tampering itself or introducing artifacts. This suggests that learning-based approaches are needed to model advanced concealment techniques and to properly handle context-specific post-processing. We consider our current approach a first step toward human-quality tampering generation, and leave this direction for future work.

\paragraph{Rotated documents, photographed documents, and layout diversity.}
Our pipeline makes one structural assumption: pages are not rotated by a large angle when line segments are constructed. We make this choice to avoid unnecessary complexity and keep line construction simple, by grouping adjacent segments only when their vertical offset is smaller than a small threshold; more details on line construction are given in \cref{appendix:getlines}. In practice, most source documents in our corpus exhibit little to no rotation, as discussed in \cref{sec:data}. Large rotations are instead handled through the rotation and perspective augmentations applied when training the localization models, with details provided in \cref{appendix:dataaug}. The method also works on photographed and mildly deformed documents, provided that large page rotation is absent. More generally, our pipeline does not assume a specific document layout. It operates on OCR-derived text and blank segments and has been observed to work across diverse layouts, fonts, image qualities, and backgrounds; examples of generated documents are provided in \cref{appendix:exemplesgeenration}. Beyond the low-rotation constraint, we have not observed other systematic limitations. Our choice to make it only support slightly rotated documents is fully justified, as we make our approach simpler without sacrificing localization performance of trained models, as rotation is applied as a data augmentation along other perspective augmentations. We note also that deskewing methods can be used prior to generation to support all input images. Moreover, one change can make our full pipeline work on rotated documents. We simply need to add a method or a model that outputs the document angle, and then calculate all distances relative to the rotated basis. Formally we have:
\[
\mathbf{e}_{\parallel}=(\cos\theta,\sin\theta), \qquad
\mathbf{e}_{\perp}=(-\sin\theta,\cos\theta),
\]
and measure distances in this basis instead of the canonical horizontal/vertical one. For two segment centers \(\mathbf{p}_i\) and \(\mathbf{p}_j\), the within-line criterion would then become
\[
\left|(\mathbf{p}_j-\mathbf{p}_i)^\top \mathbf{e}_{\perp}\right| < \tau_{\perp},
\]
while ordering and spacing along the line would be measured through
\[
(\mathbf{p}_j-\mathbf{p}_i)^\top \mathbf{e}_{\parallel}.
\]
In other words, the definitions of ``same line'', vertical distance, and normal distance would all be computed relative to the rotated basis rather than the image axes.

\paragraph{Assumptions in contrastive mining.}
The contrastive mining strategy relies on a simple document prior: on average, adjacent segments on the same line are more visually similar to each other than to segments on different lines beyond a vertical threshold. This is used only as a natural property of documents, not as a strict claim that always holds. We also do not rely solely on this assumption. Hard negatives are added by perturbing the anchor, with the corresponding transformations detailed in \cref{appendix:metadataaug}, and negatives can also be sampled from other documents when needed, as described in \cref{f}. To avoid trivial cues such as padding or obvious size differences, negatives are resized to the anchor width. At the same time, unrealistic distortions are limited by prioritizing aspect-ratio matches and bounding the mismatch with a tolerance parameter. During generation, candidates that require visible stretching can look unrealistic, so it is reasonable that the similarity network leverages such cues, among others, to reject them.

\paragraph{Training prior and generation-time selection.}
The same-line property is used only to construct positive and negative pairs for training $\mathcal{F}_\theta$. It is not imposed as a constraint during tampered-image generation. Once trained, $\mathcal{F}_\theta$ directly evaluates the visual similarity of candidate source-target regions regardless of whether they belong to the same line, different lines, or different documents.
\paragraph{Metrics.}
Different works on document tampering detection and localization use different evaluation metrics. For example, \citet{forensichub} reports the F1 score by averaging per-image F1 values while skipping images with no positive pixels. In our case, we adopt a direct global aggregation scheme: we first accumulate TP, TN, FP, and FN over all images, and then compute precision, recall, and F1 from these totals. The same global aggregation is used for image-level evaluation: the per-image decisions are accumulated into TP, TN, FP, and FN counts over the evaluation set, from which precision, recall, and F1 are computed.

\subsection{Data provenance, privacy, and societal impact}

\paragraph{Data provenance and privacy.}
Our dataset is generated from existing publicly available
document corpora and benchmarks, as detailed in
\cref{sec:data}. We additionally incorporate approximately
175{,}000 public-domain government documents collected
from the Official Journal of the French Republic,
as described in \cref{appendix:scraping}.
Our approach operates on these existing document
collections without requiring the acquisition of private
documents from individuals. To ensure independent
evaluation, we apply the source-overlap filtering
procedure described in \cref{appendix:outofdomain}.

\paragraph{Societal impact.}

Our work aims to advance document authenticity verification
by enabling the generation of large-scale, realistic
tampering examples for training and evaluating forensic
models. By improving the generalization of tampering
detectors to human-made manipulations, the proposed
framework can contribute to more reliable document
forensics. We make our generated dataset, generation
code, and pretrained model weights publicly available
to support reproducible research and further progress
in document tampering detection and localization.

\end{document}